\documentclass[final]{cvpr}

\usepackage{times}
\usepackage{epsfig}
\usepackage{graphicx}
\usepackage{amsmath}
\usepackage{amssymb}
\usepackage{pifont}
\usepackage{color}
\usepackage{enumitem}
\usepackage{mathtools}
\usepackage{bm}
\usepackage{bbding}
\usepackage{booktabs}
\usepackage{blindtext}
\usepackage{caption}
\usepackage{multirow}
\usepackage{subfig}
\usepackage{footmisc}

\renewcommand{\arraystretch}{1.2}

\newcommand*\samethanks[1][\value{footnote}]{\footnotemark[#1]}

\usepackage{xcolor}
\usepackage{colortbl}
\definecolor{gray}{gray}{0.95}

\usepackage[pagebackref=true,breaklinks=true,colorlinks,bookmarks=false]{hyperref}

\def\cvprPaperID{4083} 
\def\confYear{CVPR 2021}
\begin{document}

\title{Style-based Point Generator with Adversarial Rendering for \\Point Cloud Completion}

\author{Chulin Xie$^{1}$\thanks{Equal contribution.}  \qquad Chuxin Wang$^{2}$\samethanks \qquad Bo Zhang$^3$\qquad Hao Yang$^3$\qquad Dong Chen$^3$\qquad Fang Wen$^3$ \\ 
$^1$University of Illinois at Urbana-Champaign  \qquad $^2$University of Science and Technology of China \qquad \\ $^3$Microsoft Research Asia\\
{\tt\small chulinx2@illinois.edu} \qquad {\tt\small wcx0602@mail.ustc.edu.cn}   \qquad   {\tt\small \{zhanbo,haya,doch,fangwen\}@microsoft.com}
}

\maketitle

\begin{abstract}

In this paper, we proposed a novel Style-based Point Generator with Adversarial Rendering (SpareNet) for point cloud completion. 
Firstly, we present the channel-attentive EdgeConv to fully exploit the local structures as well as the global shape in point features. 
Secondly, we observe that the concatenation manner used by vanilla foldings limits its potential of generating a complex and faithful shape. Enlightened by the success of StyleGAN, we regard the shape feature as style code that modulates the normalization layers during the folding, which considerably enhances its capability. 
Thirdly, we realize that existing point supervisions, e.g., Chamfer Distance or Earth Mover's Distance, cannot faithfully reflect the perceptual quality of the reconstructed points. To address this, we propose to project the completed points to depth maps with a differentiable renderer and apply adversarial training to advocate the perceptual realism under different viewpoints. Comprehensive experiments on ShapeNet and KITTI prove the effectiveness of our method, which achieves state-of-the-art quantitative performance while offering superior visual quality. Code is available at \href{https://github.com/microsoft/SpareNet}{https://github.com/microsoft/SpareNet}.  

\end{abstract}

\section{Introduction}

As the 3D scanning devices such as depth camera and LiDAR become ubiquitous, point clouds get easier to acquire and have recently attracted a surge of research interest in the vision and robotics community. 
However, raw points directly captured by those devices are usually sparse and incomplete due to the limited sensor resolution and occlusions. Hence, it is essential to infer the complete shape from the partial observation so as to facilitate various downstream tasks~\cite{guo2020deep} such as classification and shape manipulation as required in real-world applications.

Due to the irregularity and unorderedness of point clouds, one workaround is to leverage intermediate representations, \eg, depth map \cite{hu2019render4completion} or voxels \cite{xie2020grnet}, that are more amenable to neural networks. However, the representation transform may result in information loss, so the detailed structures can not be well preserved. With the emergence of point-based networks~\cite{qi2017pointnet,qi2017pointnet++,thomas2019kpconv,dgcnn,guo2020deep}, predominant methods~\cite{foldingnet_2018_CVPR,Yuan-2018-pcn,atlasnet2018,wen2020point,liu2019morphing,topnet_2019_CVPR,chen2019unpaired,cascaded_2020_CVPR, pfnet_2020_CVPR, detailpreserved_eccv2020} nowadays digest the partial inputs directly and estimate the complete point clouds in an end-to-end manner. These methods typically follow the encoder-decoder paradigm and adopt permutation invariant losses~\cite{fan2017pointsetgeneration}, \eg, Chamfer Distance or Earth Mover's Distance, for regressing the groundtruth. 

Ideally, the point completion network should simultaneously meet the following needs: 1) The output is desired to faithfully preserve the detailed structures of the partial input; 2) The network has a strong imaginative power to infer the global shape from the partial clue; 3) the local structure should be sharp, accurate, and free from the corruption by the noise points. Nonetheless, existing methods fail to achieve the above goals because of the neglecting of the global context during the feature extraction, the insufficient capability of modeling the structural details, and the lack of perceptual metrics for measuring the visual quality.

\begin{figure*}[t]
\centering
 \includegraphics[width=0.7\linewidth]{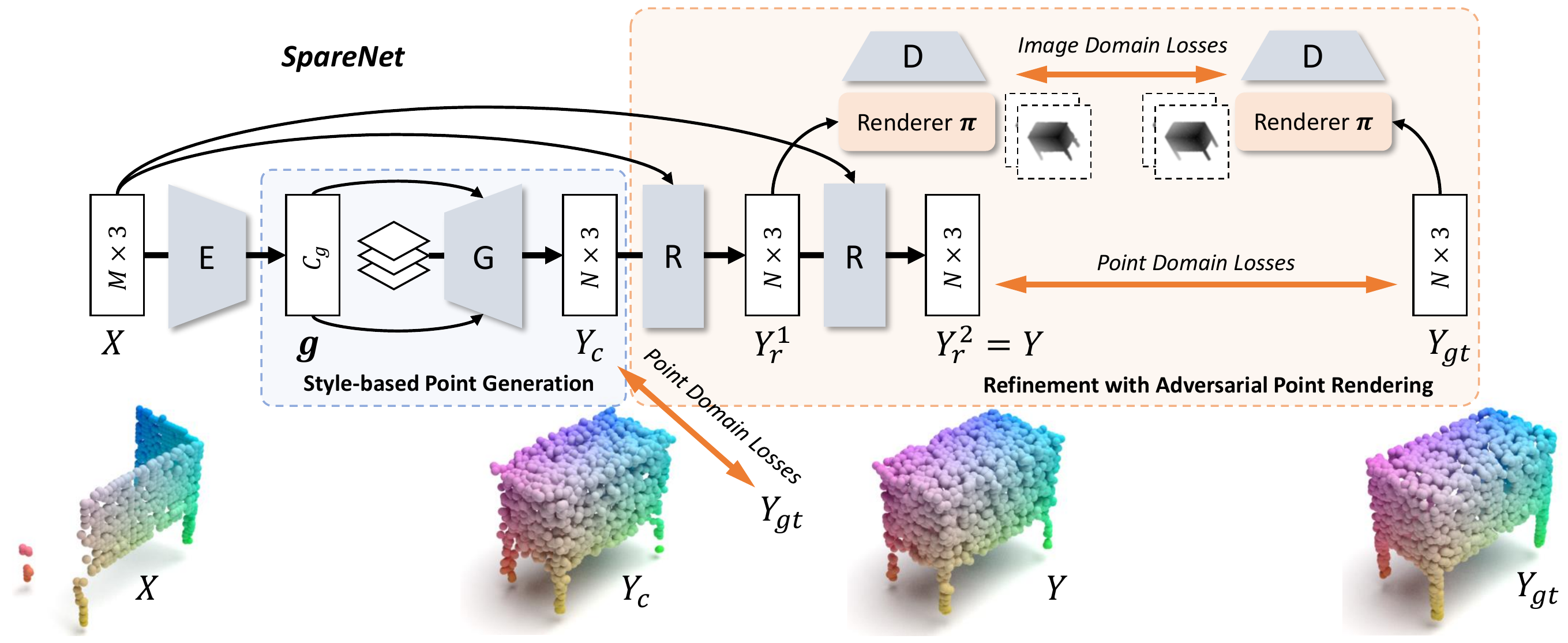}
 \footnotesize
    \caption{
    The architecture of SpareNet. An encoder $\mathtt{E}$ encodes the partial points $X$ into a shape code $\mathbf{g}$, leveraged by a style-based generator $\texttt{G}$ to synthesize a coarse completion $Y_c$, which is recurrently improved with refiner $\texttt{R}$ into the final result $Y$. Adversarial point rendering is applied to advocate the perceptual realism of completed points under different views.
    }
\label{fig:overview}
\vspace{-1em}
\end{figure*}

In this paper, we propose  \emph{Style-based Point generator with Adversarial REndering}, \ie, the SpareNet, to circumvent the above issues. We have made improvements from encoder, generator, and loss function, and proposed 3 new modules: channel-attentive EdgeConv, Style-based Point Generator, and Adversarial Point Rendering. Firstly, while previous works employ PointNet or PointNet++ to learn point-wise or local features, we propose channel-attentive EdgeConv (Section~\ref{ssec:edgeconv}), which not only considers the local information within the $k$-nearest neighbors but also wisely leverages the global context by aggregating the global features and weighting the feature channel attention for each point accordingly. The fusion of local and global context enriches the learnt representation, so the network is more powerful to characterize fine structures of the input. 

Further, we claim that the vanilla folding module~\cite{foldingnet_2018_CVPR} in conventional methods, which outputs the 3D shapes by morphing the 2D surfaces through multilayer perceptrons (MLP), has limited modeling capability due to the improper usage of the features, \ie, the features are tiled and concatenated to each location of the 2D lattice. Drawn by the success of StyleGAN~\cite{karras2019style} in image synthesis, we boost the folding capability by regarding the learnt features as style codes, which can be used to modulate the feature normalization within the folding MLPs. The resulting style-based generator, as elaborated in Section~\ref{ssec:generator}, shows considerably improved capability of modeling structural details. 

Last but not least, in order to generate visually-pleasing results, we propose to project the generated point clouds to view images (Section~\ref{sec:rendering}), whose realism is further examined by adversarial discriminators. Since the renderer we use is differentiable, the gradient from the discriminators will guide the network to learn the completion with high perceptual quality when viewed at different angles. We conduct extensive experiments on ShapeNet~\cite{chang2015shapenet} and KITTI~\cite{geiger2013vision} datasets, and our SpareNet performs favorably over state-of-the-art methods both quantitatively and qualitatively.

\section{Related Work}

\noindent\textbf{Point cloud processing.}
Pioneer works \cite{qi2017pointnet,zaheer2017deep} propose max-pooling to aggregate the features of individual points to ensure permutation invariance. Such aggregation, however, neglects the contextual relationships among different points, resulting in the representation incapable to characterize the fine structures. To amend this, \cite{qi2017pointnet++} hierarchically groups the points and extracts features for the local context. Inspired by the tremendous success of 2D convolution, a surge of conv-based methods~\cite{hua2018pointwise,xu2018spidercnn,groh2018flex,atzmon2018point,li2018pointcnn,thomas2019kpconv,dgcnn} has recently emerged, which generalizes the convolution to irregular coordinate space. Meanwhile, graph-based methods~\cite{dgcnn,wang2018local} regard the point clouds as graph structures where points are treated as nodes, and their local connectivity is denoted by edges. In~\cite{dgcnn}, the EdgeConv is proposed to process the $k$-nearest neighbor graph and dynamically models the locality according to not only the coordinate distance but also the semantic affinity. While these networks are powerful to characterize the local structure, they fail to simultaneously consider the global shape, \ie, the local feature extraction is unaware of the global information.

\noindent\textbf{Point cloud reconstruction.}
To hallucinate the complete 3D coordinates, previous works design point decoders in various forms: multilayer perceptrons (MLPs) \cite{chen2019unpaired}, hierarchical structures like a tree \cite{topnet_2019_CVPR} or a multi-level pyramid \cite{pfnet_2020_CVPR}, or through iterative refinement \cite{cascaded_2020_CVPR}.


While some works resort to proxy representations other than points for shape completion, like volumetric grids \cite{dai2017shape,xie2020grnet} or depth maps \cite{hu2019render4completion}, FoldingNet \cite{foldingnet_2018_CVPR} uses 2D manifolds to represent 3D point clouds, which also inspires others to model the target point clouds as non-linear foldings of 2D grids. 
PCN \cite{Yuan-2018-pcn} predicts the folding of a single 2D patch, whereas AtlasNet \cite{atlasnet2018} and MSN~\cite{liu2019morphing} generate the output with multiple patches. Recently, SA-Net \cite{wen2020point} proposes a hierarchical folding that progressively hallucinates the detailed structures. Instead of improving the folding mechanism, SFA-Net~\cite{detailpreserved_eccv2020} addresses the information loss during the global feature extraction. Albeit effective, these folding-based methods feed the partial features to folding modules via concatenation, but we claim that such a concatenation manner would impair the capacity of the foldings. In comparison, our style-based point generator greatly enhances the capacity for modeling structural details.

\section{The SpareNet}

The overview architecture of SpareNet is exhibited in Figure \ref{fig:overview}. Given a partial and low-res point cloud $X$ as input, SpareNet first completes $X$ with a coarse point cloud $Y_c$ through an encoder and a generator: the encoder $\mathtt{E}$ embeds $X$ into a shape code $\mathbf{g}$, the style-based generator $\mathtt{G}$ exploits the shape code $\mathbf{g}$ and synthesizes the coarse output $Y_c$. In addition, SpareNet adopts a refinement part that employs adversarial point rendering to further refine the coarse points $Y_c$ and outputs a final complete and high-res point cloud $Y$ with improved visual quality.


\begin{figure}[t]
\centering
 \includegraphics[width=0.9\linewidth]{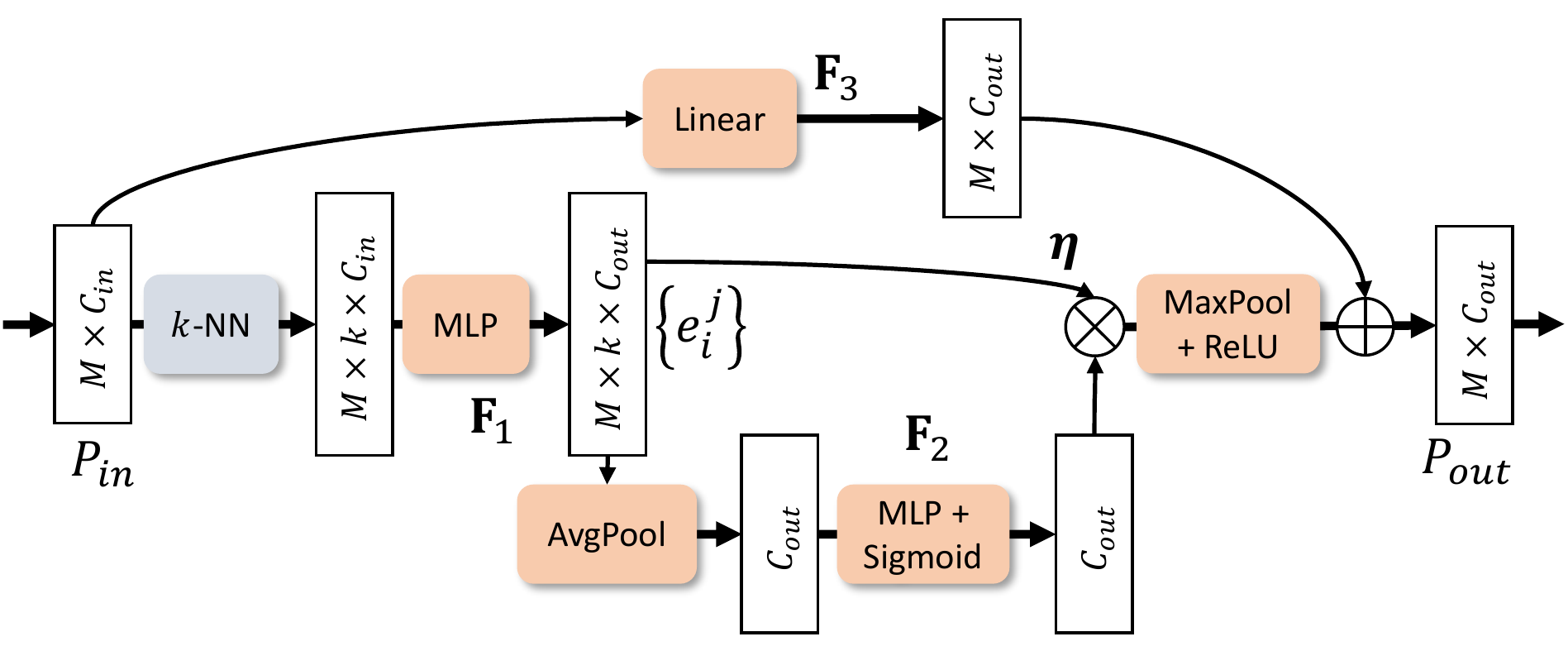}
 \footnotesize
    \caption{The structure of the Channel-Attentive EdgeConv.}
\label{fig:encoder}
\vspace{-1em}
\end{figure}

\subsection{Channel-Attentive EdgeConv}
\label{ssec:edgeconv}

We devise the \emph{\textbf{C}hannel-\textbf{A}ttentive \textbf{E}dgeConv} (CAE) to simultaneously integrate both local and global context from point features. It is inspired by the EdgeConv \cite{dgcnn} that captures a local context and the Squeeze-and-Excitation blocks \cite{hu2018senet} for capturing a global context. Our point encoder $\mathtt{E}$ heavily relies on the CAE blocks. 

Let $P_{in}$ be the input of a CAE block. Suppose it has $M$ points with feature dimension $C_{in}$. 
For each point $p_i \in P_{in}$, we first find its \emph{k}-nearest neighbors in $P_{in}$ with respect to the Euclidean distance defined in the $C_{in}$-dimensional feature space ($k=8$ in experiments). 
Denote these neighbors as $\{q_i^j, 1\leq j \leq k\}$, we have $k$ directional edges on $p_i$, with each edge represented as $(p_i, q_i^j - p_i)$. Then we use a multilayer perceptron (MLP) (denoted as $\mathbf{F}_1$) to compute a new feature $e_i^j= \mathbf{F}_1(p_i, q_i^j - p_i)$ from each edge.

In order to leverage a global context from the $k$-NN graph, we feed the global average of all edge features $\{e_i^j, 1{\leq}i{\leq}M, 1{\leq}j{\leq}k\}$ into a second MLP (denoted as $\mathbf{F}_2$) to calculate a gating vector $\bm{\eta}$:
\vspace{-0.5em}
\begin{equation}
\vspace{-0.5em}
\bm{\eta} = \sigma \circ \mathbf{F}_2 \left[ \frac{1}{kM} \times \sum_{i,j}^{M,k} e_i^j \right],
\label{equ:eta}
\end{equation}
where $\sigma$ represents \texttt{sigmoid}.
We re-calibrate every edge feature $e_i^j$ by multiplying it with $\bm{\eta}$.
Finally, for each point, we reduce its \emph{k} edge features into a new point feature through maximum pooling and \texttt{ReLU} activation.
We also add an additional linear layer $\mathbf{F}_3$ that skip-connects the output with the input to make the block residual.

Unlike the T-net presented in \cite{qi2017pointnet} that predicts an affine transformation to apply on point coordinates,
we predict a gating vector that applies on edge features, in order to leverage both the local and global context for feature activation.

Our point encoder $\mathtt{E}$ is built with four sequential CAE blocks. The input $X$ is fed into the first one and passes through all the rest, resulting in four point features with different dimensions and different receptive fields. We concatenate them together and compress the feature dimension through a final MLP. The output shape code $\mathbf{g}$ is derived as a concatenation of two global poolings of the above result: a maximum pooling and an average pooling. It has a dimension of $C_g$, which is set to $4,096$ in our experiments.

\subsection{Style-based Point Generator} 
\label{ssec:generator}

We present a \emph{style-based point generator} $\mathtt{G}$ to generate a completed point cloud from the shape code $\mathbf{g}$ through novel style-based folding layers.
In previous folding methods \cite{foldingnet_2018_CVPR,Yuan-2018-pcn,atlasnet2018,liu2019morphing}, the shape code is tiled and concatenated with $N$ 2D coordinates all sampled from a unit square $[0, 1]^2$. They learn a mapping from such combination into the 3D space using MLP, to emulate the morphing of a 2D grid into a 3D surface. Under such foldings, the shape code determines the morphing through the input concatenations ahead but hardly affects all the layers behind in an effective way. Such concatenation-based folding induces bottleneck that limits its capacity to represent different 3D surfaces.

\begin{figure}[t]
\centering
 \includegraphics[width=\linewidth]{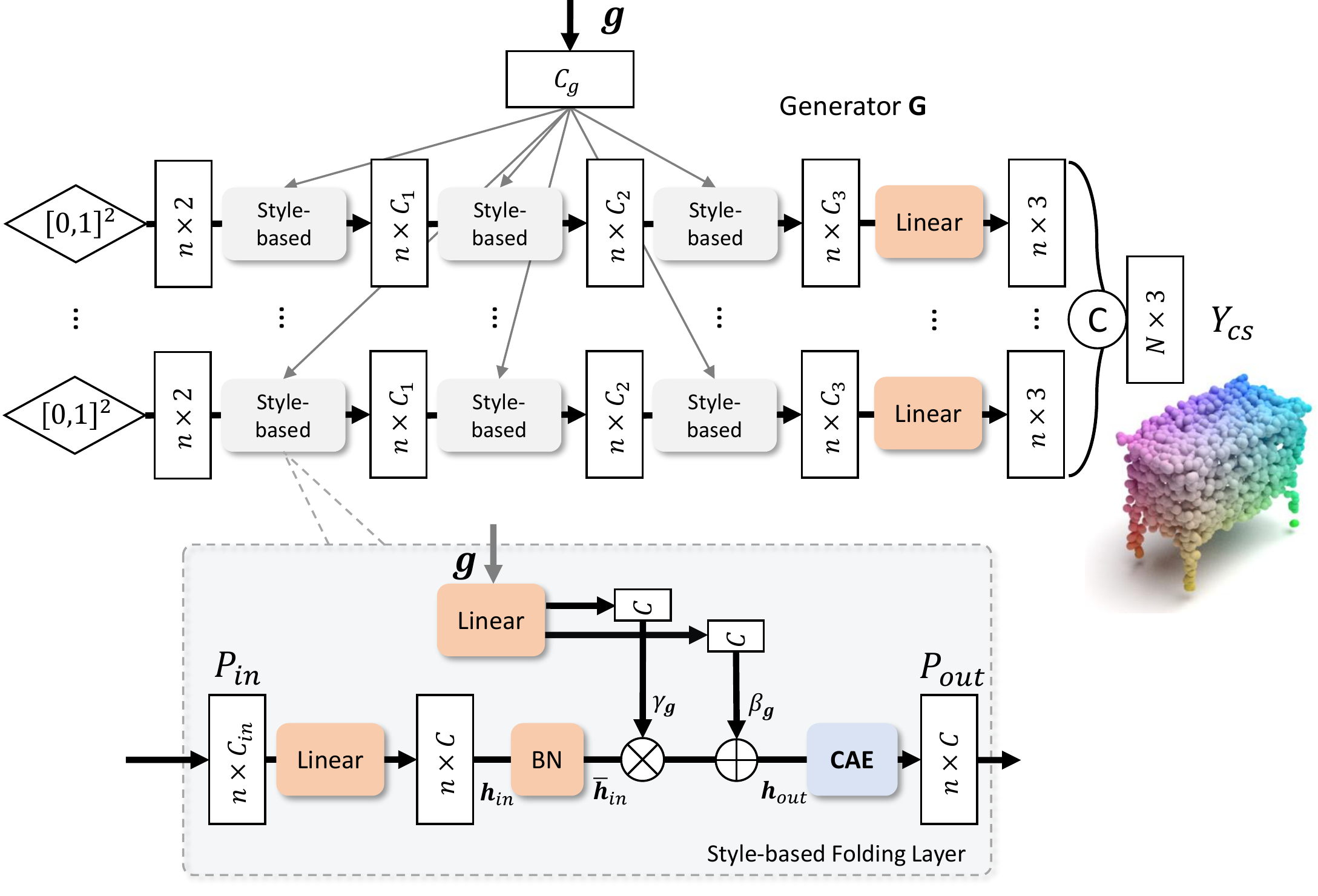}
 \footnotesize
    \caption{The styled-based folding and the generator $\mathtt{G}$.}
\label{fig:generator}
\vspace{-1em}
\end{figure}

Enlightened by the success of StyleGAN \cite{karras2019style} in image generation, we propose style-based folding to circumvent these disadvantages in point cloud generation. 
Figure \ref{fig:generator} exhibits the structure of our generator $\mathtt{G}$.
Instead of combining the shape code $\mathbf{g}$ with grid coordinates as an input to vanilla folding MLPs, we directly inject $\mathbf{g}$ into the generator $\mathtt{G}$'s internal layers, to ensure a more extensive information aggregation in point cloud synthesis.
We design novel style-based folding layers to accomplish such injection. 

A style-based folding layer transforms the input point features $P_{in}$ into new point features $P_{out}$ under the modulation of the shape code $\mathbf{g}$.
In specific, let $\mathbf{h}_{in} \in \mathbb{R}^{B \times M\times C}$ be a mini-batch of point activations that are linearly transformed from $P_{in}$, with $B$ being the batch size, $M$ the number of points and $C$ the dimension of point features. We first normalize $\mathbf{h}_{in}$ to be $\bar{\mathbf{h}}_{in} \in \mathbb{R}^{B \times M\times C}$ batch-wisely: 
\vspace{-0.5em}
\begin{equation}
\vspace{-0.5em}
\bar{\mathbf{h}}_{in} = \frac{\mathbf{h}_{in} - \bm{\mu}_{\mathbf{h}_{in}}}{\bm{\sigma}_{\mathbf{h}_{in}}},
\end{equation}
with $\bm{\mu}_{\mathbf{h}_{in}}, \bm{\sigma}_{\mathbf{h}_{in}} \in \mathbb{R}^{1\times 1\times C}$ being the means and standard deviations of $\mathbf{h}_{in}$'s channel-wise activations.
In order to integrate the shape code, we compute new activation $\mathbf{h}_{out}$ by denormalizing the normalized $\bar{\mathbf{h}}_{in}$ according to the shape code $\mathbf{g}$, with the formulation
\vspace{-0.5em}
\begin{equation}
\vspace{-0.5em}
\mathbf{h}_{out} = \gamma_{\mathbf{g}} \otimes \bar{\textbf{h}}_{in} + \beta_{\mathbf{g}},
\end{equation} 
where $\gamma_{\mathbf{g}}$ and $\beta_{\mathbf{g}}$ are two modulation parameters both transformed from $\mathbf{g}$ through linear layers.
Finally we append a CAE block to $\mathbf{h}_{out}$ to compute the output $P_{out}$.

Same with \cite{atlasnet2018,liu2019morphing}, our generator $\mathtt{G}$ employs $K$ (32 in experiments) surface elements to form a complex shape, as depicted in Figure \ref{fig:generator}. For each surface element, the generator learns a mapping from a unit square $[0, 1]^2$ to a 3D surface through three sequential style-based folding layers and one linear layer. We sample $n=N/K$ points for each surface element. Finally, the $K$ 3D surfaces (each surface represented as $n=N/K$ points) are directly merged together forming the coarse output point cloud $Y_c$ with $N$ points.

\subsection{Adversarial Point Rendering}
\label{sec:rendering}
After generating a coarse point cloud $Y_c$, we additionally refine it to acquire a final result $Y$ with improved quality. But unlike some previous works~\cite{liu2019morphing,cascaded_2020_CVPR}, our refinement guarantees an advantageous visual quality of our final point cloud $Y$ by employing a novel \emph{adversarial point rendering}. 

By \emph{point rendering}, we mean a fully differentiable point renderer that enables end-to-end rendering from 3D point cloud to 2D depth maps. The renderer makes it possible that we can supervise the training not only in the point domain but also in the image domain. By \emph{adversarial}, we mean to adversarially improve the point cloud quality with a discriminator $\mathtt{D}$, which not directly discriminates the 3D point clouds, but their rendered 2D depth maps. Observing the success of image-based convolutional networks, we believe that an image-based convolutional discriminator, combined with our differentiable point renderer, can better capture geometric details in point clouds than the point-based discriminators used by \cite{cascaded_2020_CVPR,chen2019unpaired}. 

\begin{figure}[t]
\centering
 \includegraphics[width=0.69\linewidth]{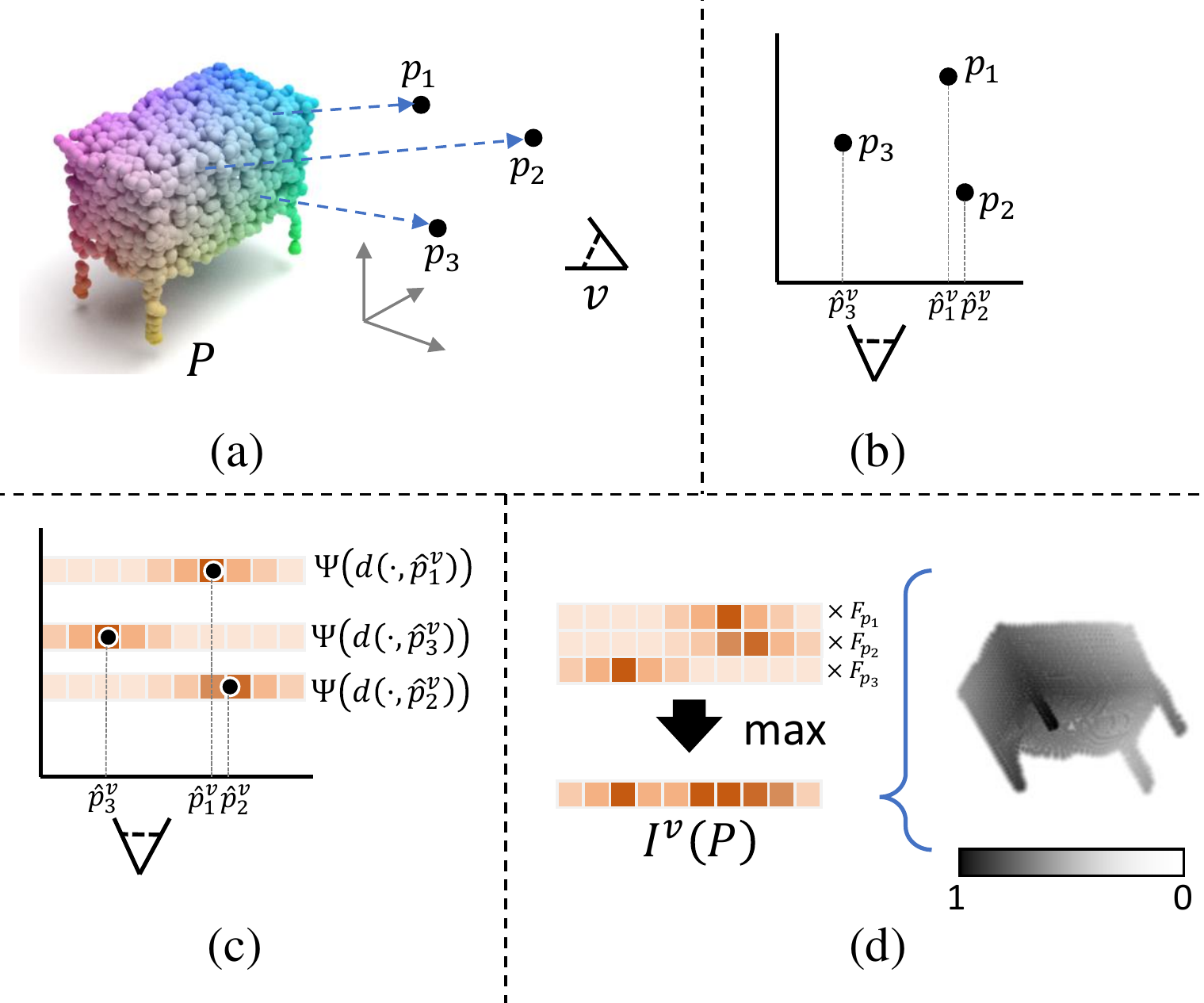}
 \footnotesize
    \caption{Pipeline of our differentiable point rendering. (a) Given 3D points $P$ and a camera view $v$, (b) the 3D points $p_i \in P$ are firstly projected as 2D points $\hat{p}_i^v$ with depths according to $v$. (c) We regard each 2D point as a smooth density function modeled by kernel $\Psi$. (d) A depth map $I^v(P)$ is generated through a pixel-wise maximum reduction of negated point depths $F$ weighted by point densities.}
\label{fig:render}
\vspace{-1em}
\end{figure}

\noindent\textbf{Renderer}
Let $P$ be a point cloud and $v$ be a camera pose,
the point renderer aims to generate a 2D depth map $I^v(P)$ whose pixels reflect $P$'s geometry that is visible in $v$.

As shown in Figure \ref{fig:render}, we start the rendering pipeline by projecting every 3D point $p=(p_x,p_y,p_z) \in P$ onto a projection plane, retrieving a 2D pixel location $(\hat{p}_x^v, \hat{p}_y^v)$ and a depth $\hat{p}_z^v$ (\emph{a.k.a.} distance from the projection). We calculate such projection with a projective transformation $T_v$, which is derived from camera pose $v$ by combining both its extrinsic and intrinsic parameters. 

We then rasterize these 2D points $\hat{p}^v=(\hat{p}_x^v, \hat{p}_y^v)$ with depths $\hat{p}_z^v$ to generate a rendered image.
To guarantee differentiability, we regard each point not as a hard image pixel, but like a density function that spread smooth influence around its center. Let $F_{p}$ be the point feature used for rendering, the rasterization can be formulated as
\vspace{-0.5em}
\begin{equation}
\vspace{-0.5em}
I^v_{x, y}(P) = \max_{p \in P}\left\{\Psi \left( \left\|(x, y), \hat{p}^v \right\|_2  \right) \times F_{p}, 0 \right\},
\end{equation}
where $\Psi(x)=\exp(-{x^2}/{2\rho^2})$ is a Gaussian-shape kernel that models the density function of a point, with $\rho$ being a hyper-parameter that controls the radius of its influence.
In order to render $I^v(P)$ as a depth map, we define $F_{p}=1 - (\hat{p}_z^v - \min_{p\in P} \hat{p}_z^v) / (\max_{p\in P} \hat{p}_z^v - \min_{p\in P} \hat{p}_z^v)$ as a negative point depth normalized within $[0, 1]$. Choosing a different $F$ for rendering is also supported, our implementation is able to render any point features without limits.

Comparing with previous differentiable point renderers \cite{insafutdinov2018unsupervised,yifan2019differentiable,aliev2019neural,wiles2020synsin,dai2020neural}, ours is much simpler yet effective for rendering depth maps. Unlike \cite{dai2020neural}, our renderer does not attach additional parameters for training. We don't need to perform z-buffer sorting like \cite{wiles2020synsin,insafutdinov2018unsupervised}, since the maximum reduction among negative depths can automatically locate the nearest point. 
More importantly, unlike \cite{aliev2019neural} where point coordinates are fixed, our renderer is fully differentiable: it supports gradients to be back-propagated not only to the feature $F_p$, but also to the 2D coordinates $(\hat{p}_x^v, \hat{p}_y^v)$.

To reduce information loss in rendering, we further propose the \emph{multi-view point renderer} $\bm{\pi}$, which utilizes our differentiable point renderer to simultaneously render a point cloud $P$ into eight depth maps, each observed from a different viewpoint. 
The resulting $\bm{\pi}(P)$ with shape $H\times W \times 8$ is an ordered concatenation of all eight depth maps in the channel dimension. 
The size $H\times W$ is set to $256\times 256$ in experiments; the eight viewpoints are set as the eight corners of a cube: $[\pm1,\pm1,\pm1]$, to cover a wide angle of observation. 
Unlike the multi-view depth maps used by \cite{hu2019render4completion}, our multi-view depth maps are rendered in a differentiable way: gradients can be back-propagated from depths maps on to the rendered points. Hence, our renderer enables end-to-end training with both image and point supervisions, which considerably promotes the perceptual quality of results.


\noindent\textbf{Refiner}
Our refiner $\mathtt{R}$ shares similar structure with \cite{liu2019morphing}: they both consist of a minimum density sampling and a residual network that resembles PointNet \cite{qi2017pointnet}. But different from \cite{liu2019morphing}, we add CAE blocks to the residual network for enhanced capability. Moreover, we recurrently deploy the refiner twice upon the coarse result $Y_c$ to get a first and a second refining result, $Y_r^1$ and $Y_r^2$, with $Y=Y_r^2$ being the final output, as illustrated by Figure \ref{fig:overview}.


\noindent\textbf{Discriminator}
We render three point clouds during training: the partial input $X$, the groundtruth $Y_{gt}$ and the first refined result $Y_r^1$. Our discriminator $\mathtt{D}$ utilizes a cGAN \cite{mirza2014conditional} strategy: the real sample is a concatenation of $\bm{\pi}(Y_{gt})$ and $\bm{\pi}(X)$ in channel dimension; the fake sample is a concatenation of $\bm{\pi}(Y_r^1)$ and $\bm{\pi}(X)$. This makes the adversarial update of $Y_r^1$ to be conditioned on the input $X$.
We implement $\mathtt{D}$ as a sequence of 2D convolution layers with spectral normalizations \cite{miyato2018spectral} and \texttt{LeakyReLU} activations. 

\subsection{Training Losses}

During training, the \emph{reconstruction loss} $\mathcal{L}_{rec}$ is required to match the output point cloud to the ground-truth. 
Even though the Chamfer Distance (CD) is very popular among existing works due to its efficiency in computation,
we follow \cite{liu2019morphing} and implement $\mathcal{L}_{rec}$ with the Earth Mover's Distance (EMD) instead, which is more faithful to visual quality as verified by \cite{fan2017pointsetgeneration,achlioptas2018learning,liu2019morphing}. The $\mathcal{L}_{rec}$ supervises both the coarse output $Y_c$ and the final output $Y$ with
\vspace{-0.5em}
\begin{equation}
\vspace{-0.5em}
\mathcal{L}_{rec} = d_\texttt{EMD}(Y_c, Y_{gt}) + d_\texttt{EMD}(Y, Y_{gt}).
\label{equ:l_rec}
\end{equation}
The \emph{fidelity loss} $\mathcal{L}_{fd}$ is employed to preserve structures of the input $X$ within the output $Y$ as
\vspace{-0.5em}
\begin{equation}
\vspace{-0.5em}
\mathcal{L}_{fd} = \frac{1}{|X|} \sum_{p\in X}\min_{q \in Y}  \| p - q \|_2^2.
\end{equation}
We also introduce losses in the image domain. We impose the \emph{depth map matching loss} $\mathcal{L}_{depth}$ as a $\mathcal{L}$-1 distance on the multi-view depth maps. We also adopt the \emph{feature matching loss} $\mathcal{L}_{fea}$ as a $\mathcal{L}$-2 distance on the discriminator features. These two losses are formulated as
\vspace{-0.5em}
\begin{equation}
\vspace{-0.5em}
\mathcal{L}_{depth} =\frac{1}{8HW} \left\| \bm{\pi}(Y_r^1), \bm{\pi}(Y_{gt})\right\|_1,
\vspace{-0.5em}
\end{equation}
\begin{equation}
\vspace{-0.5em}
\mathcal{L}_{fea} = \sum_i^4 \frac{\alpha_i}{H_i W_i D_i} \left\| \mathtt{D}_i[\bm{\pi}(Y_r^1)], \mathtt{D}_i[\bm{\pi}(Y_{gt})]\right\|_2^2,
\end{equation}
with $\mathtt{D}_{1\leq i \leq 4}$ being features extracted from intermediate layers of discriminator $\mathtt{D}$,
$H_i W_i D_i$ the feature shape, $\alpha_i=D_i/\sum_i^4 D_i$ a re-weighting factor of each feature.
The final loss for the end-to-end SpareNet training combines all individual losses as
\begin{align}
\mathcal{L} = ~& w_{rec} \mathcal{L}_{rec}+w_{fd}\mathcal{L}_{fd}+w_{depth}\mathcal{L}_{depth}+\\\nonumber
	& w_{fea}\mathcal{L}_{fea}+w_{adv}\mathcal{L}_{adv} + w_{exp}\mathcal{L}_{exp},
\end{align} 
with $\mathcal{L}_{adv}$ the adversarial loss from the discriminator, $\mathcal{L}_{exp}$ an expansion loss also used by \cite{liu2019morphing}. We follow \cite{mao2017least} and implement $\mathcal{L}_{adv}$ as mean square in GAN training. The loss weights are set as 
$w_{rec}=200$,$w_{fd}=0.5$,$w_{adv}=0.1$,$w_{depth}=w_{fea}=1$,$w_{exp}=0.1$.

\section{Experiments}
\begin{figure*}[t]
\center
\setlength\tabcolsep{0pt}
{
\renewcommand{\arraystretch}{0.0}
\footnotesize
\begin{tabular}{@{}rcccccccccc@{}}
    & Input & AtlasNet\cite{atlasnet2018} & FCAE & FoldingNet\cite{foldingnet_2018_CVPR} &PCN\cite{Yuan-2018-pcn} 
    & MSN \cite{liu2019morphing} & GRNet \cite{xie2020grnet} & \emph{Ours} & Groundtruth\\
    
    \raisebox{0.6\height}{\rotatebox{90}{Lamp}}&
    \includegraphics[width=0.2\columnwidth,trim=30 30 30 30, clip]{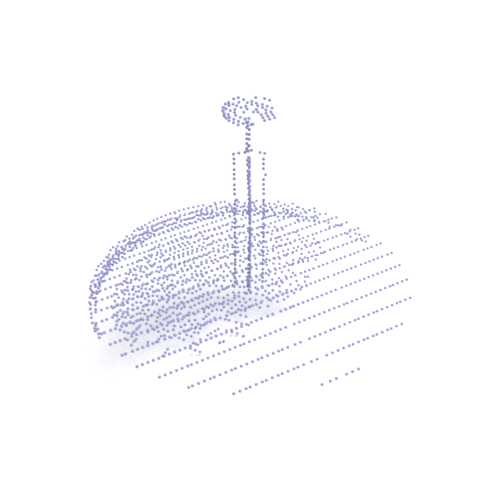}&
    \includegraphics[width=0.2\columnwidth,trim=30 30 30 30, clip]{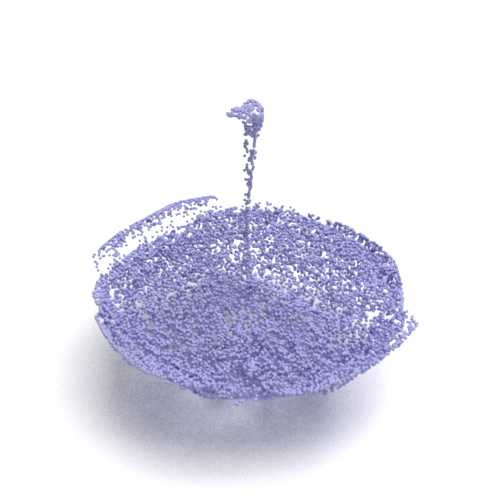}&
    \includegraphics[width=0.2\columnwidth,trim=30 30 30 30, clip]{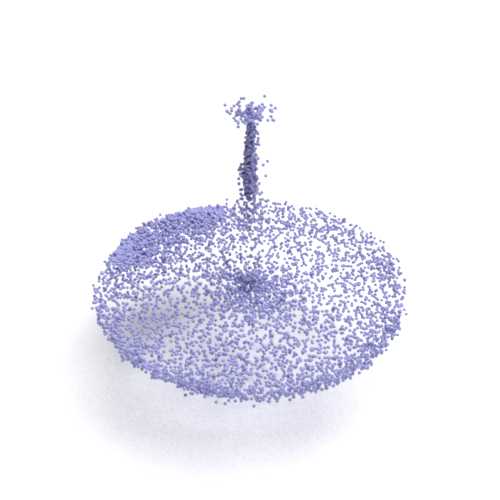}&
    \includegraphics[width=0.2\columnwidth,trim=30 30 30 30, clip]{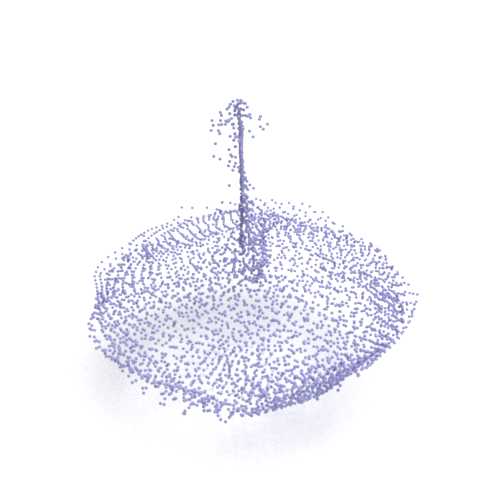}&
    \includegraphics[width=0.2\columnwidth,trim=30 30 30 30, clip]{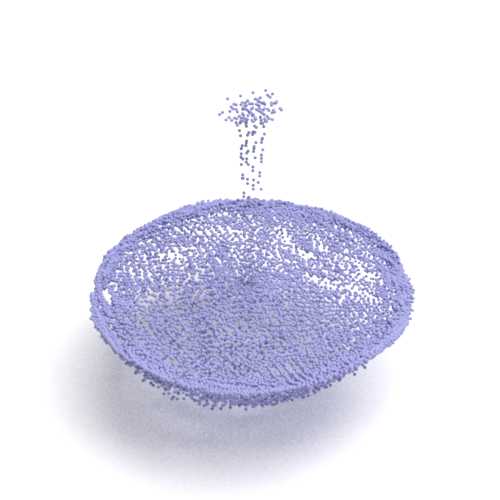}&
    \includegraphics[width=0.2\columnwidth,trim=30 30 30 30, clip]{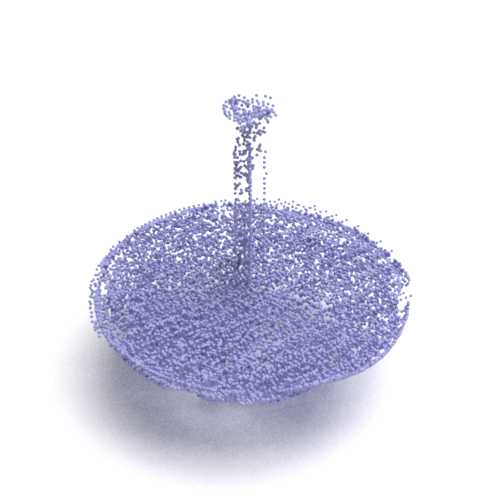}&
    \includegraphics[width=0.2\columnwidth,trim=30 30 30 30, clip]{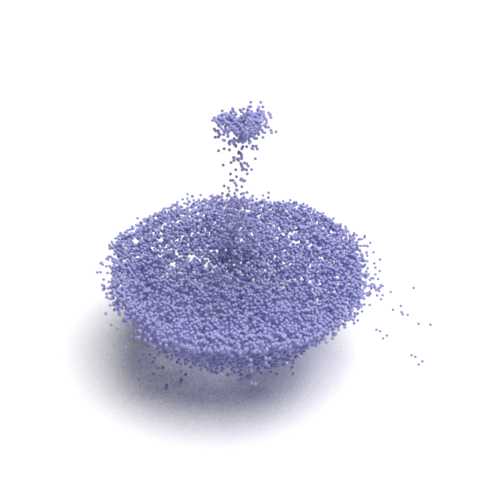}&
    \includegraphics[width=0.2\columnwidth,trim=30 30 30 30, clip]{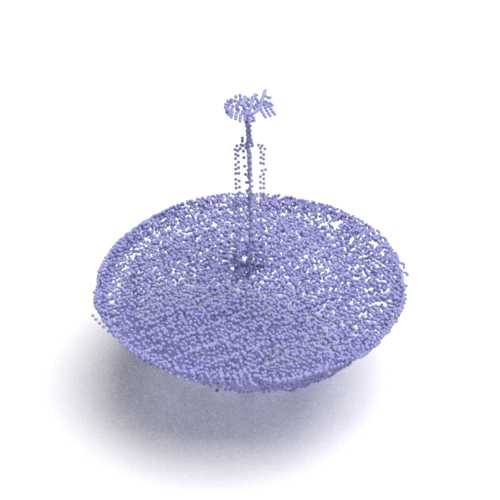}&
    \includegraphics[width=0.2\columnwidth,trim=30 30 30 30, clip]{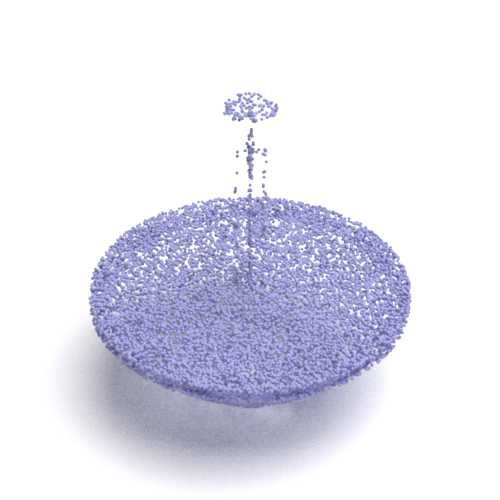}\\
    
    \raisebox{1.0\height}{\rotatebox{90}{Chair}}&
    \includegraphics[width=0.2\columnwidth,trim=30 30 30 30, clip]{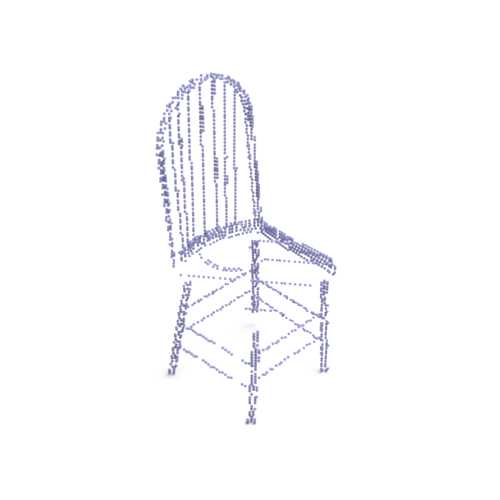}&
    \includegraphics[width=0.2\columnwidth,trim=30 30 30 30, clip]{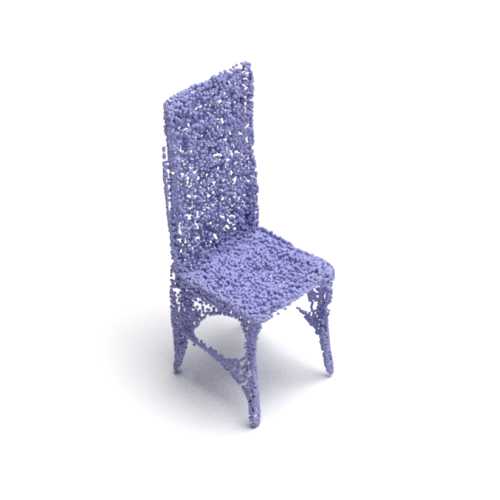}&
    \includegraphics[width=0.2\columnwidth,trim=30 30 30 30, clip]{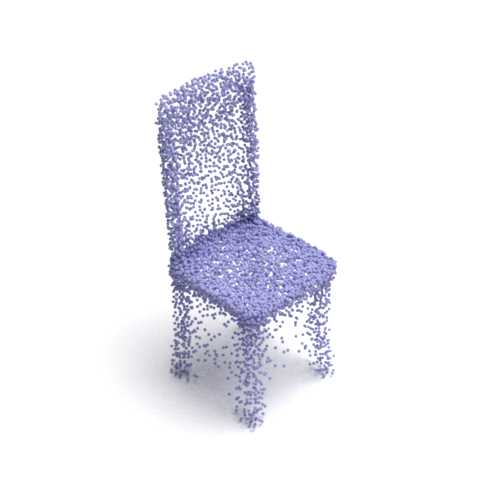}&
    \includegraphics[width=0.2\columnwidth,trim=30 30 30 30, clip]{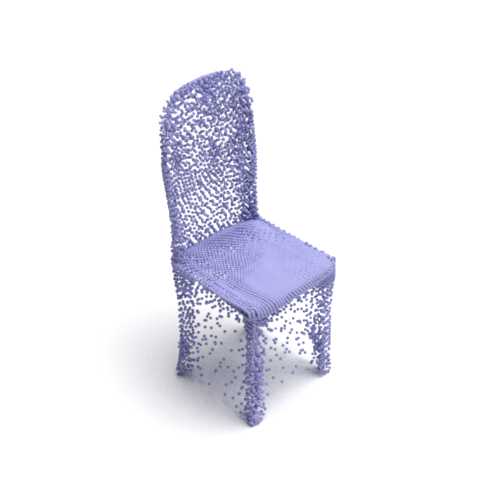}&
    \includegraphics[width=0.2\columnwidth,trim=30 30 30 30, clip]{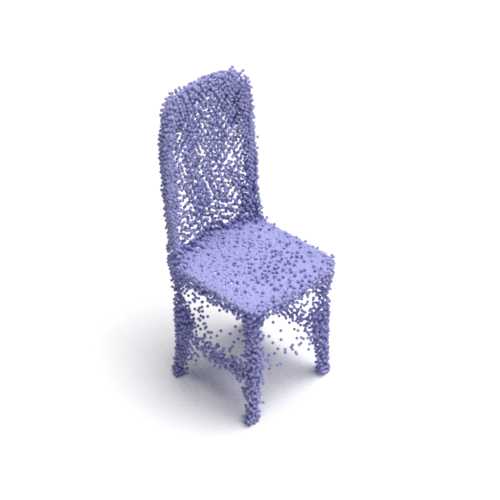}&
    \includegraphics[width=0.2\columnwidth,trim=30 30 30 30, clip]{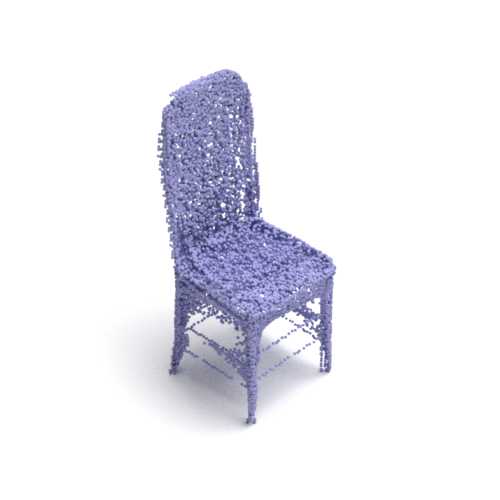}&
    \includegraphics[width=0.2\columnwidth,trim=30 30 30 30, clip]{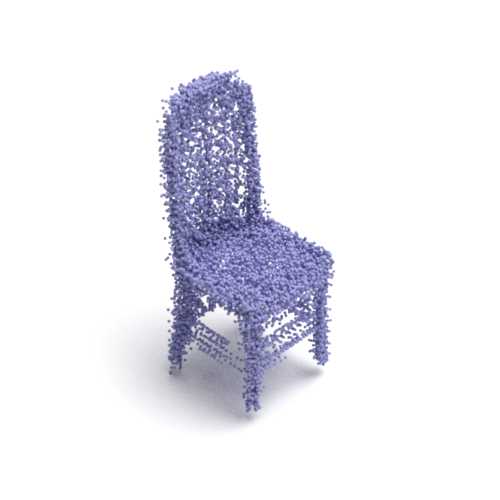}&
    \includegraphics[width=0.2\columnwidth,trim=30 30 30 30, clip]{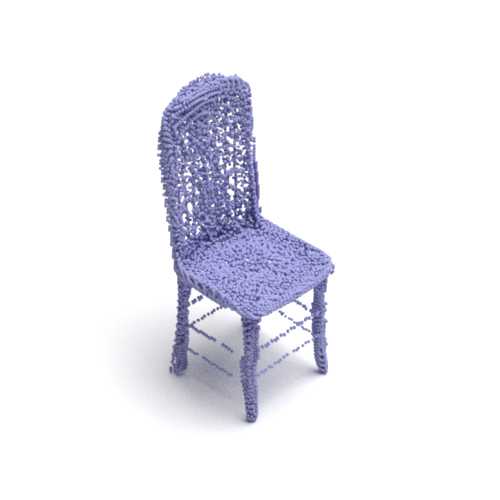}&
    \includegraphics[width=0.2\columnwidth,trim=30 30 30 30, clip]{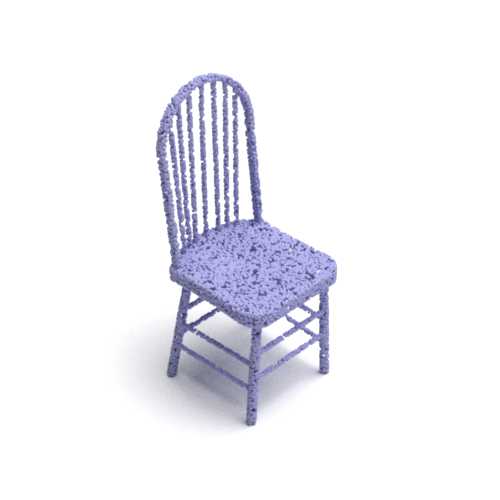}\\
    
    \raisebox{1.0\height}{\rotatebox{90}{Sofa}}&
    \includegraphics[width=0.2\columnwidth,trim=30 30 30 30, clip]{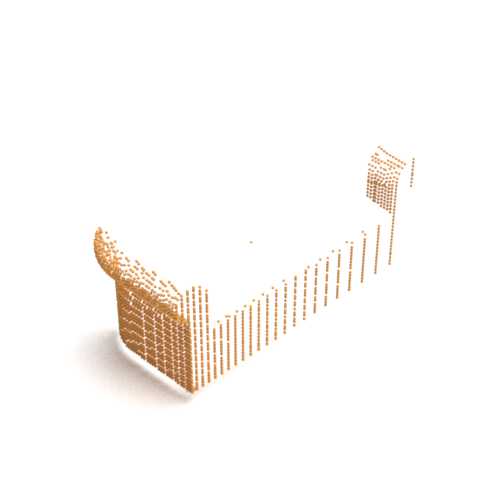}&
    \includegraphics[width=0.2\columnwidth,trim=30 30 30 30, clip]{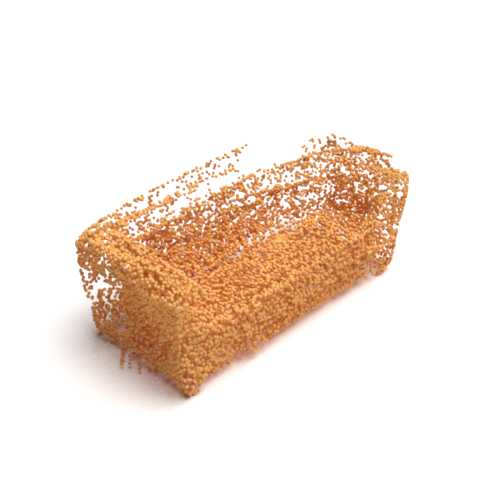}&
    \includegraphics[width=0.2\columnwidth,trim=30 30 30 30, clip]{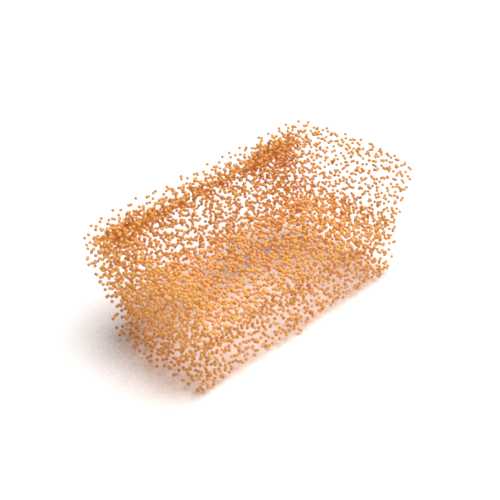}&
    \includegraphics[width=0.2\columnwidth,trim=30 30 30 30, clip]{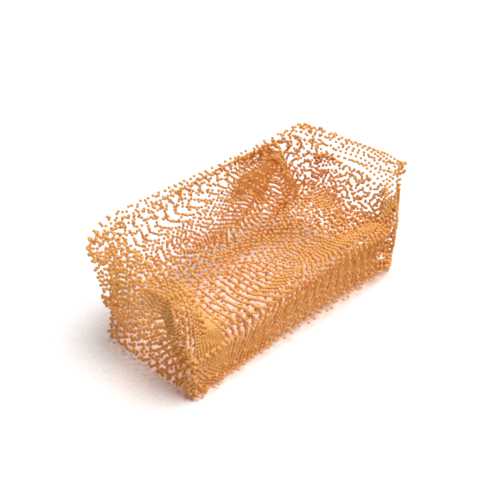}&
    \includegraphics[width=0.2\columnwidth,trim=30 30 30 30, clip]{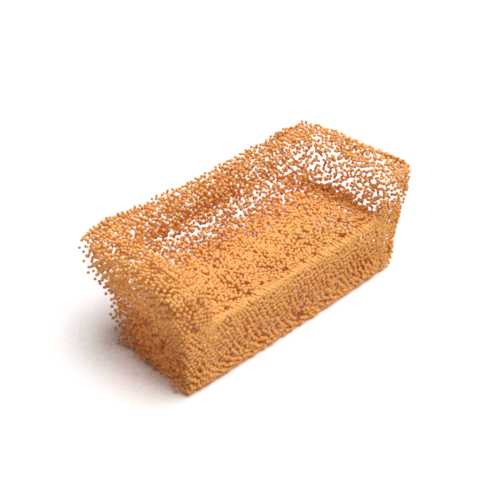}&
    \includegraphics[width=0.2\columnwidth,trim=30 30 30 30, clip]{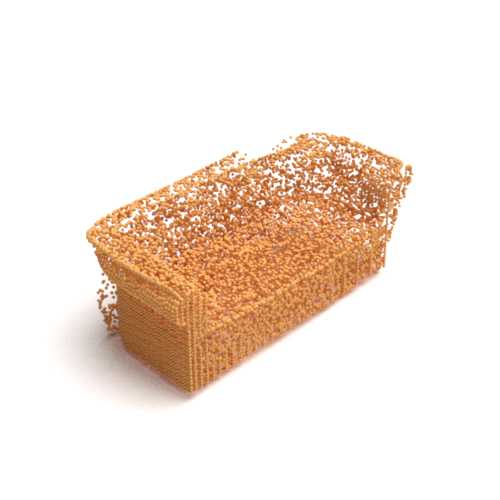}&
    \includegraphics[width=0.2\columnwidth,trim=30 30 30 30, clip]{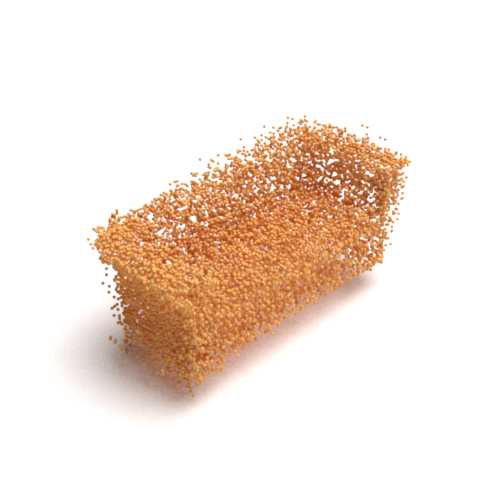}&
    \includegraphics[width=0.2\columnwidth,trim=30 30 30 30, clip]{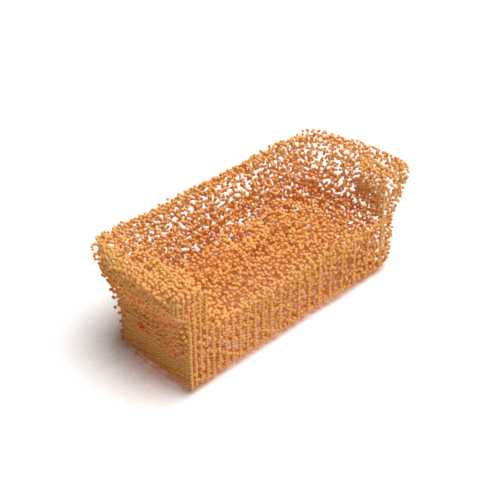}&
    \includegraphics[width=0.2\columnwidth,trim=30 30 30 30, clip]{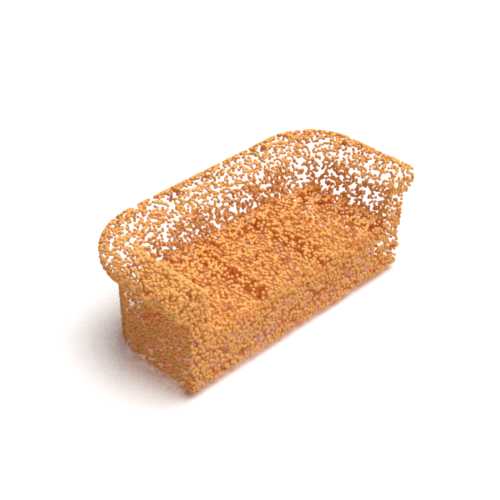}\\
    
    \raisebox{1\height}{\rotatebox{90}{Vessel}}&
    \includegraphics[width=0.2\columnwidth,trim=30 30 30 30, clip]{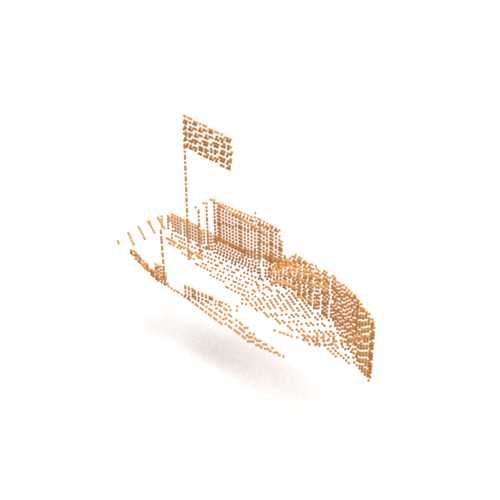}&
    \includegraphics[width=0.2\columnwidth,trim=30 30 30 30, clip]{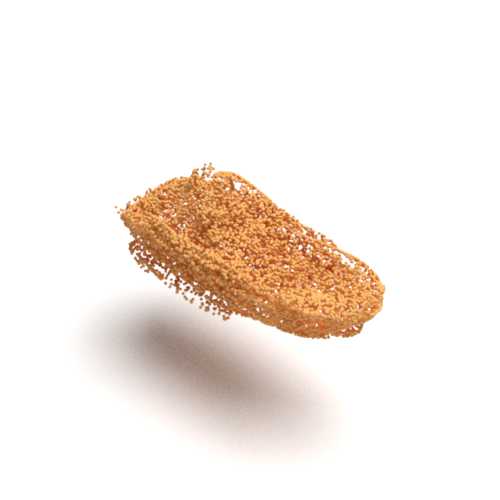}&
    \includegraphics[width=0.2\columnwidth,trim=30 30 30 30, clip]{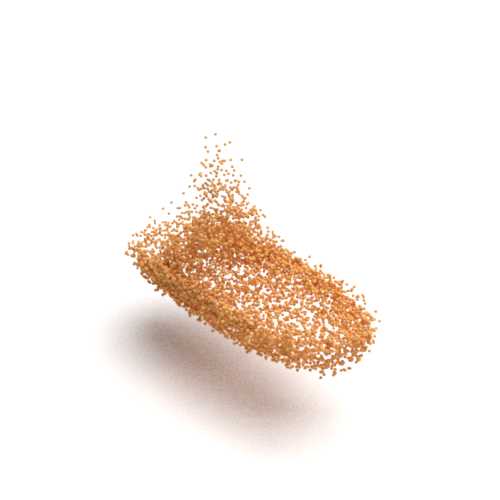}&
    \includegraphics[width=0.2\columnwidth,trim=30 30 30 30, clip]{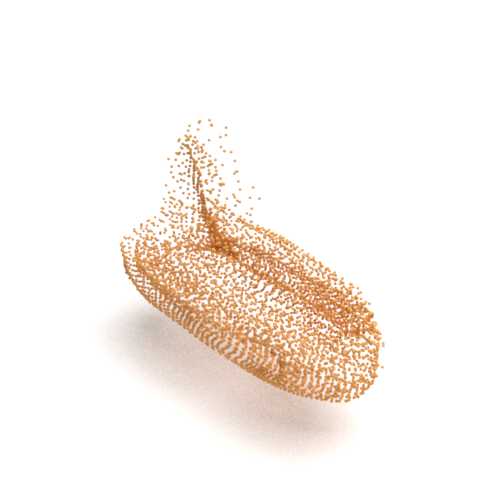}&
    \includegraphics[width=0.2\columnwidth,trim=30 30 30 30, clip]{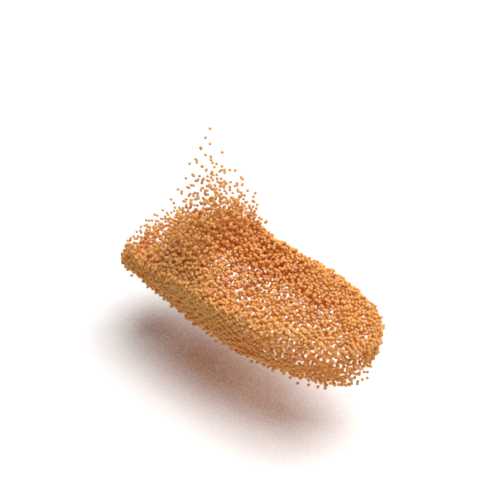}&
    \includegraphics[width=0.2\columnwidth,trim=30 30 30 30, clip]{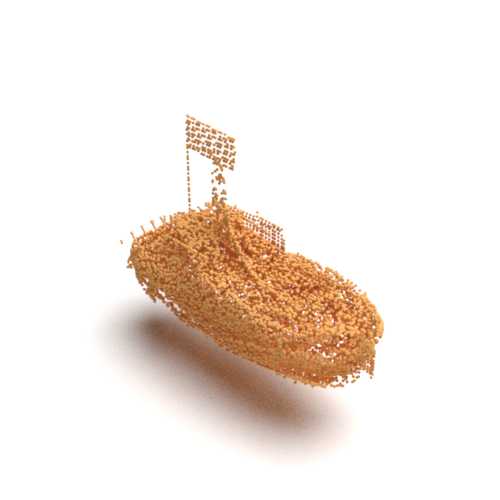}&
    \includegraphics[width=0.2\columnwidth,trim=30 30 30 30, clip]{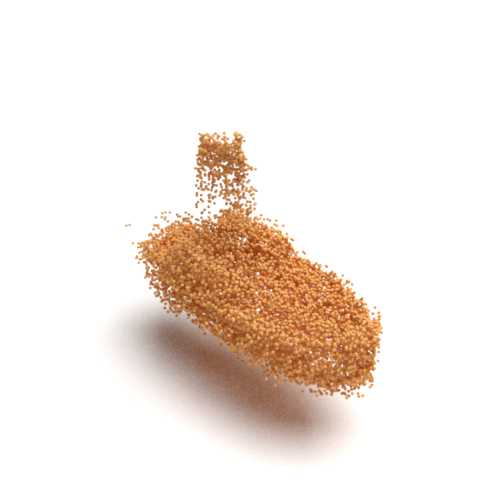}&
    \includegraphics[width=0.2\columnwidth,trim=30 30 30 30, clip]{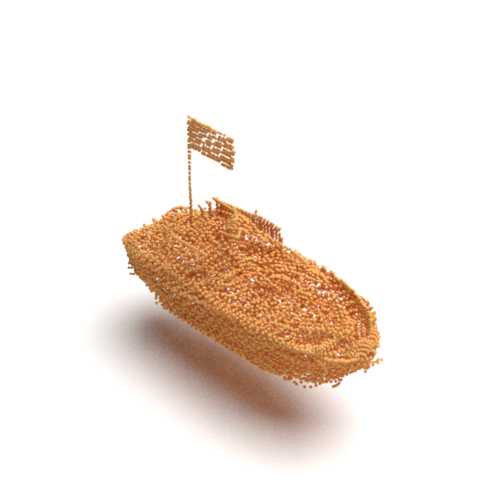}&
    \includegraphics[width=0.2\columnwidth,trim=30 30 30 30, clip]{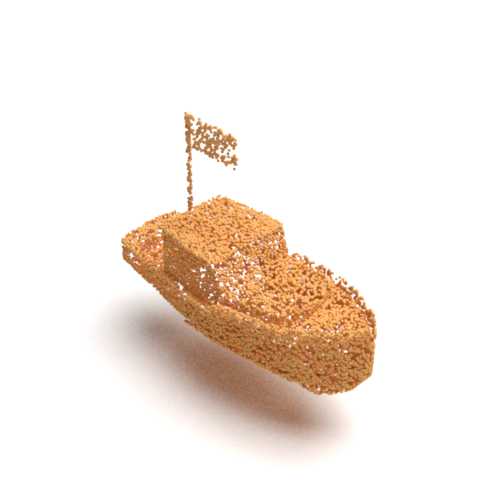}\\
    
    \raisebox{0.6\height}{\rotatebox{90}{Airplane}}&
    \includegraphics[width=0.2\columnwidth,trim=30 30 30 30, clip]{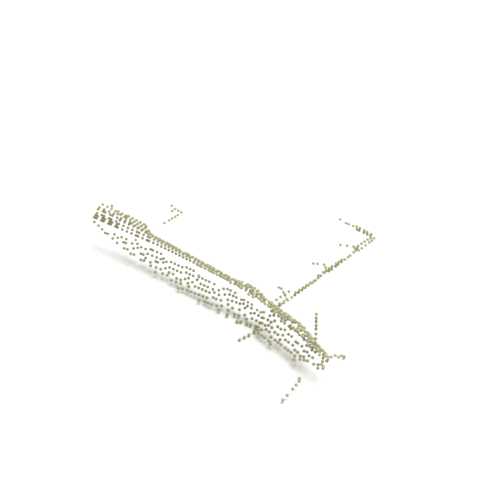}&
    \includegraphics[width=0.2\columnwidth,trim=30 30 30 30, clip]{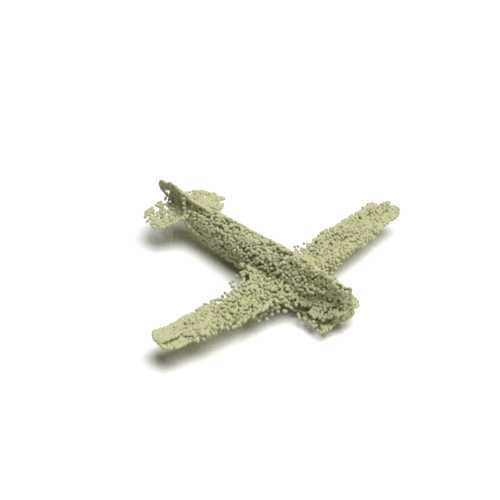}&
    \includegraphics[width=0.2\columnwidth,trim=30 30 30 30, clip]{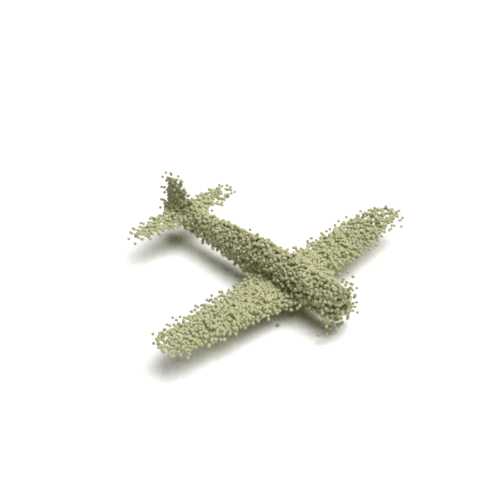}&
    \includegraphics[width=0.2\columnwidth,trim=30 30 30 30, clip]{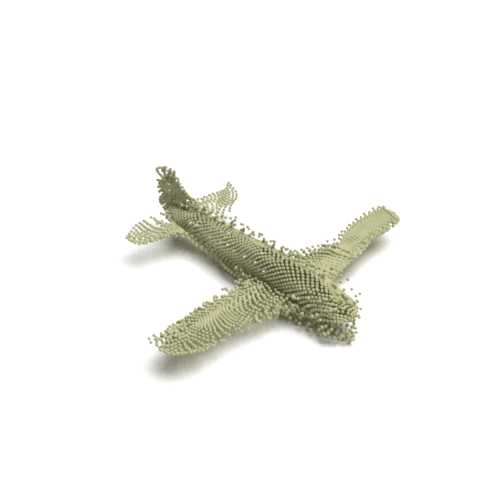}&
    \includegraphics[width=0.2\columnwidth,trim=30 30 30 30, clip]{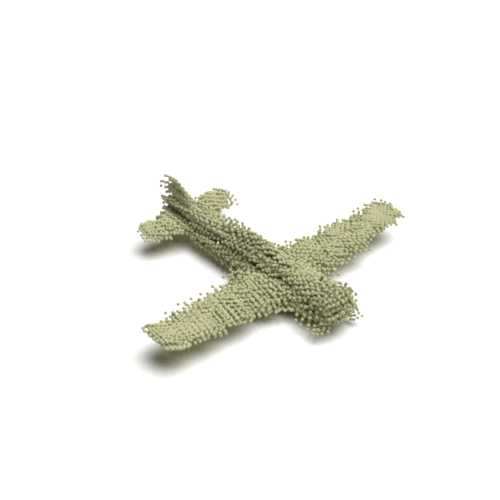}&
    \includegraphics[width=0.2\columnwidth,trim=30 30 30 30, clip]{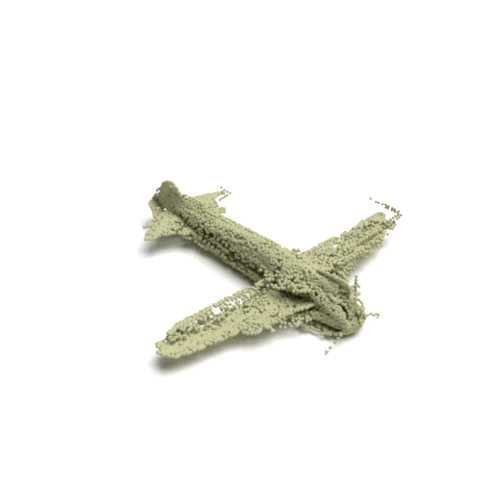}&
    \includegraphics[width=0.2\columnwidth,trim=30 30 30 30, clip]{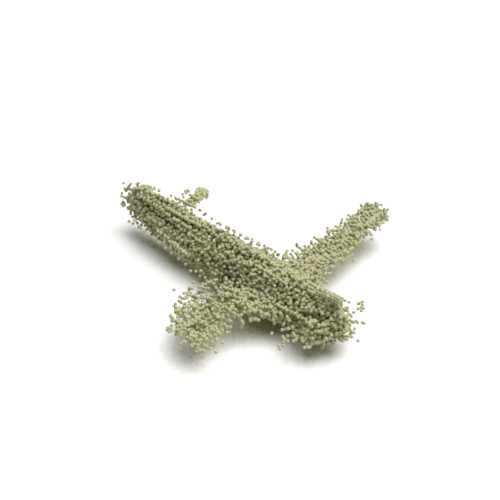}&
    \includegraphics[width=0.2\columnwidth,trim=30 30 30 30, clip]{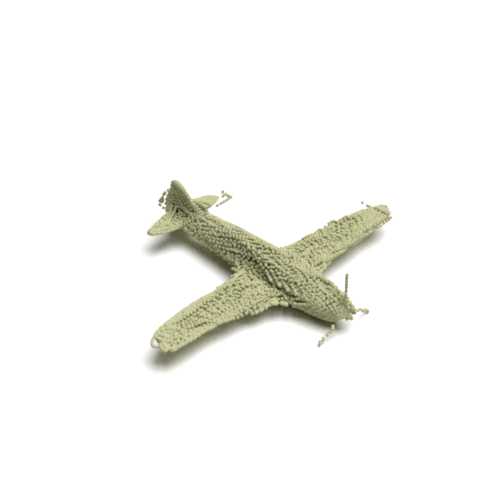}&
    \includegraphics[width=0.2\columnwidth,trim=30 30 30 30, clip]{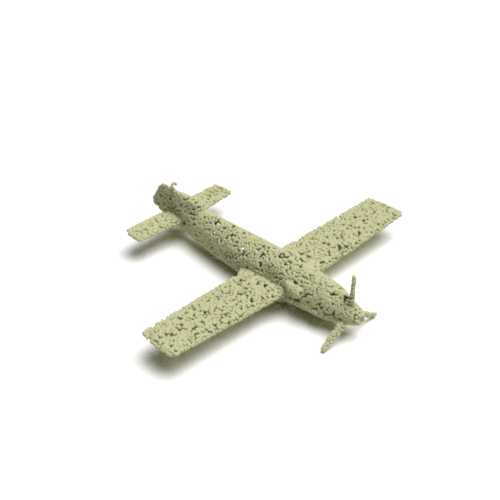}\\
    
    \raisebox{0.6\height}{\rotatebox{90}{Table}}&
    \includegraphics[width=0.2\columnwidth,trim=30 30 30 30, clip]{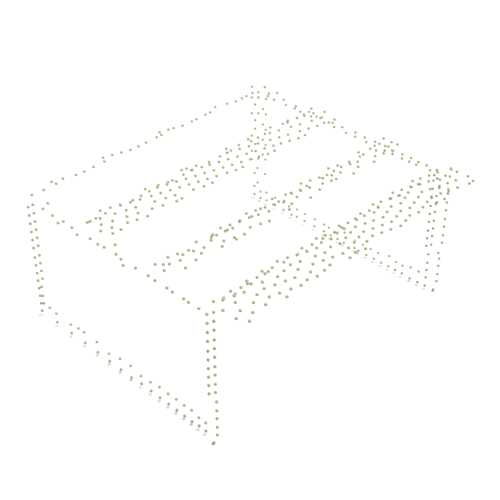}&
    \includegraphics[width=0.2\columnwidth,trim=30 30 30 30, clip]{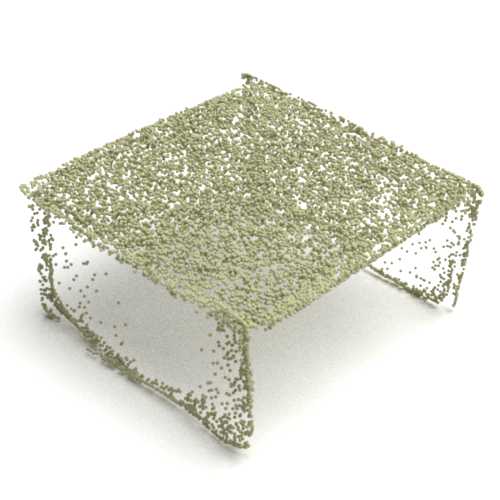}&
    \includegraphics[width=0.2\columnwidth,trim=30 30 30 30, clip]{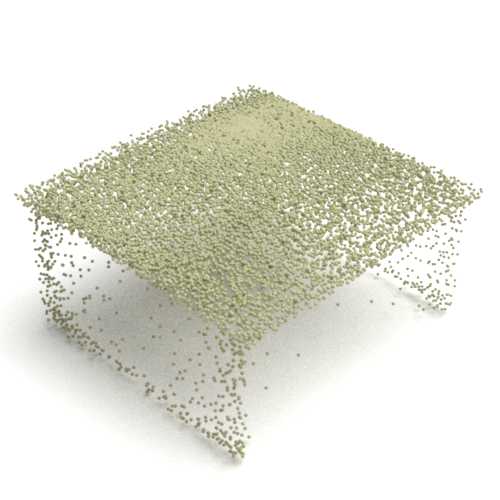}&
    \includegraphics[width=0.2\columnwidth,trim=30 30 30 30, clip]{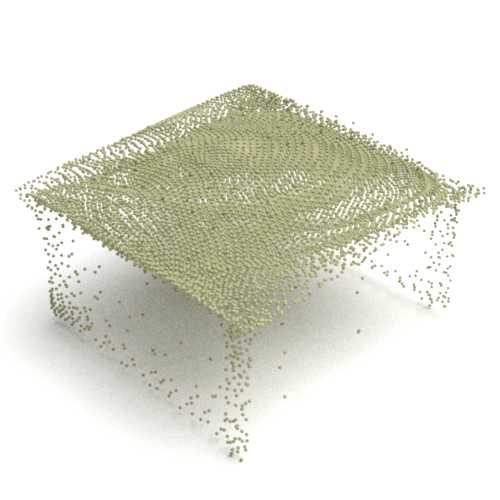}&
    \includegraphics[width=0.2\columnwidth,trim=30 30 30 30, clip]{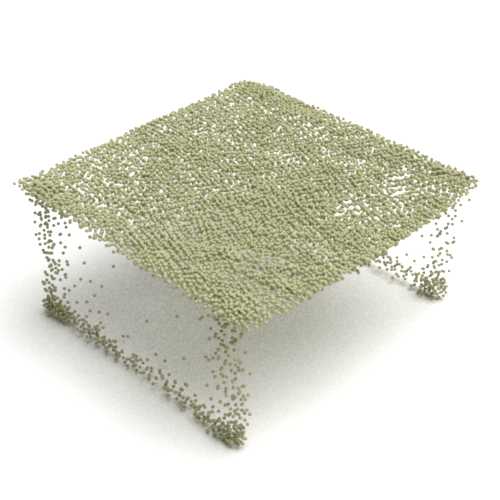}&
    \includegraphics[width=0.2\columnwidth,trim=30 30 30 30, clip]{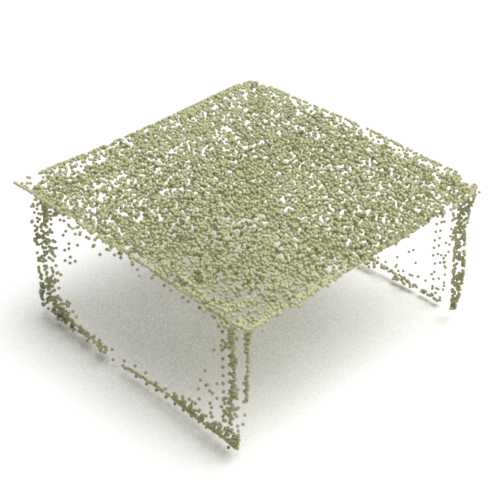}&
    \includegraphics[width=0.2\columnwidth,trim=30 30 30 30, clip]{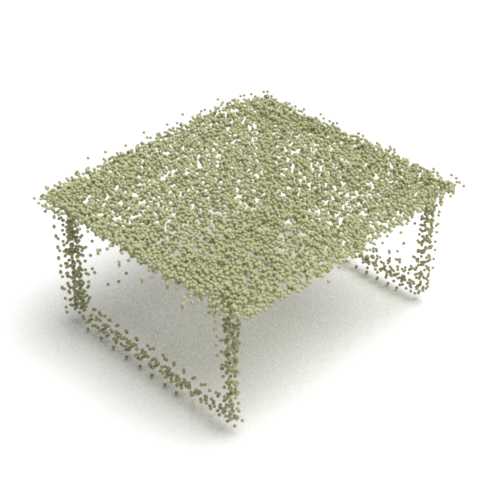}&
    \includegraphics[width=0.2\columnwidth,trim=30 30 30 30, clip]{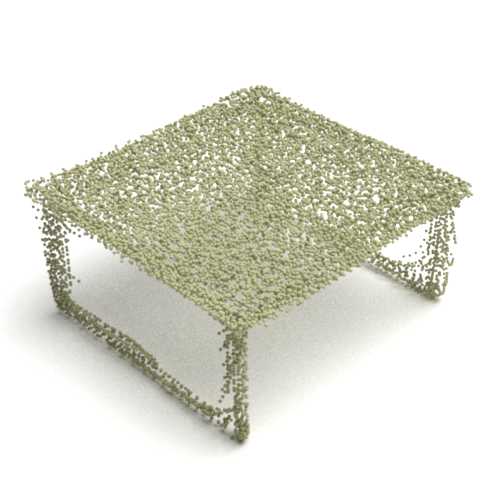}&
    \includegraphics[width=0.2\columnwidth,trim=30 30 30 30, clip]{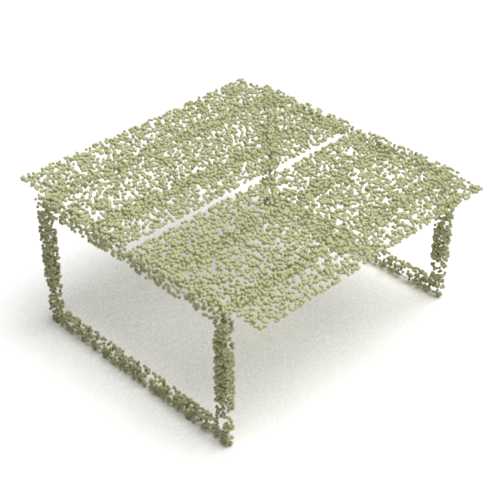}\\
    
\end{tabular}
}
\caption{Visualized completion comparison on ShapeNet.}
\label{figure:qualitative}
\vspace{-1em}
\end{figure*}

\begin{table}
\begin{center}
\footnotesize
\setlength\tabcolsep{1.5pt}
\begin{tabular}{@{}l|cccccccc|c@{}}
\toprule
Method&plane&cabinet&car&chair&lamp&sofa&table&vessel&avg\\
\midrule
PointFCAE & 1.554 & 2.631 & 2.132 & 2.954 & 4.067 & 2.997 & 2.899 & 2.619 & 2.732\\
FoldingNet & 1.682 & 2.576 & 2.183 & 2.847 & 3.062 & 3.003 & 2.500 & 2.357 & 2.526 \\
AtlasNet & 1.324 & 2.582 & 2.085 & 2.442 & 2.718 & 2.829 & 2.160 & 2.114 & 2.282 \\
PCN\footnote{} & 2.426 & \textbf{1.888} & 2.744 & 2.200 & 2.383 & \textbf{2.062} & \textbf{1.242} & 2.208 & 2.144 \\
MSN & 1.334 & 2.251 & 2.062 & 2.346 & 2.449 & 2.712 & 1.977 & 2.001 & 2.142 \\
GRNet & 1.376 & 2.128 & 1.918 & 2.127 & 2.150 & 2.468 & 1.852 & 1.876 & 1.987 \\
\emph{Ours} & \textbf{1.131} & 2.014 & \textbf{1.783} & \textbf{2.050} & \textbf{2.063} & 2.333 & 1.729 & \textbf{1.790} & \textbf{1.862}\\
\bottomrule
\end{tabular}
\end{center}
\caption{Completion comparison on ShapeNet in terms of EMD $\times 10^3$ (lower is better).}
\label{table:shapenet_emd}
\end{table}

\footnotetext{The PCN scores are calculated from a released PCN\cite{Yuan-2018-pcn} model that is trained on a fully sized ShapeNet, which is $8$ times larger than the training set of all the other models reported in Tables \ref{table:shapenet_emd}, \ref{table:shapenet_fpd}, \ref{table:shapenet_cd}. Even though, our model demonstrates superiority over it in terms of both EMD and FPD.\label{pcn}}

\begin{table}
\begin{center}
\footnotesize
\setlength\tabcolsep{1.5pt}
\begin{tabular}{@{}l|cccccccc|c@{}}
\toprule
Method&plane&cabinet&car&chair&lamp&sofa&table&vessel&avg\\
\midrule
PointFCAE & 1.424 & 3.878 & 0.519 & 2.404 & 6.989 & 2.594 & 2.673 & 8.998 & 3.683\\
FoldingNet & 1.593 & 5.918 & 0.649 & 1.355 & 4.344 & 2.400 & 2.243 & 5.508 & 3.001 \\
AtlasNet & 0.512 & 2.536 & 0.706 & 1.181 & 2.295 & 2.460 & 1.810 & 2.475 & 1.747 \\
PCN\footref{pcn} & 0.484 & 1.221 & 0.200 & 1.417 & 0.947 & 1.680 & \textbf{0.566} & 0.926 & 0.930 \\
MSN & 0.324 & 1.858 & 0.412 & \textbf{0.968} & 1.652 & 2.409 & 0.917 & 0.744 & 1.161 \\
GRNet & 4.649 & 1.002 & 1.152 & 1.712 & 2.977 & 2.717 & 1.713 & 5.528 & 2.681 \\
\emph{Ours} & \textbf{0.307} & \textbf{0.691} & \textbf{0.142} & 1.113 & \textbf{0.774} & \textbf{0.945} & 0.668 & \textbf{0.523} & \textbf{0.645}\\
\bottomrule
\end{tabular}
\end{center}
\caption{
Completion comparison on ShapeNet in terms of FPD $\times 0.1$ (lower is better).}
\label{table:shapenet_fpd}
\end{table}

\begin{table}
\begin{center}
\footnotesize
\setlength\tabcolsep{1.5pt}
\begin{tabular}{@{}l|cccccccc|c@{}}
\toprule
Method&plane&cabinet&car&chair&lamp&sofa&table&vessel&avg\\
\midrule
PointFCAE & 0.340 & 1.170 & 0.480 & 1.240 & 2.294 & 1.312 & 1.433 & 0.893 & 1.145 \\
FoldingNet & 0.622 & 1.608 & 0.619 & 1.553 & 2.025 & 1.543 & 1.534 & 0.910 & 1.302 \\
AtlasNet & 0.301 & 0.967 & 0.440 & 0.858 & 1.126 & 1.174 & 0.813 & 0.639 & 0.790 \\
PCN\footref{pcn} & 0.559 & \textbf{0.389} & 0.581 & \textbf{0.466} & 0.684 & \textbf{0.263} & \textbf{0.156} & 0.395 & \textbf{0.437} \\
MSN & 0.252 & 0.974 & 0.445 & 0.770 & 0.933 & 1.152 & 0.669 & 0.491 & 0.711 \\
GRNet & 0.293 & 0.560 & 0.363 & 0.583 & 0.690 & 0.935 & 0.532 & 0.389 & 0.543 \\
\emph{Ours} & \textbf{0.176} & 0.664 & \textbf{0.362} & 0.616 & \textbf{0.631} & 0.789 & 0.498 & \textbf{0.384} & 0.515\\
\bottomrule
\end{tabular}
\end{center}
\caption{Completion comparison on ShapeNet in terms of CD $\times 10^3$ (lower is better).}
\label{table:shapenet_cd}
\vspace{-1em}
\end{table}

During training, we use the Adam optimizer with $\beta_1=0$ and $\beta_2=0.9$. Following TTUR~\cite{heusel2017gans}, we set imbalanced initial learning rates, $1\times 10^{-4}$ for the generator and $4\times 10^{-4}$ for the discriminator. We train the network for 200 epochs, with the learning rates decayed by $0.1$ at $100$ and $150$ epochs. 
With the batch size $32$, it takes 5 days for training on the ShapeNet dataset with 4 Tesla V100 GPUs. 

\subsection{Datasets}
We conduct experiments on two datasets commonly used for point cloud completion: ShapeNet \cite{chang2015shapenet} and KITTI \cite{geiger2013vision}.

\noindent\textbf{ShapeNet.} The ShapeNet dataset derived from \cite{Yuan-2018-pcn} consists of 30,974 3D models that belong to 8 categories: airplane, cabinet, car, chair, lamp, sofa, table, and vessel. Each groundtruth point cloud comprises 16,384 points that are uniformly sampled from the corresponding 3D model. The partial point clouds are constructed by projecting the 2.5D depth maps of the model back into 3D. This leads to 8 different partial point clouds for each groundtruth in the training set.
Following \cite{xie2020grnet,cascaded_2020_CVPR}, we randomly select only one partial point cloud from the 8 to build our training pairs. 
This results in a training set only $1/8$ the size of the one used by PCN \cite{Yuan-2018-pcn}.
For a fair comparison, we use the same train/val/test splits as \cite{xie2020grnet,cascaded_2020_CVPR}.

\noindent\textbf{KITTI.} The KITTI dataset comprises a sequence of real LiDAR scans. 
For each frame, the cars are extracted according to the labeled bounding box, resulting in 2,401 point partial inputs. Since no groundtruth exist in this dataset, we cannot rely on paired metrics for evaluation.

\subsection{Comparison on ShapeNet}

\noindent\textbf{Metrics.} We employ three evaluation metrics to measure completion accuracy on ShapeNet: the \emph{Chamfer Distance} (CD), the \emph{Earth Mover's Distance} (EMD) and the \emph{Fr\'echet Point cloud Distance} (FPD).
Among these distances, CD is a widely used one due to its computation efficiency. 
EMD is more discriminative to the local details and density distribution. Previous works \cite{liu2019morphing,fan2017pointsetgeneration,achlioptas2018learning} have claimed EMD to be a more reliable measure for visual quality than CD.
The FPD \cite{shu20193d}, inspired by the FID \cite{heusel2017gans}, is a metric that computes the Fr\'echet distance between Gaussian fitted distributions. It evaluates not the accuracy of an individual point cloud, but the overall perceptual quality of all predicted point clouds.

\noindent\textbf{Completion evaluation.}
We compare SpareNet with the following state-of-the-art approaches: 1) PointFCAE, a simple autoencoder adopting PointNet as encoder and MLPs as decoder; 2) AtalasNet~\cite{atlasnet2018}, which generates points by sampling from the parametric surface elements; 3) FoldingNet~\cite{foldingnet_2018_CVPR}, which proposes folding-based decoder to directly generate points;  4) PCN~\cite{Yuan-2018-pcn} that uses stacked PointNet for feature extraction; 5) MSN~\cite{liu2019morphing}, the baseline of this work; 6) GRNet~\cite{xie2020grnet}, a recent leading approach that operates on volumetric grids. For a fair comparison, we train all the models using the EMD loss with their released codes. All models utilize the same training set except PCN\footref{pcn}. 

Tables~\ref{table:shapenet_emd} and \ref{table:shapenet_cd} show that our method outperforms prior works in terms of both EMD and CD metric. In particular, our method is more advantageous in terms of the FPD score as shown in Table~\ref{table:shapenet_fpd}, reducing the FPD from the second-best $0.930$ to $0.645$. The qualitative results in Figure~\ref{figure:qualitative} further corroborate the perceptual advantage of our method. In comparison, our method is able to generate fine structures with shapes whereas other methods are prone to give blurry results. Thin structures, such as the mast of the vessel, can also be faithfully generated. Overall, our completed result appears much less noisy and visually pleasing.

\begin{figure}[t]
\centering
 \includegraphics[width=0.8\linewidth]{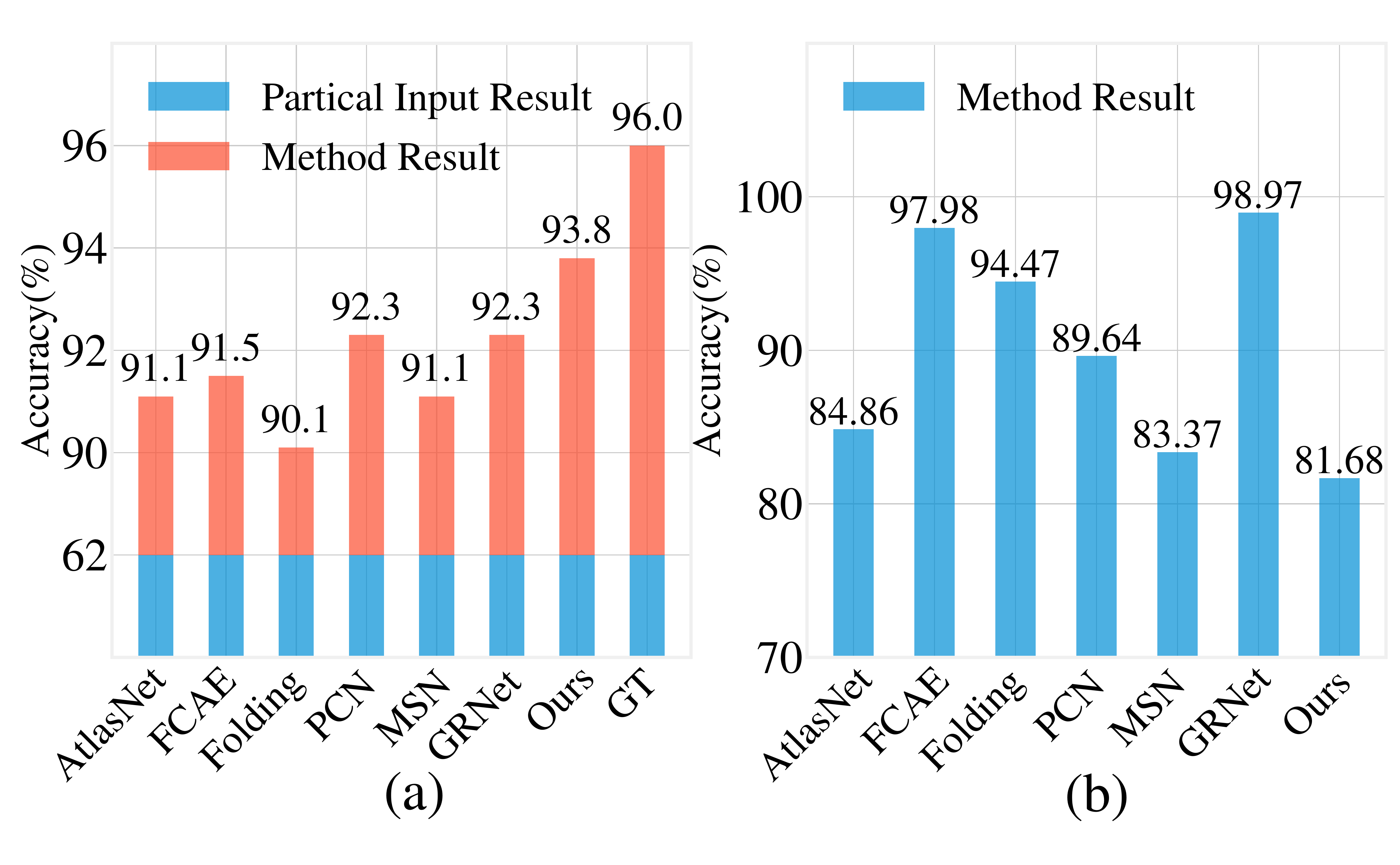}
 \footnotesize
\caption{(a) Comparing informativeness. Higher is better. (b) Comparing fake detection accuracies. Lower is better. }
\label{fig:pointnet_classification}
\vspace{-1em}
\end{figure}

\begin{table}
\begin{center}
\footnotesize
\setlength\tabcolsep{3pt}
\begin{tabular}{@{}l|ccccccc@{}}
\toprule
Method&Atlasnet&PCN&Folding&TopNet&MSN&GRNet&\emph{Ours}\\

\midrule
Consistency$\downarrow$ & 0.700 & 1.557 & 1.053 & 0.568 & 1.951 & 0.313 & \textbf{0.249}\\
Fidelity$\downarrow$ & 1.759 & 2.235 & 7.467 & 5.354 & \textbf{0.434} & 0.816 & 1.461\\
MMD$\downarrow$ & 2.108 & 1.366 & 0.537 & 0.636 & 2.259 & 0.568 & \textbf{0.368}\\
\bottomrule
\end{tabular}
\end{center}
\caption{Quantitative comparison on KITTI dataset in terms of consistency, fidelity and minimum matching distance (MMD). The best results are highlighted in bold.
\label{table:kitti}
}
\vspace{-1em}
\end{table}

\noindent\textbf{Comparison via classification.}
To better quantify the completion quality, we propose two novel metrics for comparison. First, we employ a pretrained PointNet model and test its classification accuracy on the completed point clouds by different methods. A pretrained classifier should perform better when the input is more resemblance to the real point clouds. In Figure~\ref{fig:pointnet_classification} (a), the PointNet achieves the highest classification accuracy when using our completed results as input, showing that our method can well preserve the semantics of the partial input and is able to complete the points as the real ones. Besides, we compare the perceptual quality of various methods by examining their rendered view images. As we do not have the labeling of quality scores, we train an image classifier (ResNet-18) to differentiate the fake or real data based on the complete results of all seven methods, and then make use of this fake detector to test each method. Here we assume that different methods share similar types of artifacts and the fake detection trained on one method is generalizable to other approaches. Figure~~\ref{fig:pointnet_classification} (b) shows that our method can better mislead the fake detector by giving a lower detection accuracy, proving that our methods suffer from less noticeable artifacts when observed at different viewpoints.

\subsection{Comparison on KITTI}
This dataset contains the LiDAR scans from auto-driving scenes, and the objects are much sparser than the ShapeNet dataset. For better transferred performance, we finetune all the models using the ShapeNetCars subset that only contains the cars from the ShapeNet so that the prior knowledge for this category can be better leveraged. 

Since the groundtruth point clouds are not available in this real-world dataset, we cannot apply the full-reference metrics (CD, EMD and FPD) for evaluation. As such, we follow the practice in~\cite{Yuan-2018-pcn}, and use the following metrics: 1) the \emph{Temporal Consistency} that measures the Chamfer distance for the consecutive frames; 2) the \emph{Fidelity}, a single-directional Chamfer distance that measures how the input structures are preserved in the output;  3) the \emph{Minimum Matching Distance} (MMD), the CD between the output and the point cloud in ShapetNetCars that is closest to the output in terms of CD. This metric makes sure that the output resembles a car model in ShapeNet. Table~\ref{table:kitti} shows the quantitative results. Our method shows superior consistency and MMD, indicating that our method can complete the real point points with high stability and quality. Yet, MSN and GRNet outperform in fidelity, possibly because our adversarial rendering that enforces high perceptual quality brings a slight sacrifice of the fidelity relative to the input.

\subsection{Ablation Study}

\begin{table}[t]
\begin{center}
\footnotesize
\setlength\tabcolsep{3.0pt}
\begin{tabular}{@{}lc|ccc@{}}
\toprule
& Ablation &  EMD & CD & FPD\\
\midrule
{[A]} & w/o $\bm{\eta}$ in CAE  &  2.060 & 0.606 &1.012\\
{[B]} & w/o EdgeConv in CAE &  1.972 & 0.542 &0.692\\
{[C]} & w/o Style-based Folding &  3.274 & 1.779 & 3.111\\
{[D]} & w/o Recurrent Refine &  2.184 & 0.708 &1.856\\
  & \emph{Ours Full} & \textbf{1.862} & \textbf{0.515} & \textbf{0.645}\\
\bottomrule
\end{tabular}
\end{center}
\caption{Ablation results of SpareNet.}
\label{table:ablation}
\vspace{-1em}
\end{table}

\begin{figure}[t]
\centering
 \includegraphics[width=0.8\linewidth]{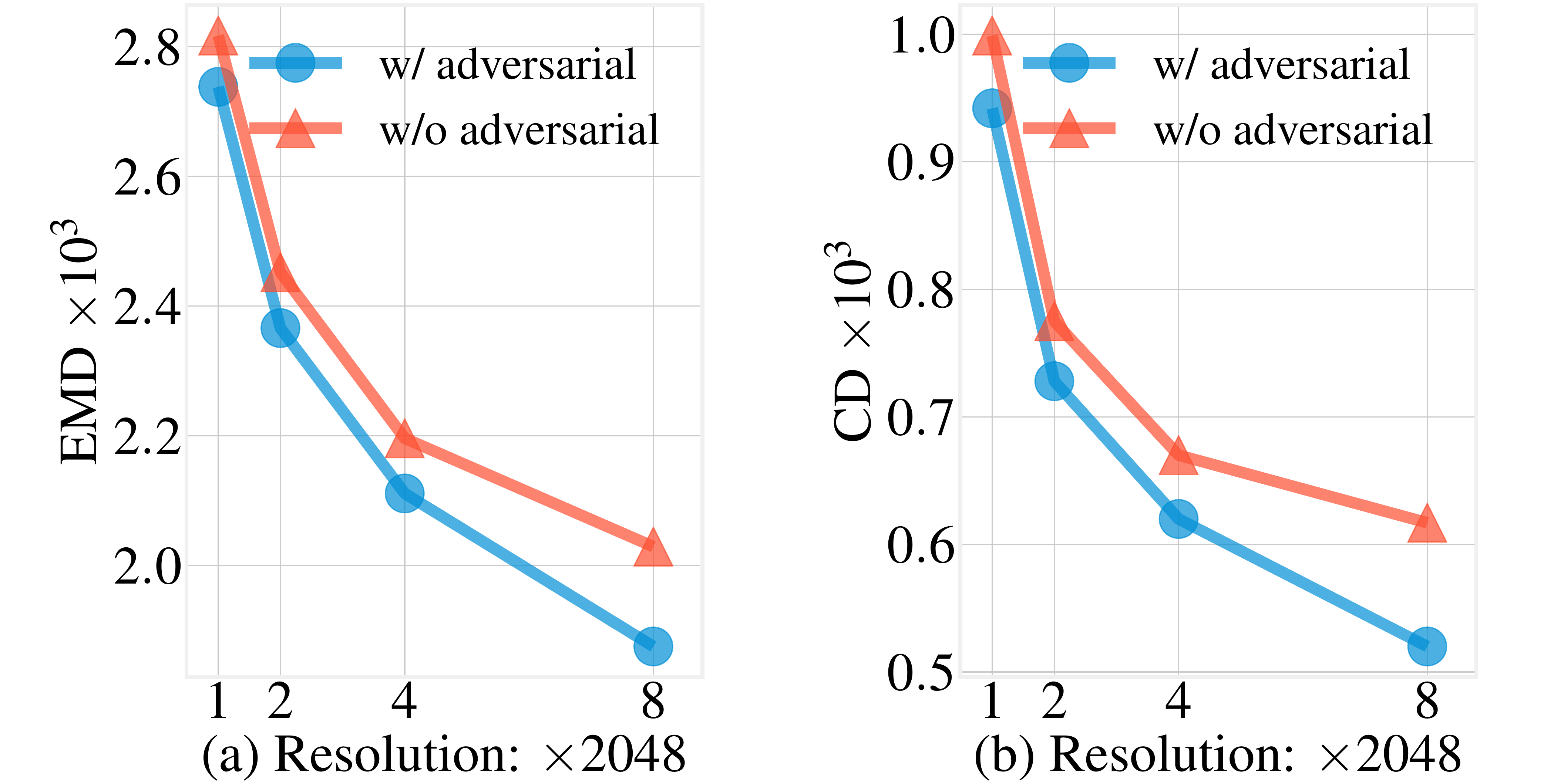}
 \footnotesize
\caption{Ablation of adversarial rendering under different point resolutions. Our adversarial rendering demonstrates larger improvement when predicting denser point clouds.}
\label{fig:differnet_resolution_ablation_emd}
\vspace{-1em}
\end{figure}

\begin{figure}[t]
\centering
 \includegraphics[width=0.7\linewidth]{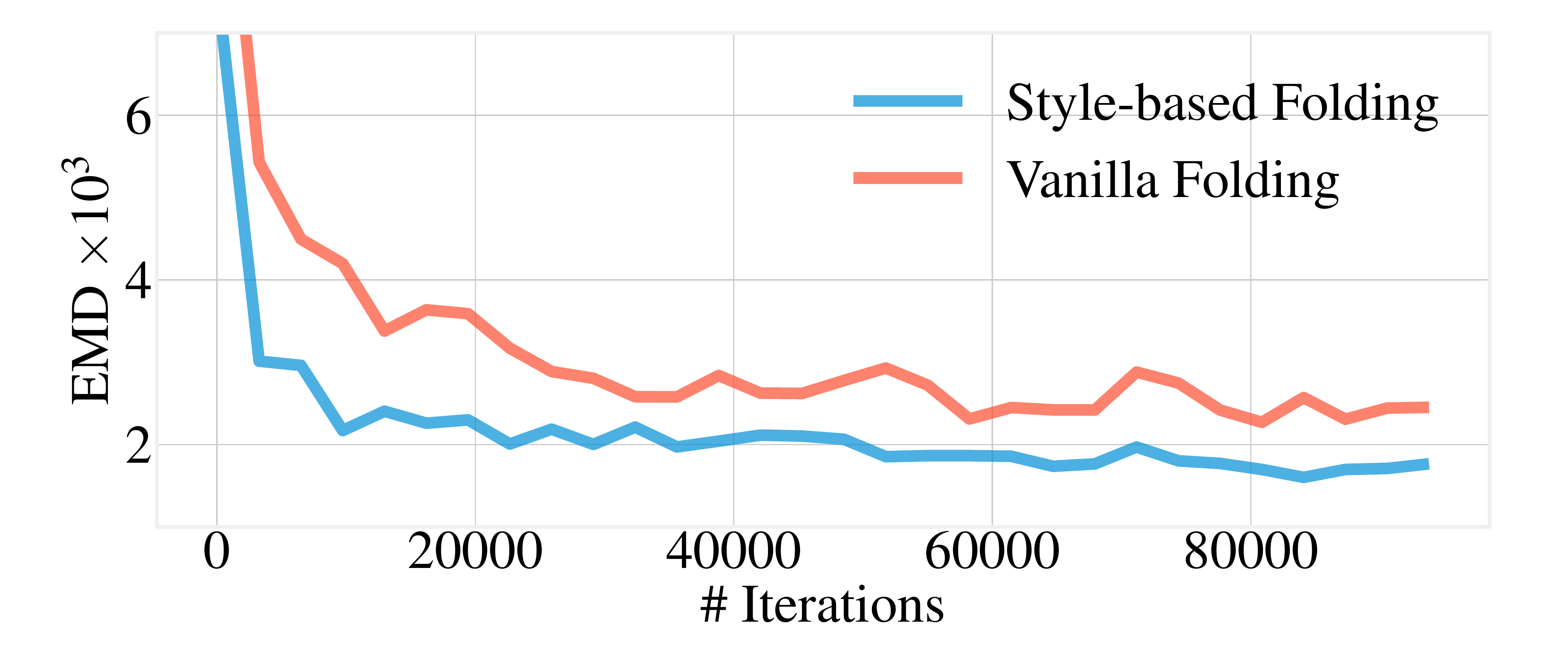}
 \footnotesize
\caption{Loss curves of training two models: one uses the vanilla folding (red); the other applies style-based folding (blue). The latter converges much faster than the former.}
\label{fig:adain_ablation}
\vspace{-1em}
\end{figure}

We report the results of extensive ablation experiments in Table \ref{table:ablation}, with each row corresponding to one ablation setting. We observe that the ablation settings all cause the drop of performance (\ie raise of the three distance numbers) comparing to our full model shown at the bottom. It in turn exhibits the effectiveness of each ablated component we propose.
For example, in \textbf{[A]}, the structure that computes and applies the channel attention vector $\bm{\eta}$ (Equation \ref{equ:eta}) is removed from CAE blocks. The raise of metric numbers reflects the importance of our proposed channel-attentive functionality. 
On the other hand, the advantage of edge features over point features can be observed by comparing the full model with \textbf{[B]} where its CAE blocks abandon the EdgeConv structure but preserve the channel attention $\bm{\eta}$ on point features.
We also ablate on the style-based generator in \textbf{[C]}, where the generator $\mathtt{G}$ adopts the vanilla concatenation-based folding instead of our style-based folding. The large-margin performance drop indicates the criticalness of our style-based folding in point cloud completion. It is also verified by Figure \ref{fig:adain_ablation}, where the style-base folding leads to much faster convergence of training loss than the vanilla folding.
In addition, we use a single-step refinement (with renderings applied on the final output) instead of the recurrent refinement and report results in \textbf{[D]}, the performance drop indicates a necessity of recurrent refinement in our framework.

\begin{figure}[t]
\center
\setlength\tabcolsep{0pt}
{
\renewcommand{\arraystretch}{0.0}
\footnotesize
\begin{tabular}{@{}rccccc@{}}
    & Input & $w_{adv}=0.0$ & $w_{adv}=0.1$ & $w_{adv}=5$ & Groundtruth\\
    
    \raisebox{1\height}{\rotatebox{90}{airplane}}&
    \includegraphics[width=0.2\columnwidth,trim=30 30 30 30, clip]{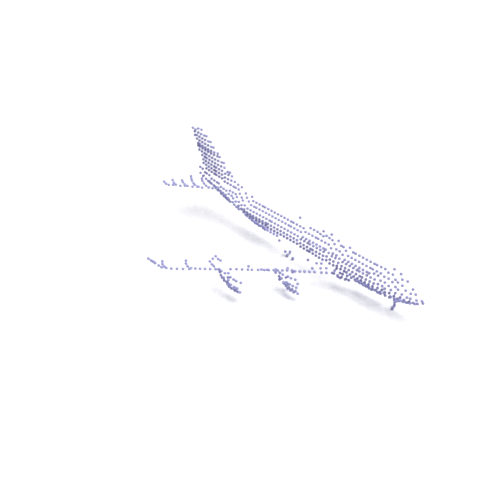}&
    \includegraphics[width=0.2\columnwidth,trim=30 30 30 30, clip]{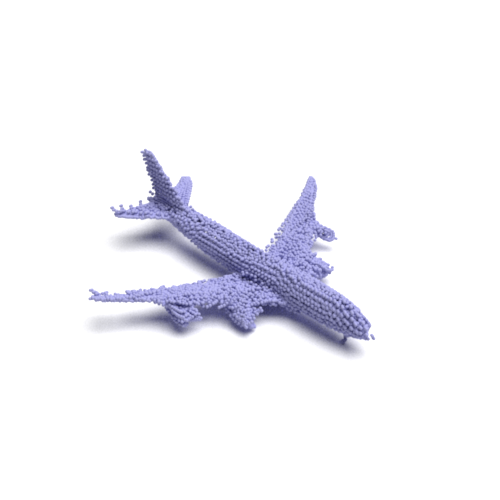}&
    \includegraphics[width=0.2\columnwidth,trim=30 30 30 30, clip]{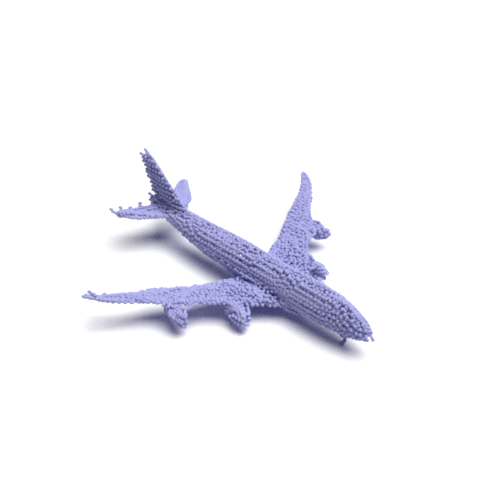}&
    \includegraphics[width=0.2\columnwidth,trim=30 30 30 30, clip]{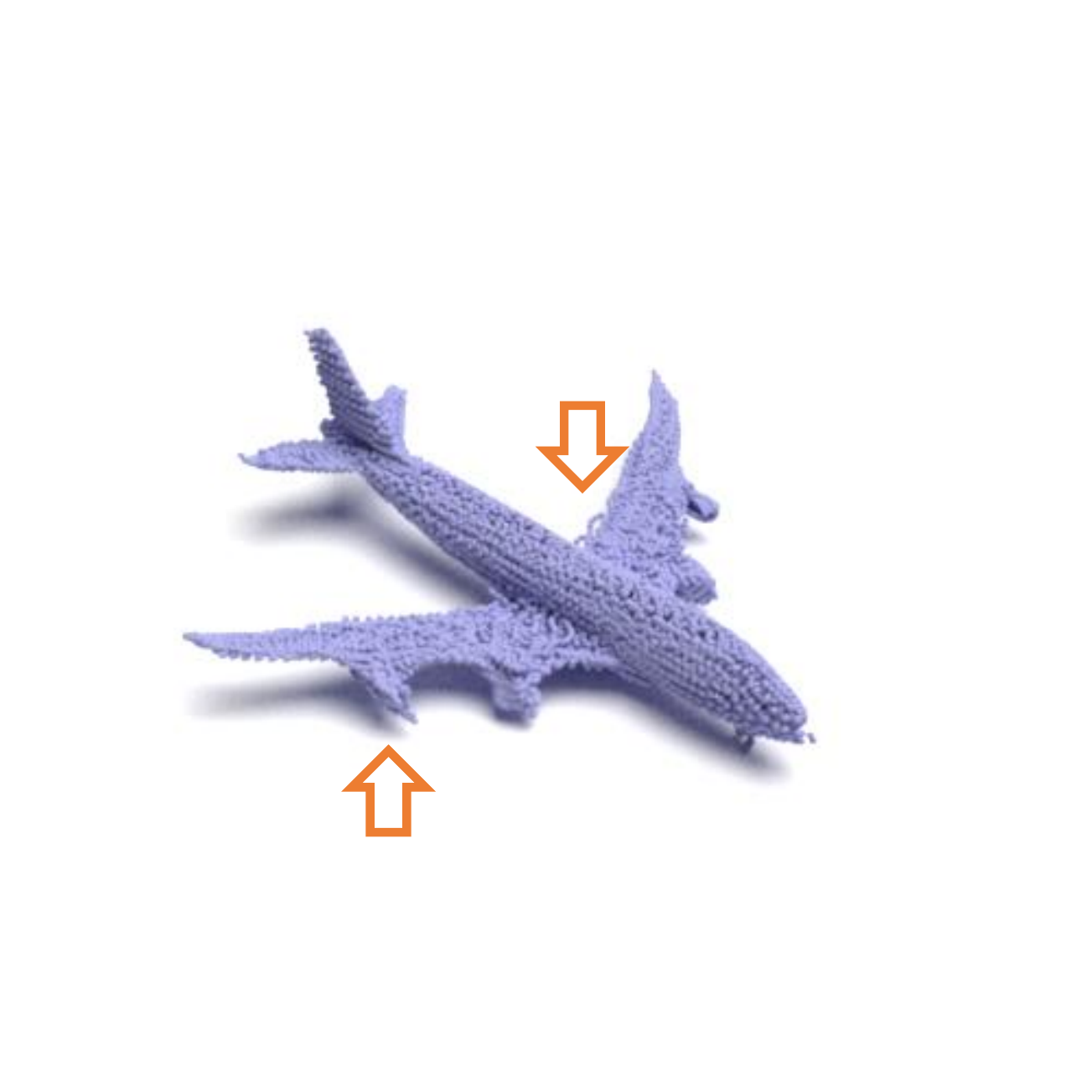}&
    \includegraphics[width=0.2\columnwidth,trim=30 30 30 30, clip]{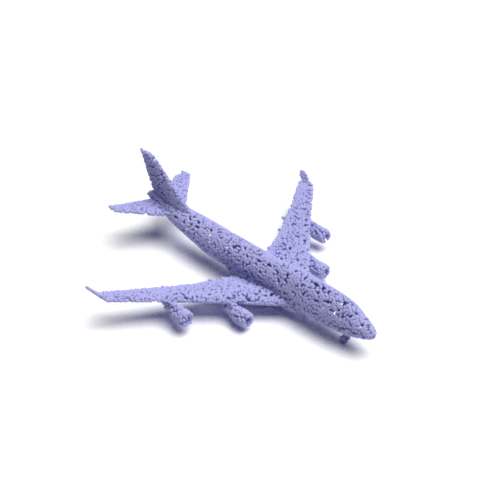}\\
    
    \raisebox{1.0\height}{\rotatebox{90}{chair}}&
    \includegraphics[width=0.2\columnwidth,trim=30 30 30 30, clip]{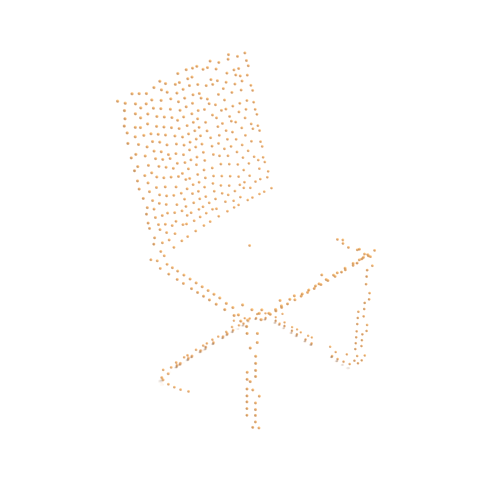}&
    \includegraphics[width=0.2\columnwidth,trim=30 30 30 30, clip]{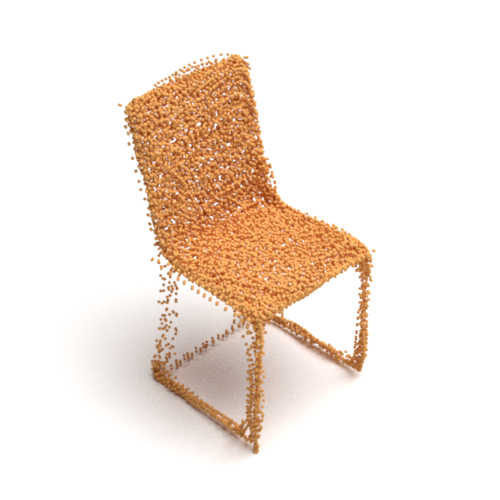}&
    \includegraphics[width=0.2\columnwidth,trim=30 30 30 30, clip]{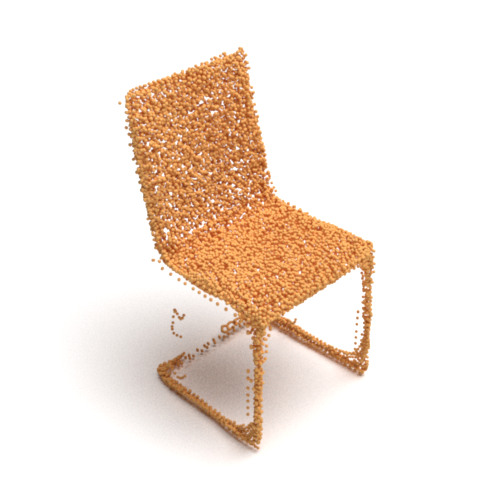}&
    \includegraphics[width=0.2\columnwidth,trim=30 30 30 30, clip]{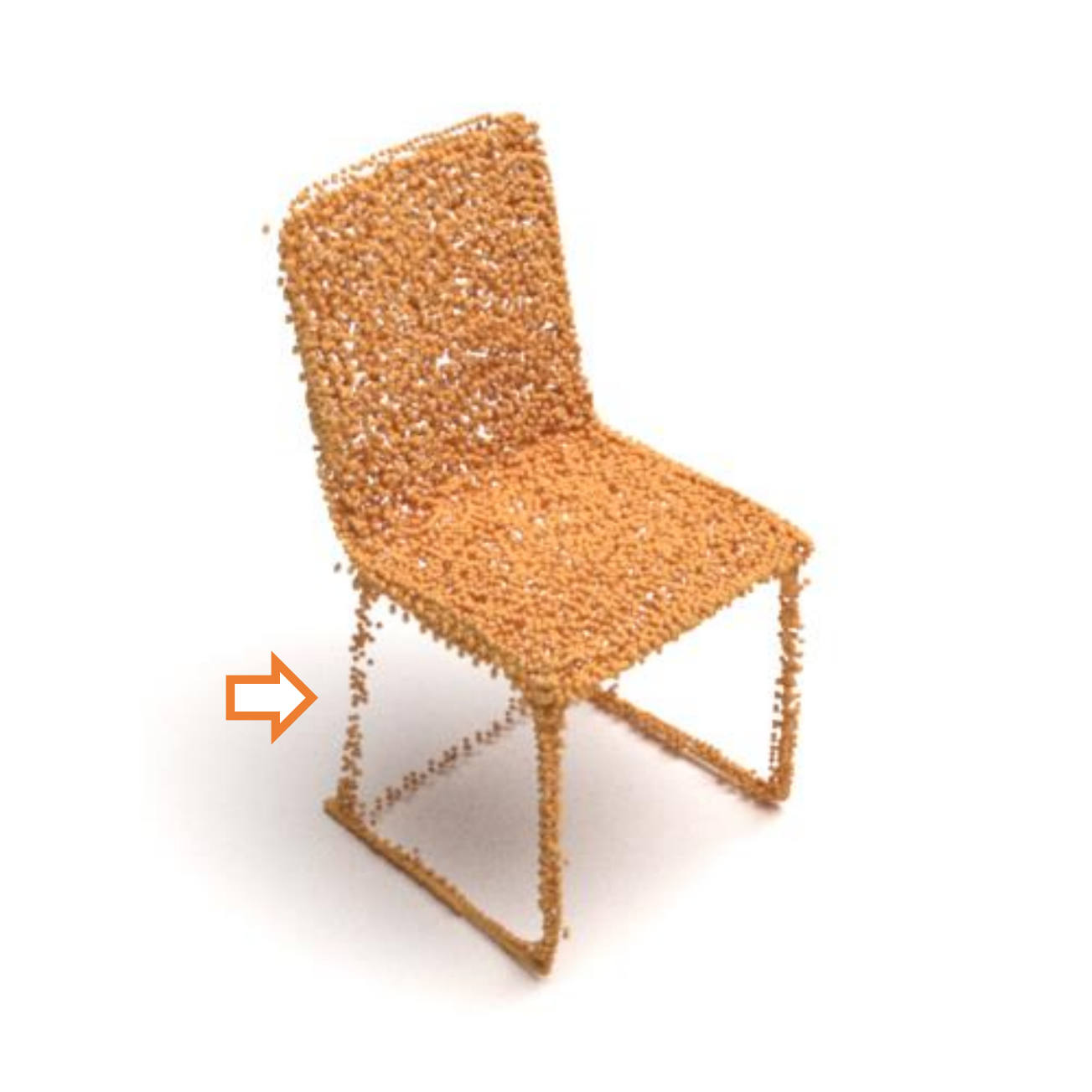}&
    \includegraphics[width=0.2\columnwidth,trim=30 30 30 30, clip]{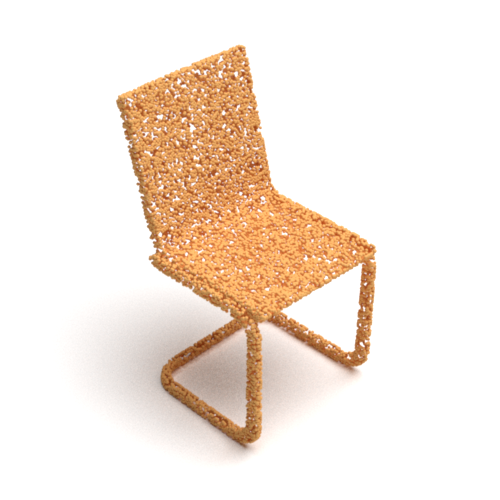}\\
    
    \raisebox{1.0\height}{\rotatebox{90}{lamp}}&
    \includegraphics[width=0.2\columnwidth,trim=30 30 30 30, clip]{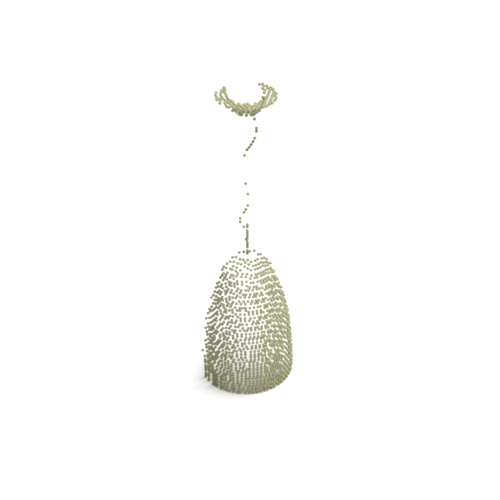}&
    \includegraphics[width=0.2\columnwidth,trim=30 30 30 30, clip]{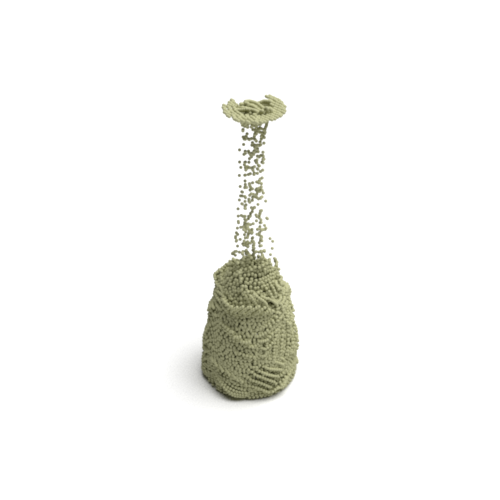}&
    \includegraphics[width=0.2\columnwidth,trim=30 30 30 30, clip]{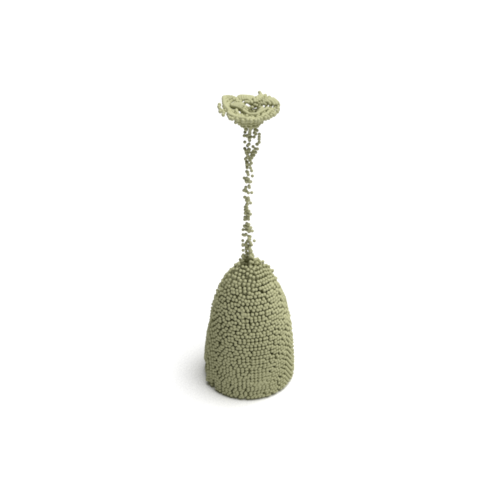}&
    \includegraphics[width=0.2\columnwidth,trim=30 30 30 30, clip]{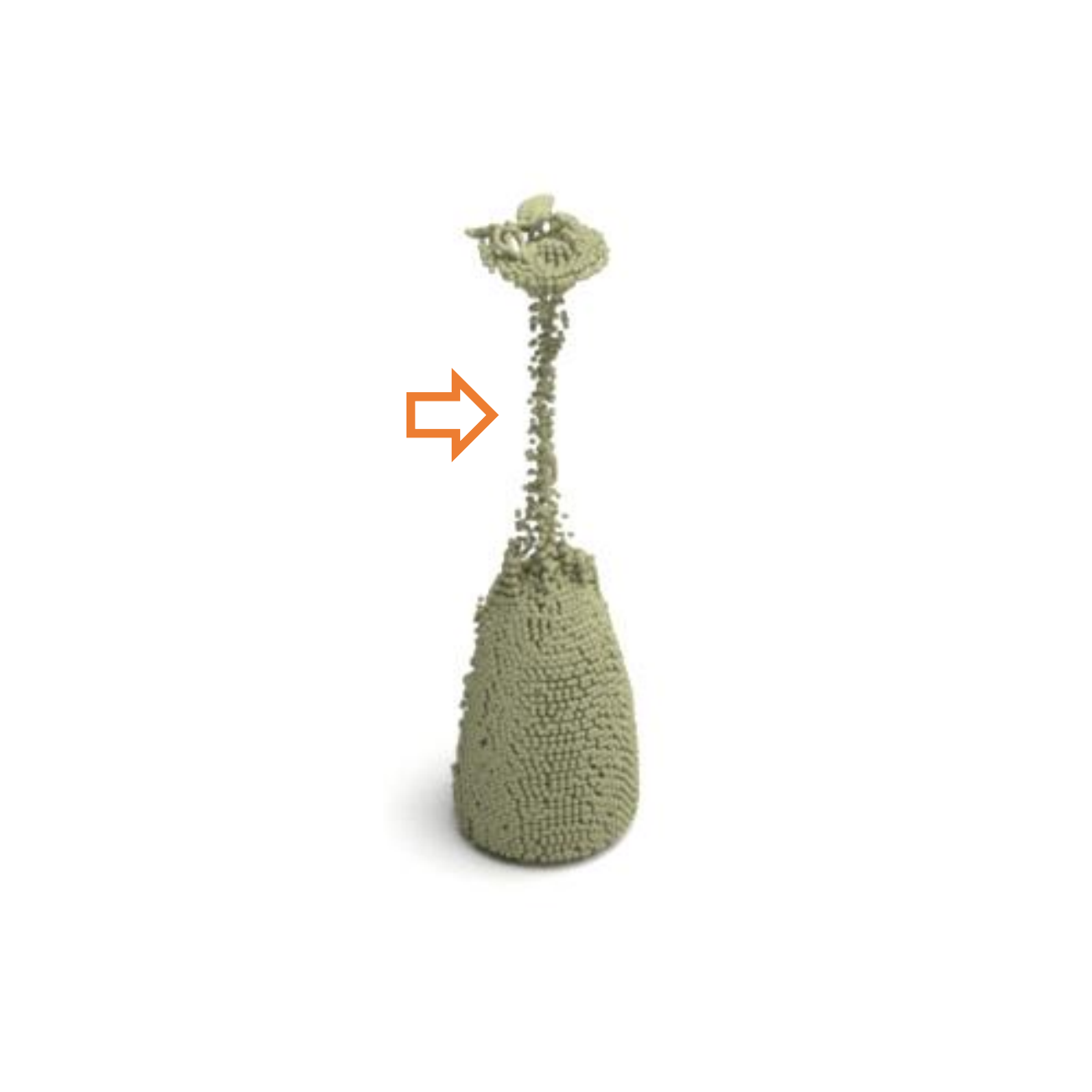}&
    \includegraphics[width=0.2\columnwidth,trim=30 30 30 30, clip]{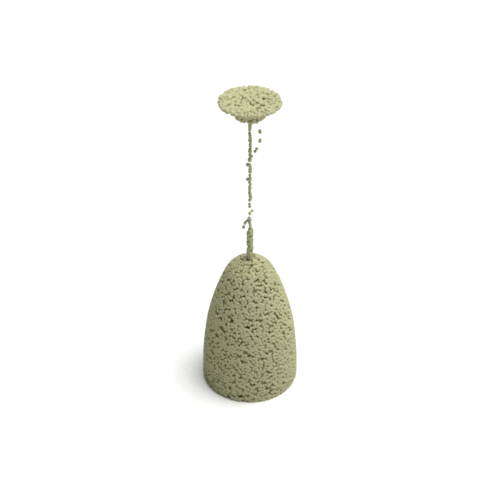}\\
       
\end{tabular}
}
\caption{Results with different adversarial loss weights.}
\vspace{-0.5em}
\label{fig:larger_gan_weight}
\end{figure}

More experiments are conducted to study the adversarial training settings. We compare the accuracies with or without adversarial training in terms of both EMD and CD in Figure \ref{fig:differnet_resolution_ablation_emd}. It shows that 1) adversarial rendering brings significant improvement under all point resolutions, from 2,048 to 16,384; 2) a denser predicted point cloud leads to a larger improvement. We believe it is due to the rendering quality: a denser point cloud renders smoother depth maps with less point scattering effects, aiding the discriminator to better capture the intrinsic geometry of point clouds.

Let $\mathcal{Y}_{adv}$ be the point clouds which we apply adversarial rendering on; let $w_{adv}$ be the weight of the adversarial loss. 
Different settings of $\mathcal{Y}_{adv}$ and $w_{adv}$ are exploited in Table \ref{table:ablation_gan} showing that:
1) when adversarial rendering involves the final output, \ie when $Y_r^2 \in \mathcal{Y}_{adv}$, the completion accuracy is slightly impaired; 2) a large adversarial loss weight ($w_{adv}=5$) also hinders the completion accuracy.
We deem this a conflict between the adversarial loss $\mathcal{L}_{adv}$ and the reconstruction loss $\mathcal{L}_{rec}$ (defined by Equation \ref{equ:l_rec}) when they are applied on the same output, or when their loss weights are comparable: 
$\mathcal{L}_{rec}$ encourages the output to match the single corresponding groundtruth, whereas the $\mathcal{L}_{adv}$ promotes an improved visual quality through the discriminator, which is learned from a larger distribution instead of a single groundtruth. This conflict is also visualized by Figure \ref{fig:larger_gan_weight}, where a larger adversarial loss weight $w_{adv}$, though helps forge some structures (\eg the airplane engine, the chair leg) reasonable for the category, nevertheless often distracts the resulting shape from its true groundtruth.

\begin{table}[t]
\begin{center}
\footnotesize
\setlength\tabcolsep{2.0pt}
\begin{tabular}{@{}c|ccc||c|ccc@{}}
\toprule
 $\mathcal{Y}_{adv}$  &  EMD & CD & FPD & $w_{adv}$ & EMD & CD & FPD \\
\midrule
$\varnothing$ & 2.029 & 0.617 & 1.308 & $0$ & 2.029 & 0.617 & 1.308 \\
$\{Y_r^1\}$ & \textbf{1.862} & \textbf{0.515} & \textbf{0.645} & $0.1$ & \textbf{1.862} & \textbf{0.515} & \textbf{0.645} \\
$\{Y_r^2\}$  & 1.937 & 0.601 & 0.874 & $0.2$ & 1.898 & 0.535 & 0.654\\
$\{Y_r^1, Y_r^2\}$ & 1.898 & 0.535 & 0.736 & $5$ & 1.980 & 0.593 & 1.013\\
\bottomrule
\end{tabular}
\end{center}
\caption{Ablation of adversarial training settings. By default we apply discriminator only on $Y_r^1$ with $w_{adv}=0.1$.}
\label{table:ablation_gan}
\vspace{-1em}
\end{table}


\section{Conclusion}
This paper presents a novel framework, SpareNet, for point cloud completion. 
It comprises channel-attentative Edge-Convs for the fusion of local and global context in point feature extraction. 
It also performs style-based folding for an enhanced capability in point cloud synthesis.
In addition, adversarial point rendering is adopted: by leveraging a fully differentiable point renderer, an image-based discriminator is utilized for capturing geometric details in point clouds.
Extensive experiments on ShapeNet and KITTI verify the state-of-the-art performance of SpareNet, as well as the effectiveness of each proposed component.


{\small

\bibliographystyle{ieee_fullname}
}
\onecolumn 
\appendix

\section*{Appendix}

\section{Network Architectures}
We first elaborate on the network architectures of our SpareNet.

\subsection{Encoder}
The input for our point encoder $\mathtt{E}$ is a partial and low-res point cloud $X$ with $M$ points in 3D.
As shown by Table~\ref{table:network_point_encoder}, the encoder $\mathtt{E}$ consists of four sequential Channel-Attentive EdgeConv (CAE) blocks with layer output sizes $256$, $256$, $512$, $1024$. The $k$ of $k$-NN is set as $8$. The slope of all LeakyReLU layers is set to $0.2$.
We concatenate the outputs of the four layers, and feed them into a final shared MLP-BN-ReLU layer with dimension $2048$. 
The output shape code $\mathbf{g}$ with dimension $4096$ is the concatenation of two global poolings of the above result: a maximum pooling and an average pooling.

\begin{table}[h]
\begin{center}
\footnotesize
\setlength\tabcolsep{1.5pt}
\begin{tabular}{@{}l|cccccccc|c@{}}
\toprule
Layer & $C_{in}$ & $\mathbf{F}_1$ & $\mathbf{F}_2$ & $\mathbf{F}_3$ \\
\midrule
1&3&[256]&[16, 256]&/ \\
2&256&[256]&[16, 256]& [256]\\
3&256&[512]&[32, 512]&[512]\\
4&512&[1024]&[64, 1024]&[1024] \\
\bottomrule
\end{tabular}
\end{center}
\caption{Channels of the CAE blocks in point encoder $\mathtt{E}$.}
\label{table:network_point_encoder}
\end{table}

\subsection{Generator}
Our {style-based point generator} $\mathtt{G}$ employs $K$ ($K$=32) surface elements to form a complex shape. For each surface element, the generator maps a $n \times  2$ unit square  ($n=N/K$) into a $n \times 3$ surface through three sequential style-based folding layers and one linear layer. The output sizes of the four layers are $4096$, $2048$, $1024$ and $3$. 

We use two linear layers with $\{4096, 3059\}$ neurons to transform the shape code $\mathbf{g}$ into modulation parameters $\gamma_{\mathbf{g}}$ and $\beta_{\mathbf{g}}$ for the three style-based folding layers, with $3059$ being the total size of modulation parameters in the three style-based folding layers. We partition the modulation parameters into parts according to the size of activation $\bar{\mathbf{h}}_{in}$ in each style-based folding layer, and assign them to each layer respectively. The learned $\gamma_{\mathbf{g}}$ and $\beta_{\mathbf{g}}$ are shared among all the $K$ surface elements. 

\subsection{Renderer}
The size of a rendered depth map $H\times W$ is set to $256\times 256$ in experiments. 
The eight viewpoints for the multi-view rendering are set as the eight corners of a cube: $(\pm1,\pm1,\pm1)$.
We adopt the radius $\rho=3$ in point rendering.

\subsection{Refiner}
In each refiner $\mathtt{R}$, instead of directly concatenating the previous output points (containing $N$ points) and the partial input points (denoted by $X$, containing $M$ points) into one point cloud, we attach a flag to each point before concatenation: a $0$ is attached if the point comes from $X$ while a $1$ label is attached otherwise. This results in a point cloud with $N+M$ $4$-dimensional points,
which is fed into a minimum density sampling \cite{liu2019morphing} that samples $N$ points out of the $N+M$ points. 
A residual network then learns point-wise residuals for the re-sampled $N$ points to adjust their positions. The residual network consists of 7 CAE blocks as depicted in Table~\ref{table:network_residual_network}.
We concatenate the outputs of the first and the third CAE blocks, 
which is fed into the fourth CAE block and outputs residuals for the refined point coordinates.

\begin{table}[h]
\begin{center}
\footnotesize
\setlength\tabcolsep{1.5pt}
\begin{tabular}{@{}l|cccccccc|c@{}}
\toprule
Layer & $C_{in}$ & $\mathbf{F}_1$ & $\mathbf{F}_2$  \\
\midrule
1& 4 & [64] &[4, 64] \\
2& 64 & [128] &[8, 128]\\
3& 128 & [1024]&[64, 1024]\\
4& 1088 & [512] &[32, 512] \\
5& 512 & [256] &[16, 256] \\
6& 256 & [128] &[8, 128] \\
7& 128 & [3] & / \\
\bottomrule
\end{tabular}
\end{center}
\caption{Channels of the CAE blocks in the residual network of refiner.}
\label{table:network_residual_network}
\end{table}

\subsection{Discriminator}
\begin{figure}[h]
\centering
 \includegraphics[width=0.45\linewidth]{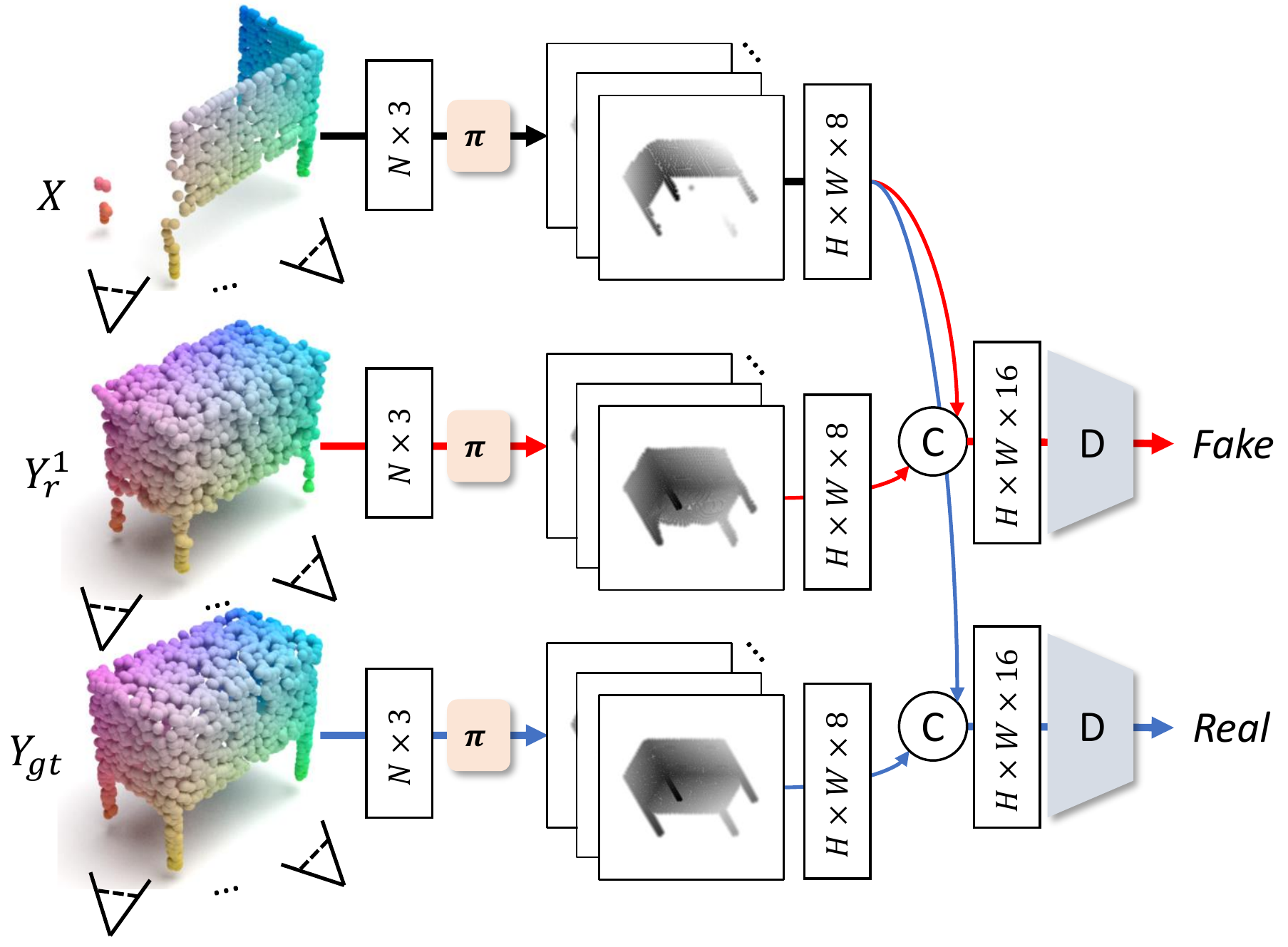}
 \footnotesize
    \caption{Discriminator $\mathtt{D}$ is trained using depth maps generated from the multi-view point renderer $\bm{\pi}$.}
\label{fig:gan}
\end{figure}

The real samples for training the discriminator $\mathtt{D}$ is the concatenation of the depth maps rendered from $X$ and $Y_r^1$,
while the fake samples come from concatenating the depth maps of $X$ and $Y_{gt}$, as illustrated in Figure~\ref{fig:gan}. 
The discriminator consists of four Conv-LeakyReLu-Dropouts and one Linear, with spectral normalization \cite{miyato2018spectral} applied on the Conv and Linear weights. The number of channels are $16$, $32$, $64$, $128$ for the four convolutional layers. Each Conv has a kernel size of 3, a stride of 2 and a padding size of 1. The slope of LeakyReLU is 0.2. The drop rate of all Dropouts is 0.75.
A last Linear outputs a scalar for discrimination.

\section{Importance of the Image Domain Supervisions}
We remove all the image-domain losses but instead implement the discriminator with a PointNet. We denote such setting as the \emph{Point-based}, and report its comparisons with our proposed setting in Tables~\ref{table:shapenet_emd},~\ref{table:shapenet_fpd},~\ref{table:shapenet_cd} and Figure~\ref{fig:image_domain}. These results all verify the effectiveness of our proposed image-domain losses in point cloud completion.  

\begin{table}[h]
\begin{center}
\footnotesize
\setlength\tabcolsep{1.5pt}
\begin{tabular}{@{}l|cccccccc|c@{}}
\toprule
Method&plane&cabinet&car&chair&lamp&sofa&table&vessel&avg\\
\midrule
Proposed & \textbf{1.131} & \textbf{2.014} & \textbf{1.783} & \textbf{2.050} & \textbf{2.063} & \textbf{2.333} & \textbf{1.729} & \textbf{1.790} & \textbf{1.862} \\
Point-based & 1.563 & 2.355 & 2.144 & 2.379 & 2.508 & 2.886 & 2.133 & 2.170 & 2.267 \\

\bottomrule
\end{tabular}
\end{center}
\caption{Completion comparison on ShapeNet in terms of EMD $\times 10^3$ (lower is better).}
\label{table:shapenet_emd}
\end{table}

\begin{table}[h]
\begin{center}
\footnotesize
\setlength\tabcolsep{1.5pt}
\begin{tabular}{@{}l|cccccccc|c@{}}
\toprule
Method&plane&cabinet&car&chair&lamp&sofa&table&vessel&avg\\
\midrule
Proposed & \textbf{0.307} & \textbf{0.691} & \textbf{0.142} & \textbf{1.113} & \textbf{0.774} & \textbf{0.945} & \textbf{0.668} & \textbf{0.523} & \textbf{0.645} \\
Point-based & 1.961 & 0.826 & 1.592 & 4.275 & 1.406 & 5.153 & 0.673 & 1.074 & 2.120 \\
\bottomrule
\end{tabular}
\end{center}
\caption{
Completion comparison on ShapeNet in terms of FPD $\times 0.1$ (lower is better).}
\label{table:shapenet_fpd}
\end{table}

\begin{table}[h]
\begin{center}
\footnotesize
\setlength\tabcolsep{1.5pt}
\begin{tabular}{@{}l|cccccccc|c@{}}
\toprule
Method&plane&cabinet&car&chair&lamp&sofa&table&vessel&avg\\
\midrule
Proposed & \textbf{0.176} & \textbf{0.664} & \textbf{0.362} & \textbf{0.616} & \textbf{0.631} & \textbf{0.789} & \textbf{0.498} & \textbf{0.384} & \textbf{0.515} \\
Point-based & 0.331 & 0.809 & 0.471 & 0.812 & 0.834 & 1.187 & 0.649 & 0.557 & 0.706 \\
\bottomrule
\end{tabular}
\end{center}
\caption{Completion comparison on ShapeNet in terms of CD $\times 10^3$ (lower is better).}
\label{table:shapenet_cd}
\end{table}

\section{More Qualitative Results}
\begin{figure*}[h!]
\center
\setlength\tabcolsep{0pt}
{
\renewcommand{\arraystretch}{0.0}
\small
\begin{tabular}{@{}rcccccccccc@{}}
    & Input & AtlasNet\cite{atlasnet2018} & FCAE & FoldingNet\cite{foldingnet_2018_CVPR} &PCN\cite{Yuan-2018-pcn} 
    & MSN \cite{liu2019morphing} & GRNet \cite{xie2020grnet} & \emph{Ours} & Groundtruth\\
    
    \raisebox{0.6\height}{\rotatebox{90}{Airplane}}&
    \includegraphics[width=0.1\columnwidth,trim=30 30 30 30, clip]{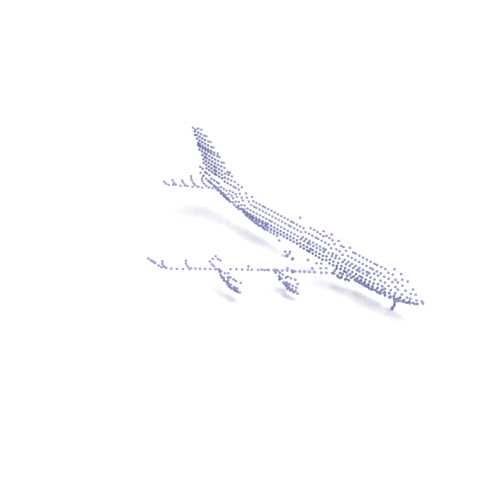}&   
    \includegraphics[width=0.1\columnwidth,trim=30 30 30 30, clip]{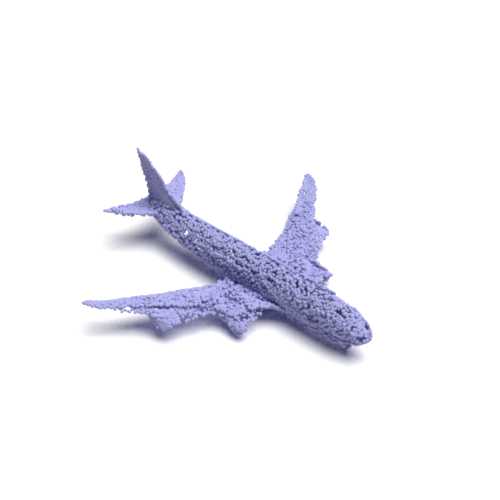}&
    \includegraphics[width=0.1\columnwidth,trim=30 30 30 30, clip]{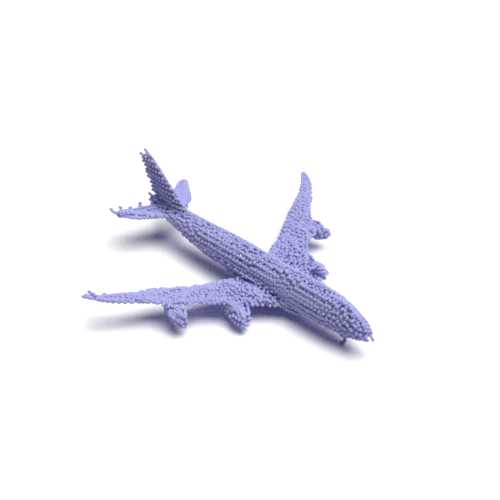}&
    \includegraphics[width=0.1\columnwidth,trim=30 30 30 30, clip]{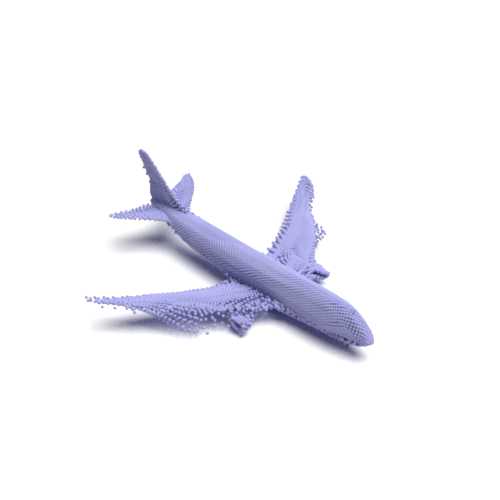}&
    \includegraphics[width=0.1\columnwidth,trim=30 30 30 30, clip]{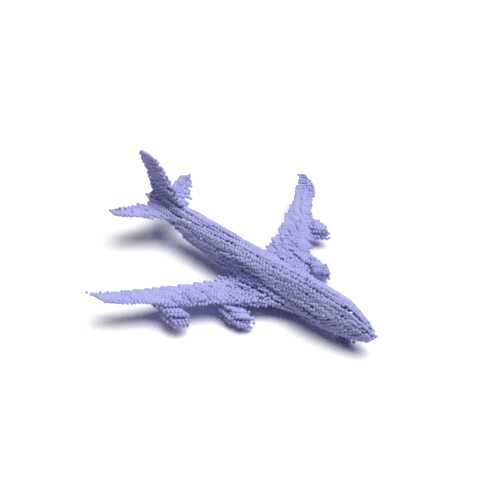}&
    \includegraphics[width=0.1\columnwidth,trim=30 30 30 30, clip]{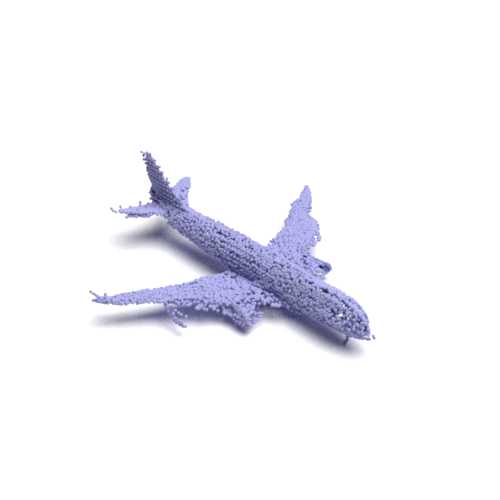}&
    \includegraphics[width=0.1\columnwidth,trim=30 30 30 30, clip]{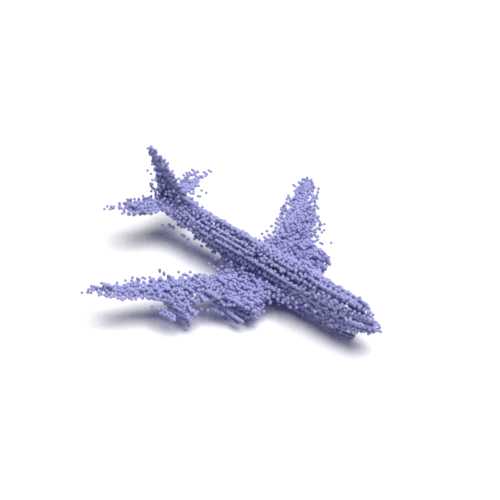}&
    \includegraphics[width=0.1\columnwidth,trim=30 30 30 30, clip]{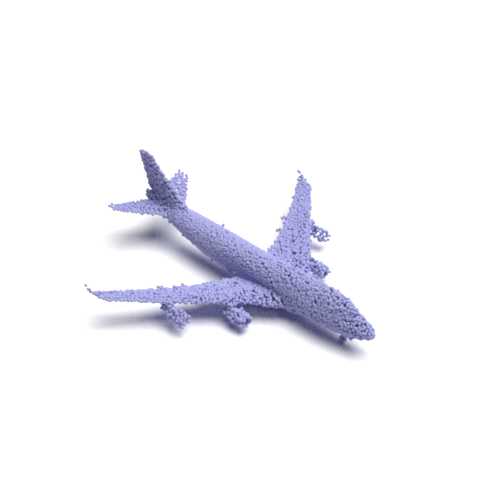}&
    \includegraphics[width=0.1\columnwidth,trim=30 30 30 30, clip]{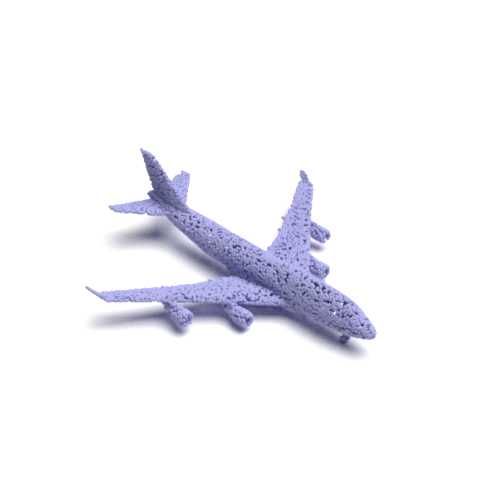}\\
    
    \raisebox{0.6\height}{\rotatebox{90}{Airplane}}&
    \includegraphics[width=0.1\columnwidth,trim=30 30 30 30, clip]{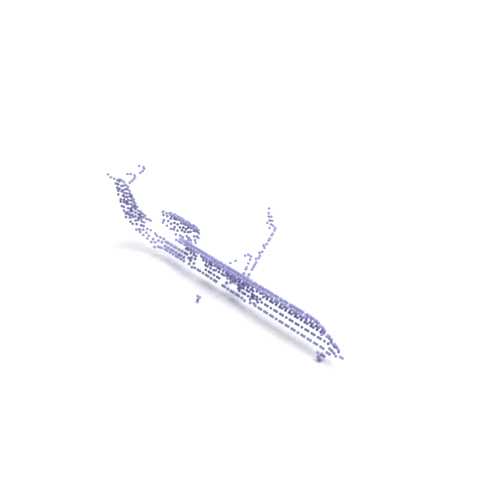}&
    \includegraphics[width=0.1\columnwidth,trim=30 30 30 30, clip]{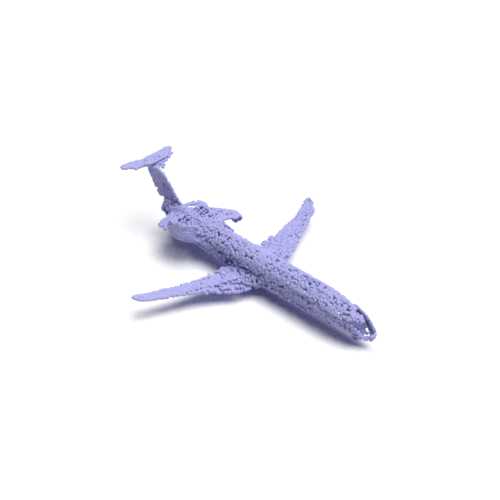}&
    \includegraphics[width=0.1\columnwidth,trim=30 30 30 30, clip]{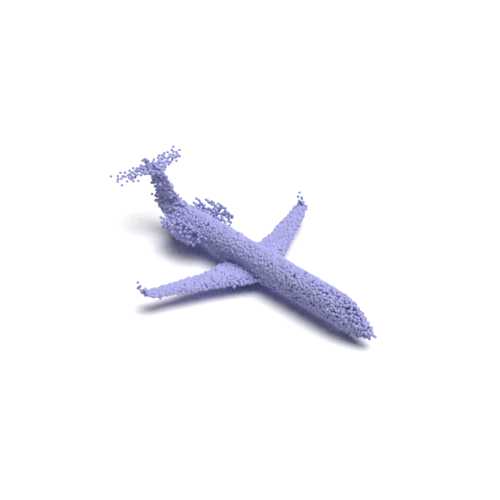}&
    \includegraphics[width=0.1\columnwidth,trim=30 30 30 30, clip]{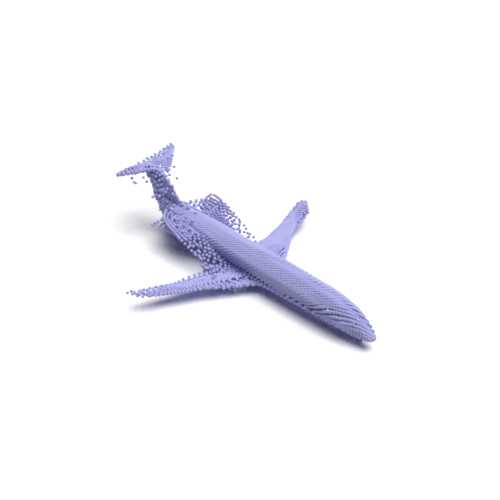}&
    \includegraphics[width=0.1\columnwidth,trim=30 30 30 30, clip]{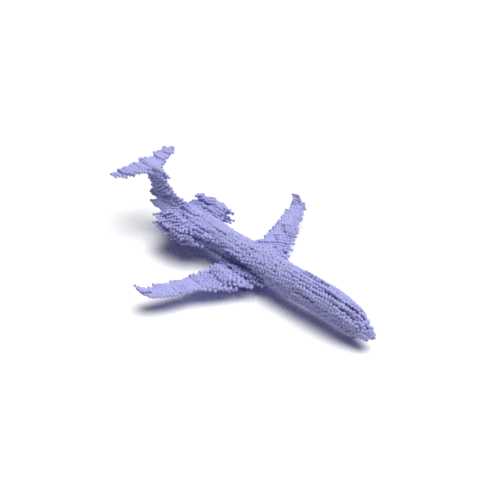}&
    \includegraphics[width=0.1\columnwidth,trim=30 30 30 30, clip]{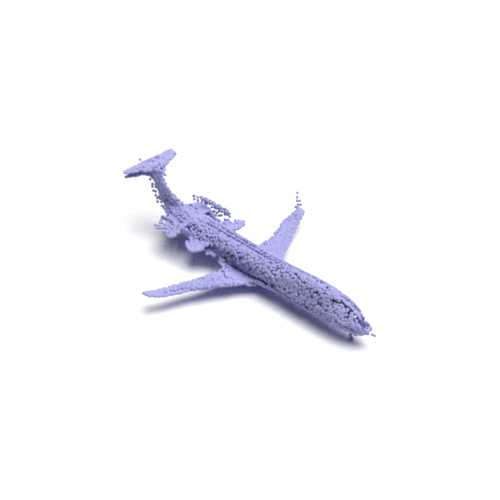}&
    \includegraphics[width=0.1\columnwidth,trim=30 30 30 30, clip]{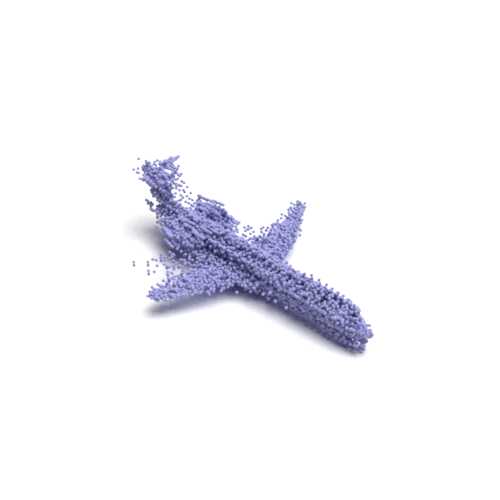}&
    \includegraphics[width=0.1\columnwidth,trim=30 30 30 30, clip]{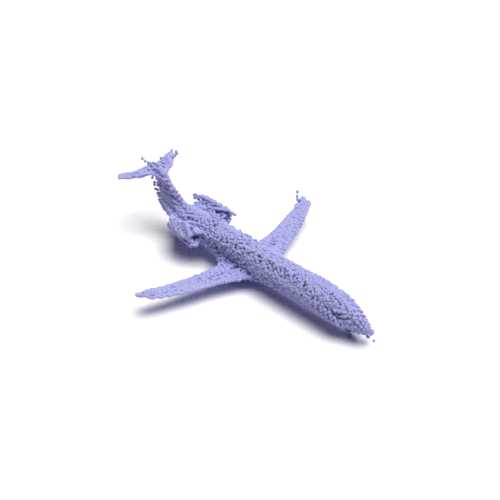}&
    \includegraphics[width=0.1\columnwidth,trim=30 30 30 30, clip]{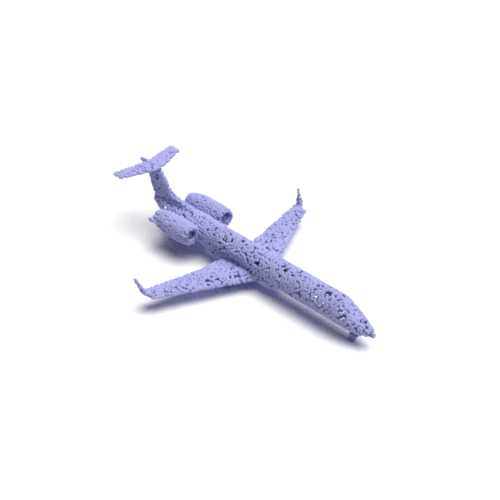}\\
    
    \raisebox{0.6\height}{\rotatebox{90}{Airplane}}&
    \includegraphics[width=0.1\columnwidth,trim=30 30 30 30, clip]{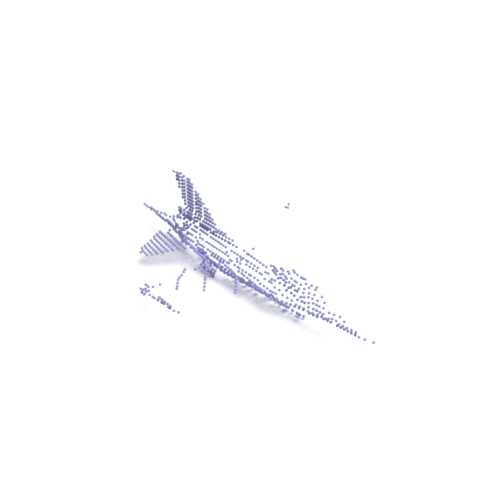}&
    \includegraphics[width=0.1\columnwidth,trim=30 30 30 30, clip]{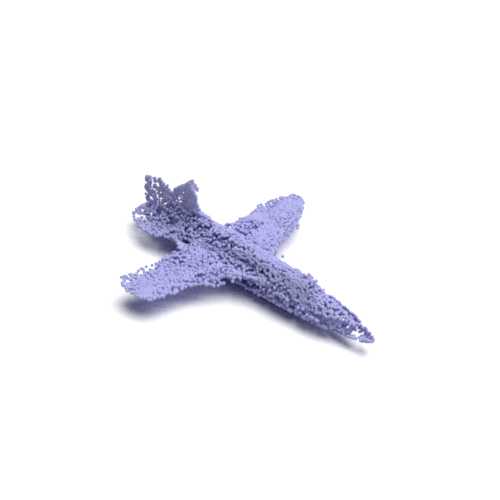}&
    \includegraphics[width=0.1\columnwidth,trim=30 30 30 30, clip]{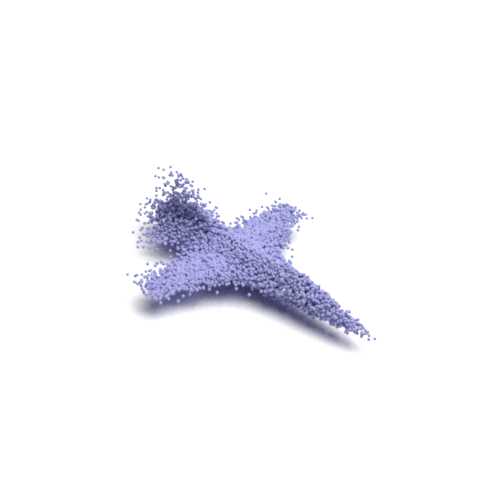}&
    \includegraphics[width=0.1\columnwidth,trim=30 30 30 30, clip]{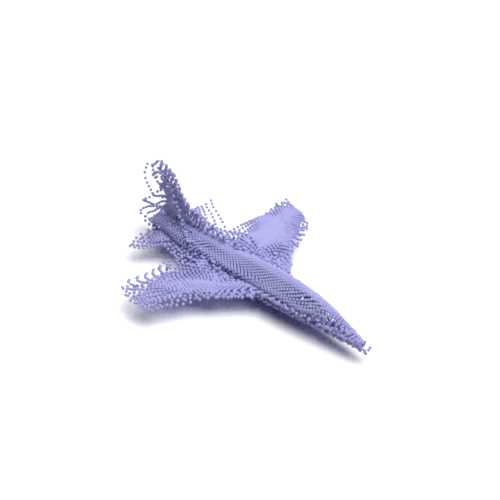}&
    \includegraphics[width=0.1\columnwidth,trim=30 30 30 30, clip]{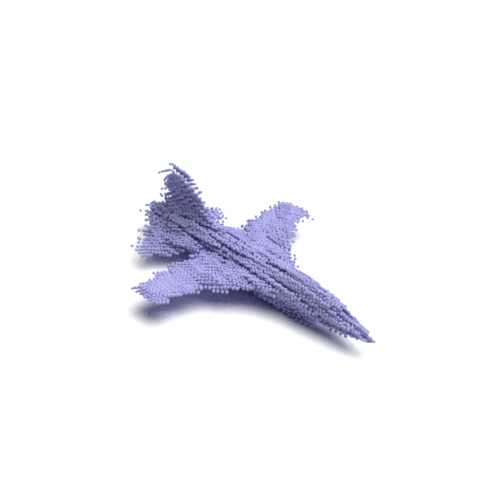}&
    \includegraphics[width=0.1\columnwidth,trim=30 30 30 30, clip]{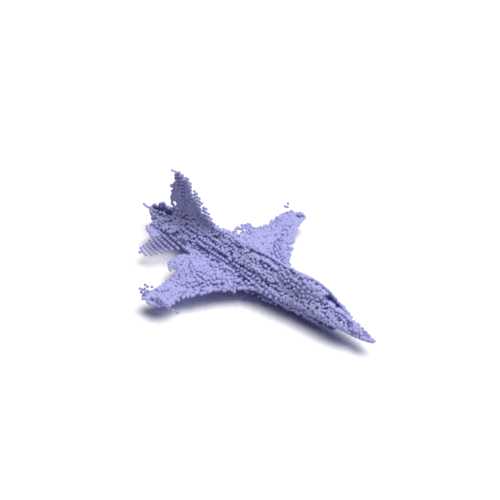}&
    \includegraphics[width=0.1\columnwidth,trim=30 30 30 30, clip]{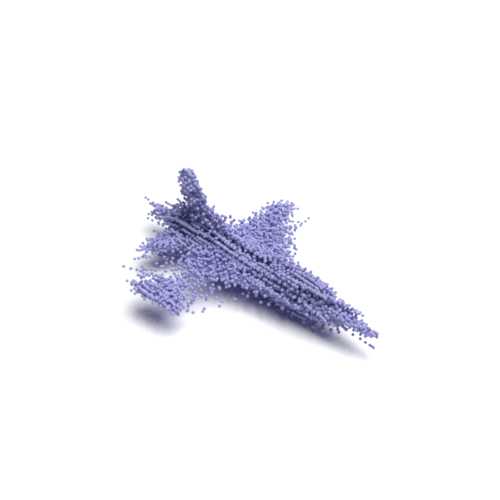}&
    \includegraphics[width=0.1\columnwidth,trim=30 30 30 30, clip]{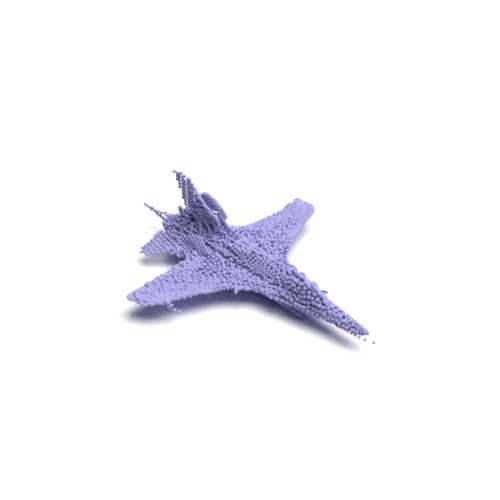}&
    \includegraphics[width=0.1\columnwidth,trim=30 30 30 30, clip]{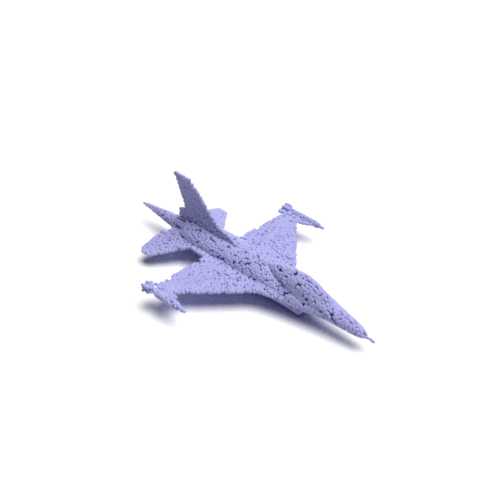}\\
    
    \raisebox{0.6\height}{\rotatebox{90}{Airplane}}&
    \includegraphics[width=0.1\columnwidth,trim=30 30 30 30, clip]{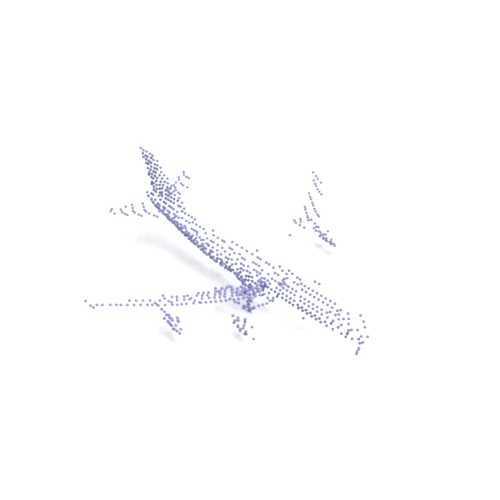}&
    \includegraphics[width=0.1\columnwidth,trim=30 30 30 30, clip]{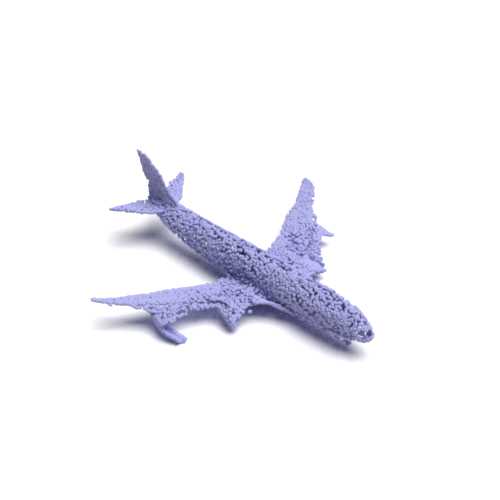}&
    \includegraphics[width=0.1\columnwidth,trim=30 30 30 30, clip]{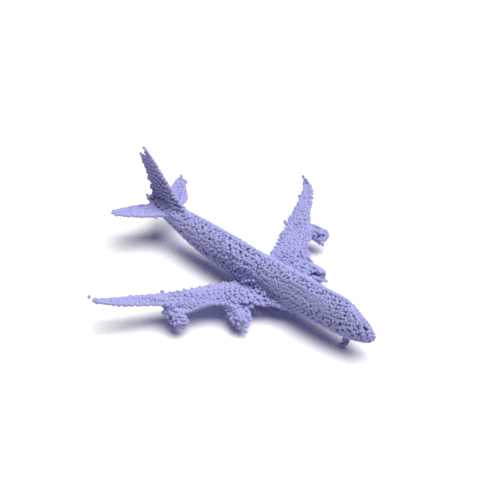}&
    \includegraphics[width=0.1\columnwidth,trim=30 30 30 30, clip]{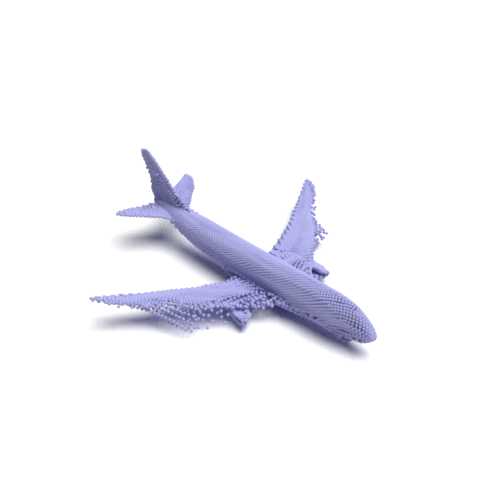}&
    \includegraphics[width=0.1\columnwidth,trim=30 30 30 30, clip]{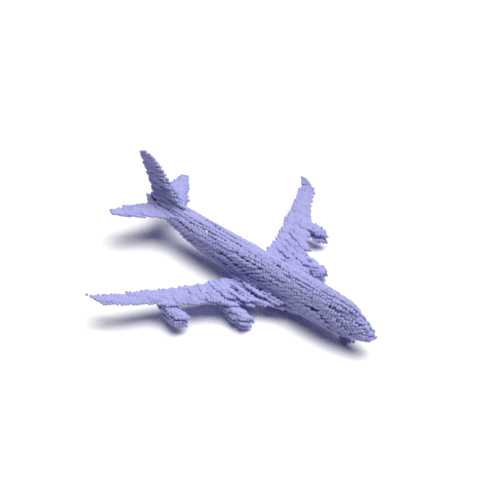}&
    \includegraphics[width=0.1\columnwidth,trim=30 30 30 30, clip]{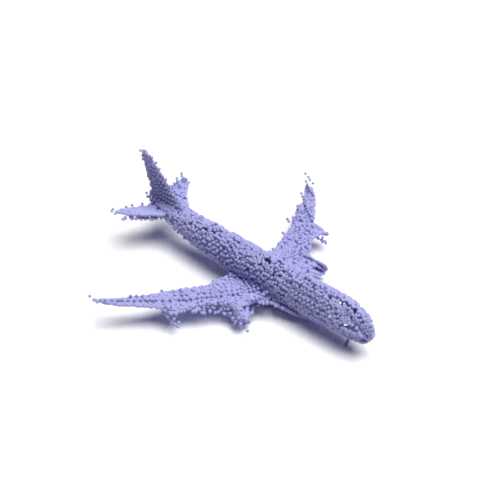}&
    \includegraphics[width=0.1\columnwidth,trim=30 30 30 30, clip]{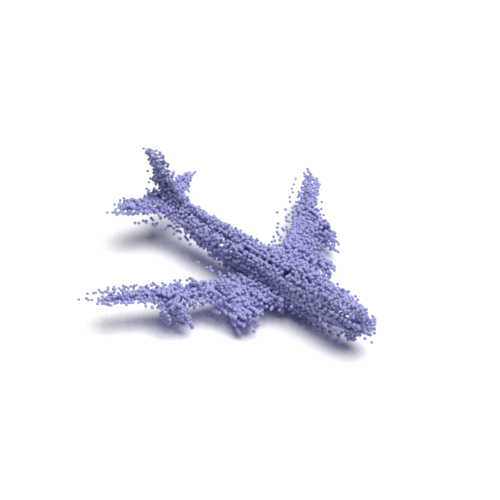}&
    \includegraphics[width=0.1\columnwidth,trim=30 30 30 30, clip]{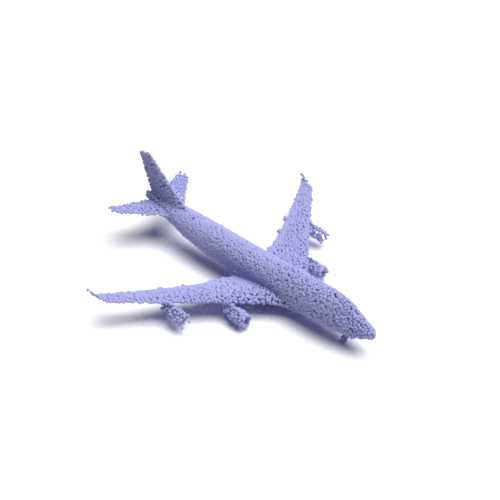}&
    \includegraphics[width=0.1\columnwidth,trim=30 30 30 30, clip]{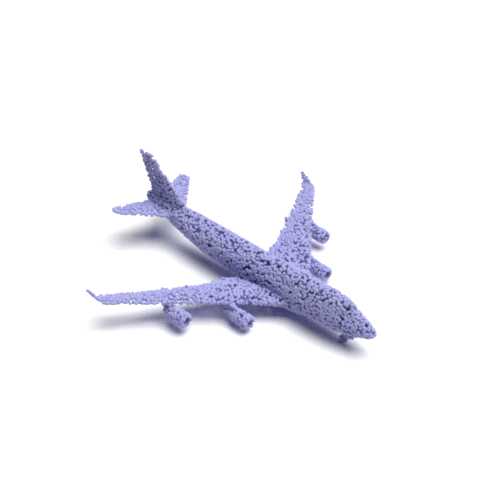}\\
    
    \raisebox{0.6\height}{\rotatebox{90}{Cabinet}}&
    \includegraphics[width=0.1\columnwidth,trim=30 30 30 30, clip]{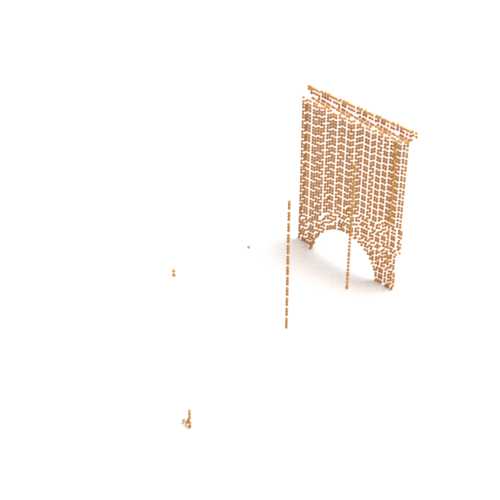}&
    \includegraphics[width=0.1\columnwidth,trim=30 30 30 30, clip]{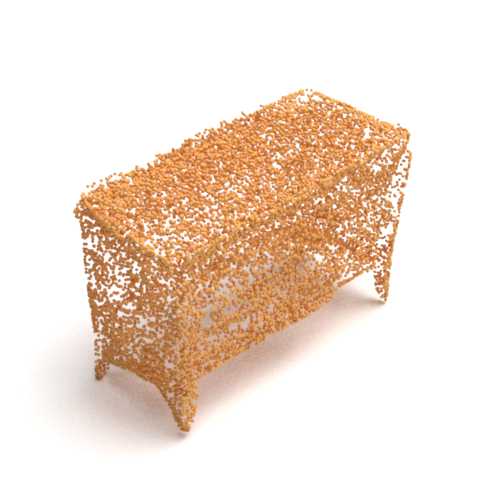}&
    \includegraphics[width=0.1\columnwidth,trim=30 30 30 30, clip]{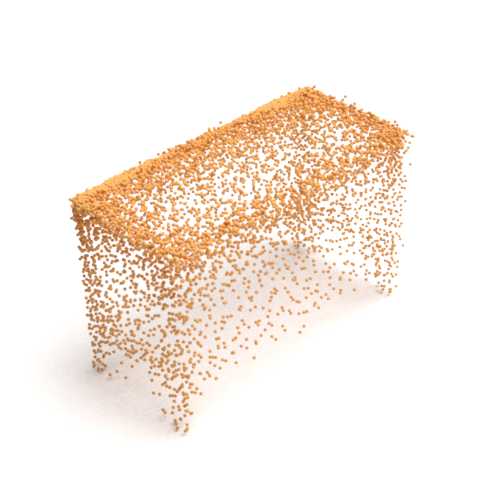}&
    \includegraphics[width=0.1\columnwidth,trim=30 30 30 30, clip]{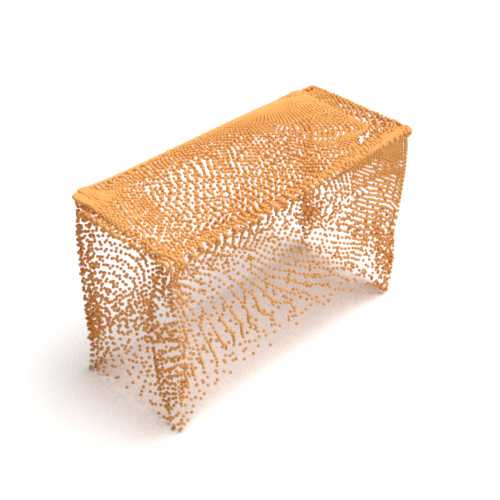}&
    \includegraphics[width=0.1\columnwidth,trim=30 30 30 30, clip]{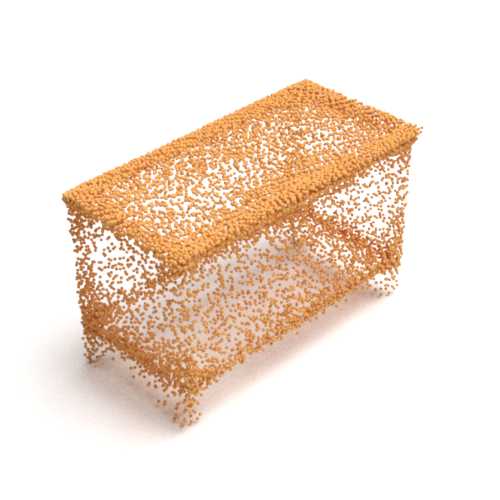}&
    \includegraphics[width=0.1\columnwidth,trim=30 30 30 30, clip]{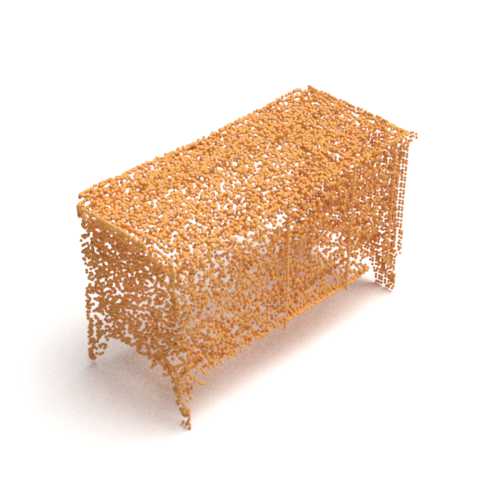}&
    \includegraphics[width=0.1\columnwidth,trim=30 30 30 30, clip]{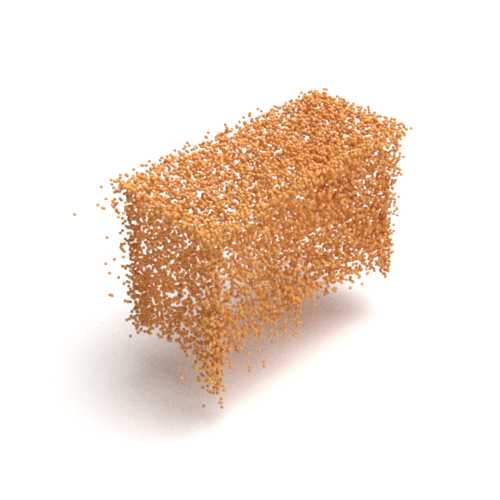}&
    \includegraphics[width=0.1\columnwidth,trim=30 30 30 30, clip]{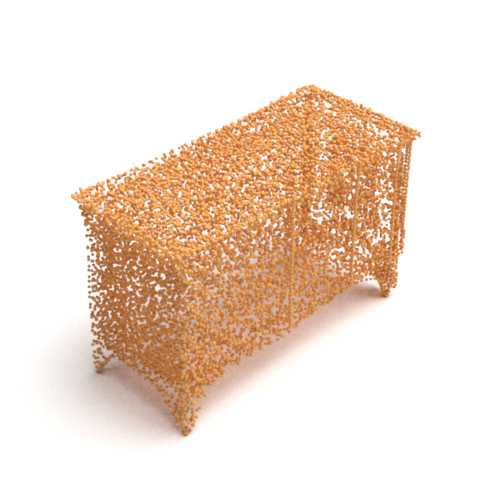}&
    \includegraphics[width=0.1\columnwidth,trim=30 30 30 30, clip]{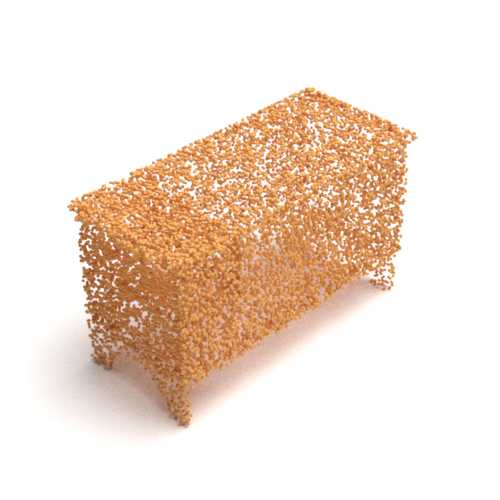}\\
    
    \raisebox{0.6\height}{\rotatebox{90}{Cabinet}}&
    \includegraphics[width=0.1\columnwidth,trim=30 30 30 30, clip]{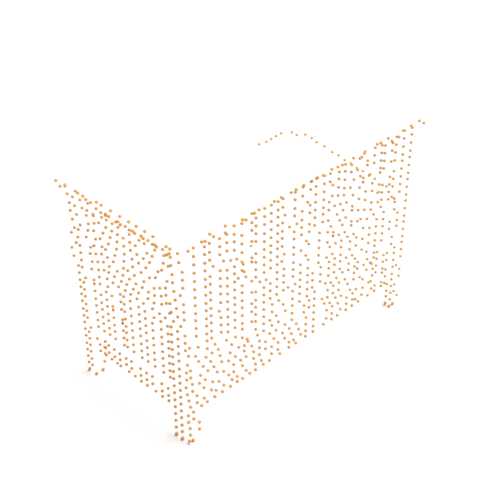}&
    \includegraphics[width=0.1\columnwidth,trim=30 30 30 30, clip]{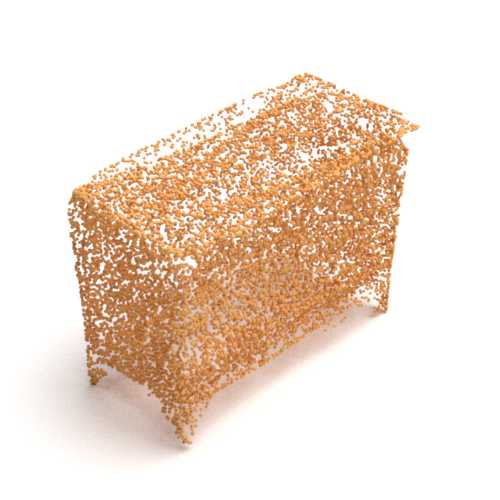}&
    \includegraphics[width=0.1\columnwidth,trim=30 30 30 30, clip]{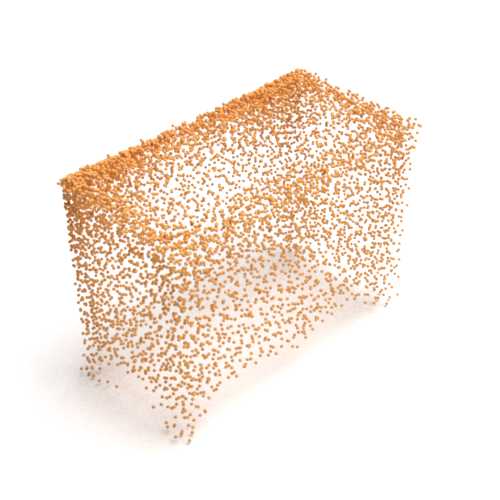}&
    \includegraphics[width=0.1\columnwidth,trim=30 30 30 30, clip]{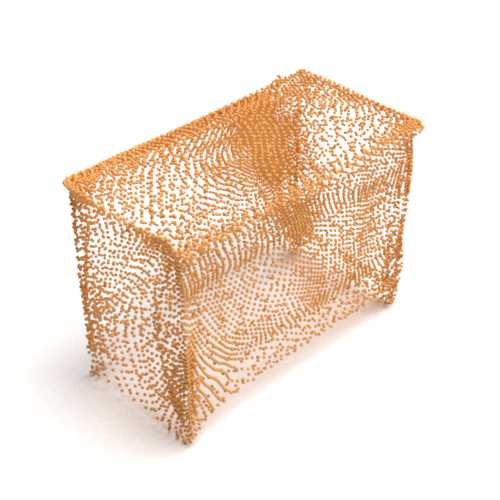}&
    \includegraphics[width=0.1\columnwidth,trim=30 30 30 30, clip]{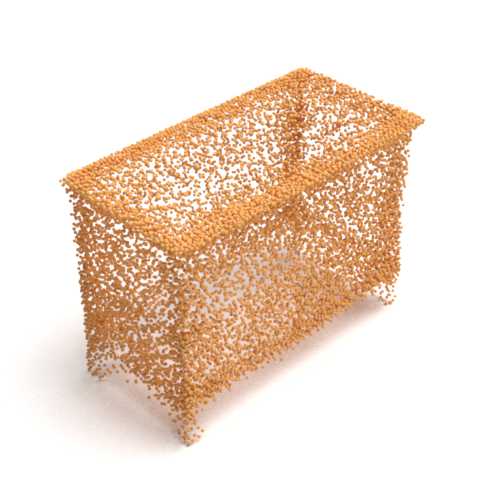}&
    \includegraphics[width=0.1\columnwidth,trim=30 30 30 30, clip]{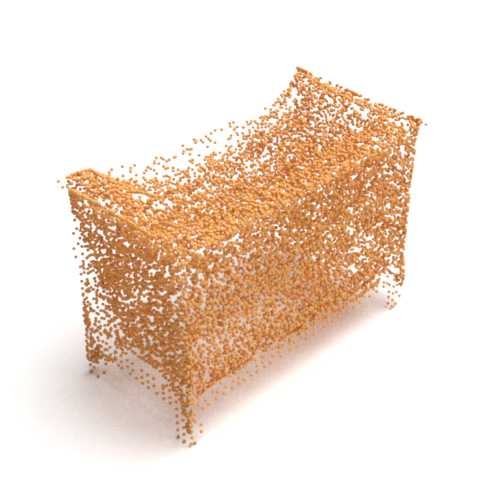}&
    \includegraphics[width=0.1\columnwidth,trim=30 30 30 30, clip]{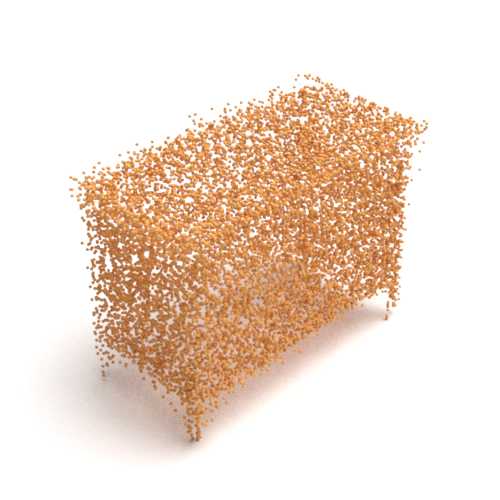}&
    \includegraphics[width=0.1\columnwidth,trim=30 30 30 30, clip]{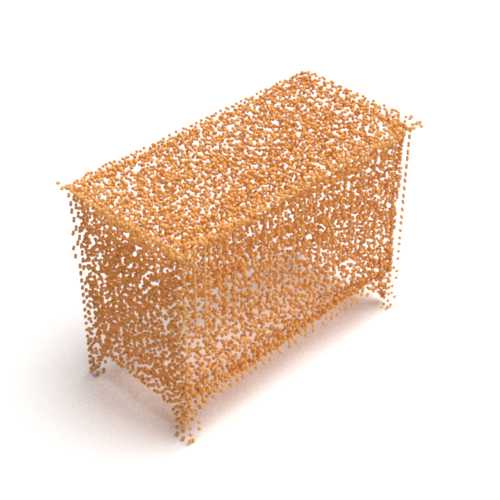}&
    \includegraphics[width=0.1\columnwidth,trim=30 30 30 30, clip]{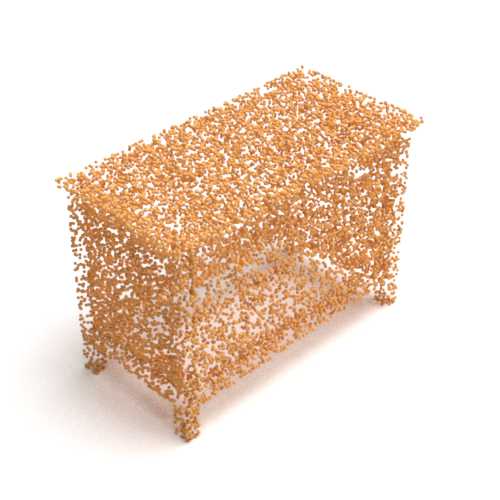}\\
    
    \raisebox{0.6\height}{\rotatebox{90}{Cabinet}}&
    \includegraphics[width=0.1\columnwidth,trim=30 30 30 30, clip]{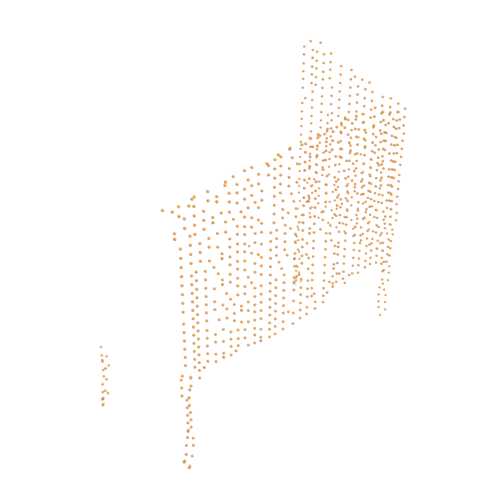}&   
    \includegraphics[width=0.1\columnwidth,trim=30 30 30 30, clip]{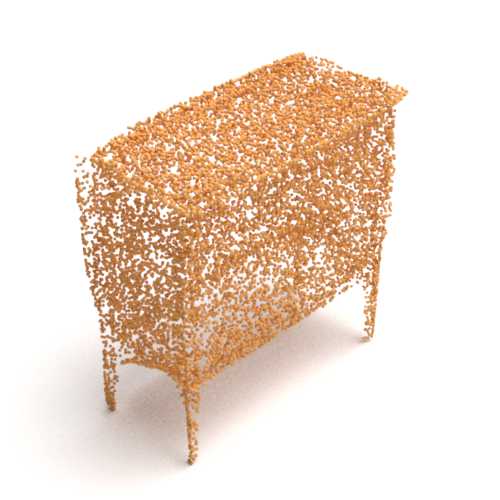}&
    \includegraphics[width=0.1\columnwidth,trim=30 30 30 30, clip]{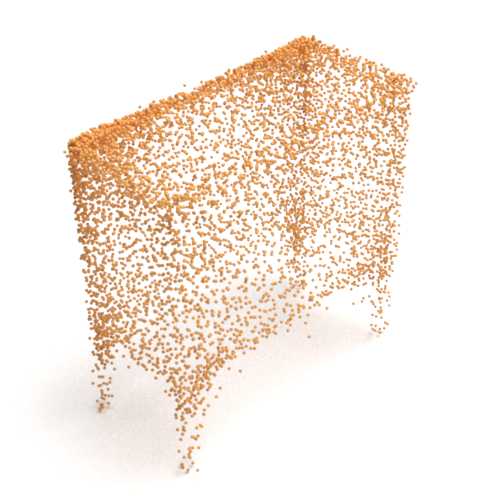}&
    \includegraphics[width=0.1\columnwidth,trim=30 30 30 30, clip]{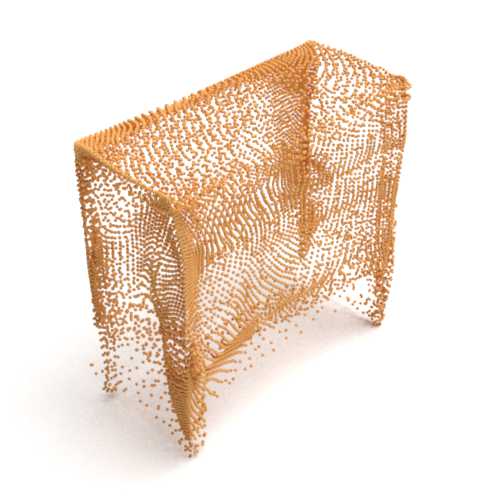}&
    \includegraphics[width=0.1\columnwidth,trim=30 30 30 30, clip]{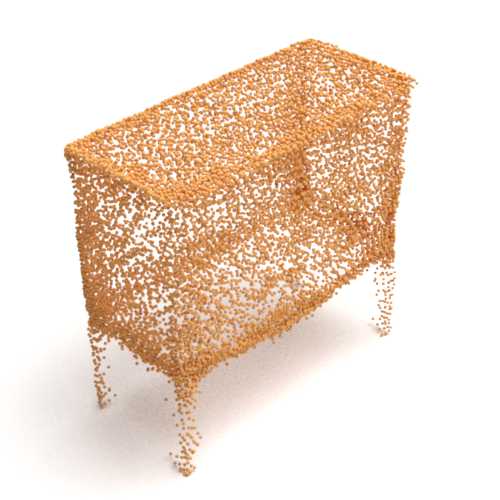}&
    \includegraphics[width=0.1\columnwidth,trim=30 30 30 30, clip]{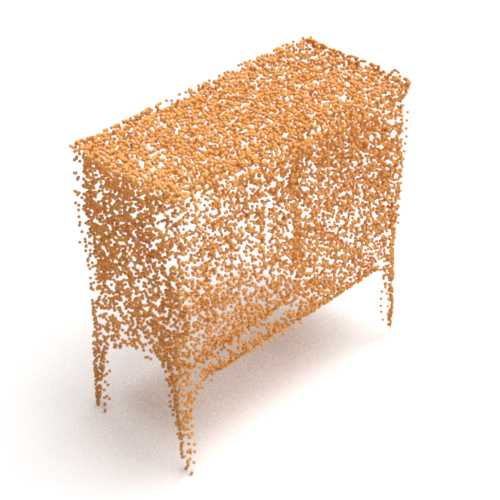}&
    \includegraphics[width=0.1\columnwidth,trim=30 30 30 30, clip]{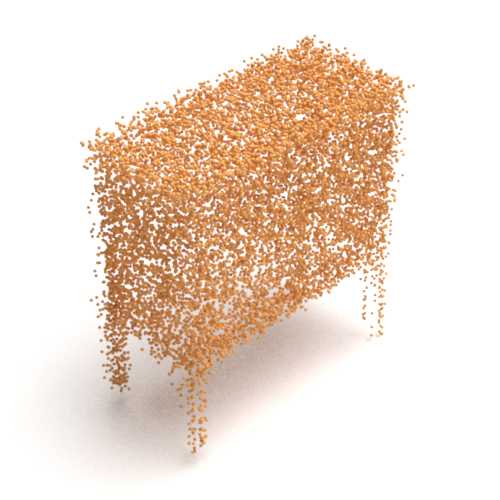}&
    \includegraphics[width=0.1\columnwidth,trim=30 30 30 30, clip]{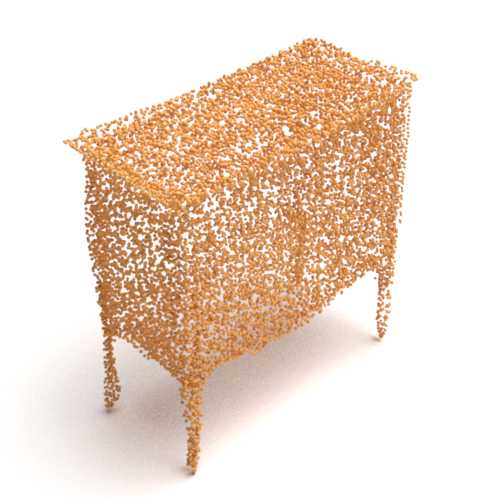}&
    \includegraphics[width=0.1\columnwidth,trim=30 30 30 30, clip]{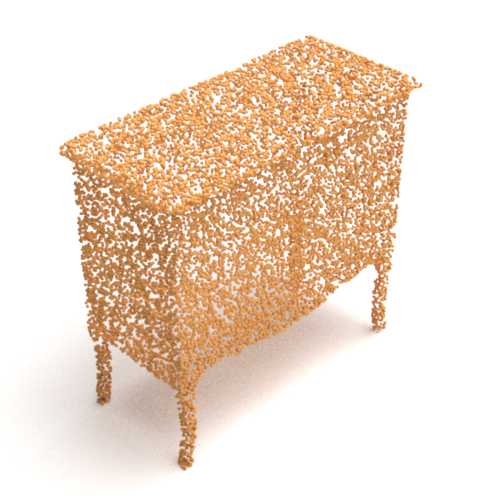}\\
    
    \raisebox{0.6\height}{\rotatebox{90}{Cabinet}}&
    \includegraphics[width=0.1\columnwidth,trim=30 30 30 30, clip]{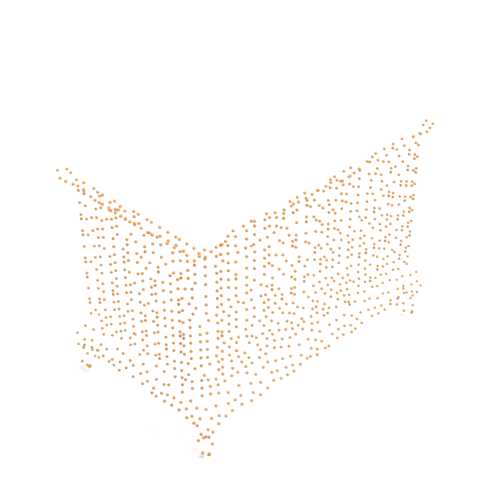}&
    \includegraphics[width=0.1\columnwidth,trim=30 30 30 30, clip]{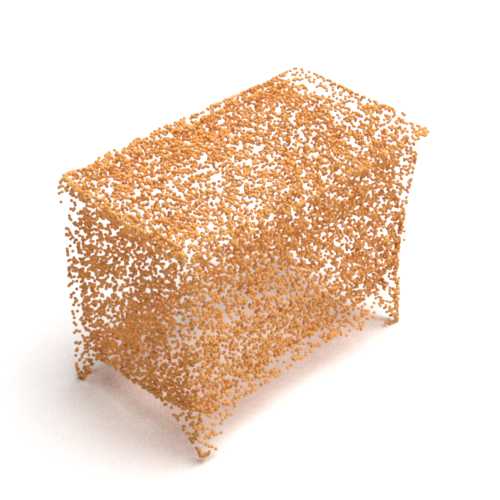}&
    \includegraphics[width=0.1\columnwidth,trim=30 30 30 30, clip]{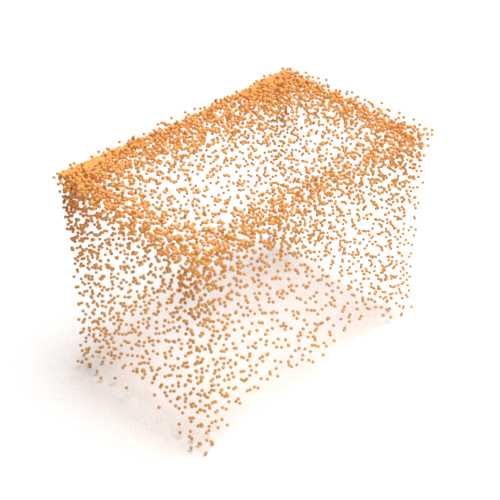}&
    \includegraphics[width=0.1\columnwidth,trim=30 30 30 30, clip]{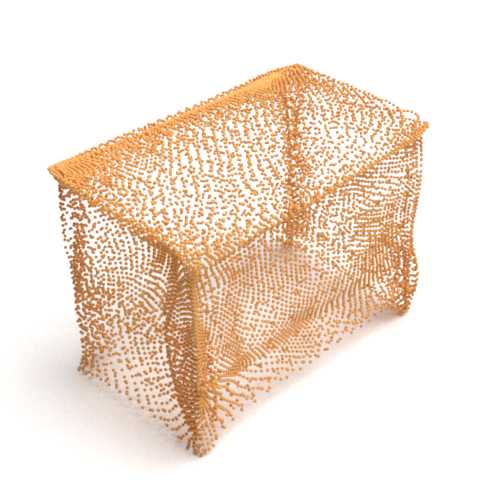}&
    \includegraphics[width=0.1\columnwidth,trim=30 30 30 30, clip]{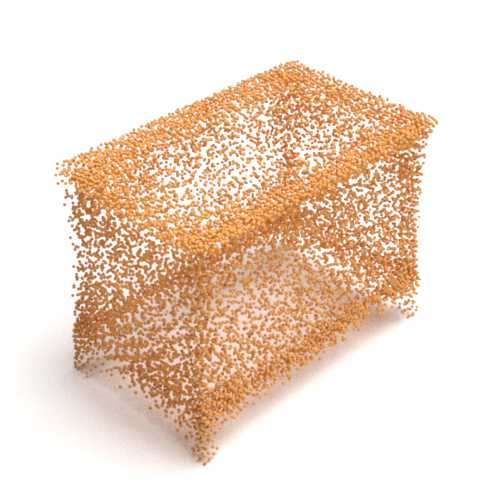}&
    \includegraphics[width=0.1\columnwidth,trim=30 30 30 30, clip]{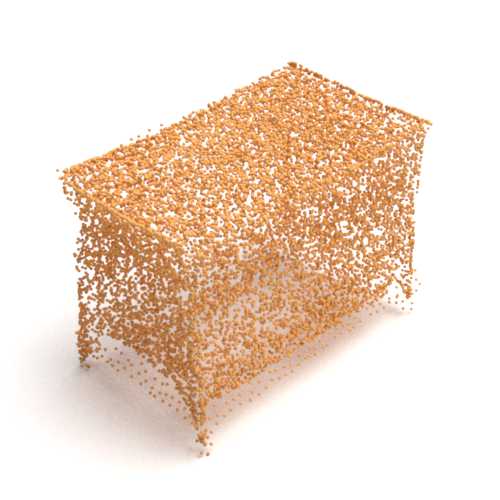}&
    \includegraphics[width=0.1\columnwidth,trim=30 30 30 30, clip]{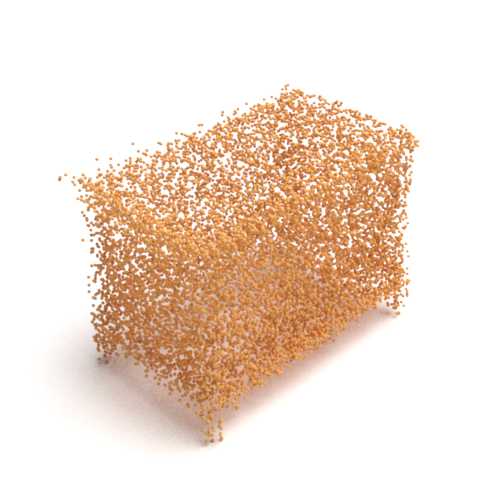}&
    \includegraphics[width=0.1\columnwidth,trim=30 30 30 30, clip]{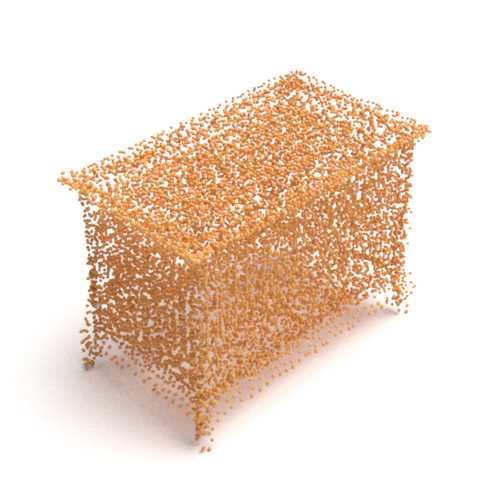}&
    \includegraphics[width=0.1\columnwidth,trim=30 30 30 30, clip]{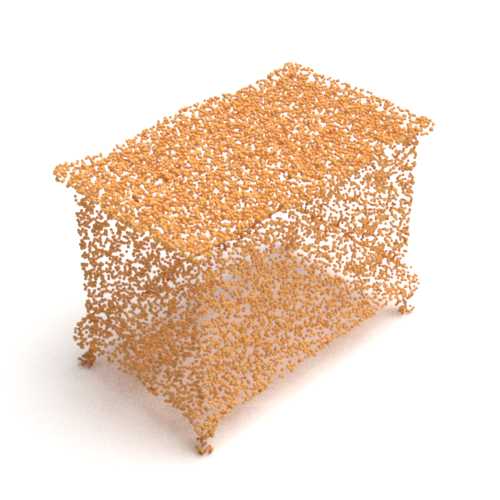}\\
    
    \raisebox{\height}{\rotatebox{90}{Car}}&
    \includegraphics[width=0.1\columnwidth,trim=30 30 30 30, clip]{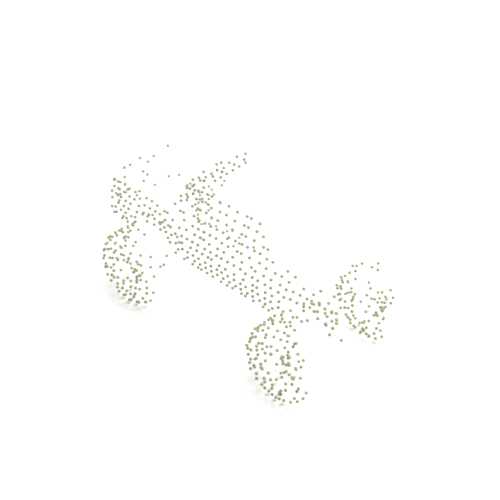}&
    \includegraphics[width=0.1\columnwidth,trim=30 30 30 30, clip]{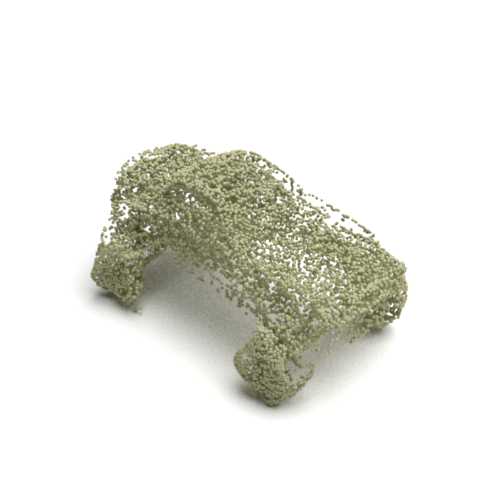}&
    \includegraphics[width=0.1\columnwidth,trim=30 30 30 30, clip]{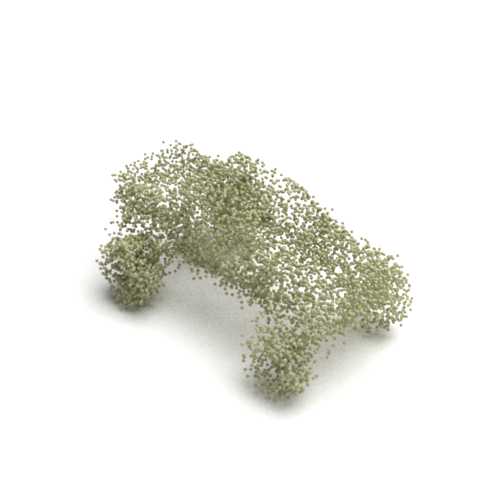}&
    \includegraphics[width=0.1\columnwidth,trim=30 30 30 30, clip]{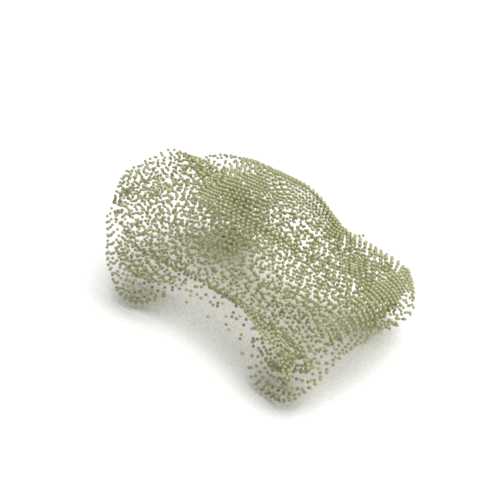}&
    \includegraphics[width=0.1\columnwidth,trim=30 30 30 30, clip]{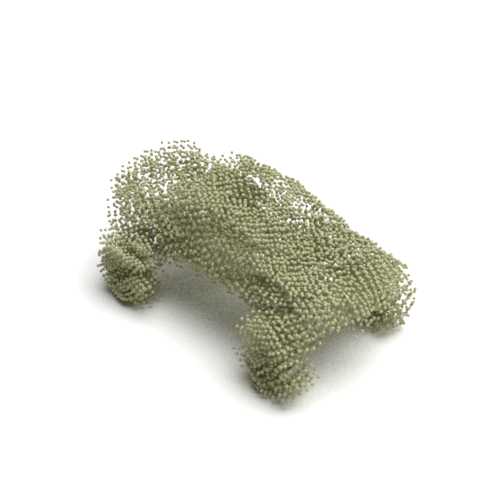}&
    \includegraphics[width=0.1\columnwidth,trim=30 30 30 30, clip]{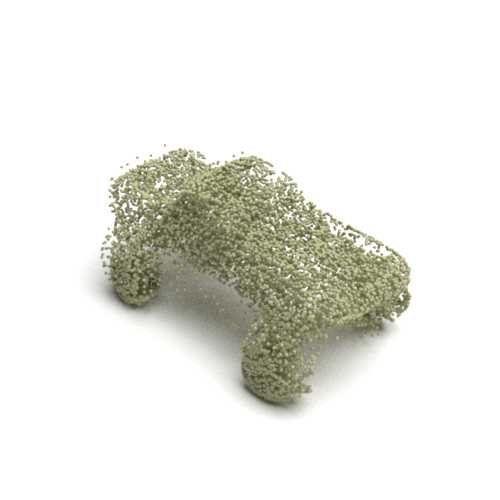}&
    \includegraphics[width=0.1\columnwidth,trim=30 30 30 30, clip]{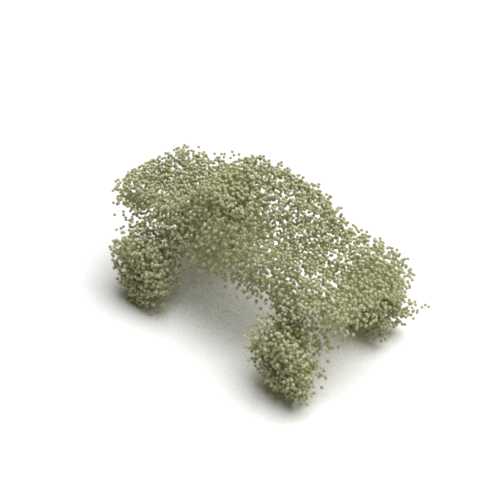}&
    \includegraphics[width=0.1\columnwidth,trim=30 30 30 30, clip]{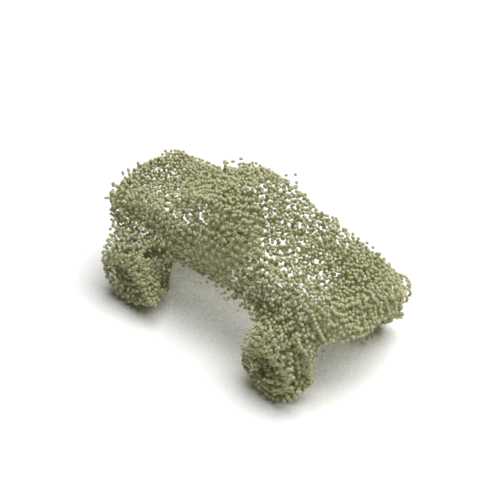}&
    \includegraphics[width=0.1\columnwidth,trim=30 30 30 30, clip]{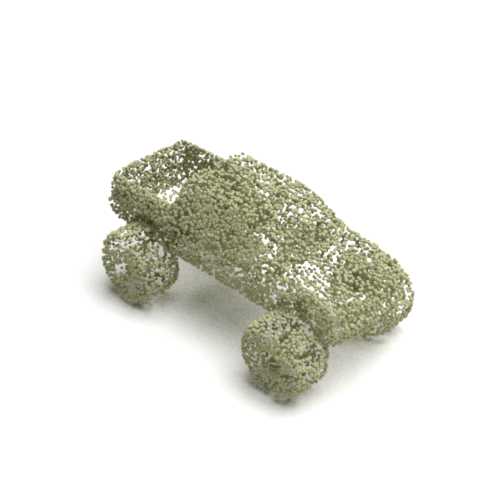}\\
    
    \raisebox{\height}{\rotatebox{90}{Car}}&
    \includegraphics[width=0.1\columnwidth,trim=30 30 30 30, clip]{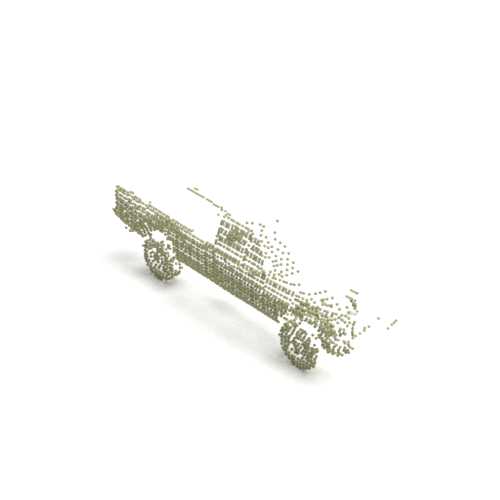}&
    \includegraphics[width=0.1\columnwidth,trim=30 30 30 30, clip]{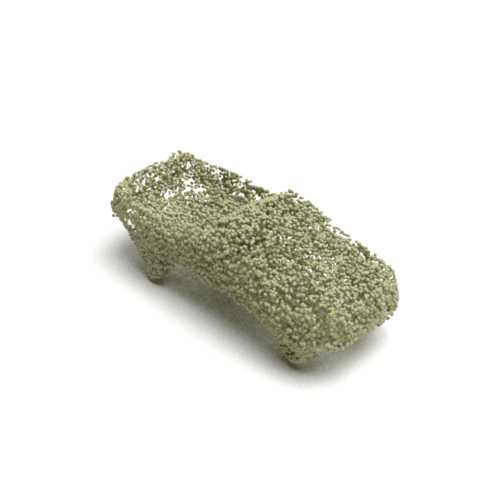}&
    \includegraphics[width=0.1\columnwidth,trim=30 30 30 30, clip]{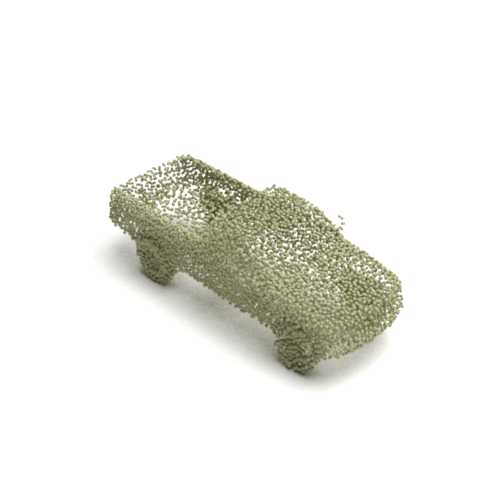}&
    \includegraphics[width=0.1\columnwidth,trim=30 30 30 30, clip]{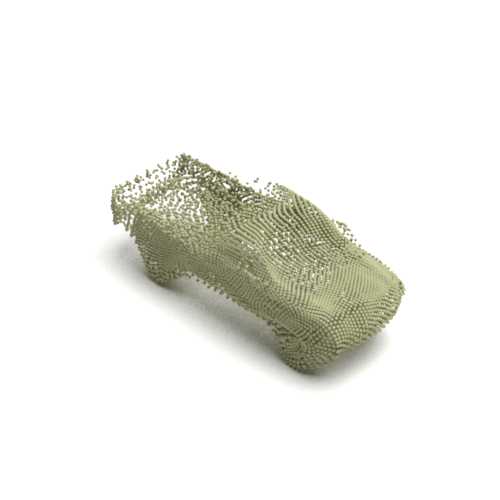}&
    \includegraphics[width=0.1\columnwidth,trim=30 30 30 30, clip]{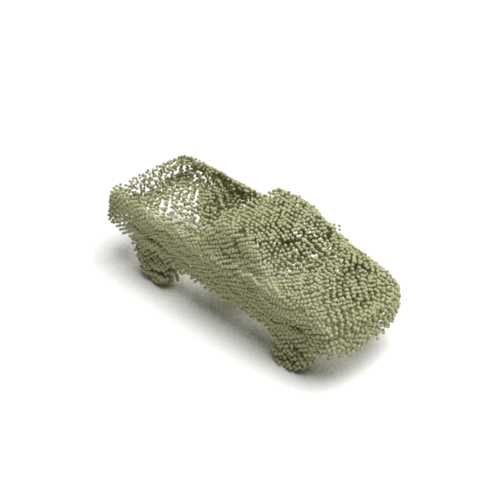}&
    \includegraphics[width=0.1\columnwidth,trim=30 30 30 30, clip]{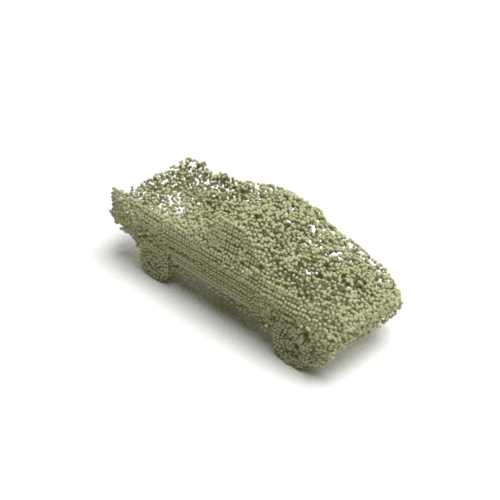}&
    \includegraphics[width=0.1\columnwidth,trim=30 30 30 30, clip]{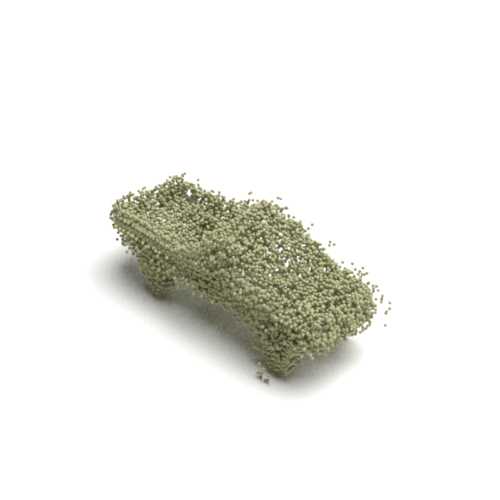}&
    \includegraphics[width=0.1\columnwidth,trim=30 30 30 30, clip]{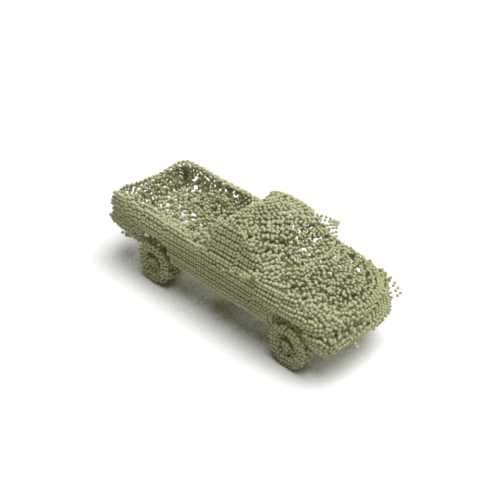}&
    \includegraphics[width=0.1\columnwidth,trim=30 30 30 30, clip]{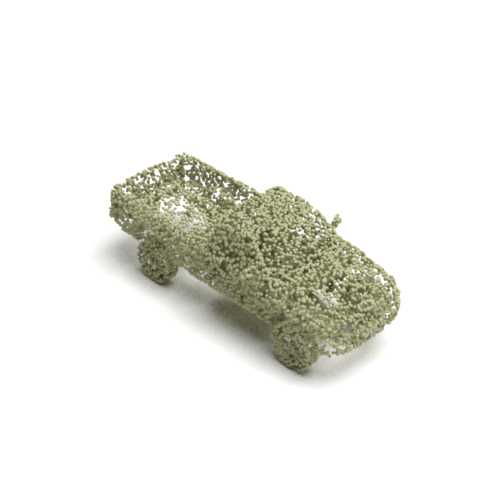}\\
    
    \raisebox{\height}{\rotatebox{90}{Car}}&
    \includegraphics[width=0.1\columnwidth,trim=30 30 30 30, clip]{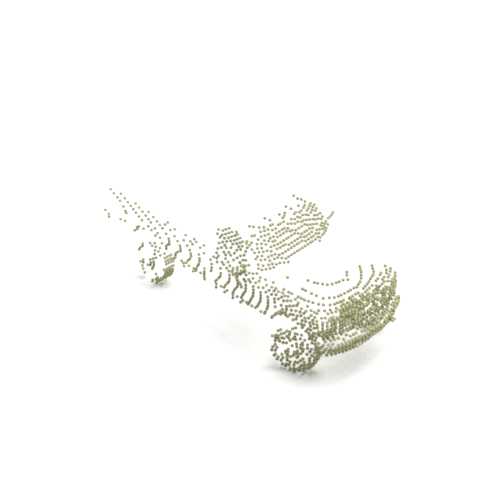}&
    \includegraphics[width=0.1\columnwidth,trim=30 30 30 30, clip]{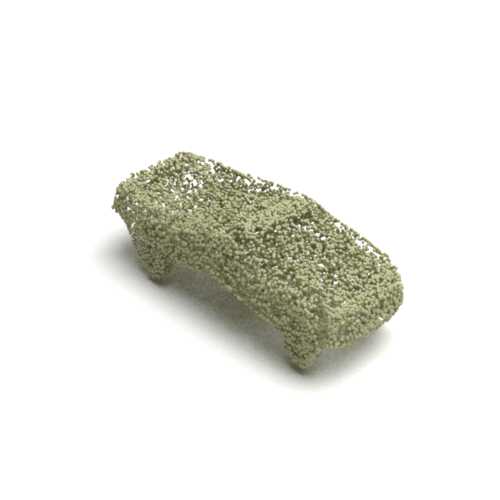}&
    \includegraphics[width=0.1\columnwidth,trim=30 30 30 30, clip]{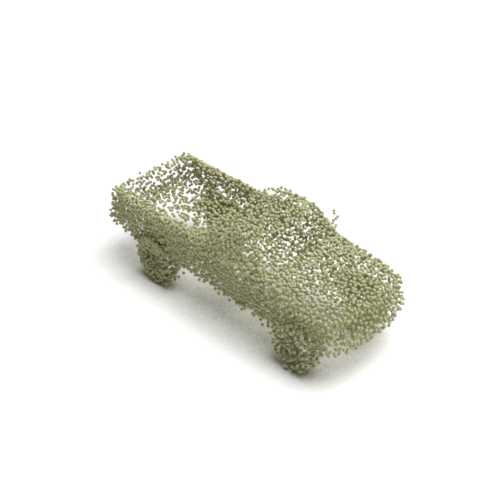}&
    \includegraphics[width=0.1\columnwidth,trim=30 30 30 30, clip]{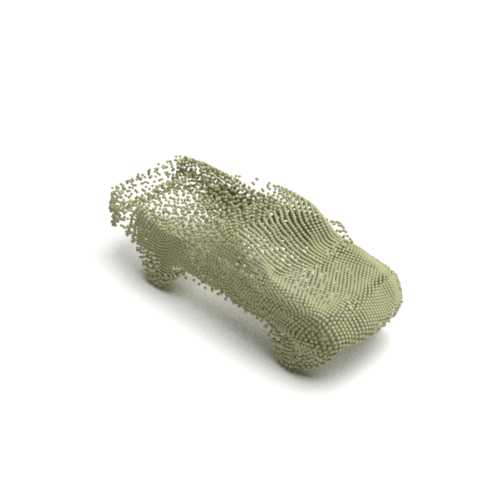}&
    \includegraphics[width=0.1\columnwidth,trim=30 30 30 30, clip]{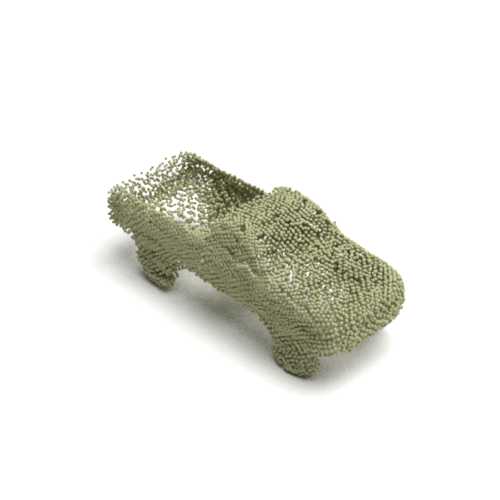}&
    \includegraphics[width=0.1\columnwidth,trim=30 30 30 30, clip]{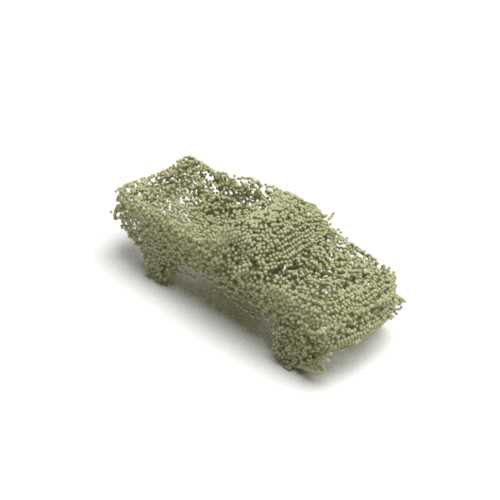}&
    \includegraphics[width=0.1\columnwidth,trim=30 30 30 30, clip]{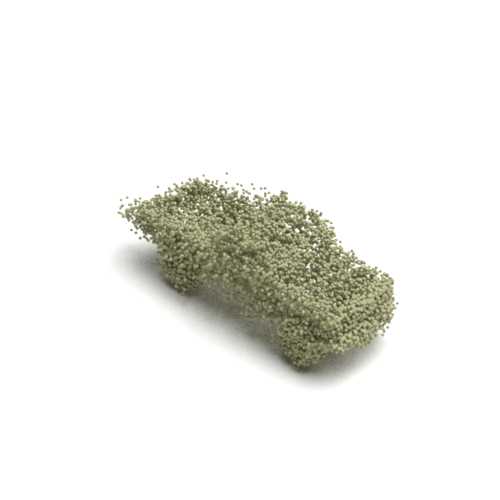}&
    \includegraphics[width=0.1\columnwidth,trim=30 30 30 30, clip]{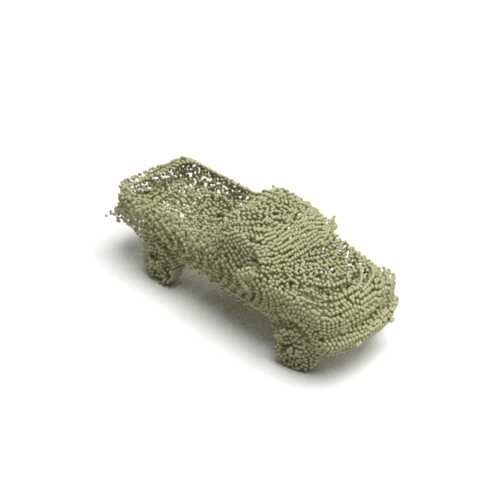}&
    \includegraphics[width=0.1\columnwidth,trim=30 30 30 30, clip]{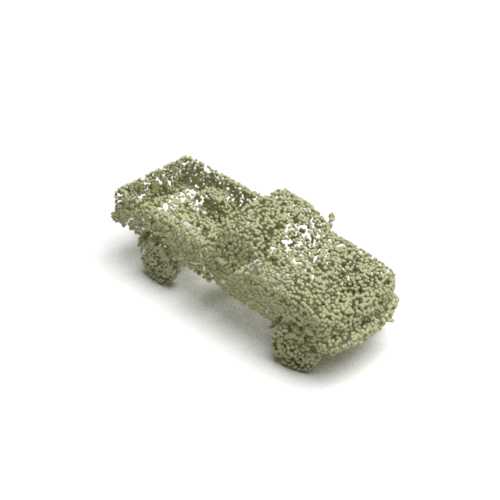}\\
    
    \raisebox{\height}{\rotatebox{90}{Car}}&
    \includegraphics[width=0.1\columnwidth,trim=30 30 30 30, clip]{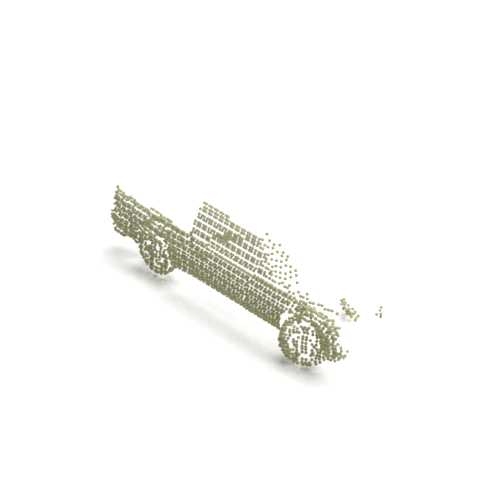}&
    \includegraphics[width=0.1\columnwidth,trim=30 30 30 30, clip]{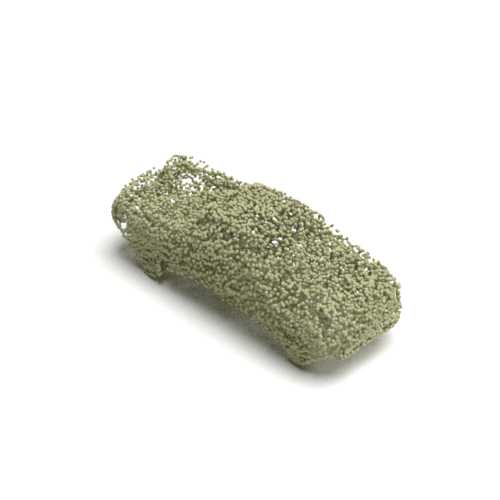}&
    \includegraphics[width=0.1\columnwidth,trim=30 30 30 30, clip]{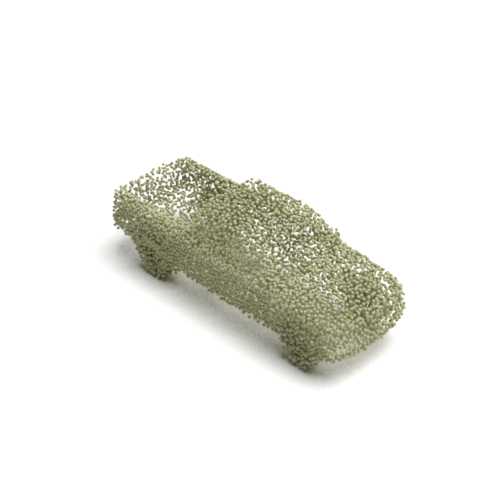}&
    \includegraphics[width=0.1\columnwidth,trim=30 30 30 30, clip]{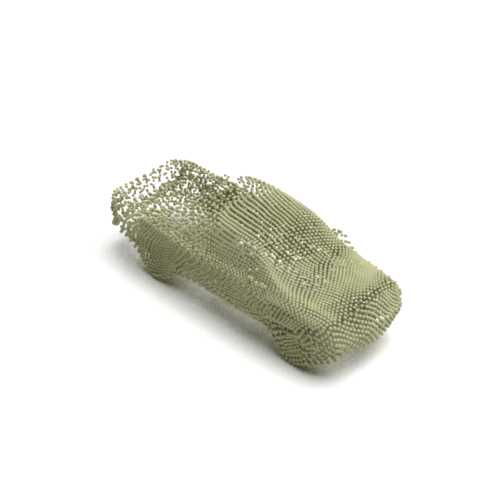}&
    \includegraphics[width=0.1\columnwidth,trim=30 30 30 30, clip]{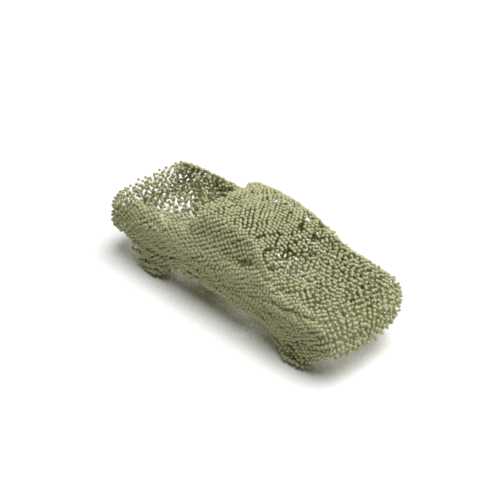}&
    \includegraphics[width=0.1\columnwidth,trim=30 30 30 30, clip]{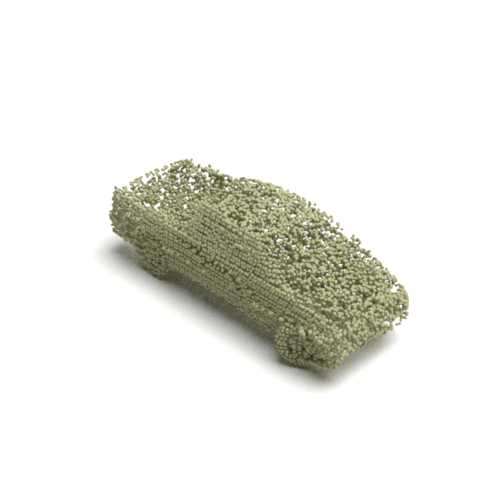}&
    \includegraphics[width=0.1\columnwidth,trim=30 30 30 30, clip]{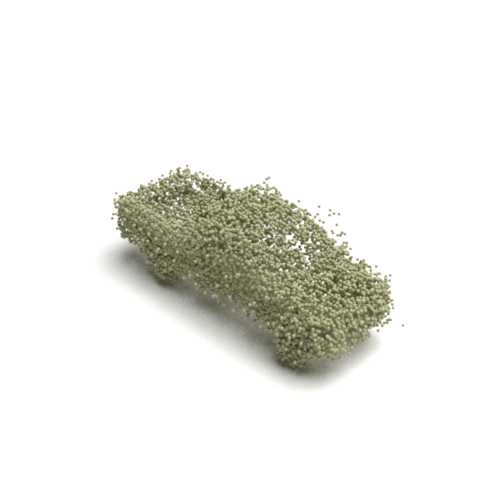}&
    \includegraphics[width=0.1\columnwidth,trim=30 30 30 30, clip]{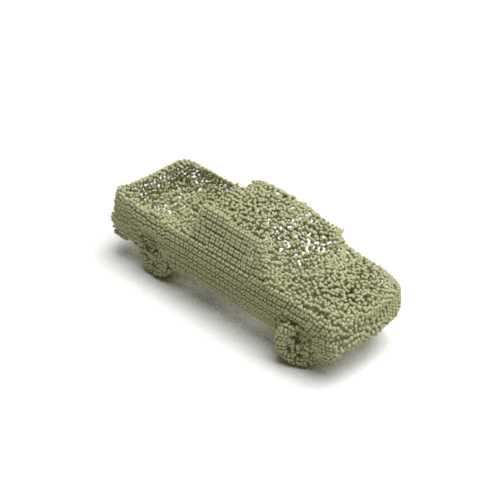}&
    \includegraphics[width=0.1\columnwidth,trim=30 30 30 30, clip]{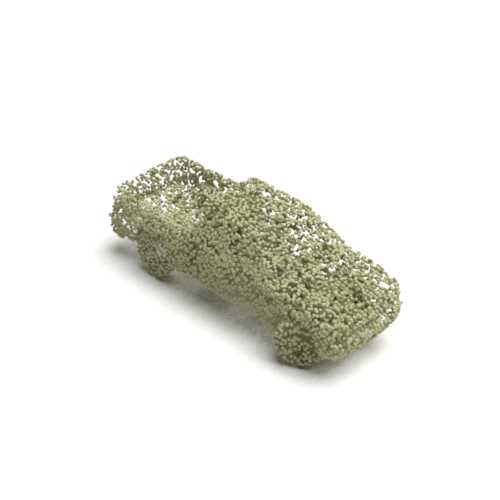}\\
    
\end{tabular}
}
\caption{Visualized completion comparison on ShapeNet.}
\label{fig:qualitative1}
\end{figure*}

\begin{figure*}[t]
\center
\setlength\tabcolsep{0pt}
{
\renewcommand{\arraystretch}{0.0}
\small
\begin{tabular}{@{}rcccccccccc@{}}
    & Input & AtlasNet\cite{atlasnet2018} & FCAE & FoldingNet\cite{foldingnet_2018_CVPR} &PCN\cite{Yuan-2018-pcn} 
    & MSN \cite{liu2019morphing} & GRNet \cite{xie2020grnet} & \emph{Ours} & Groundtruth\\
    
    \raisebox{0.6\height}{\rotatebox{90}{Chair}}&
    \includegraphics[width=0.1\columnwidth,trim=30 30 30 30, clip]{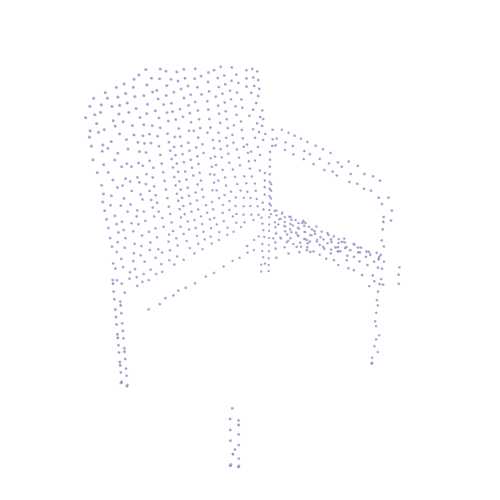}&   \includegraphics[width=0.1\columnwidth,trim=30 30 30 30, clip]{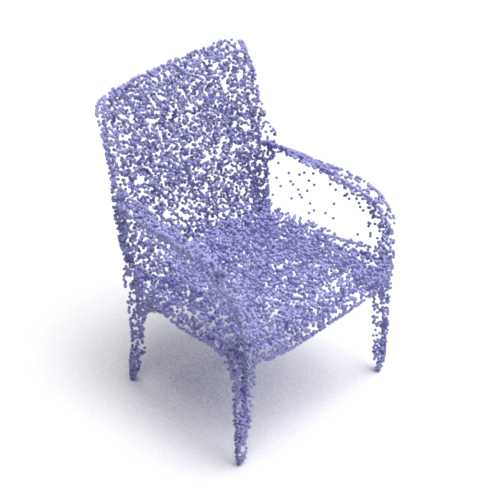}&
    \includegraphics[width=0.1\columnwidth,trim=30 30 30 30, clip]{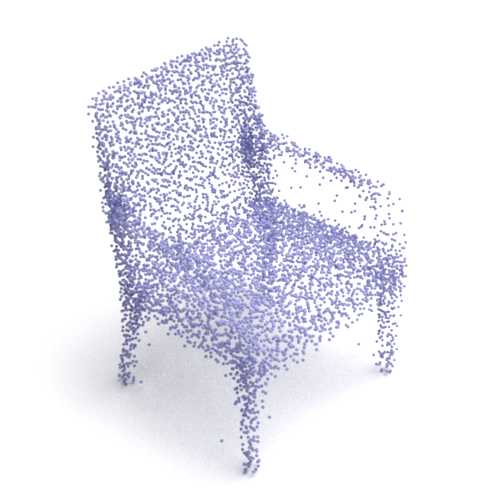}&
    \includegraphics[width=0.1\columnwidth,trim=30 30 30 30, clip]{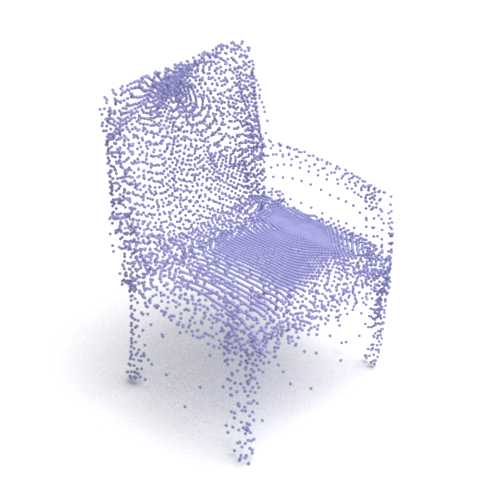}&
    \includegraphics[width=0.1\columnwidth,trim=30 30 30 30, clip]{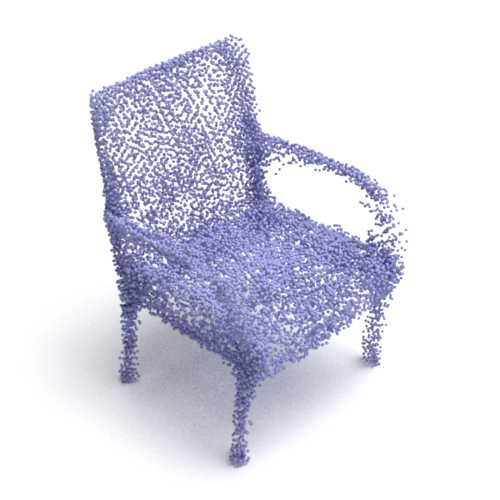}&
    \includegraphics[width=0.1\columnwidth,trim=30 30 30 30, clip]{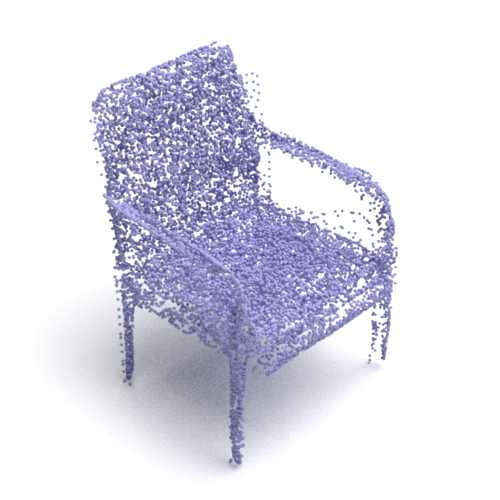}&
    \includegraphics[width=0.1\columnwidth,trim=30 30 30 30, clip]{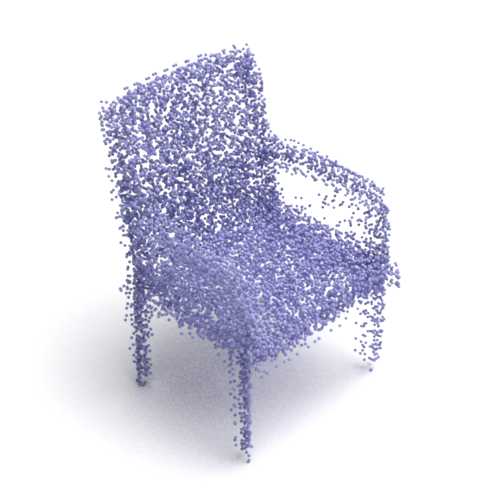}&
    \includegraphics[width=0.1\columnwidth,trim=30 30 30 30, clip]{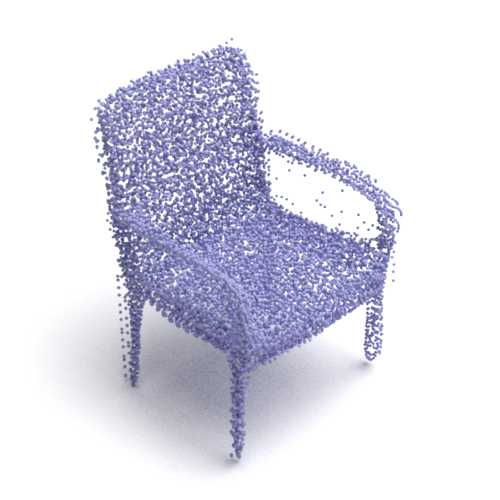}&
    \includegraphics[width=0.1\columnwidth,trim=30 30 30 30, clip]{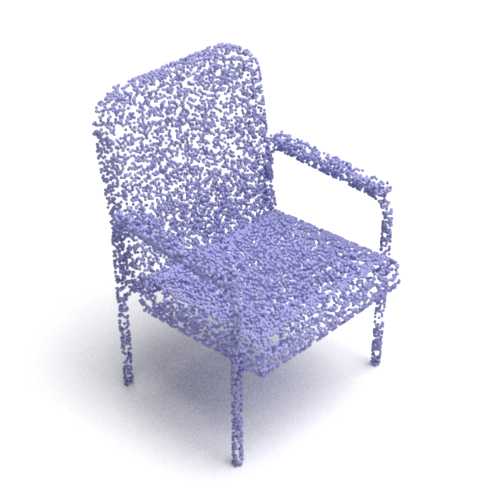}\\
    
    \raisebox{1.0\height}{\rotatebox{90}{Chair}}&
    \includegraphics[width=0.1\columnwidth,trim=30 30 30 30, clip]{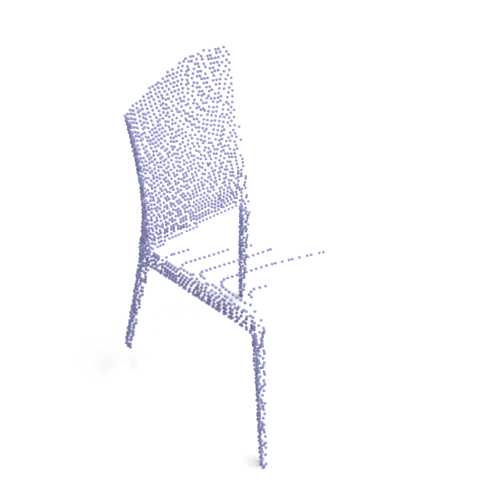}&
    \includegraphics[width=0.1\columnwidth,trim=30 30 30 30, clip]{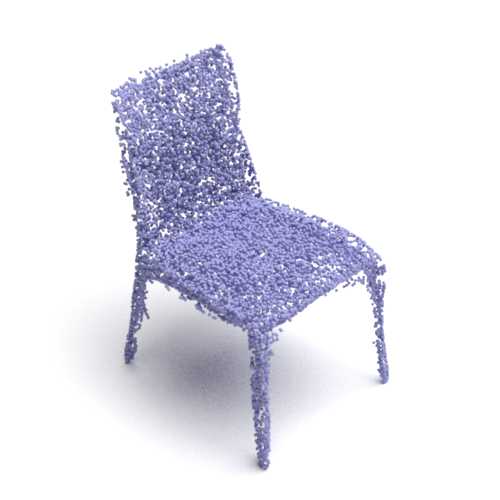}&
    \includegraphics[width=0.1\columnwidth,trim=30 30 30 30, clip]{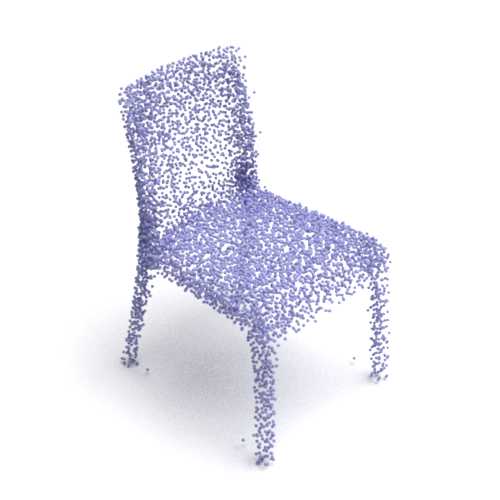}&
    \includegraphics[width=0.1\columnwidth,trim=30 30 30 30, clip]{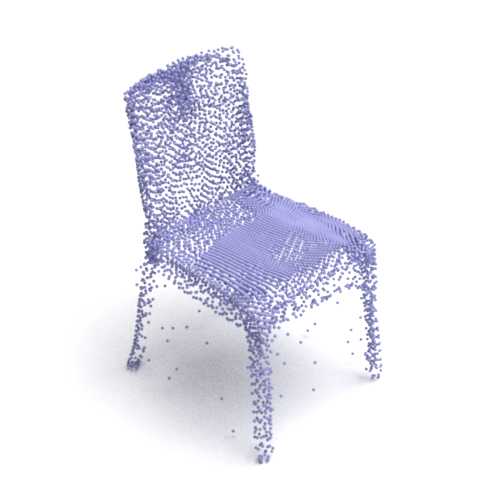}&
    \includegraphics[width=0.1\columnwidth,trim=30 30 30 30, clip]{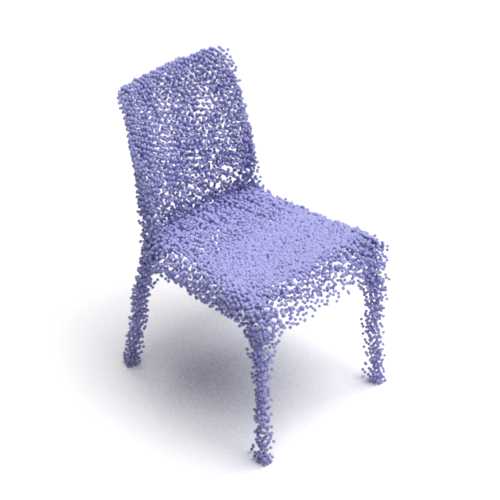}&
    \includegraphics[width=0.1\columnwidth,trim=30 30 30 30, clip]{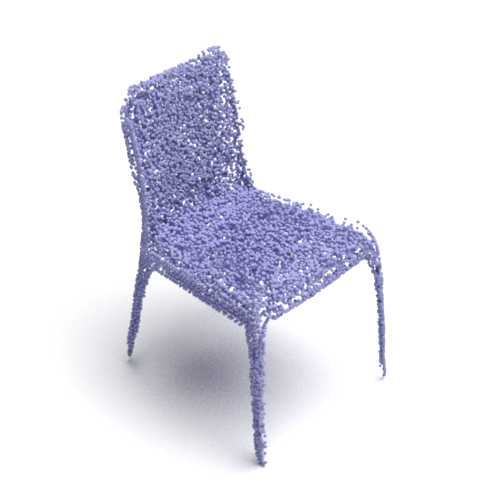}&
    \includegraphics[width=0.1\columnwidth,trim=30 30 30 30, clip]{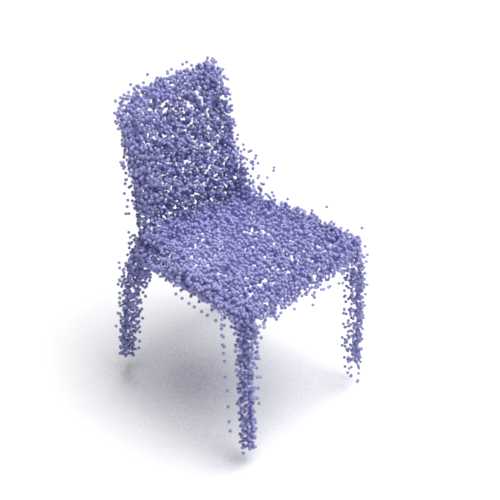}&
    \includegraphics[width=0.1\columnwidth,trim=30 30 30 30, clip]{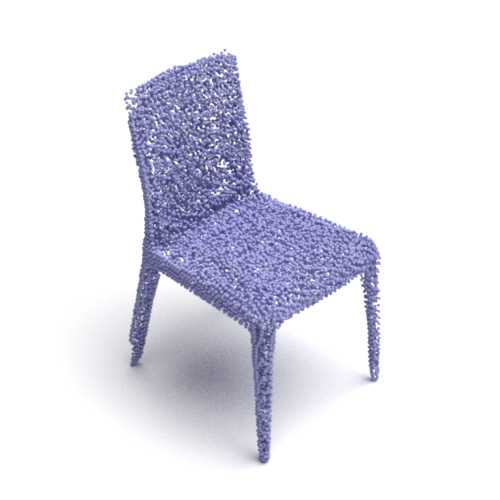}&
    \includegraphics[width=0.1\columnwidth,trim=30 30 30 30, clip]{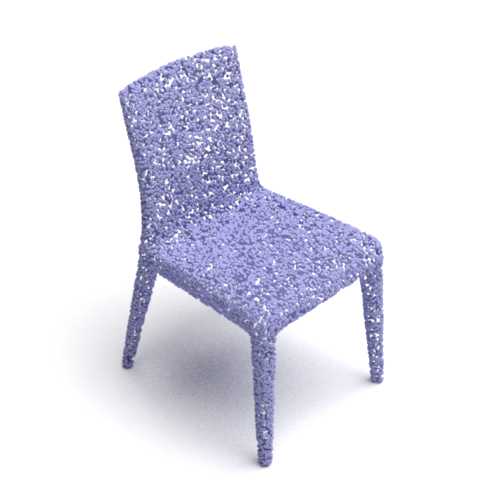}\\
    
    \raisebox{1.0\height}{\rotatebox{90}{Chair}}&
    \includegraphics[width=0.1\columnwidth,trim=30 30 30 30, clip]{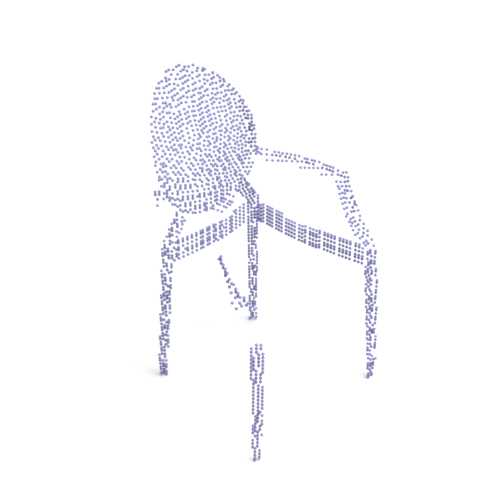}&
    \includegraphics[width=0.1\columnwidth,trim=30 30 30 30, clip]{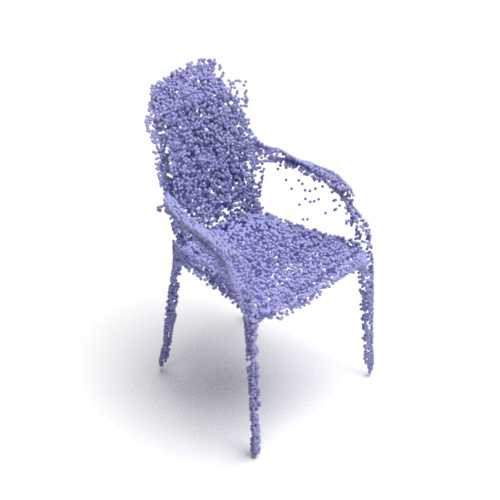}&
    \includegraphics[width=0.1\columnwidth,trim=30 30 30 30, clip]{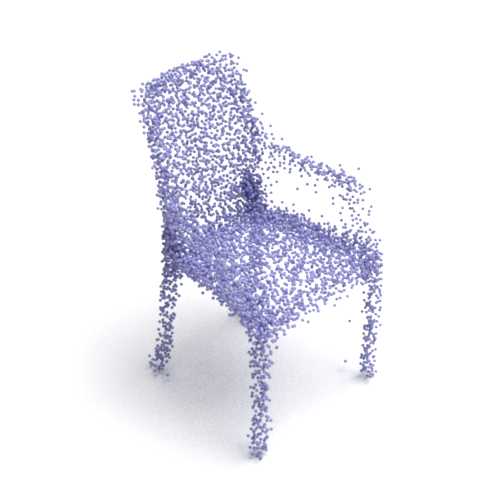}&
    \includegraphics[width=0.1\columnwidth,trim=30 30 30 30, clip]{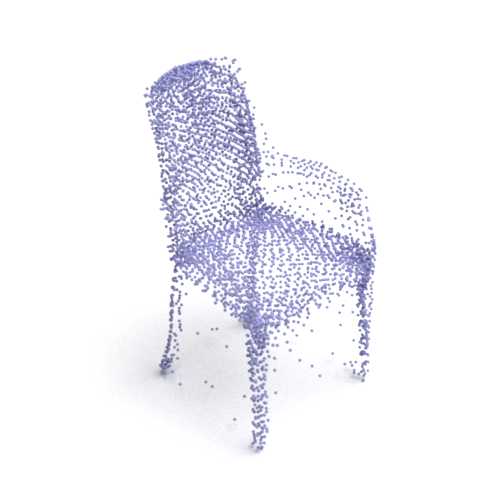}&
    \includegraphics[width=0.1\columnwidth,trim=30 30 30 30, clip]{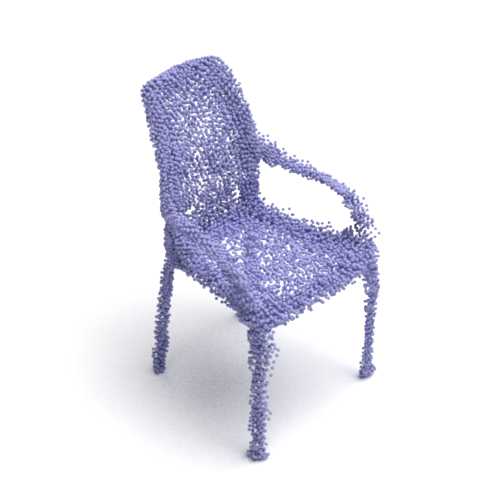}&
    \includegraphics[width=0.1\columnwidth,trim=30 30 30 30, clip]{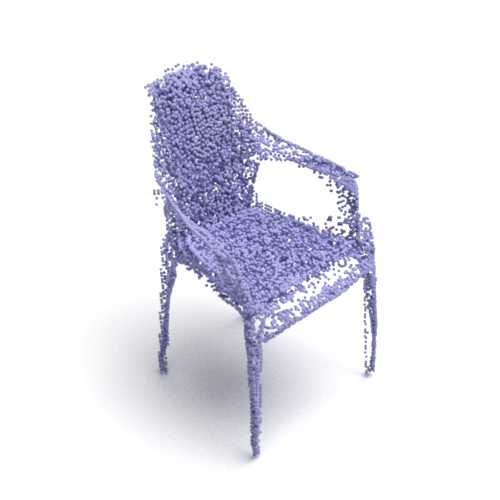}&
    \includegraphics[width=0.1\columnwidth,trim=30 30 30 30, clip]{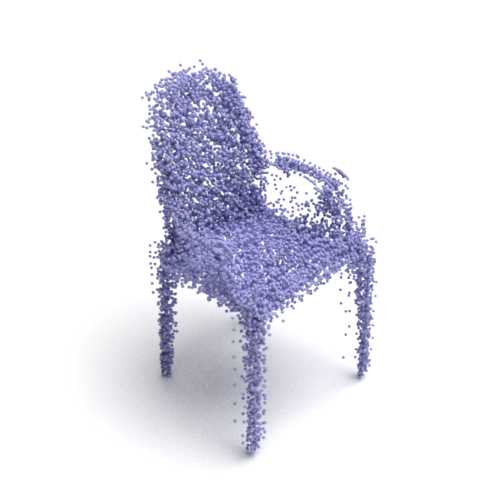}&
    \includegraphics[width=0.1\columnwidth,trim=30 30 30 30, clip]{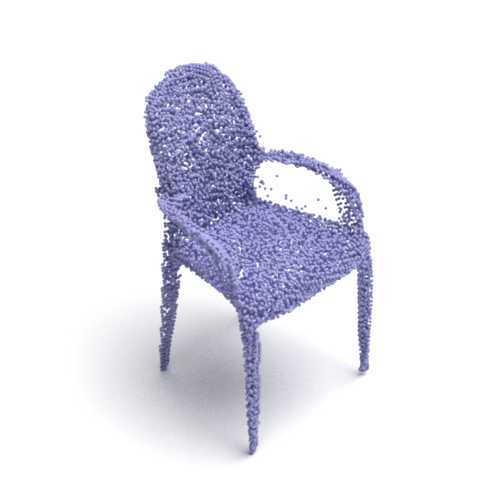}&
    \includegraphics[width=0.1\columnwidth,trim=30 30 30 30, clip]{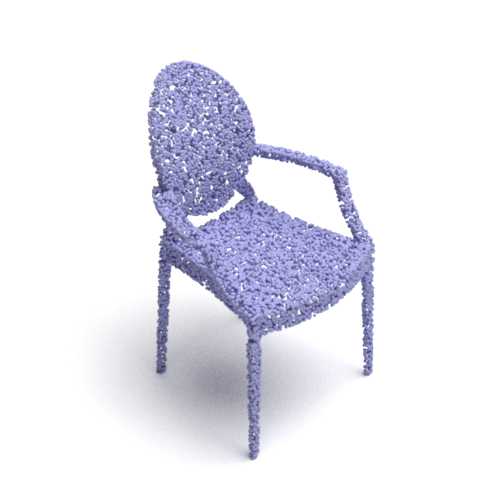}\\
    
    \raisebox{1\height}{\rotatebox{90}{Chair}}&
    \includegraphics[width=0.1\columnwidth,trim=30 30 30 30, clip]{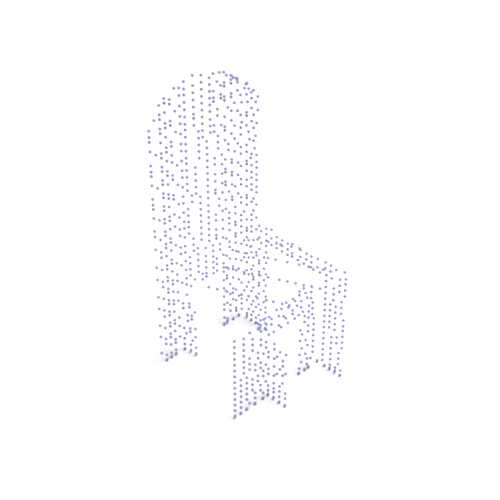}&
    \includegraphics[width=0.1\columnwidth,trim=30 30 30 30, clip]{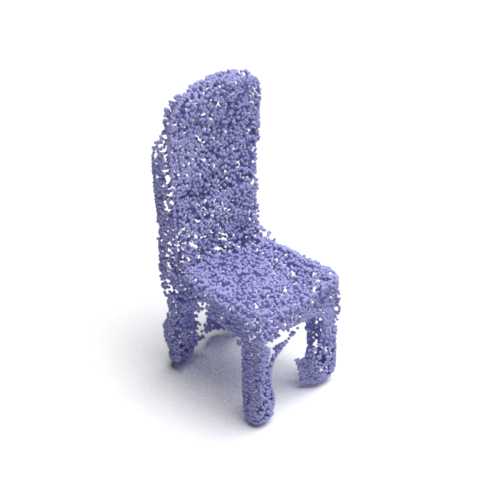}&
    \includegraphics[width=0.1\columnwidth,trim=30 30 30 30, clip]{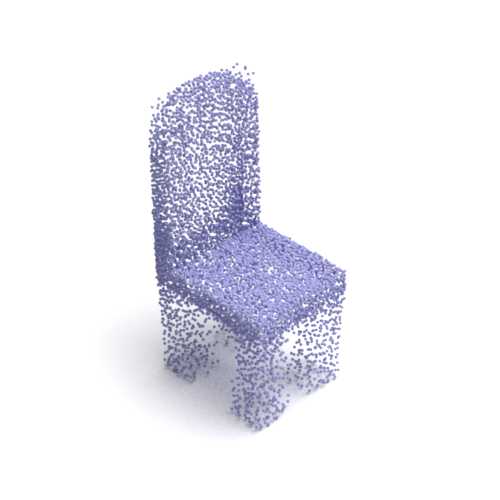}&
    \includegraphics[width=0.1\columnwidth,trim=30 30 30 30, clip]{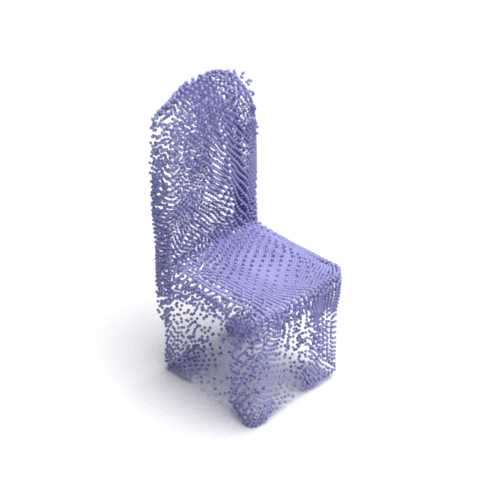}&
    \includegraphics[width=0.1\columnwidth,trim=30 30 30 30, clip]{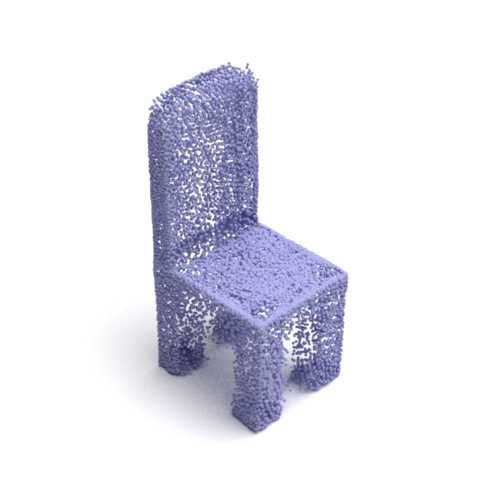}&
    \includegraphics[width=0.1\columnwidth,trim=30 30 30 30, clip]{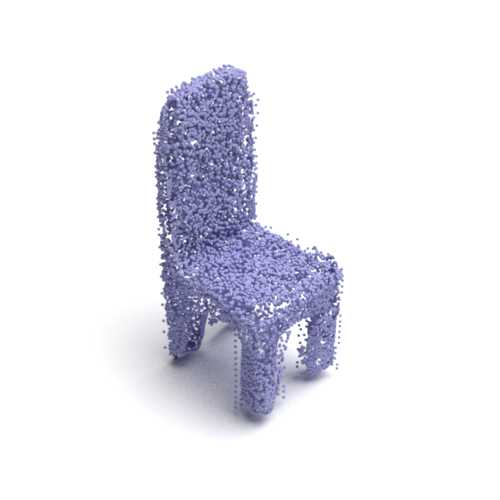}&
    \includegraphics[width=0.1\columnwidth,trim=30 30 30 30, clip]{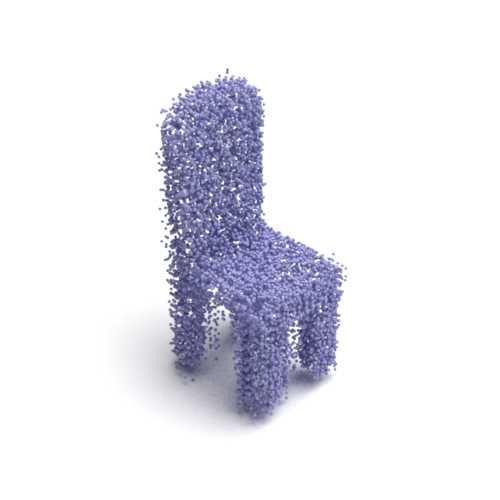}&
    \includegraphics[width=0.1\columnwidth,trim=30 30 30 30, clip]{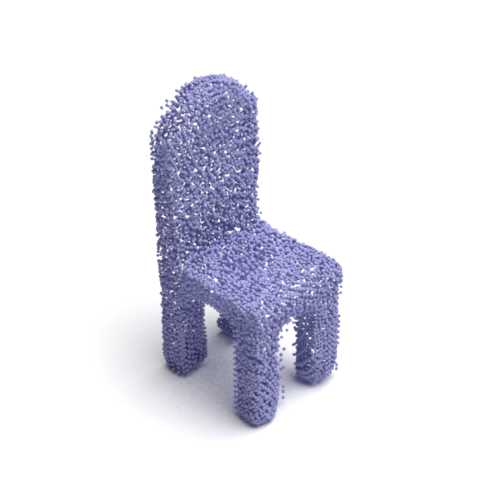}&
    \includegraphics[width=0.1\columnwidth,trim=30 30 30 30, clip]{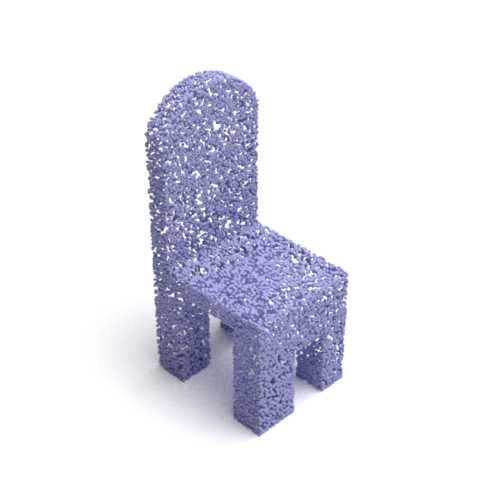}\\
    
    \raisebox{0.6\height}{\rotatebox{90}{Lamp}}&
    \includegraphics[width=0.1\columnwidth,trim=30 30 30 30, clip]{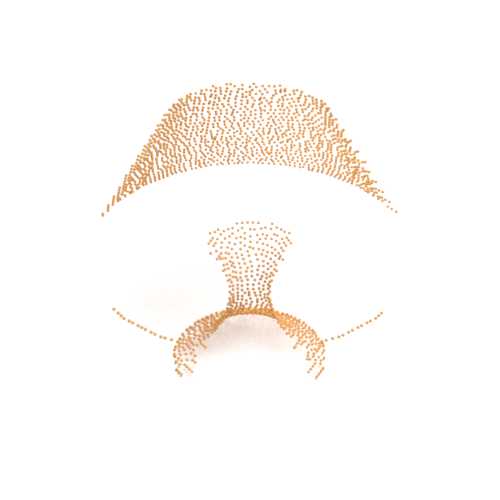}&
    \includegraphics[width=0.1\columnwidth,trim=30 30 30 30, clip]{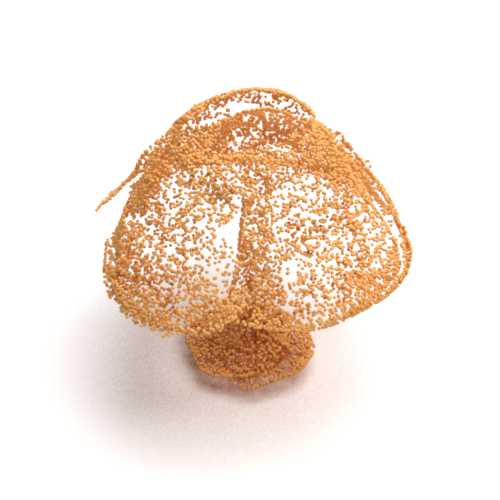}&
    \includegraphics[width=0.1\columnwidth,trim=30 30 30 30, clip]{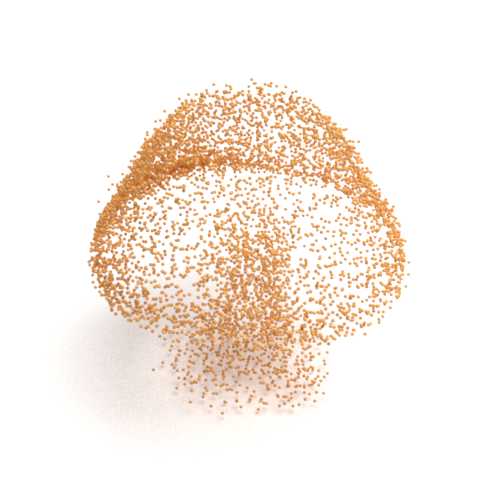}&
    \includegraphics[width=0.1\columnwidth,trim=30 30 30 30, clip]{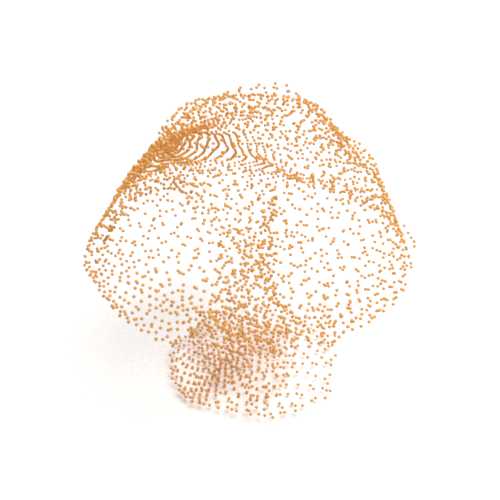}&
    \includegraphics[width=0.1\columnwidth,trim=30 30 30 30, clip]{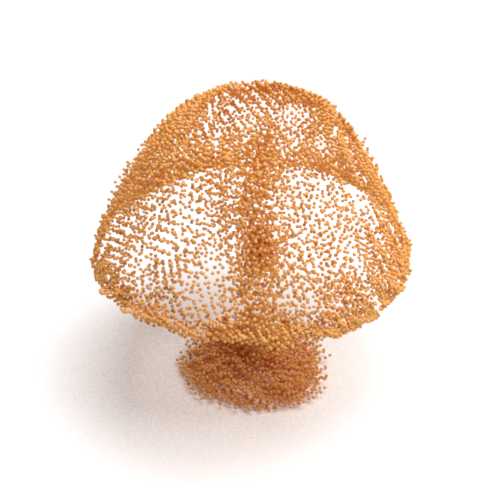}&
    \includegraphics[width=0.1\columnwidth,trim=30 30 30 30, clip]{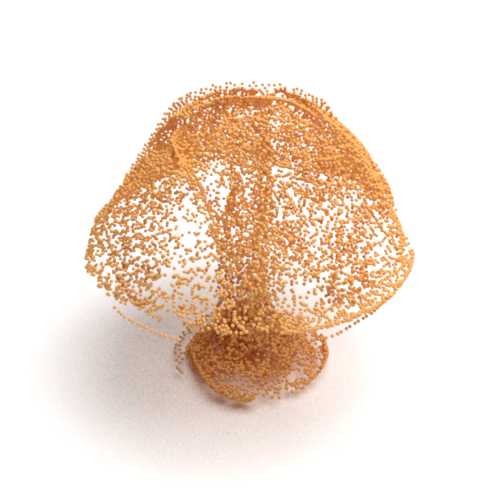}&
    \includegraphics[width=0.1\columnwidth,trim=30 30 30 30, clip]{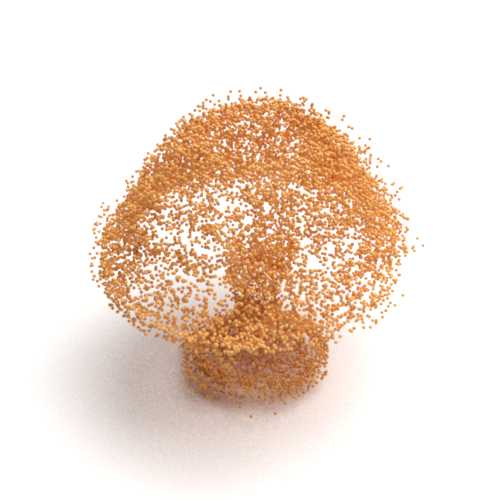}&
    \includegraphics[width=0.1\columnwidth,trim=30 30 30 30, clip]{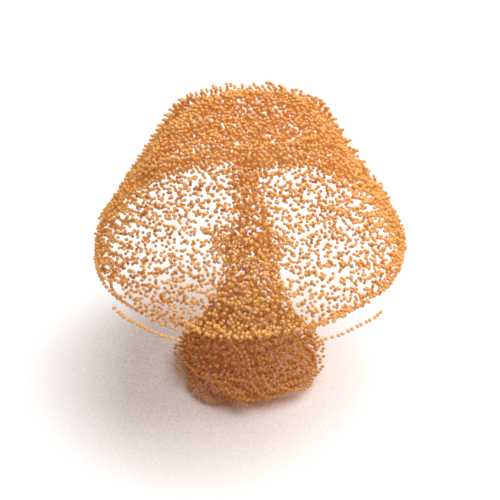}&
    \includegraphics[width=0.1\columnwidth,trim=30 30 30 30, clip]{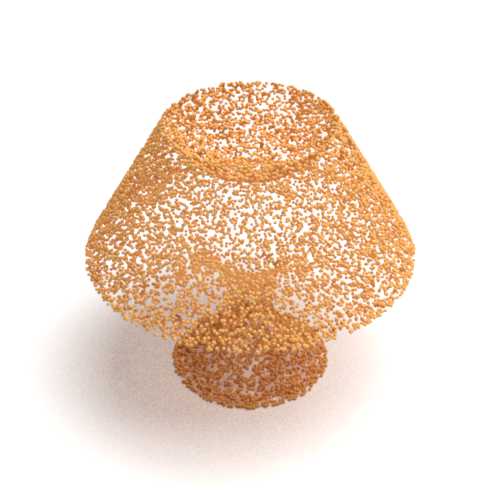}\\
    
    \raisebox{0.6\height}{\rotatebox{90}{Lamp}}&
    \includegraphics[width=0.1\columnwidth,trim=30 30 30 30, clip]{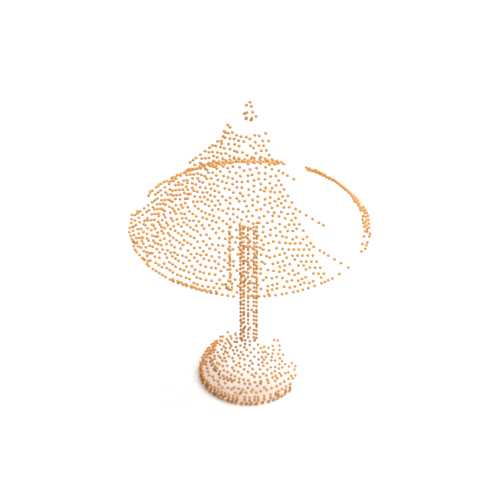}&
    \includegraphics[width=0.1\columnwidth,trim=30 30 30 30, clip]{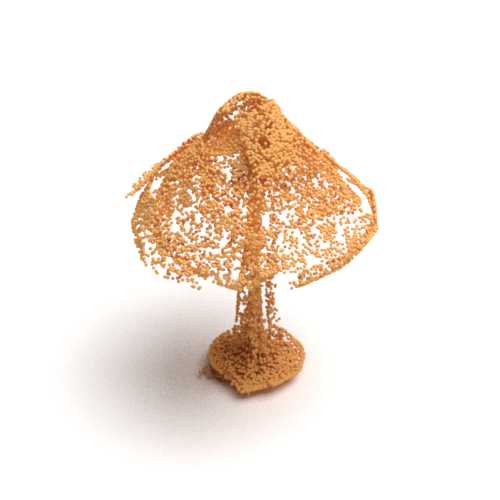}&
    \includegraphics[width=0.1\columnwidth,trim=30 30 30 30, clip]{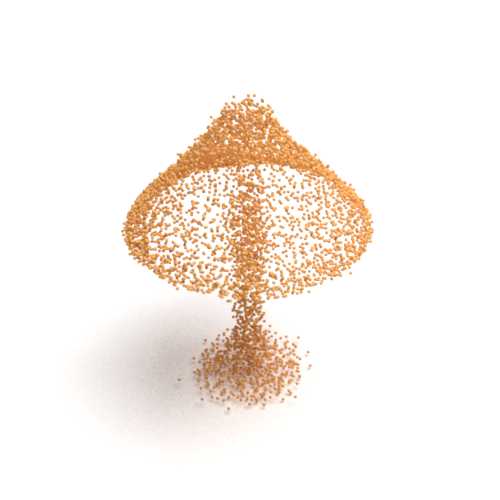}&
    \includegraphics[width=0.1\columnwidth,trim=30 30 30 30, clip]{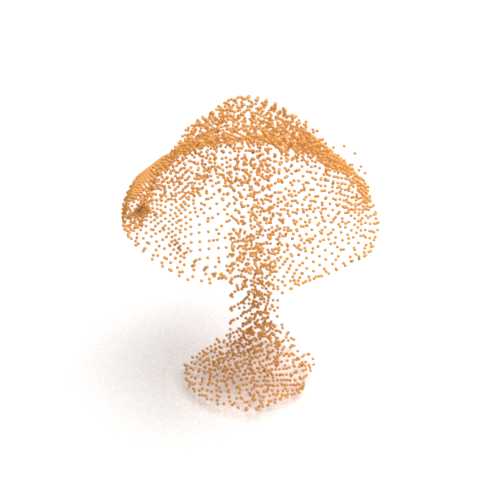}&
    \includegraphics[width=0.1\columnwidth,trim=30 30 30 30, clip]{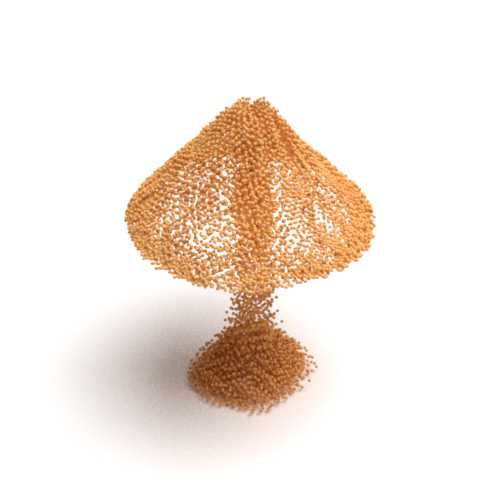}&
    \includegraphics[width=0.1\columnwidth,trim=30 30 30 30, clip]{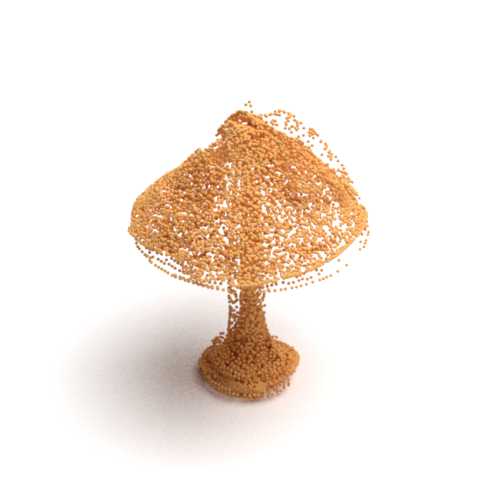}&
    \includegraphics[width=0.1\columnwidth,trim=30 30 30 30, clip]{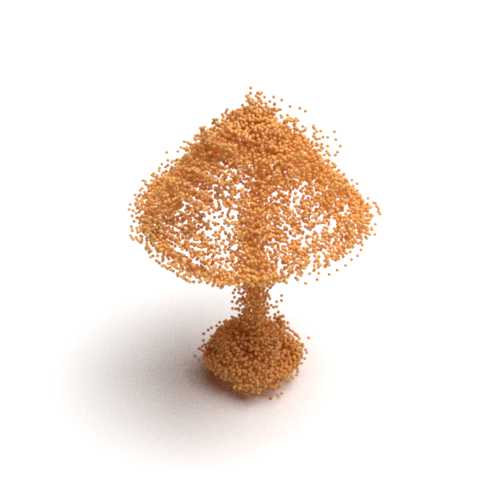}&
    \includegraphics[width=0.1\columnwidth,trim=30 30 30 30, clip]{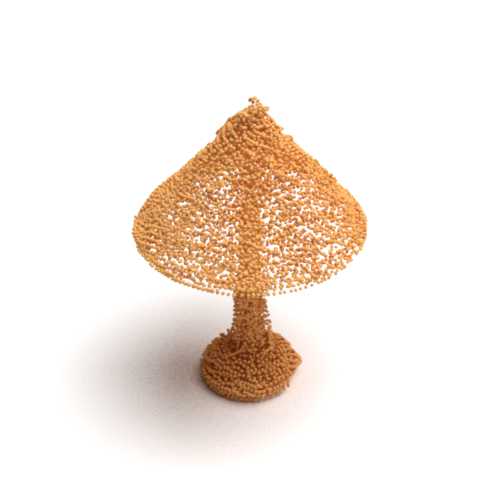}&
    \includegraphics[width=0.1\columnwidth,trim=30 30 30 30, clip]{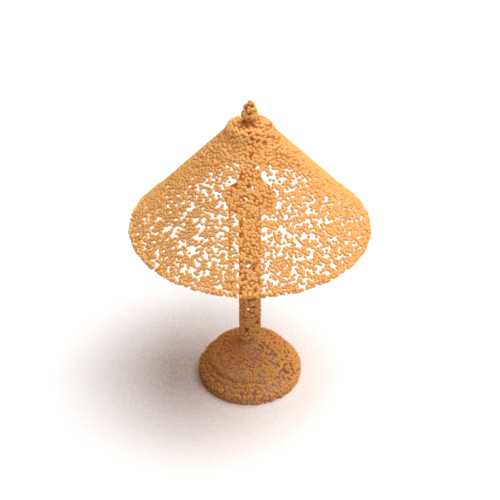}\\
    
    \raisebox{0.6\height}{\rotatebox{90}{Lamp}}&
    \includegraphics[width=0.1\columnwidth,trim=30 30 30 30, clip]{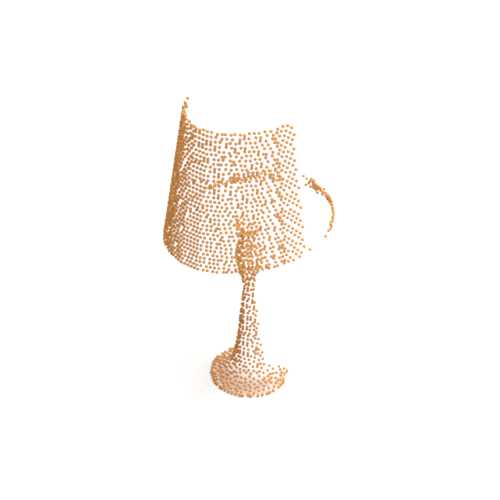}&   \includegraphics[width=0.1\columnwidth,trim=30 30 30 30, clip]{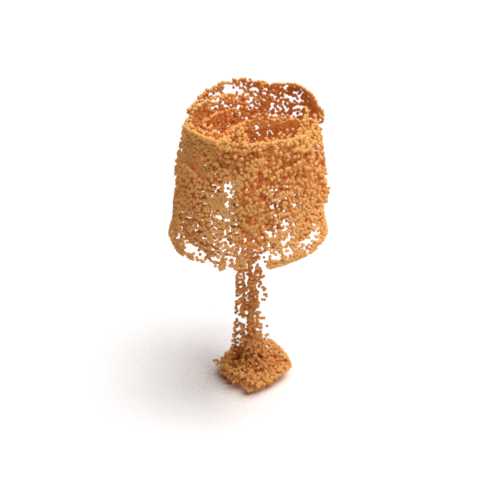}&
    \includegraphics[width=0.1\columnwidth,trim=30 30 30 30, clip]{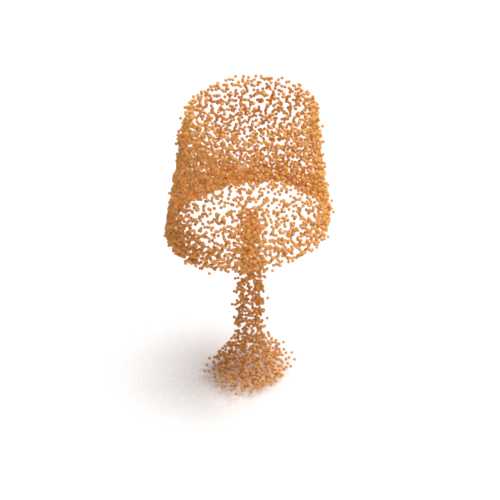}&
    \includegraphics[width=0.1\columnwidth,trim=30 30 30 30, clip]{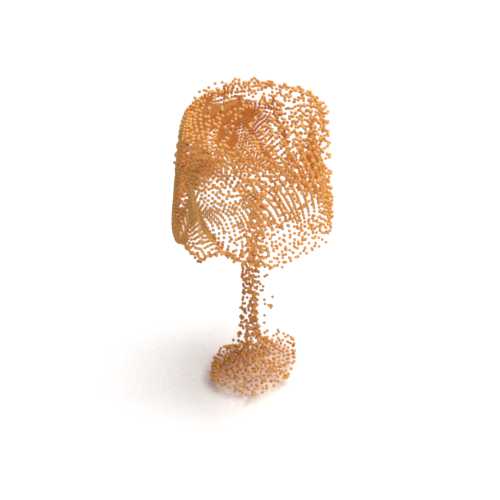}&
    \includegraphics[width=0.1\columnwidth,trim=30 30 30 30, clip]{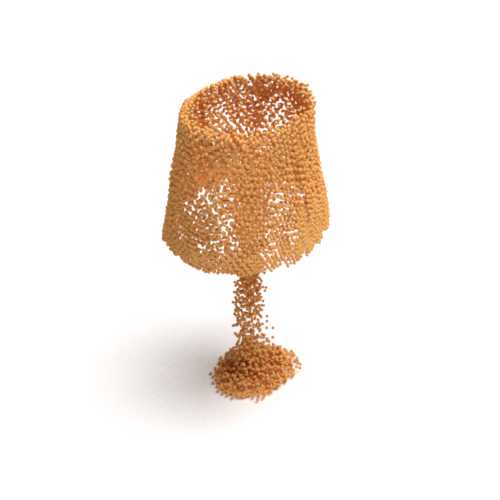}&
    \includegraphics[width=0.1\columnwidth,trim=30 30 30 30, clip]{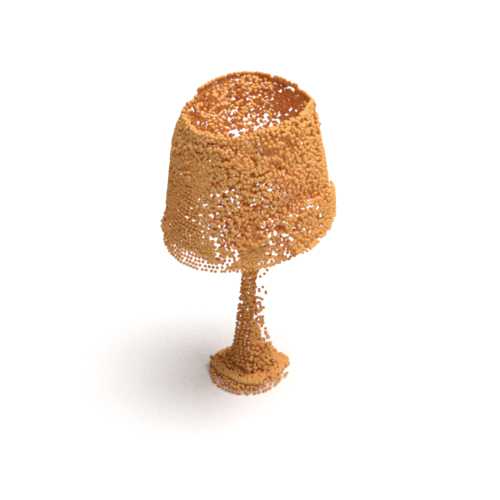}&
    \includegraphics[width=0.1\columnwidth,trim=30 30 30 30, clip]{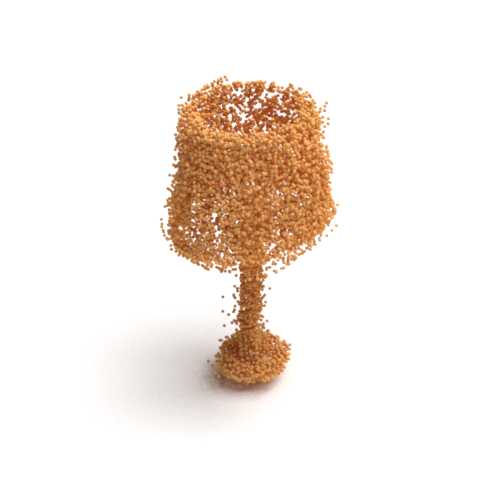}&
    \includegraphics[width=0.1\columnwidth,trim=30 30 30 30, clip]{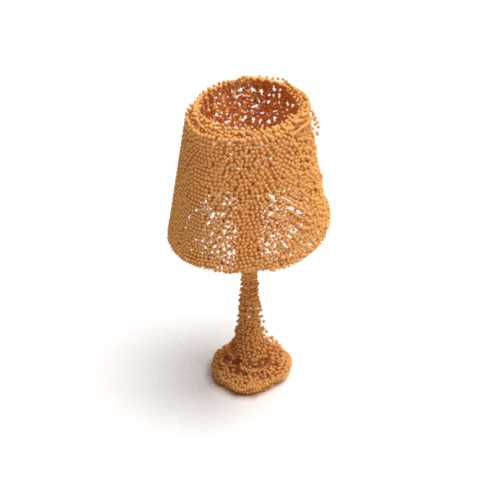}&
    \includegraphics[width=0.1\columnwidth,trim=30 30 30 30, clip]{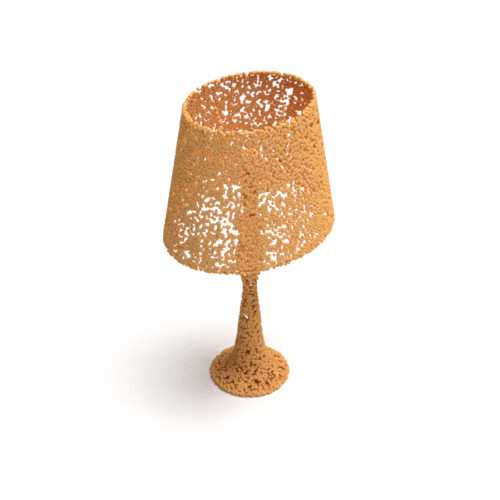}\\
    
    \raisebox{1.0\height}{\rotatebox{90}{Lamp}}&
    \includegraphics[width=0.1\columnwidth,trim=30 30 30 30, clip]{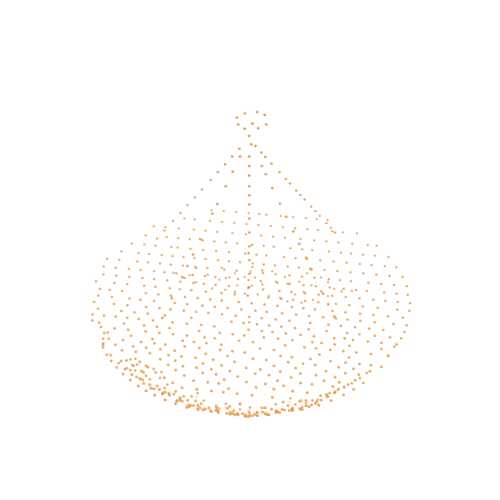}&
    \includegraphics[width=0.1\columnwidth,trim=30 30 30 30, clip]{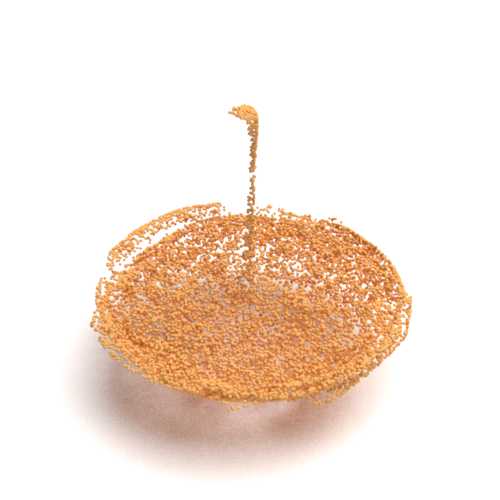}&
    \includegraphics[width=0.1\columnwidth,trim=30 30 30 30, clip]{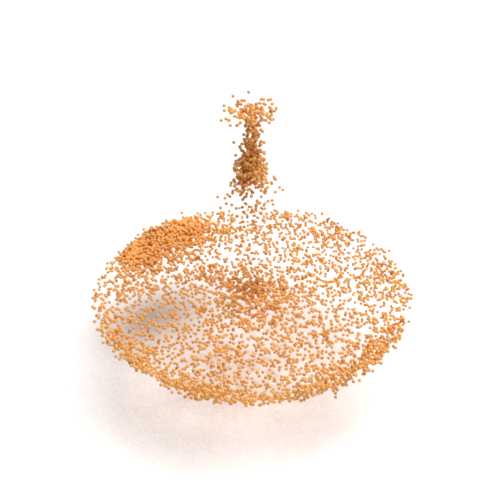}&
    \includegraphics[width=0.1\columnwidth,trim=30 30 30 30, clip]{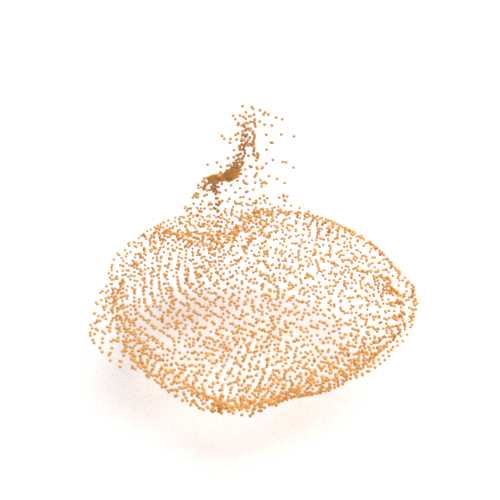}&
    \includegraphics[width=0.1\columnwidth,trim=30 30 30 30, clip]{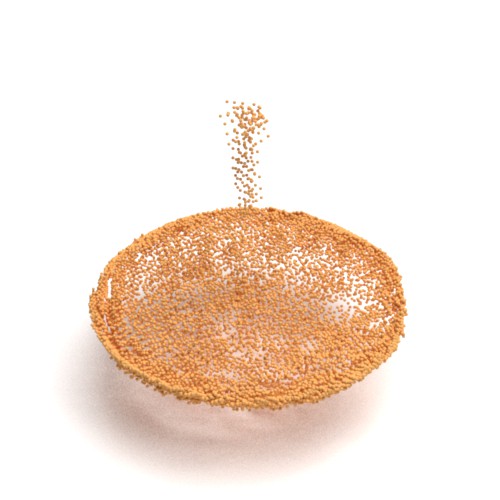}&
    \includegraphics[width=0.1\columnwidth,trim=30 30 30 30, clip]{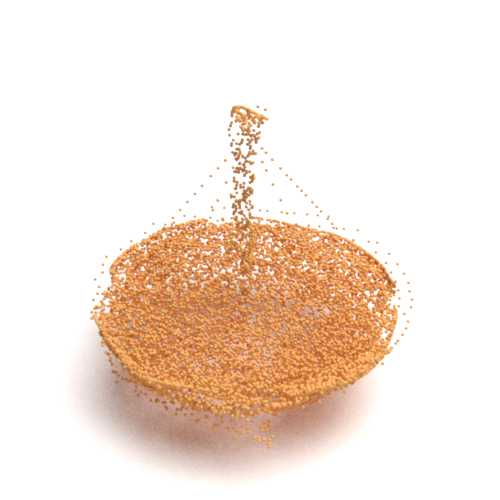}&
    \includegraphics[width=0.1\columnwidth,trim=30 30 30 30, clip]{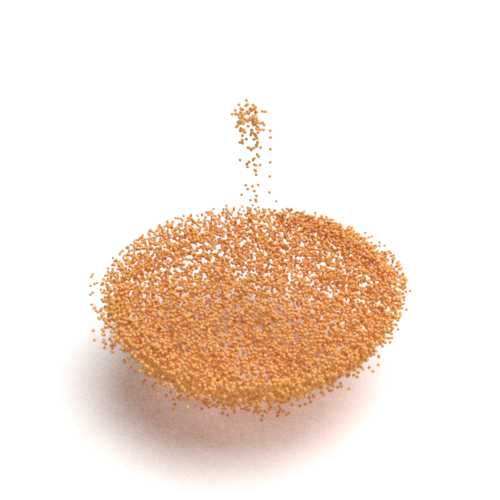}&
    \includegraphics[width=0.1\columnwidth,trim=30 30 30 30, clip]{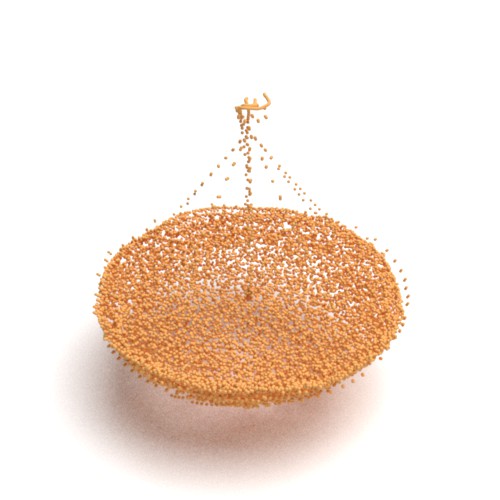}&
    \includegraphics[width=0.1\columnwidth,trim=30 30 30 30, clip]{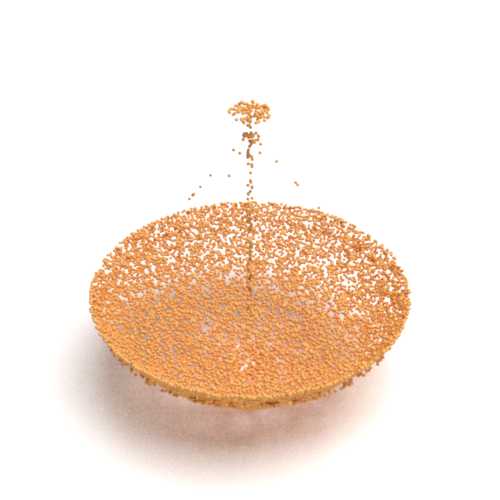}\\
    
    \raisebox{1.0\height}{\rotatebox{90}{Sofa}}&
    \includegraphics[width=0.1\columnwidth,trim=30 30 30 30, clip]{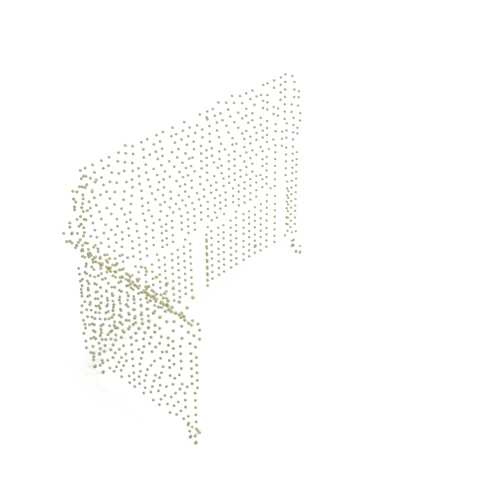}&
    \includegraphics[width=0.1\columnwidth,trim=30 30 30 30, clip]{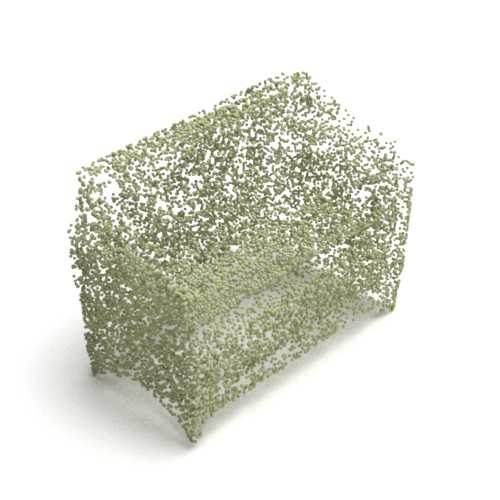}&
    \includegraphics[width=0.1\columnwidth,trim=30 30 30 30, clip]{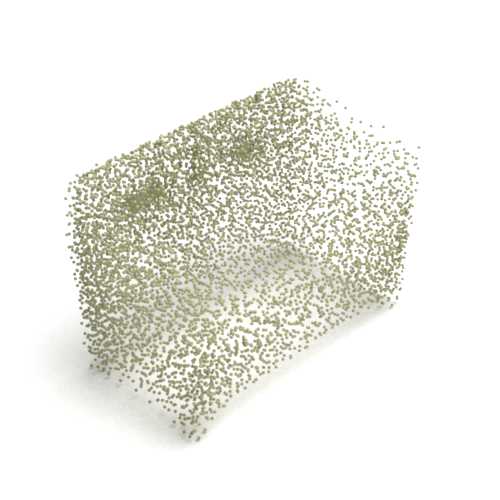}&
    \includegraphics[width=0.1\columnwidth,trim=30 30 30 30, clip]{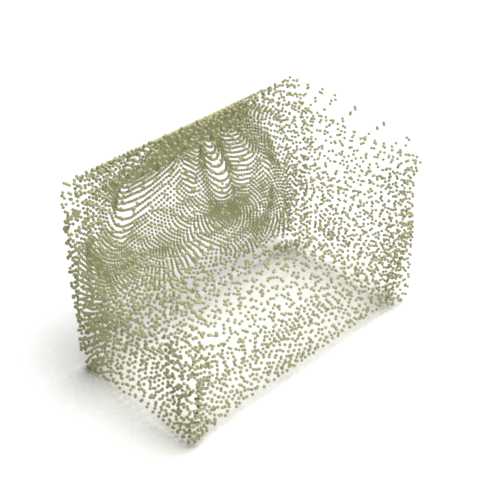}&
    \includegraphics[width=0.1\columnwidth,trim=30 30 30 30, clip]{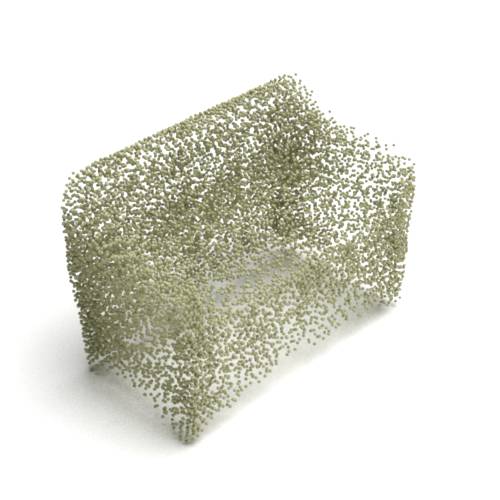}&
    \includegraphics[width=0.1\columnwidth,trim=30 30 30 30, clip]{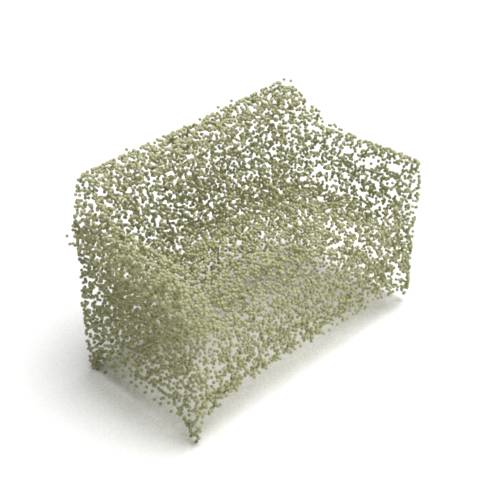}&
    \includegraphics[width=0.1\columnwidth,trim=30 30 30 30, clip]{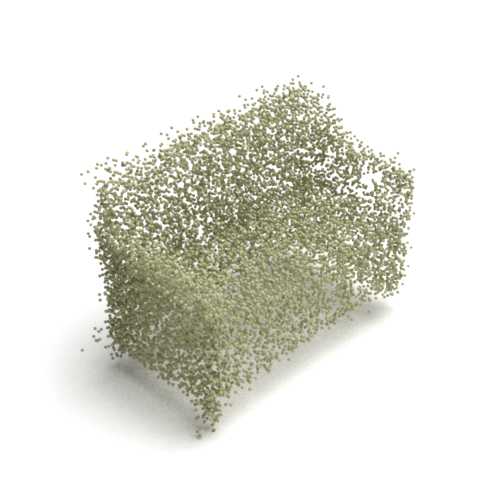}&
    \includegraphics[width=0.1\columnwidth,trim=30 30 30 30, clip]{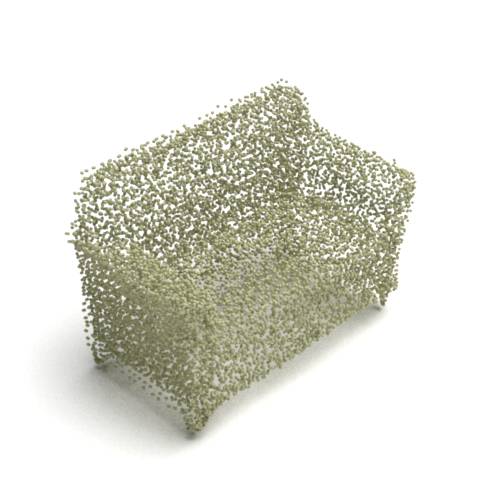}&
    \includegraphics[width=0.1\columnwidth,trim=30 30 30 30, clip]{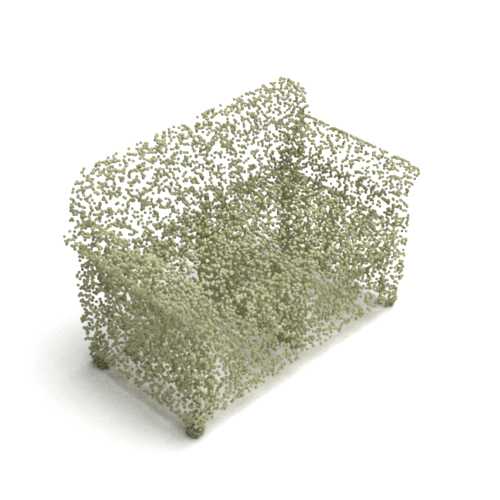}\\
    
    \raisebox{1\height}{\rotatebox{90}{Sofa}}&
    \includegraphics[width=0.1\columnwidth,trim=30 30 30 30, clip]{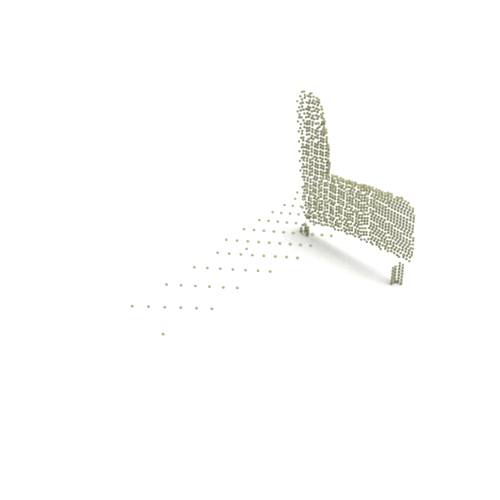}&
    \includegraphics[width=0.1\columnwidth,trim=30 30 30 30, clip]{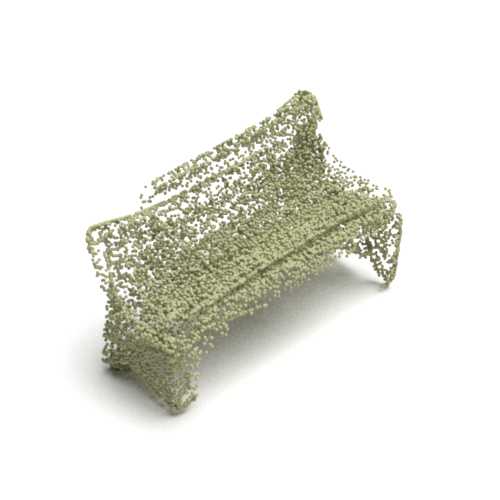}&
    \includegraphics[width=0.1\columnwidth,trim=30 30 30 30, clip]{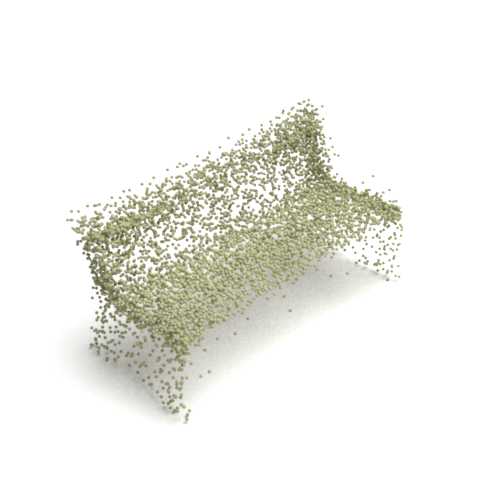}&
    \includegraphics[width=0.1\columnwidth,trim=30 30 30 30, clip]{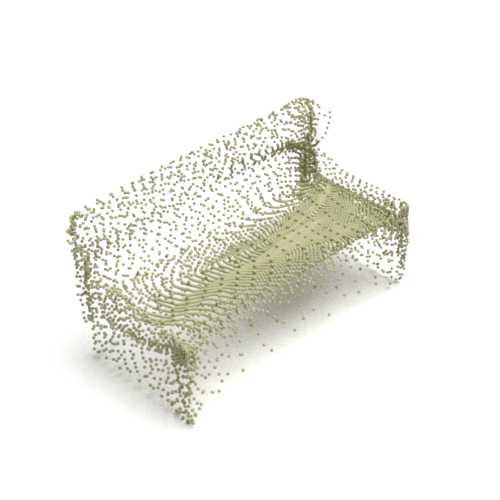}&
    \includegraphics[width=0.1\columnwidth,trim=30 30 30 30, clip]{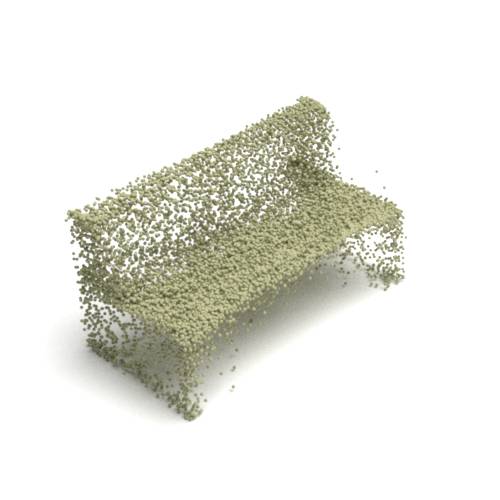}&
    \includegraphics[width=0.1\columnwidth,trim=30 30 30 30, clip]{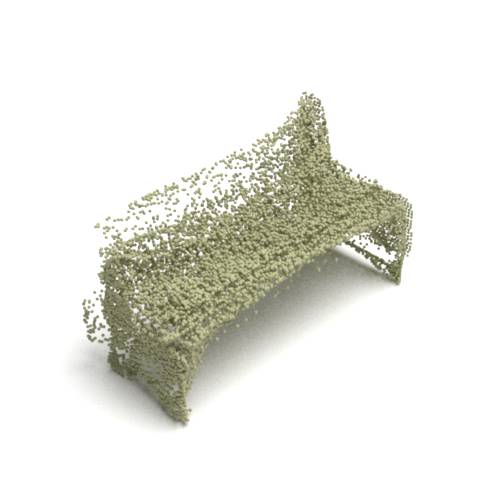}&
    \includegraphics[width=0.1\columnwidth,trim=30 30 30 30, clip]{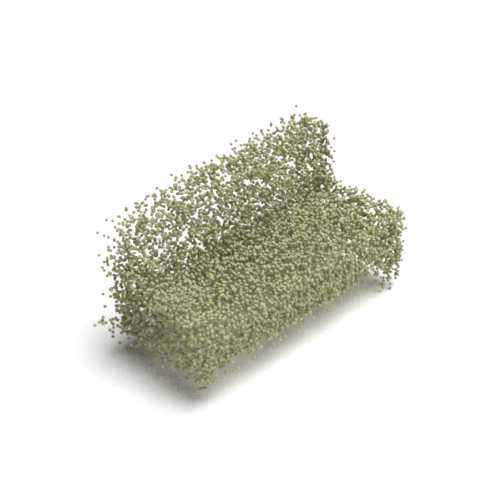}&
    \includegraphics[width=0.1\columnwidth,trim=30 30 30 30, clip]{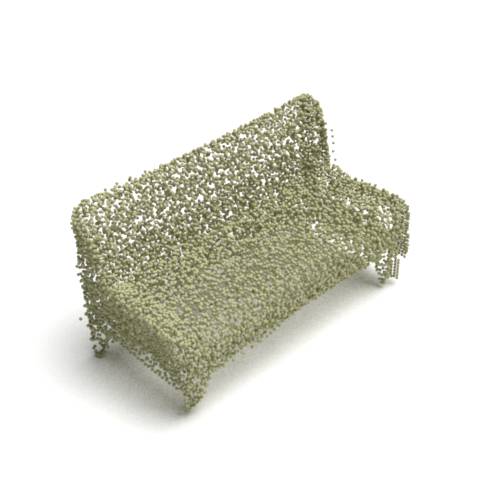}&
    \includegraphics[width=0.1\columnwidth,trim=30 30 30 30, clip]{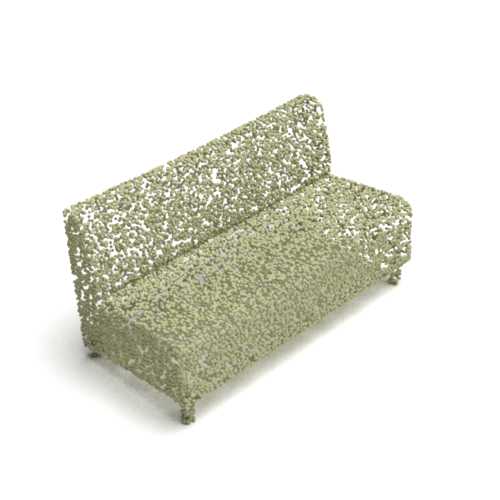}\\
    
    \raisebox{0.6\height}{\rotatebox{90}{Sofa}}&
    \includegraphics[width=0.1\columnwidth,trim=30 30 30 30, clip]{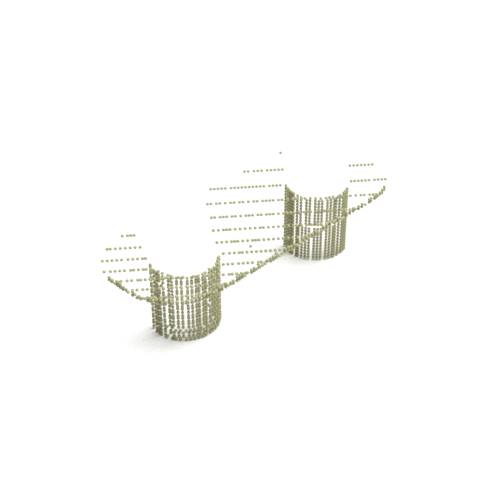}&
    \includegraphics[width=0.1\columnwidth,trim=30 30 30 30, clip]{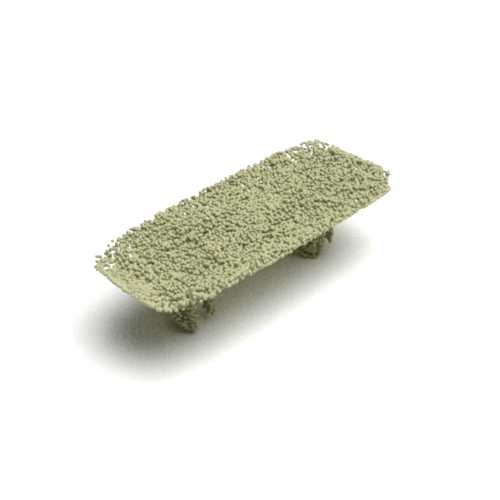}&
    \includegraphics[width=0.1\columnwidth,trim=30 30 30 30, clip]{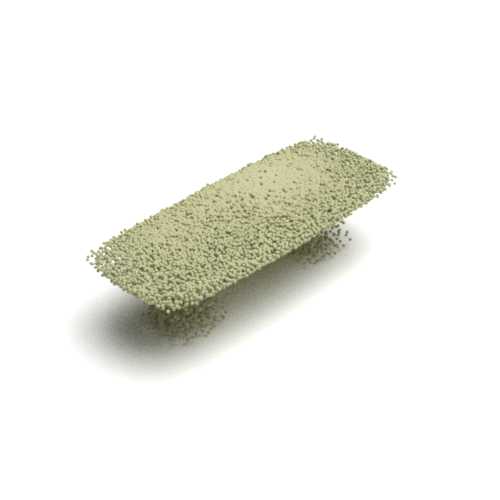}&
    \includegraphics[width=0.1\columnwidth,trim=30 30 30 30, clip]{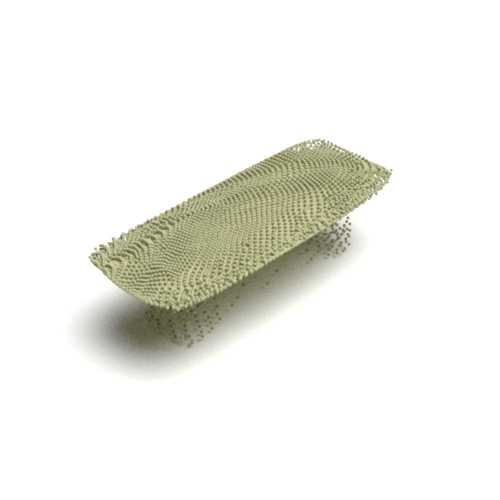}&
    \includegraphics[width=0.1\columnwidth,trim=30 30 30 30, clip]{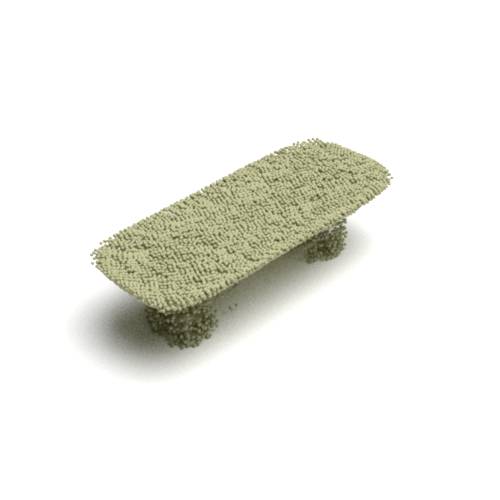}&
    \includegraphics[width=0.1\columnwidth,trim=30 30 30 30, clip]{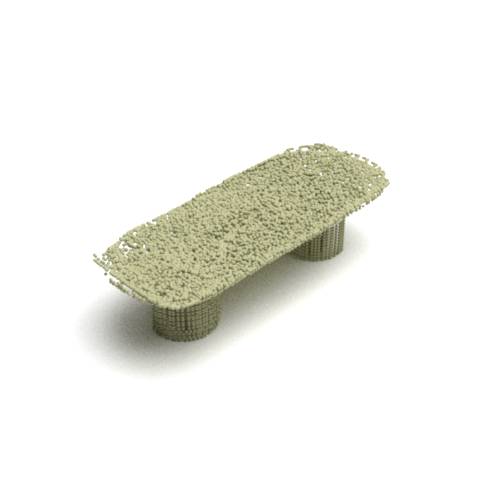}&
    \includegraphics[width=0.1\columnwidth,trim=30 30 30 30, clip]{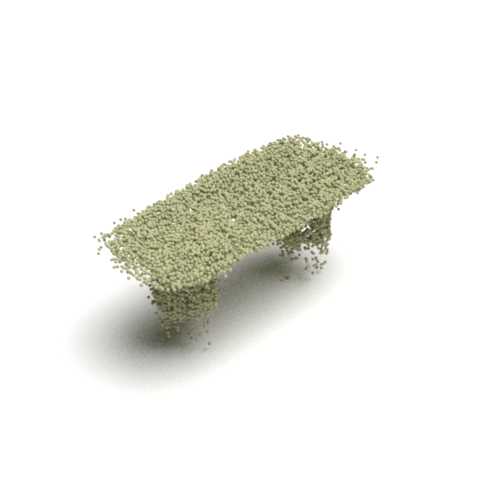}&
    \includegraphics[width=0.1\columnwidth,trim=30 30 30 30, clip]{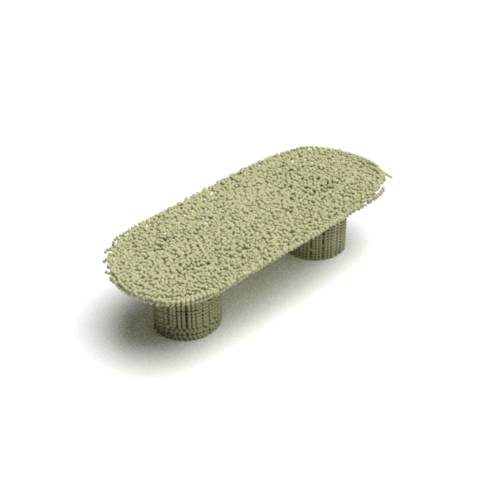}&
    \includegraphics[width=0.1\columnwidth,trim=30 30 30 30, clip]{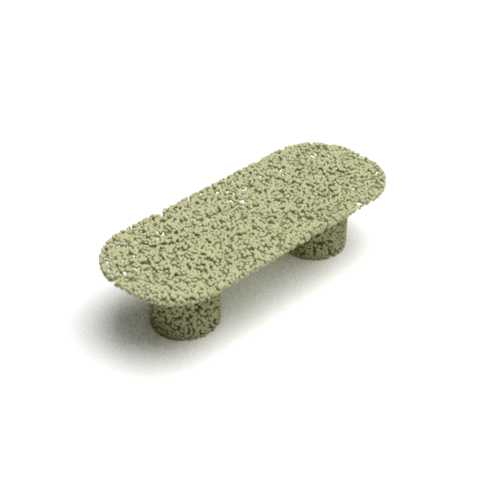}\\
    
    \raisebox{0.6\height}{\rotatebox{90}{Sofa}}&
    \includegraphics[width=0.1\columnwidth,trim=30 30 30 30, clip]{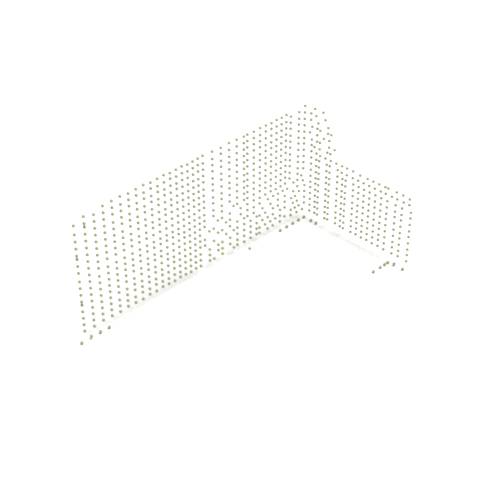}&
    \includegraphics[width=0.1\columnwidth,trim=30 30 30 30, clip]{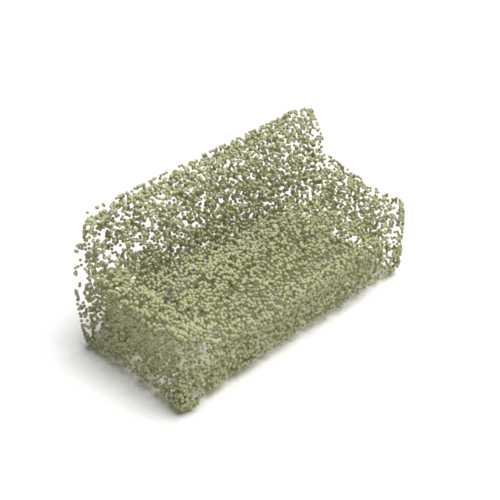}&
    \includegraphics[width=0.1\columnwidth,trim=30 30 30 30, clip]{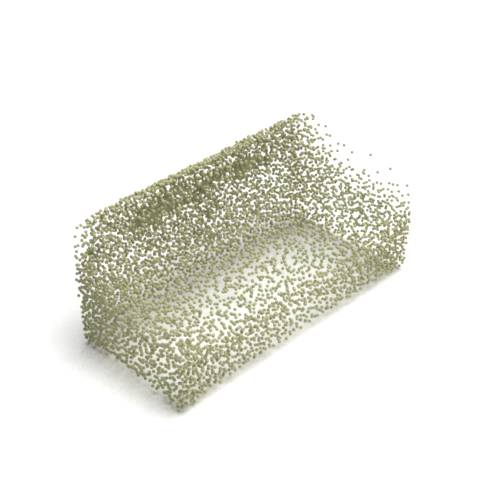}&
    \includegraphics[width=0.1\columnwidth,trim=30 30 30 30, clip]{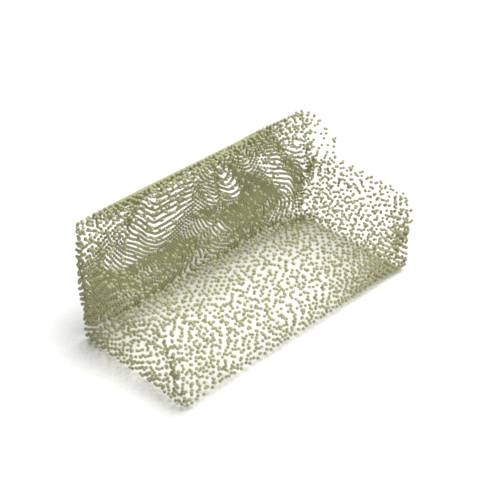}&
    \includegraphics[width=0.1\columnwidth,trim=30 30 30 30, clip]{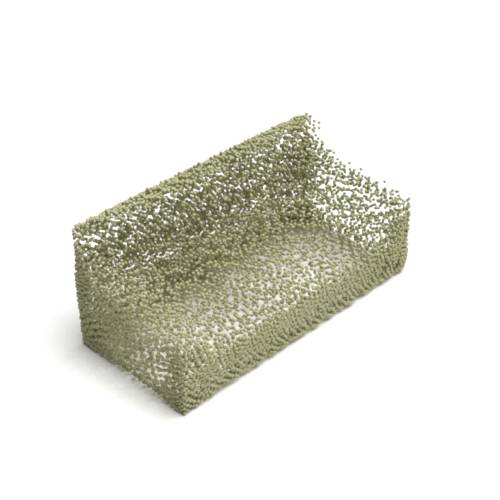}&
    \includegraphics[width=0.1\columnwidth,trim=30 30 30 30, clip]{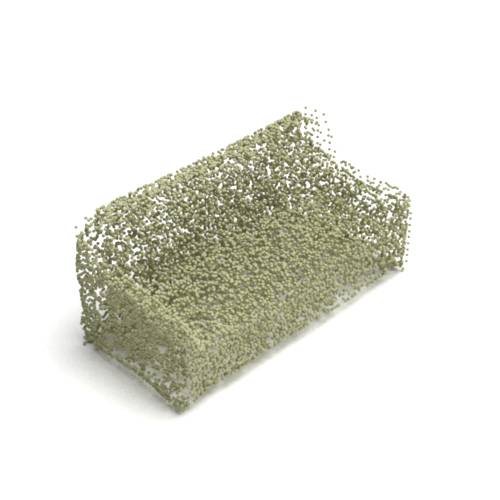}&
    \includegraphics[width=0.1\columnwidth,trim=30 30 30 30, clip]{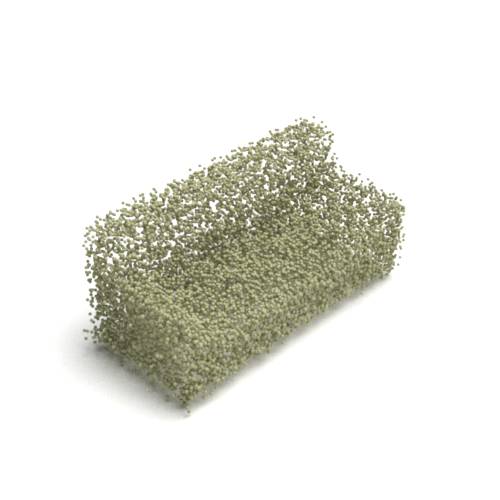}&
    \includegraphics[width=0.1\columnwidth,trim=30 30 30 30, clip]{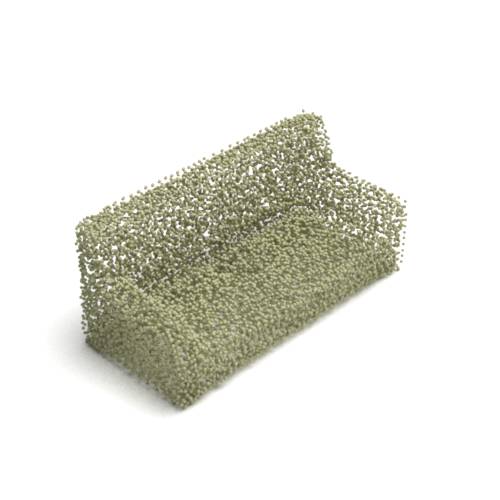}&
    \includegraphics[width=0.1\columnwidth,trim=30 30 30 30, clip]{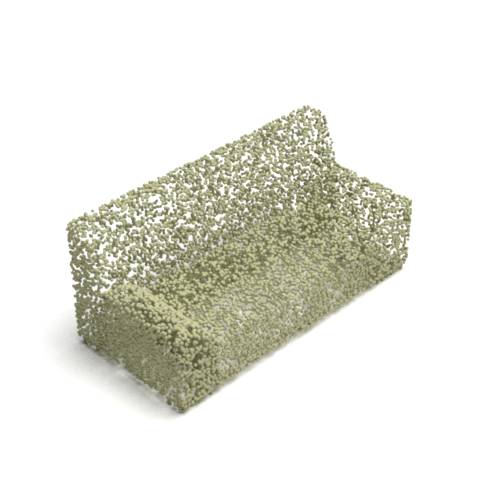}\\
    
\end{tabular}
}
\caption{Visualized completion comparison on ShapeNet.}
\label{fig:qualitative2}
\end{figure*}

\begin{figure*}[t]
\center
\setlength\tabcolsep{0pt}
{
\renewcommand{\arraystretch}{0.0}
\small
\begin{tabular}{@{}rcccccccccc@{}}
    & Input & AtlasNet\cite{atlasnet2018} & FCAE & FoldingNet\cite{foldingnet_2018_CVPR} &PCN\cite{Yuan-2018-pcn} 
    & MSN \cite{liu2019morphing} & GRNet \cite{xie2020grnet} & \emph{Ours} & Groundtruth\\
    
    \raisebox{0.6\height}{\rotatebox{90}{Table}}&
    \includegraphics[width=0.1\columnwidth,trim=30 30 30 30, clip]{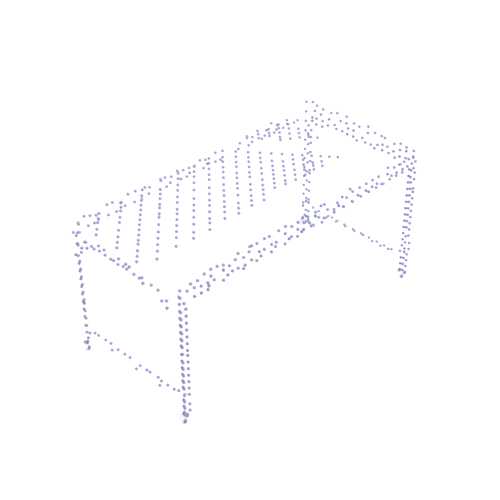}&   \includegraphics[width=0.1\columnwidth,trim=30 30 30 30, clip]{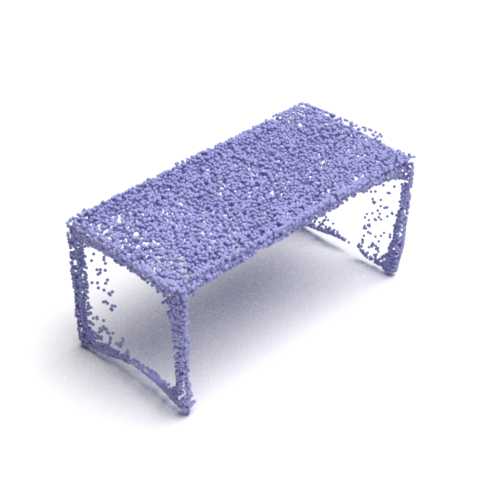}&
    \includegraphics[width=0.1\columnwidth,trim=30 30 30 30, clip]{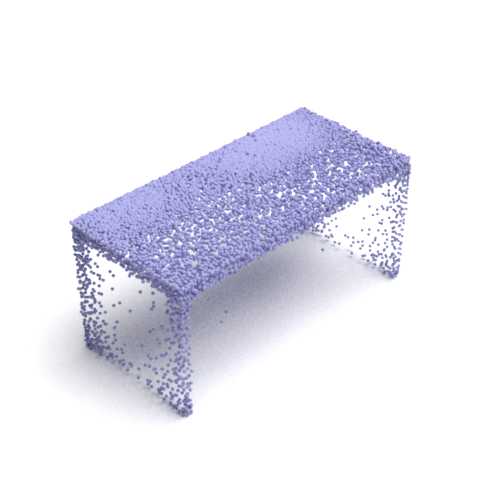}&
    \includegraphics[width=0.1\columnwidth,trim=30 30 30 30, clip]{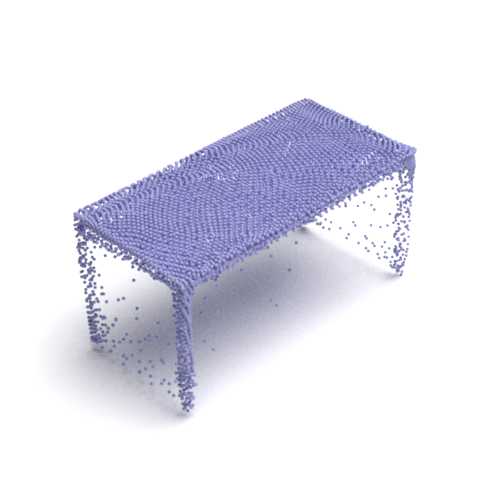}&
    \includegraphics[width=0.1\columnwidth,trim=30 30 30 30, clip]{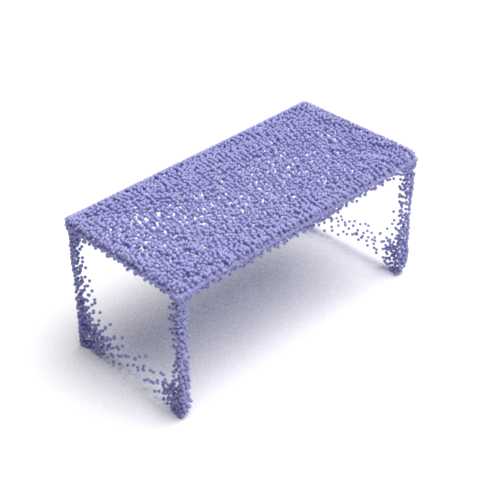}&
    \includegraphics[width=0.1\columnwidth,trim=30 30 30 30, clip]{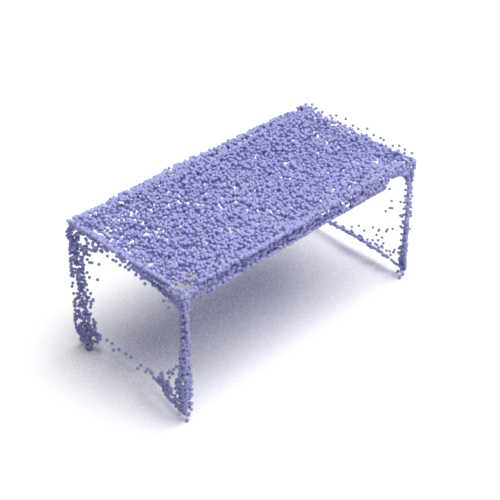}&
    \includegraphics[width=0.1\columnwidth,trim=30 30 30 30, clip]{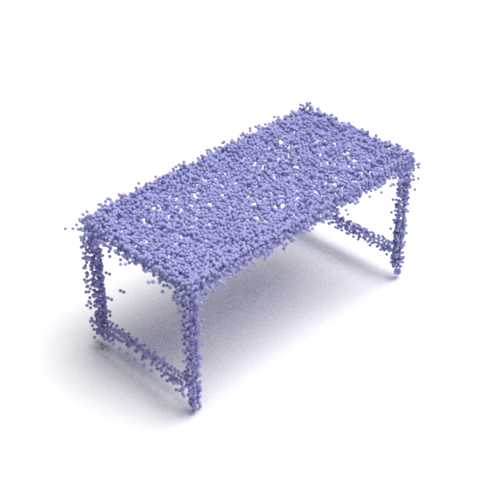}&
    \includegraphics[width=0.1\columnwidth,trim=30 30 30 30, clip]{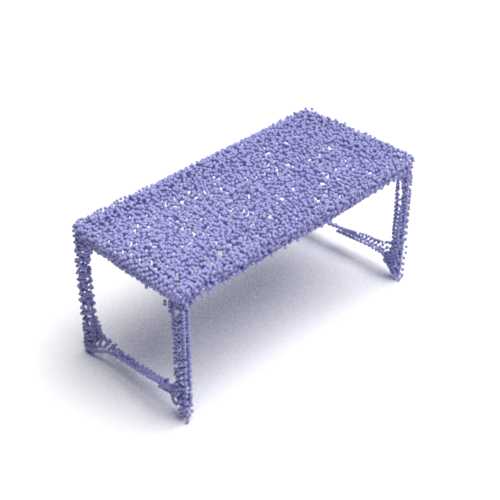}&
    \includegraphics[width=0.1\columnwidth,trim=30 30 30 30, clip]{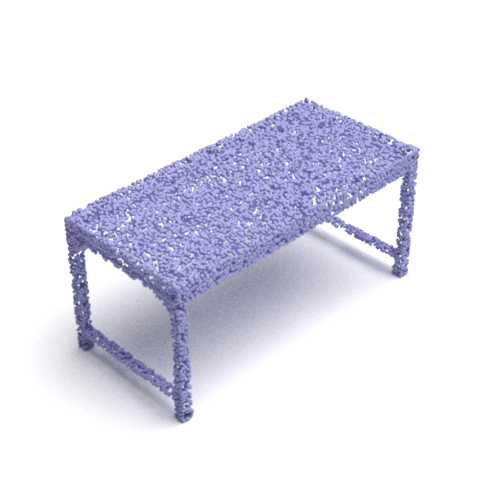}\\
    
    \raisebox{1.0\height}{\rotatebox{90}{Table}}&
    \includegraphics[width=0.1\columnwidth,trim=30 30 30 30, clip]{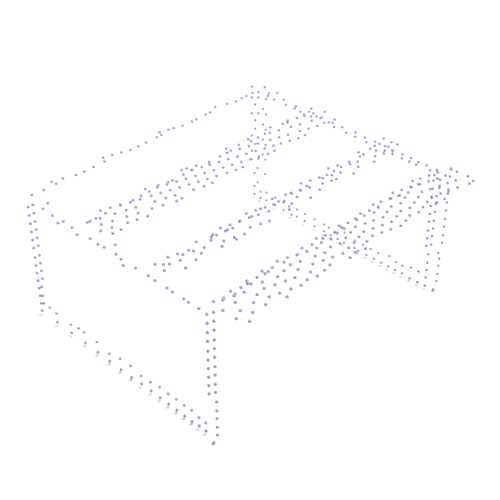}&
    \includegraphics[width=0.1\columnwidth,trim=30 30 30 30, clip]{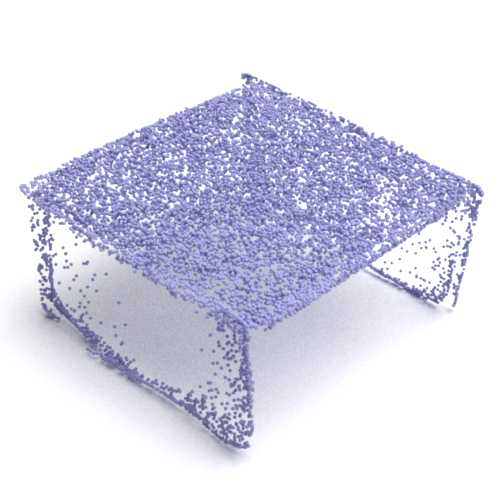}&
    \includegraphics[width=0.1\columnwidth,trim=30 30 30 30, clip]{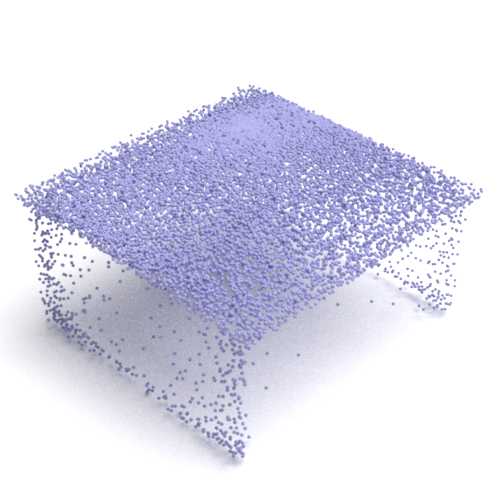}&
    \includegraphics[width=0.1\columnwidth,trim=30 30 30 30, clip]{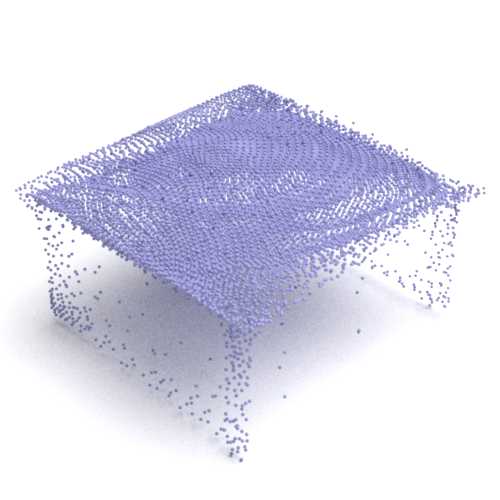}&
    \includegraphics[width=0.1\columnwidth,trim=30 30 30 30, clip]{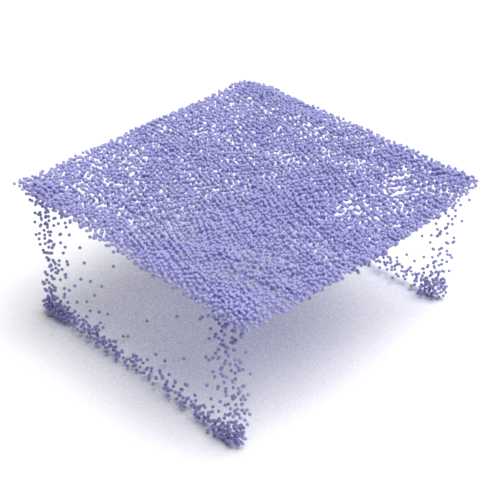}&
    \includegraphics[width=0.1\columnwidth,trim=30 30 30 30, clip]{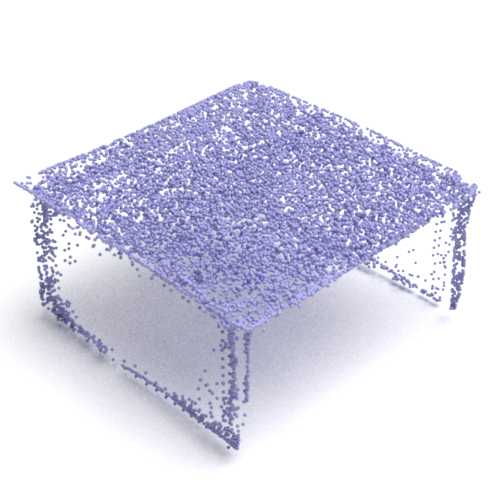}&
    \includegraphics[width=0.1\columnwidth,trim=30 30 30 30, clip]{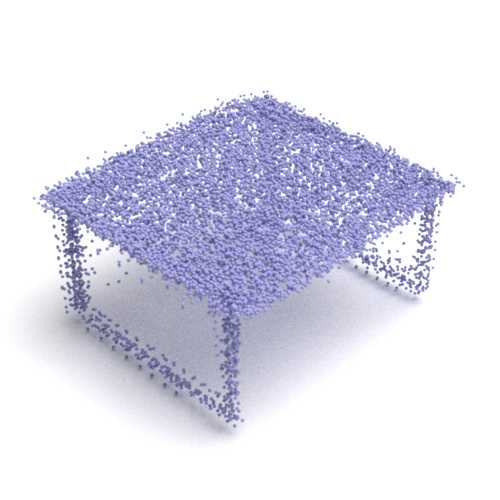}&
    \includegraphics[width=0.1\columnwidth,trim=30 30 30 30, clip]{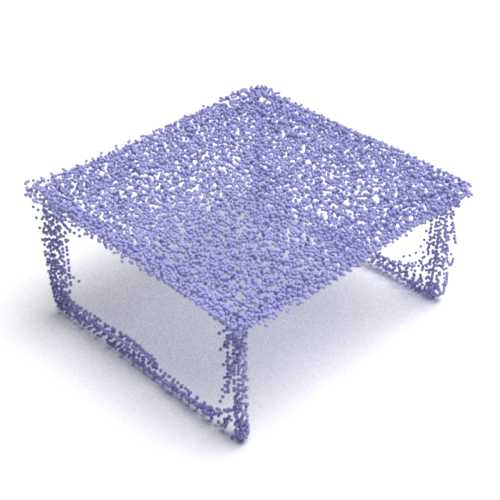}&
    \includegraphics[width=0.1\columnwidth,trim=30 30 30 30, clip]{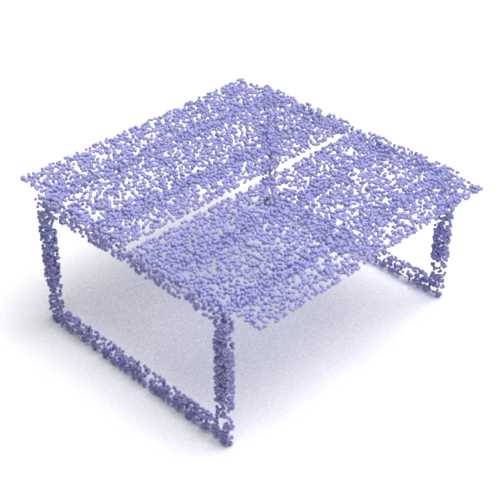}\\
    
    \raisebox{1.0\height}{\rotatebox{90}{Table}}&
    \includegraphics[width=0.1\columnwidth,trim=30 30 30 30, clip]{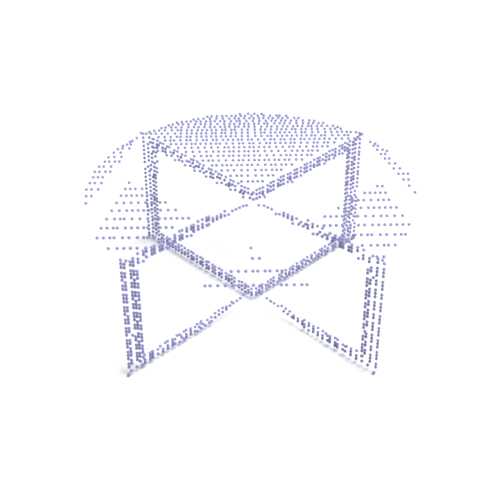}&
    \includegraphics[width=0.1\columnwidth,trim=30 30 30 30, clip]{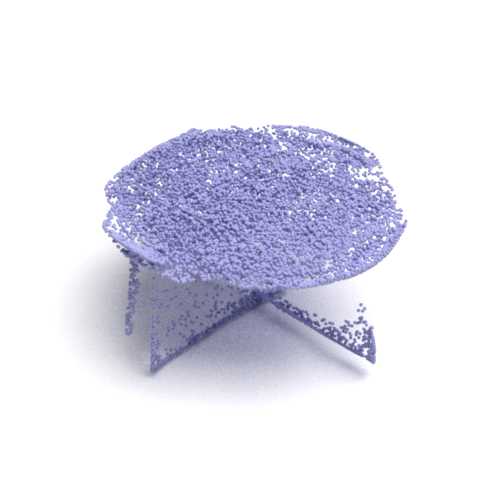}&
    \includegraphics[width=0.1\columnwidth,trim=30 30 30 30, clip]{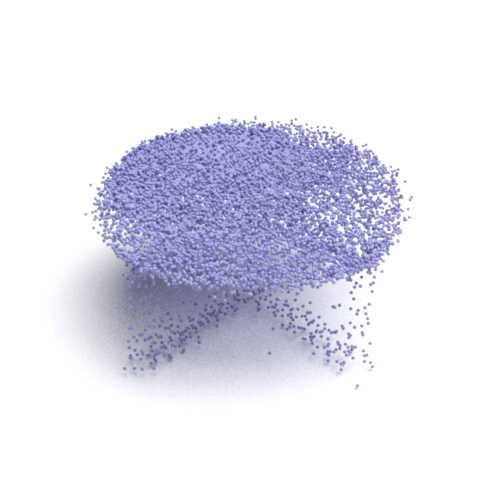}&
    \includegraphics[width=0.1\columnwidth,trim=30 30 30 30, clip]{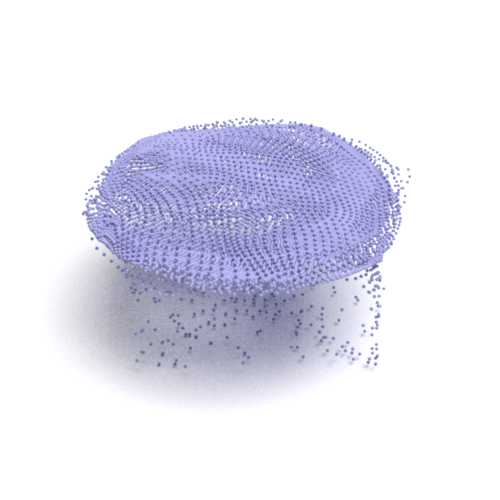}&
    \includegraphics[width=0.1\columnwidth,trim=30 30 30 30, clip]{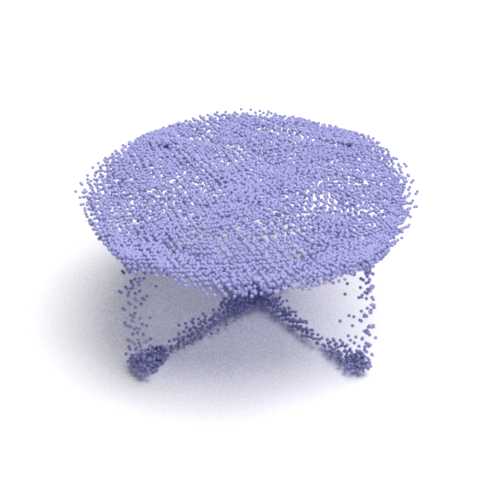}&
    \includegraphics[width=0.1\columnwidth,trim=30 30 30 30, clip]{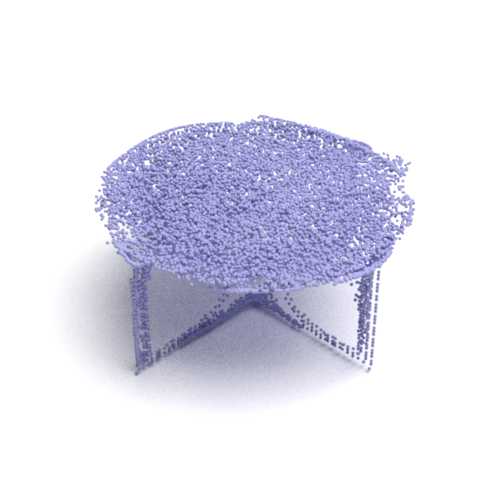}&
    \includegraphics[width=0.1\columnwidth,trim=30 30 30 30, clip]{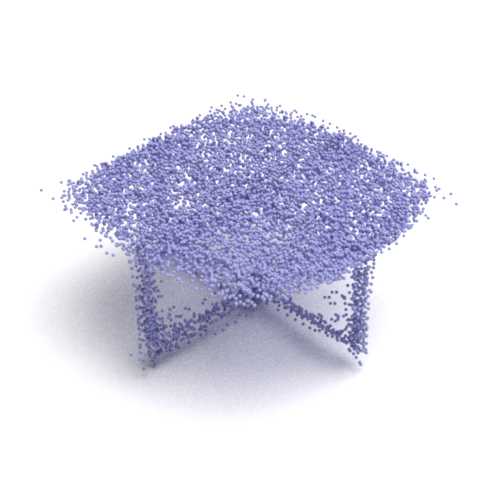}&
    \includegraphics[width=0.1\columnwidth,trim=30 30 30 30, clip]{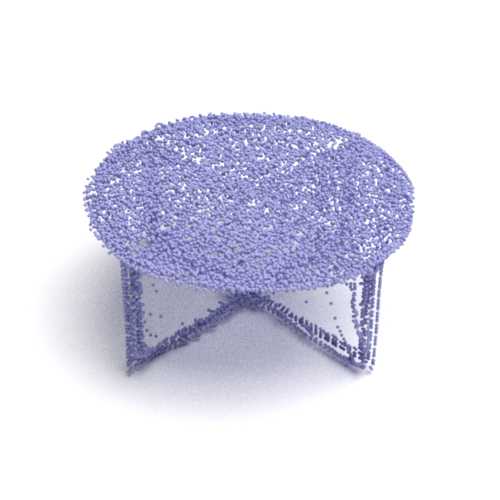}&
    \includegraphics[width=0.1\columnwidth,trim=30 30 30 30, clip]{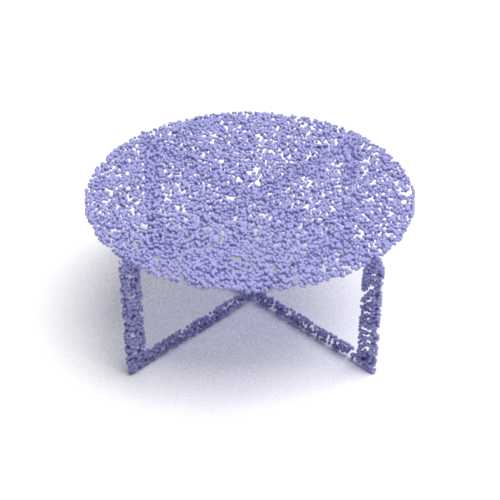}\\
    
    \raisebox{1\height}{\rotatebox{90}{Table}}&
    \includegraphics[width=0.1\columnwidth,trim=30 30 30 30, clip]{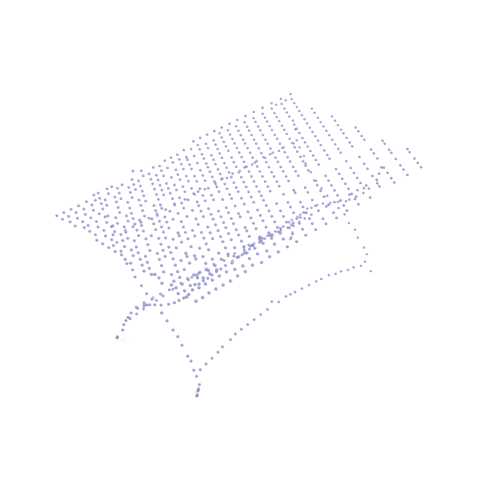}&
    \includegraphics[width=0.1\columnwidth,trim=30 30 30 30, clip]{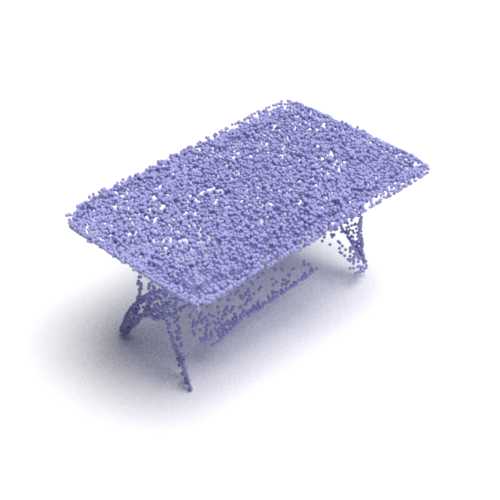}&
    \includegraphics[width=0.1\columnwidth,trim=30 30 30 30, clip]{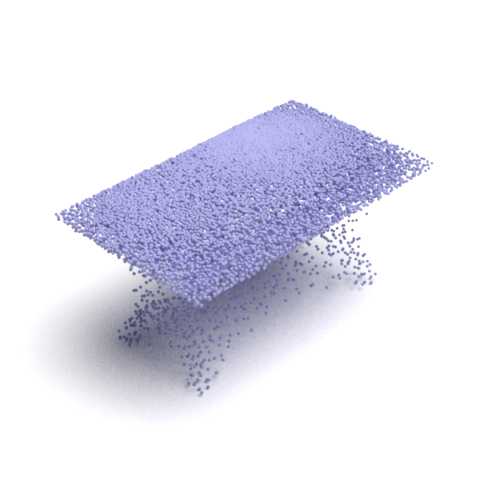}&
    \includegraphics[width=0.1\columnwidth,trim=30 30 30 30, clip]{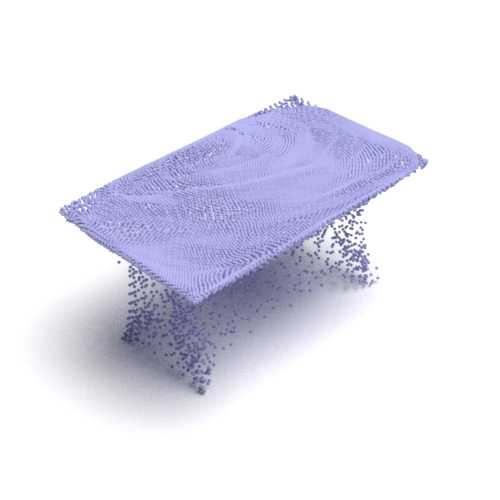}&
    \includegraphics[width=0.1\columnwidth,trim=30 30 30 30, clip]{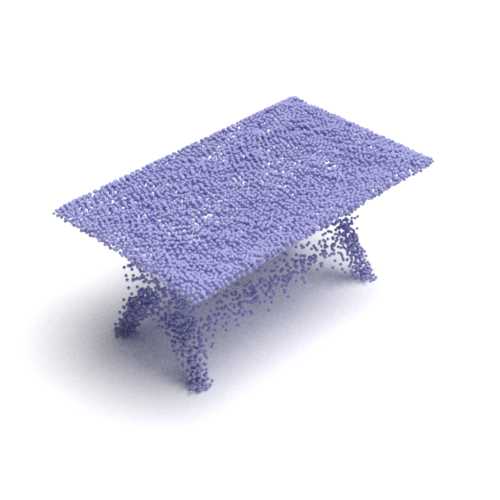}&
    \includegraphics[width=0.1\columnwidth,trim=30 30 30 30, clip]{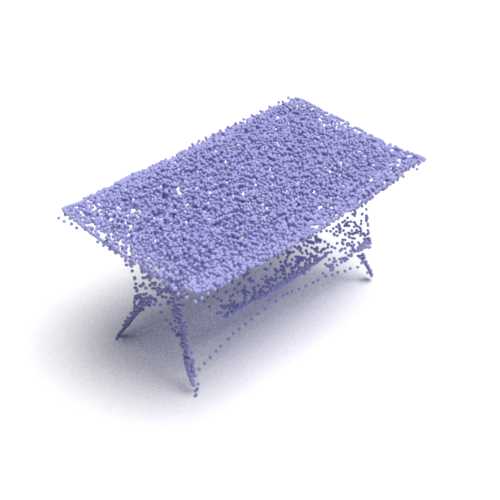}&
    \includegraphics[width=0.1\columnwidth,trim=30 30 30 30, clip]{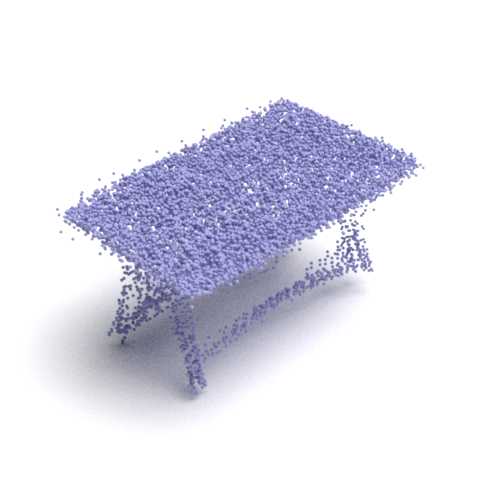}&
    \includegraphics[width=0.1\columnwidth,trim=30 30 30 30, clip]{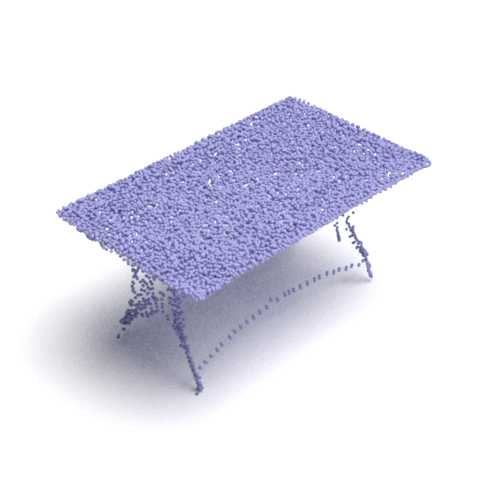}&
    \includegraphics[width=0.1\columnwidth,trim=30 30 30 30, clip]{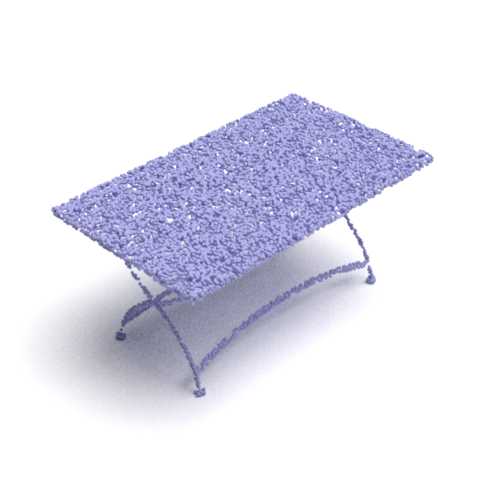}\\
    
    \raisebox{0.6\height}{\rotatebox{90}{Vessel}}&
    \includegraphics[width=0.1\columnwidth,trim=30 30 30 30, clip]{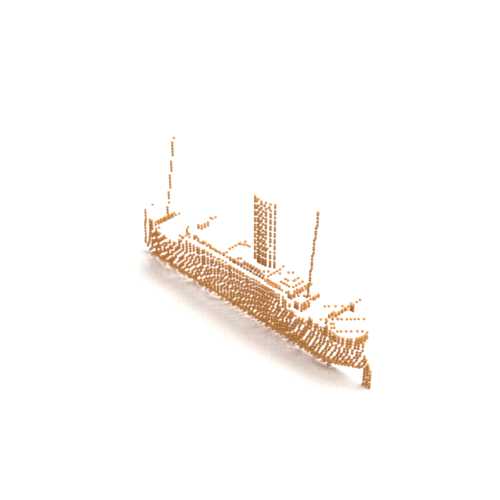}&
    \includegraphics[width=0.1\columnwidth,trim=30 30 30 30, clip]{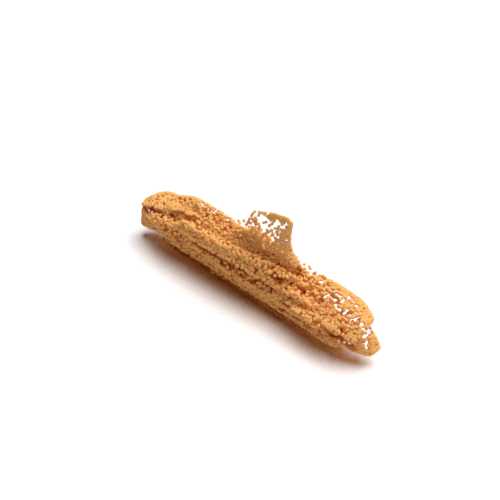}&
    \includegraphics[width=0.1\columnwidth,trim=30 30 30 30, clip]{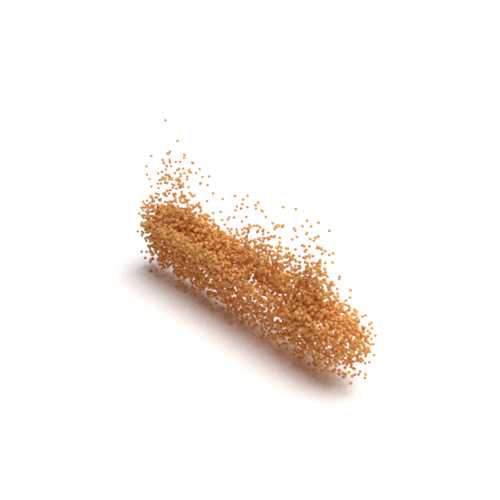}&
    \includegraphics[width=0.1\columnwidth,trim=30 30 30 30, clip]{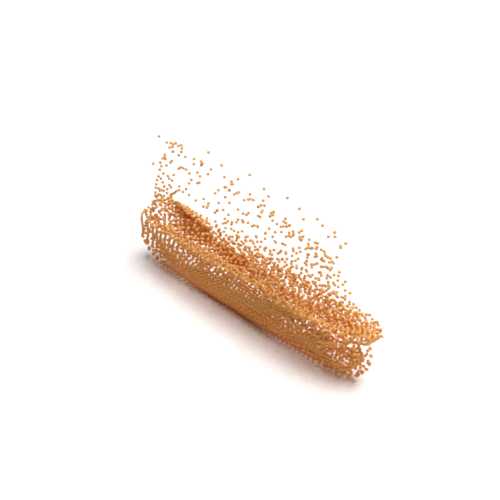}&
    \includegraphics[width=0.1\columnwidth,trim=30 30 30 30, clip]{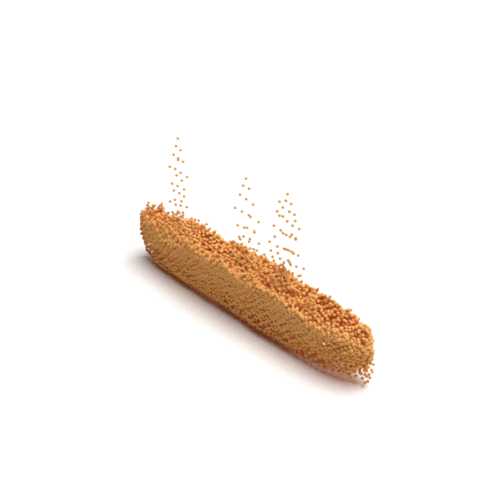}&
    \includegraphics[width=0.1\columnwidth,trim=30 30 30 30, clip]{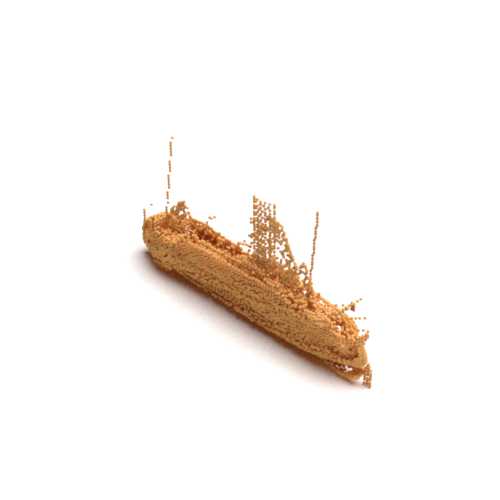}&
    \includegraphics[width=0.1\columnwidth,trim=30 30 30 30, clip]{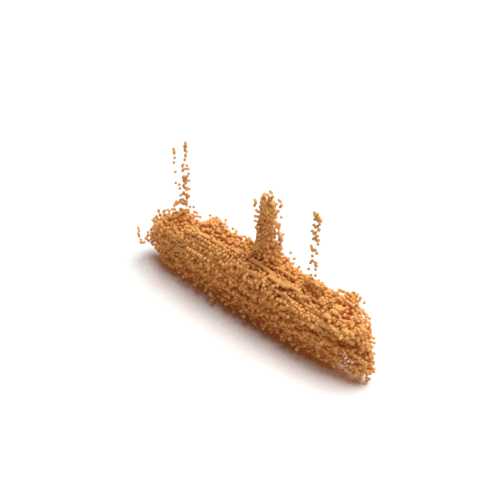}&
    \includegraphics[width=0.1\columnwidth,trim=30 30 30 30, clip]{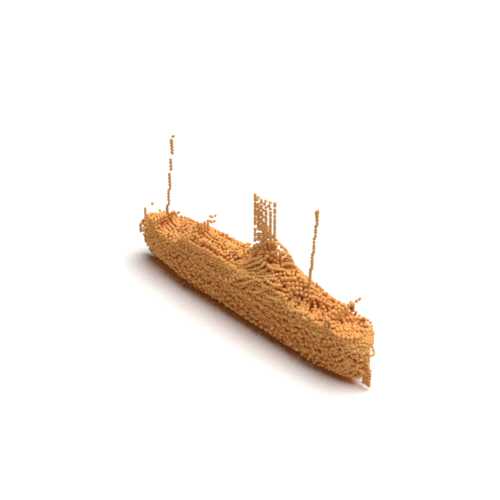}&
    \includegraphics[width=0.1\columnwidth,trim=30 30 30 30, clip]{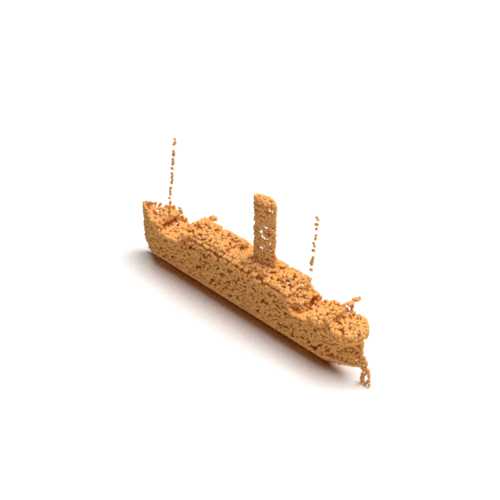}\\
    
    \raisebox{0.6\height}{\rotatebox{90}{Vessel}}&
    \includegraphics[width=0.1\columnwidth,trim=30 30 30 30, clip]{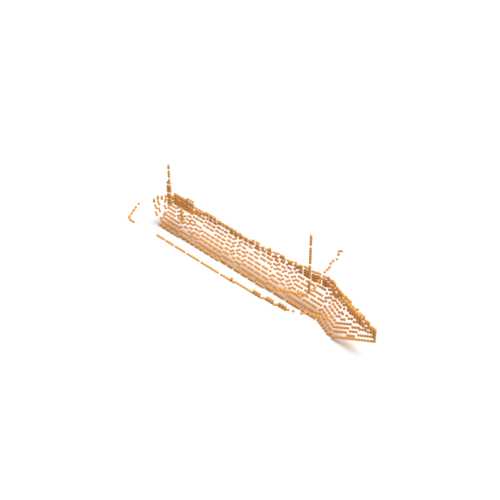}&
    \includegraphics[width=0.1\columnwidth,trim=30 30 30 30, clip]{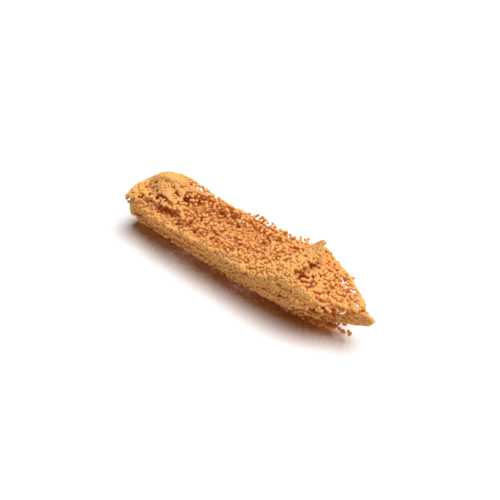}&
    \includegraphics[width=0.1\columnwidth,trim=30 30 30 30, clip]{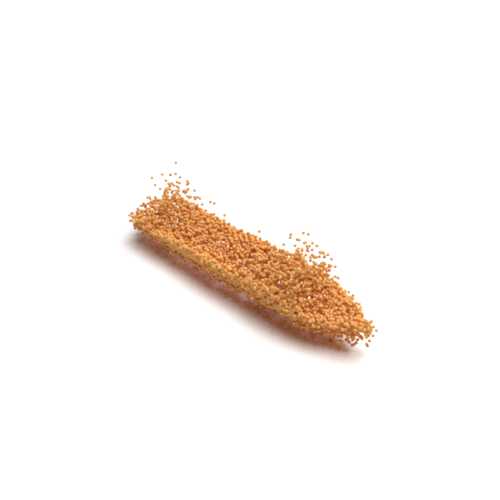}&
    \includegraphics[width=0.1\columnwidth,trim=30 30 30 30, clip]{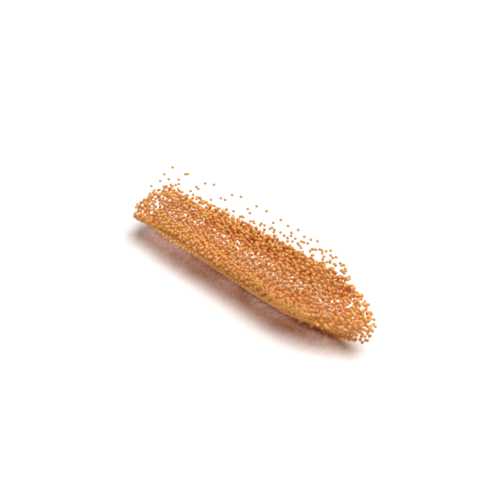}&
    \includegraphics[width=0.1\columnwidth,trim=30 30 30 30, clip]{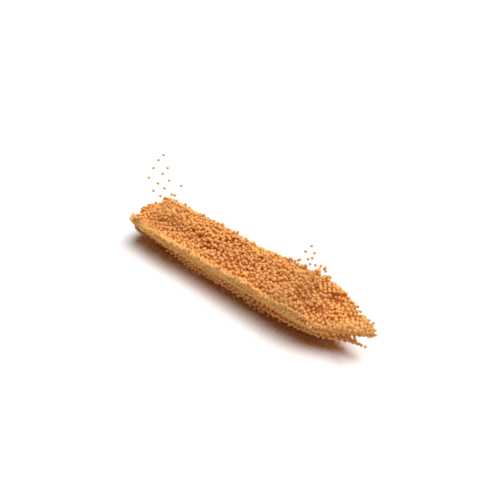}&
    \includegraphics[width=0.1\columnwidth,trim=30 30 30 30, clip]{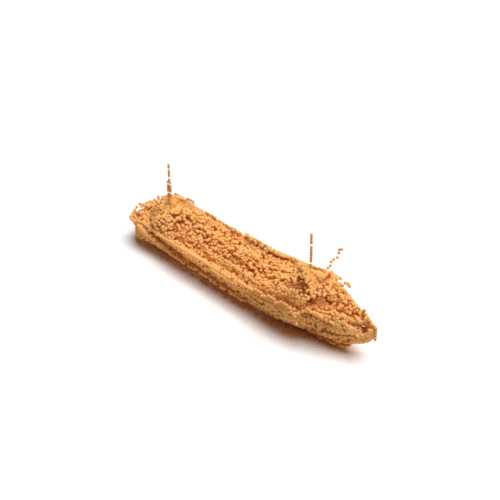}&
    \includegraphics[width=0.1\columnwidth,trim=30 30 30 30, clip]{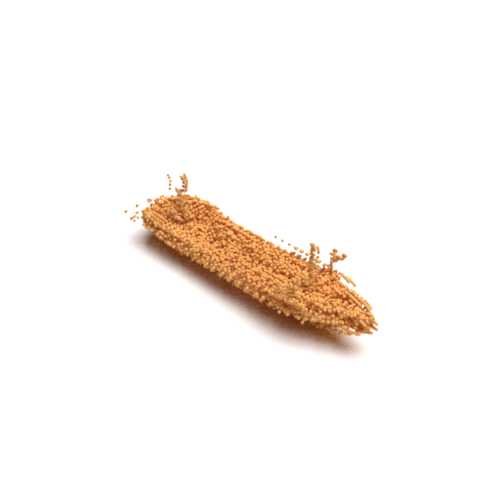}&
    \includegraphics[width=0.1\columnwidth,trim=30 30 30 30, clip]{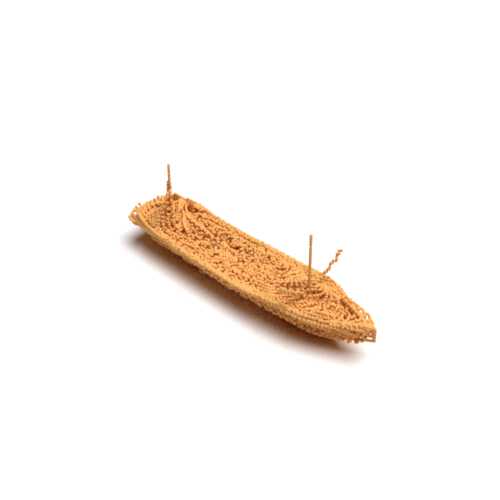}&
    \includegraphics[width=0.1\columnwidth,trim=30 30 30 30, clip]{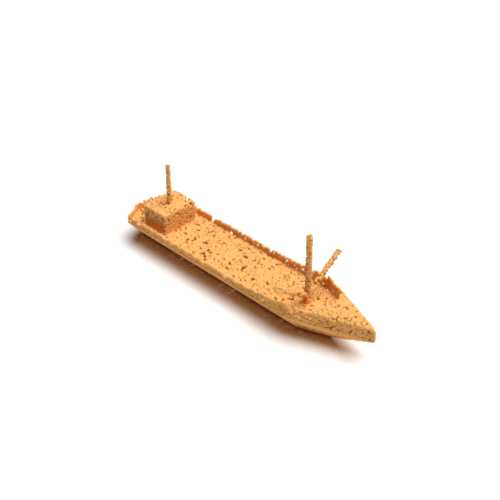}\\
    
    \raisebox{0.6\height}{\rotatebox{90}{Vessel}}&
    \includegraphics[width=0.1\columnwidth,trim=30 30 30 30, clip]{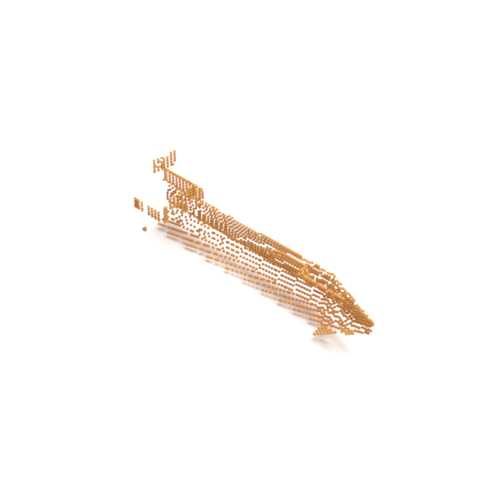}&   \includegraphics[width=0.1\columnwidth,trim=30 30 30 30, clip]{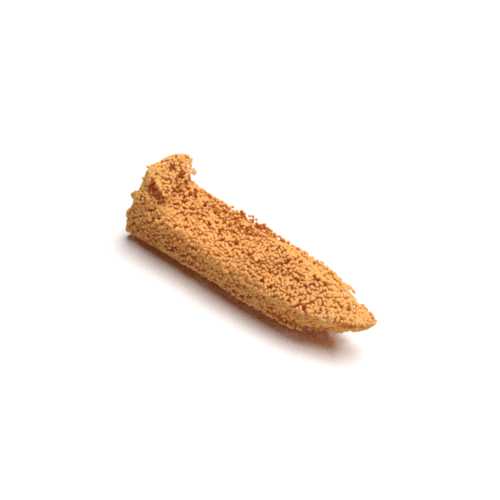}&
    \includegraphics[width=0.1\columnwidth,trim=30 30 30 30, clip]{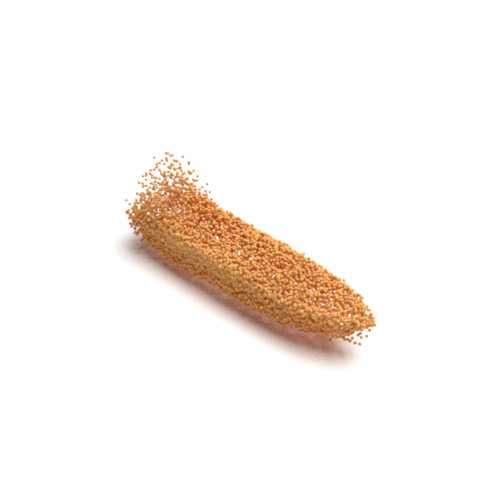}&
    \includegraphics[width=0.1\columnwidth,trim=30 30 30 30, clip]{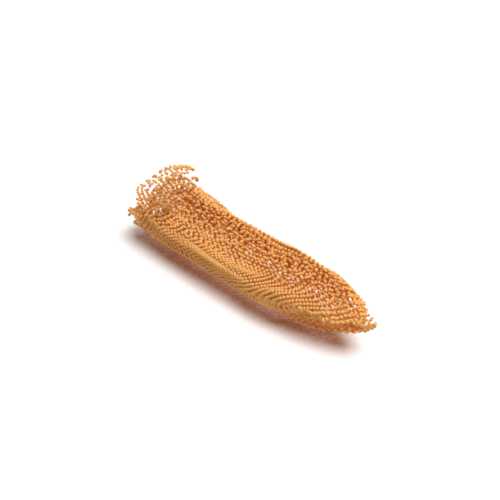}&
    \includegraphics[width=0.1\columnwidth,trim=30 30 30 30, clip]{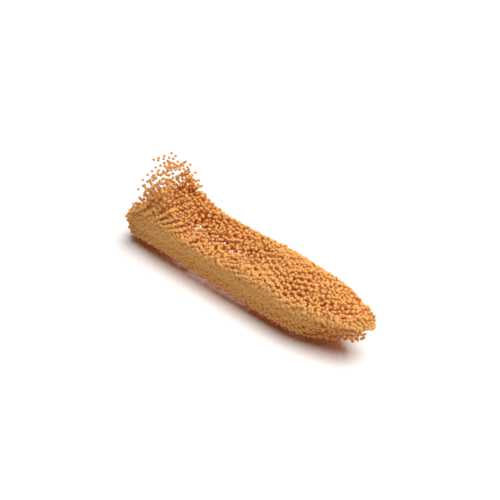}&
    \includegraphics[width=0.1\columnwidth,trim=30 30 30 30, clip]{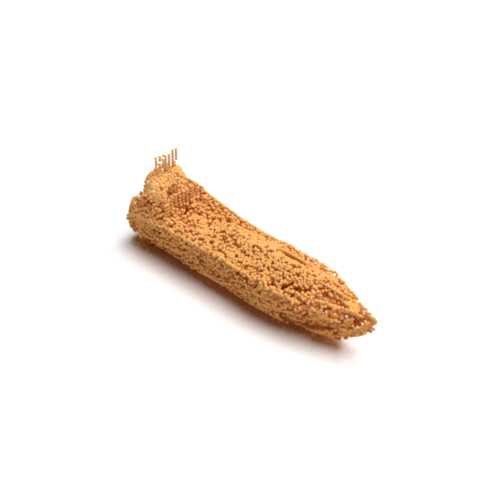}&
    \includegraphics[width=0.1\columnwidth,trim=30 30 30 30, clip]{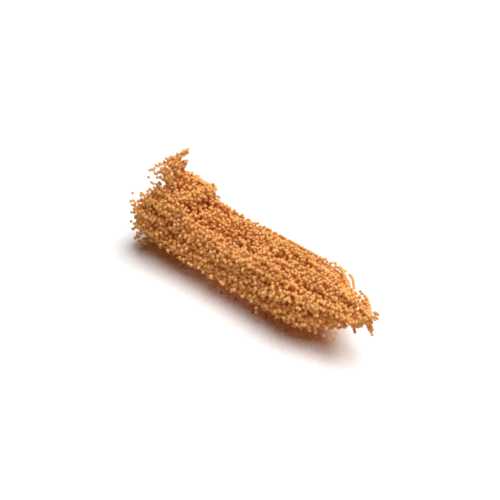}&
    \includegraphics[width=0.1\columnwidth,trim=30 30 30 30, clip]{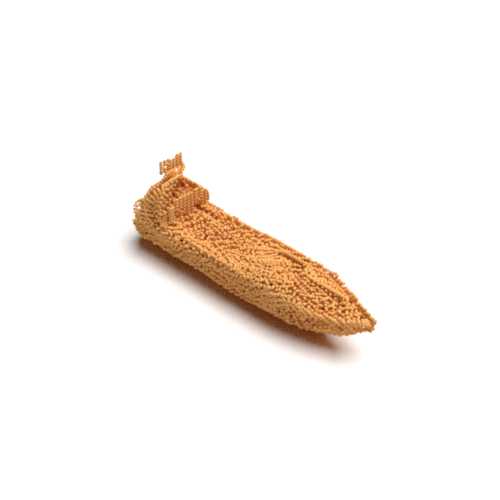}&
    \includegraphics[width=0.1\columnwidth,trim=30 30 30 30, clip]{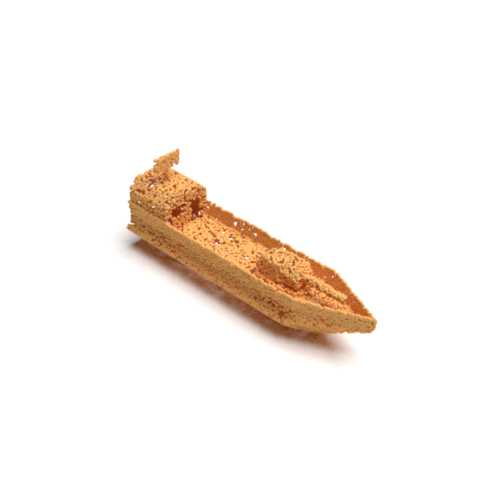}\\
    
    \raisebox{1.0\height}{\rotatebox{90}{Vessel}}&
    \includegraphics[width=0.1\columnwidth,trim=30 30 30 30, clip]{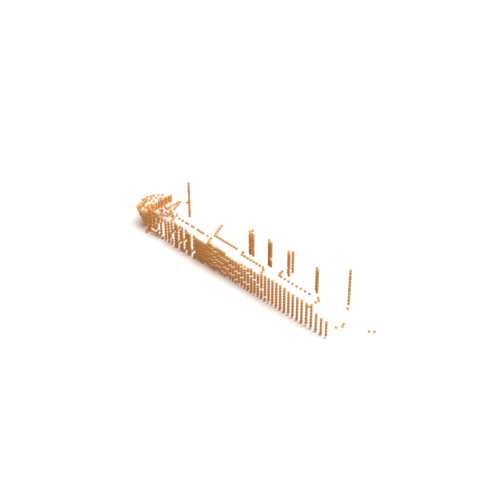}&
    \includegraphics[width=0.1\columnwidth,trim=30 30 30 30, clip]{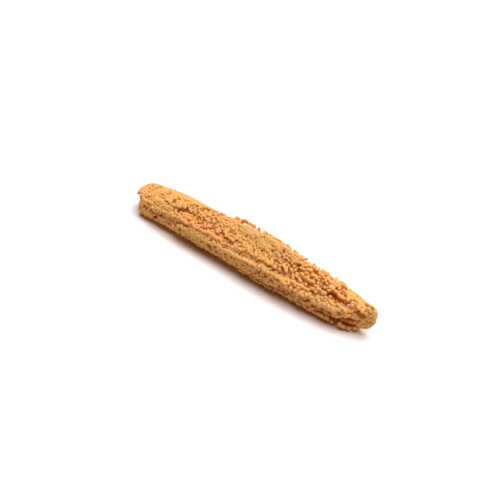}&
    \includegraphics[width=0.1\columnwidth,trim=30 30 30 30, clip]{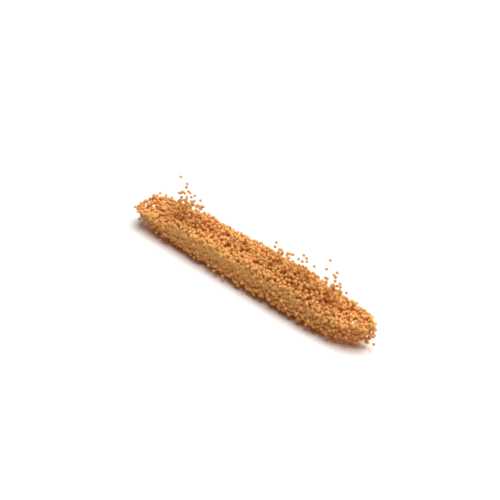}&
    \includegraphics[width=0.1\columnwidth,trim=30 30 30 30, clip]{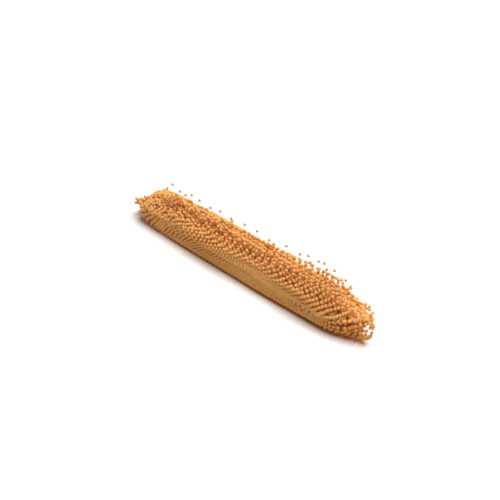}&
    \includegraphics[width=0.1\columnwidth,trim=30 30 30 30, clip]{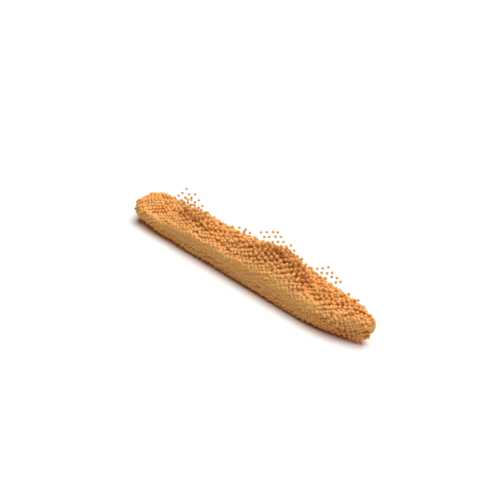}&
    \includegraphics[width=0.1\columnwidth,trim=30 30 30 30, clip]{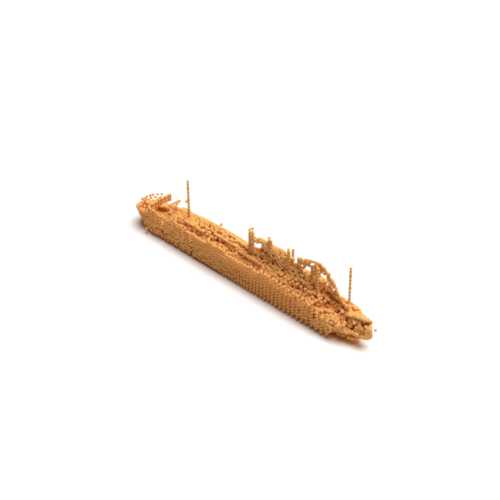}&
    \includegraphics[width=0.1\columnwidth,trim=30 30 30 30, clip]{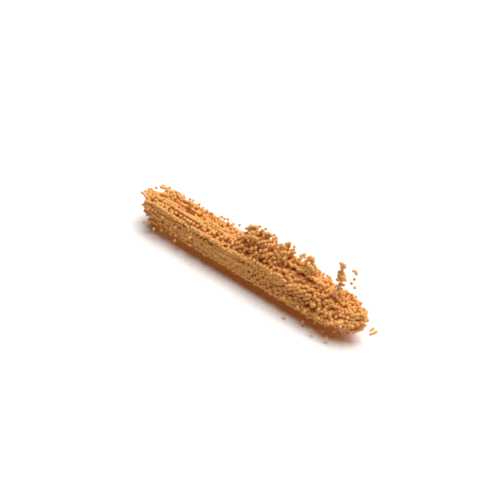}&
    \includegraphics[width=0.1\columnwidth,trim=30 30 30 30, clip]{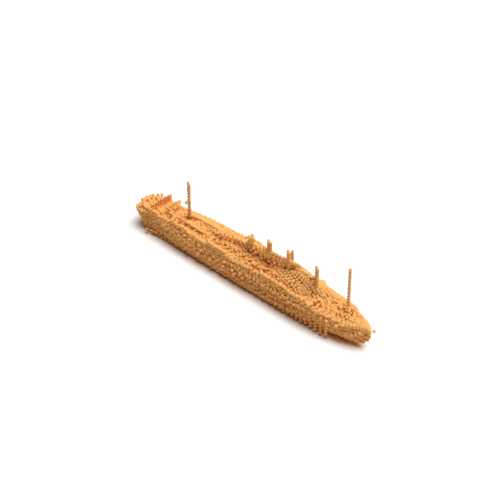}&
    \includegraphics[width=0.1\columnwidth,trim=30 30 30 30, clip]{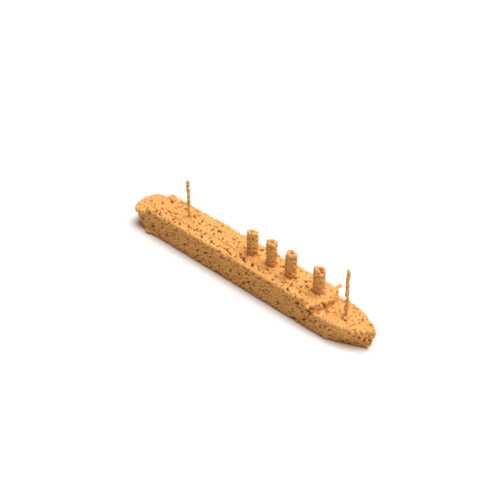}\\
    
\end{tabular}
}
\caption{Visualized completion comparison on ShapeNet.}
\label{fig:qualitative3}
\end{figure*}

\begin{figure*}[t]
\center
\setlength\tabcolsep{3.5pt}
{
\renewcommand{\arraystretch}{2.0}
\small
\begin{tabular}{@{}rcccccccc@{}}
    & view1 & view2 & view3 & view4& view5 & view6 & view7 & view8 \\
    \raisebox{0.5\height}{\rotatebox{90}{Input}}&
    \includegraphics[width=0.09\columnwidth,trim=30 30 30 30, clip]{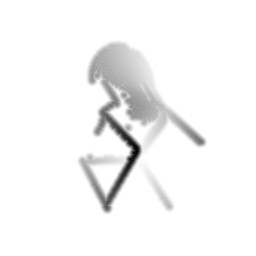}&   
    \includegraphics[width=0.09\columnwidth,trim=30 30 30 30, clip]{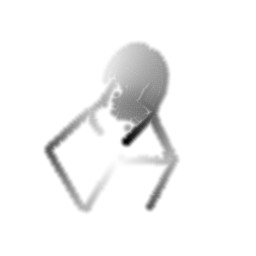}&
    \includegraphics[width=0.09\columnwidth,trim=30 30 30 30, clip]{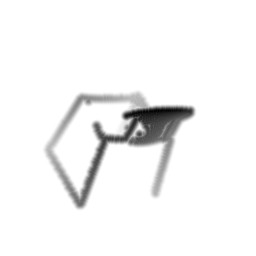}&
    \includegraphics[width=0.09\columnwidth,trim=30 30 30 30, clip]{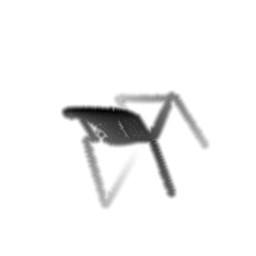}&
    \includegraphics[width=0.09\columnwidth,trim=30 30 30 30, clip]{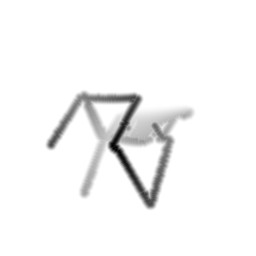}&
    \includegraphics[width=0.09\columnwidth,trim=30 30 30 30, clip]{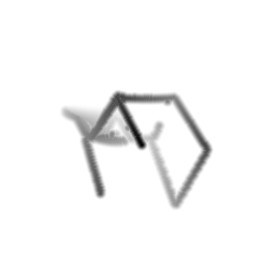}&
    \includegraphics[width=0.09\columnwidth,trim=30 30 30 30, clip]{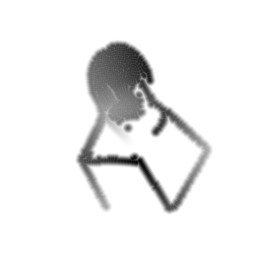}&
    \includegraphics[width=0.09\columnwidth,trim=30 30 30 30, clip]{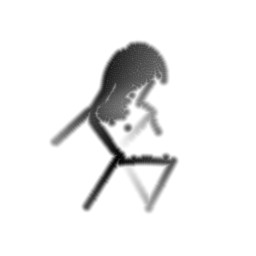}\\
    
    \raisebox{0.0\height}{\rotatebox{90}{AtlasNet \cite{atlasnet2018}}}&
    \includegraphics[width=0.09\columnwidth,trim=30 30 30 30, clip]{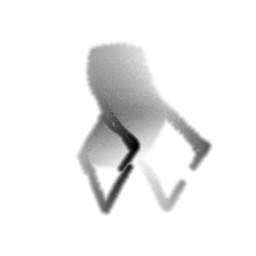}&   
    \includegraphics[width=0.09\columnwidth,trim=30 30 30 30, clip]{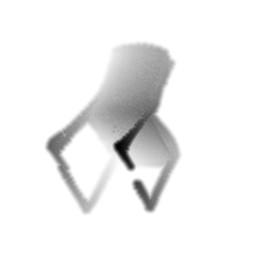}&
    \includegraphics[width=0.09\columnwidth,trim=30 30 30 30, clip]{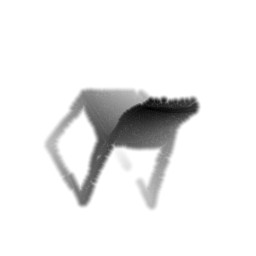}&
    \includegraphics[width=0.09\columnwidth,trim=30 30 30 30, clip]{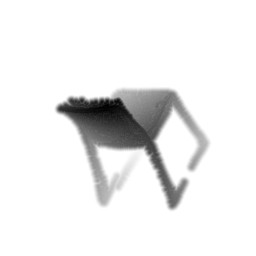}&
    \includegraphics[width=0.09\columnwidth,trim=30 30 30 30, clip]{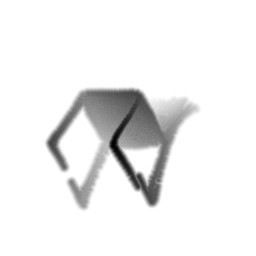}&
    \includegraphics[width=0.09\columnwidth,trim=30 30 30 30, clip]{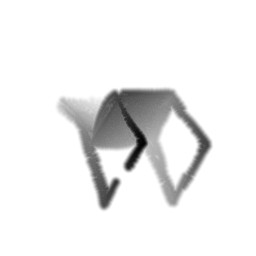}&
    \includegraphics[width=0.09\columnwidth,trim=30 30 30 30, clip]{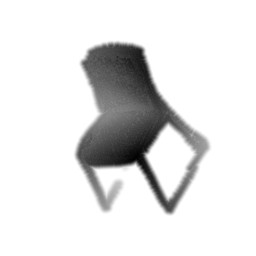}&
    \includegraphics[width=0.09\columnwidth,trim=30 30 30 30, clip]{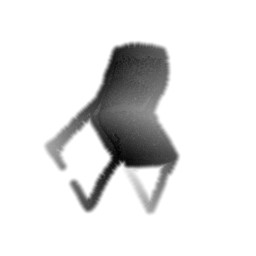}\\
    
    \raisebox{0.3\height}{\rotatebox{90}{FCAE}}&
    \includegraphics[width=0.09\columnwidth,trim=30 30 30 30, clip]{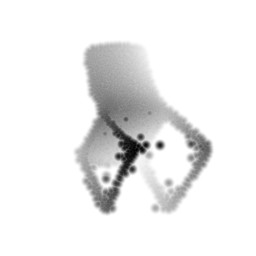}&   
    \includegraphics[width=0.09\columnwidth,trim=30 30 30 30, clip]{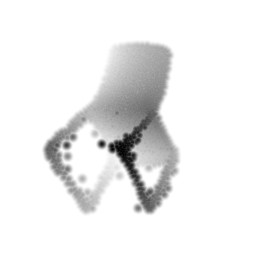}&
    \includegraphics[width=0.09\columnwidth,trim=30 30 30 30, clip]{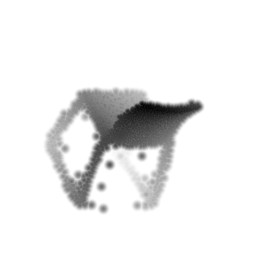}&
    \includegraphics[width=0.09\columnwidth,trim=30 30 30 30, clip]{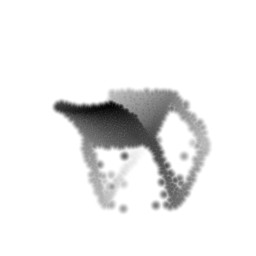}&
    \includegraphics[width=0.09\columnwidth,trim=30 30 30 30, clip]{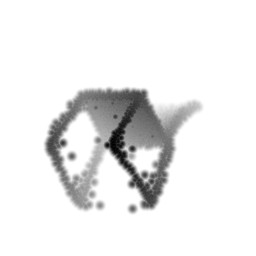}&
    \includegraphics[width=0.09\columnwidth,trim=30 30 30 30, clip]{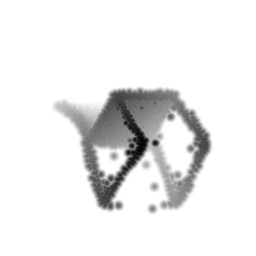}&
    \includegraphics[width=0.09\columnwidth,trim=30 30 30 30, clip]{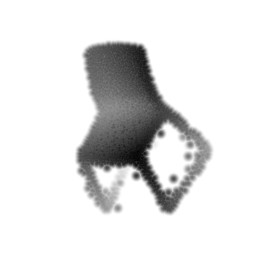}&
    \includegraphics[width=0.09\columnwidth,trim=30 30 30 30, clip]{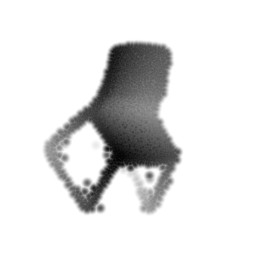}\\
    
    \raisebox{0\height}{\rotatebox{90}{FoldingNet \cite{foldingnet_2018_CVPR}}}&
    \includegraphics[width=0.09\columnwidth,trim=30 30 30 30, clip]{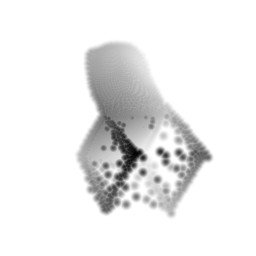}&   
    \includegraphics[width=0.09\columnwidth,trim=30 30 30 30, clip]{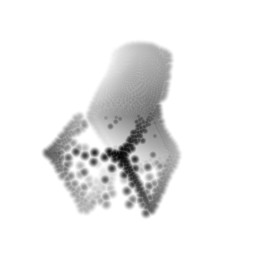}&
    \includegraphics[width=0.09\columnwidth,trim=30 30 30 30, clip]{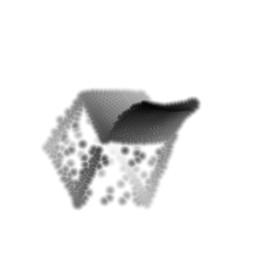}&
    \includegraphics[width=0.09\columnwidth,trim=30 30 30 30, clip]{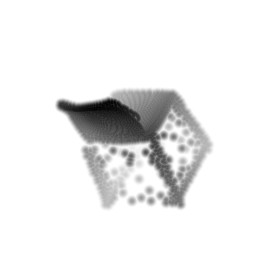}&
    \includegraphics[width=0.09\columnwidth,trim=30 30 30 30, clip]{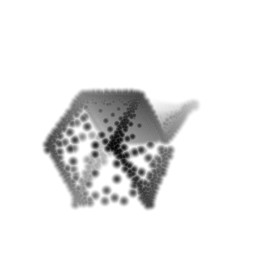}&
    \includegraphics[width=0.09\columnwidth,trim=30 30 30 30, clip]{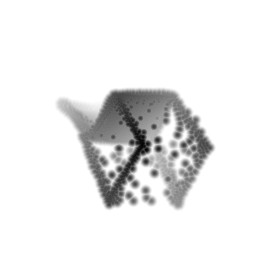}&
    \includegraphics[width=0.09\columnwidth,trim=30 30 30 30, clip]{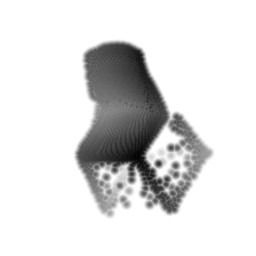}&
    \includegraphics[width=0.09\columnwidth,trim=30 30 30 30, clip]{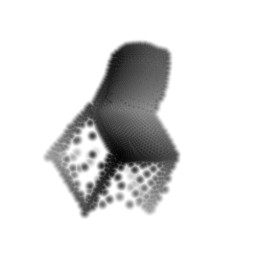}\\
    
    \raisebox{0.3\height}{\rotatebox{90}{PCN \cite{Yuan-2018-pcn}}}&
    \includegraphics[width=0.09\columnwidth,trim=30 30 30 30, clip]{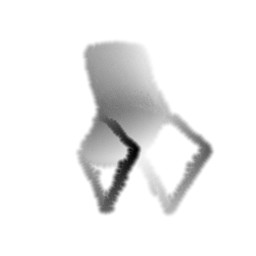}&   
    \includegraphics[width=0.09\columnwidth,trim=30 30 30 30, clip]{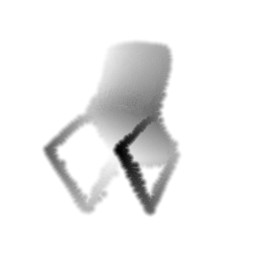}&
    \includegraphics[width=0.09\columnwidth,trim=30 30 30 30, clip]{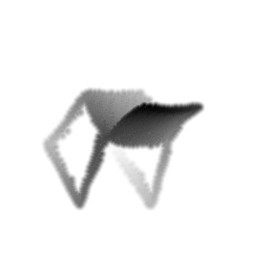}&
    \includegraphics[width=0.09\columnwidth,trim=30 30 30 30, clip]{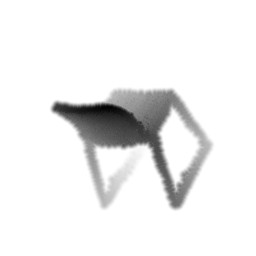}&
    \includegraphics[width=0.09\columnwidth,trim=30 30 30 30, clip]{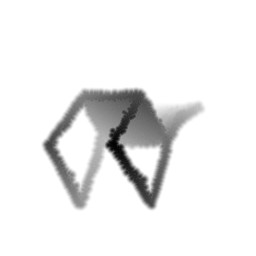}&
    \includegraphics[width=0.09\columnwidth,trim=30 30 30 30, clip]{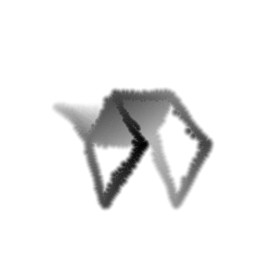}&
    \includegraphics[width=0.09\columnwidth,trim=30 30 30 30, clip]{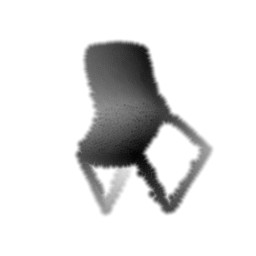}&
    \includegraphics[width=0.09\columnwidth,trim=30 30 30 30, clip]{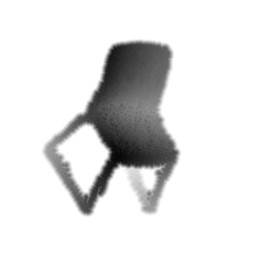}\\
    
    \raisebox{0.3\height}{\rotatebox{90}{MSN \cite{liu2019morphing}}}&
    \includegraphics[width=0.09\columnwidth,trim=30 30 30 30, clip]{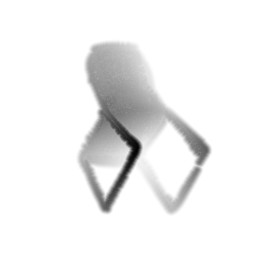}&   
    \includegraphics[width=0.09\columnwidth,trim=30 30 30 30, clip]{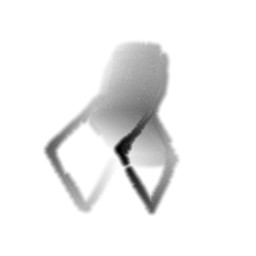}&
    \includegraphics[width=0.09\columnwidth,trim=30 30 30 30, clip]{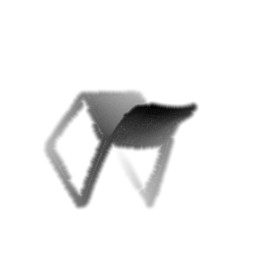}&
    \includegraphics[width=0.09\columnwidth,trim=30 30 30 30, clip]{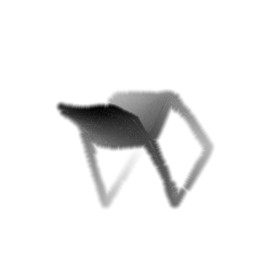}&
    \includegraphics[width=0.09\columnwidth,trim=30 30 30 30, clip]{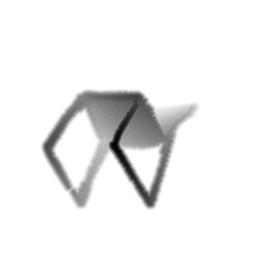}&
    \includegraphics[width=0.09\columnwidth,trim=30 30 30 30, clip]{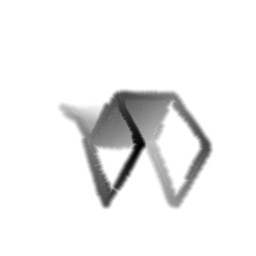}&
    \includegraphics[width=0.09\columnwidth,trim=30 30 30 30, clip]{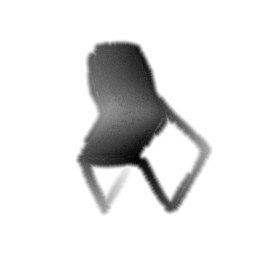}&
    \includegraphics[width=0.09\columnwidth,trim=30 30 30 30, clip]{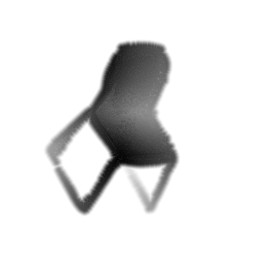}\\
    
    \raisebox{0.2\height}{\rotatebox{90}{GRNet \cite{xie2020grnet}}}&
    \includegraphics[width=0.09\columnwidth,trim=30 30 30 30, clip]{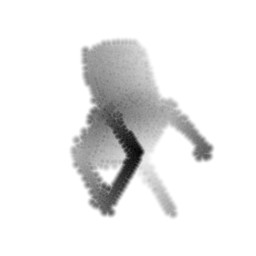}&   
    \includegraphics[width=0.09\columnwidth,trim=30 30 30 30, clip]{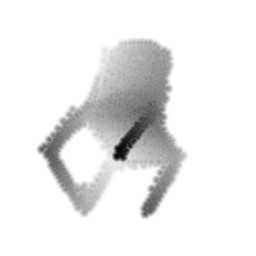}&
    \includegraphics[width=0.09\columnwidth,trim=30 30 30 30, clip]{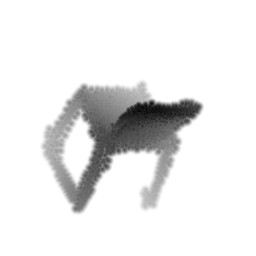}&
    \includegraphics[width=0.09\columnwidth,trim=30 30 30 30, clip]{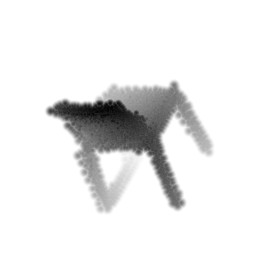}&
    \includegraphics[width=0.09\columnwidth,trim=30 30 30 30, clip]{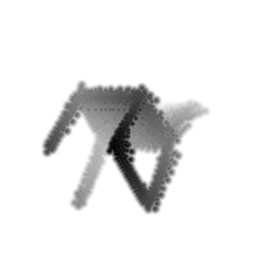}&
    \includegraphics[width=0.09\columnwidth,trim=30 30 30 30, clip]{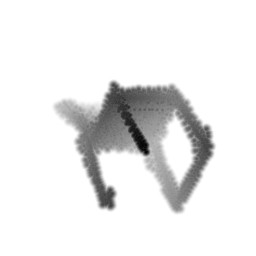}&
    \includegraphics[width=0.09\columnwidth,trim=30 30 30 30, clip]{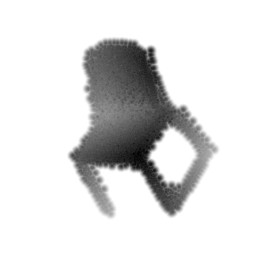}&
    \includegraphics[width=0.09\columnwidth,trim=30 30 30 30, clip]{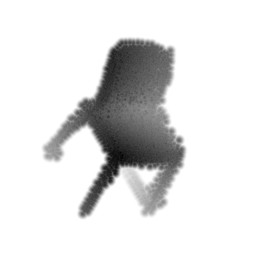}\\
    
    \raisebox{0.8\height}{\rotatebox{90}{\emph{Ours}}}&
    \includegraphics[width=0.09\columnwidth,trim=30 30 30 30, clip]{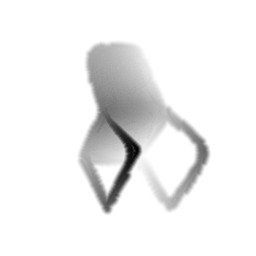}&   
    \includegraphics[width=0.09\columnwidth,trim=30 30 30 30, clip]{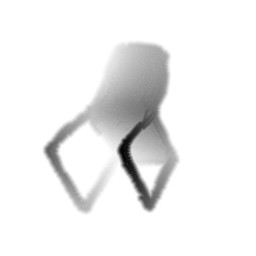}&
    \includegraphics[width=0.09\columnwidth,trim=30 30 30 30, clip]{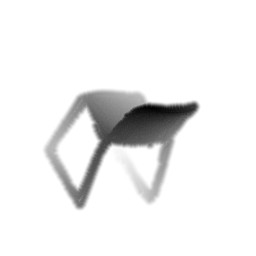}&
    \includegraphics[width=0.09\columnwidth,trim=30 30 30 30, clip]{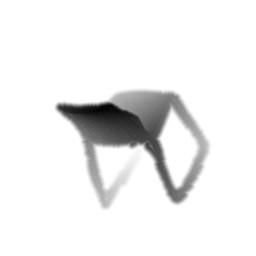}&
    \includegraphics[width=0.09\columnwidth,trim=30 30 30 30, clip]{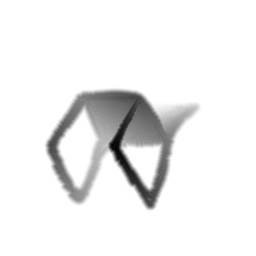}&
    \includegraphics[width=0.09\columnwidth,trim=30 30 30 30, clip]{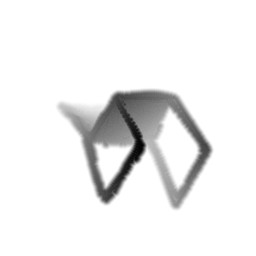}&
    \includegraphics[width=0.09\columnwidth,trim=30 30 30 30, clip]{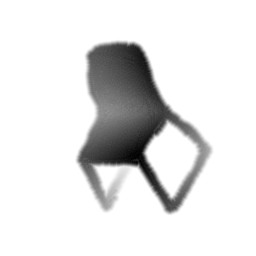}&
    \includegraphics[width=0.09\columnwidth,trim=30 30 30 30, clip]{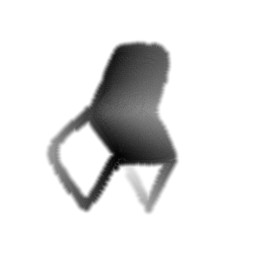}\\
    
    \raisebox{0.0\height}{\rotatebox{90}{Groundtruth}}&
    \includegraphics[width=0.09\columnwidth,trim=30 30 30 30, clip]{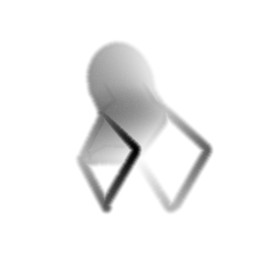}&   
    \includegraphics[width=0.09\columnwidth,trim=30 30 30 30, clip]{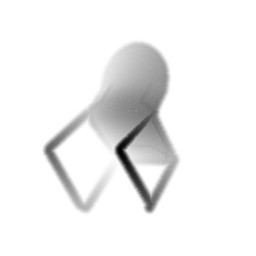}&
    \includegraphics[width=0.09\columnwidth,trim=30 30 30 30, clip]{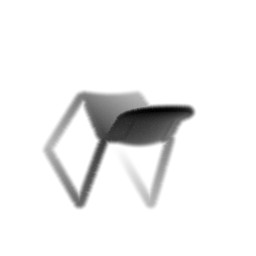}&
    \includegraphics[width=0.09\columnwidth,trim=30 30 30 30, clip]{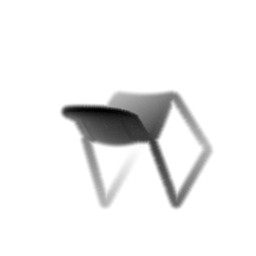}&
    \includegraphics[width=0.09\columnwidth,trim=30 30 30 30, clip]{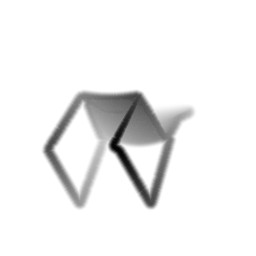}&
    \includegraphics[width=0.09\columnwidth,trim=30 30 30 30, clip]{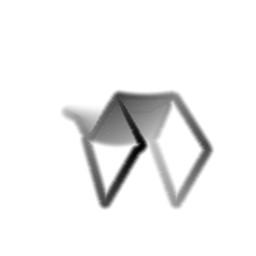}&
    \includegraphics[width=0.09\columnwidth,trim=30 30 30 30, clip]{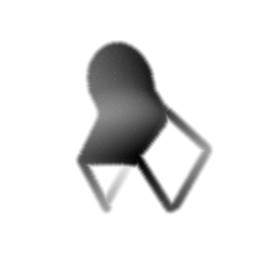}&
    \includegraphics[width=0.09\columnwidth,trim=30 30 30 30, clip]{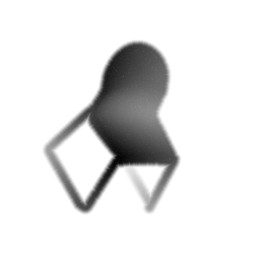}\\
\end{tabular}
}
\caption{Completion comparison visualized in rendered depth maps.}
\label{fig:depthmaps}
\end{figure*}

\begin{figure*}
\center
\setlength\tabcolsep{0pt}
{
\renewcommand{\arraystretch}{0.0}
\small
\begin{tabular}{@{}cc@{}}
    Input & Completion results\\
    
    \includegraphics[width=0.5\columnwidth,trim=40 40 40 40, clip]{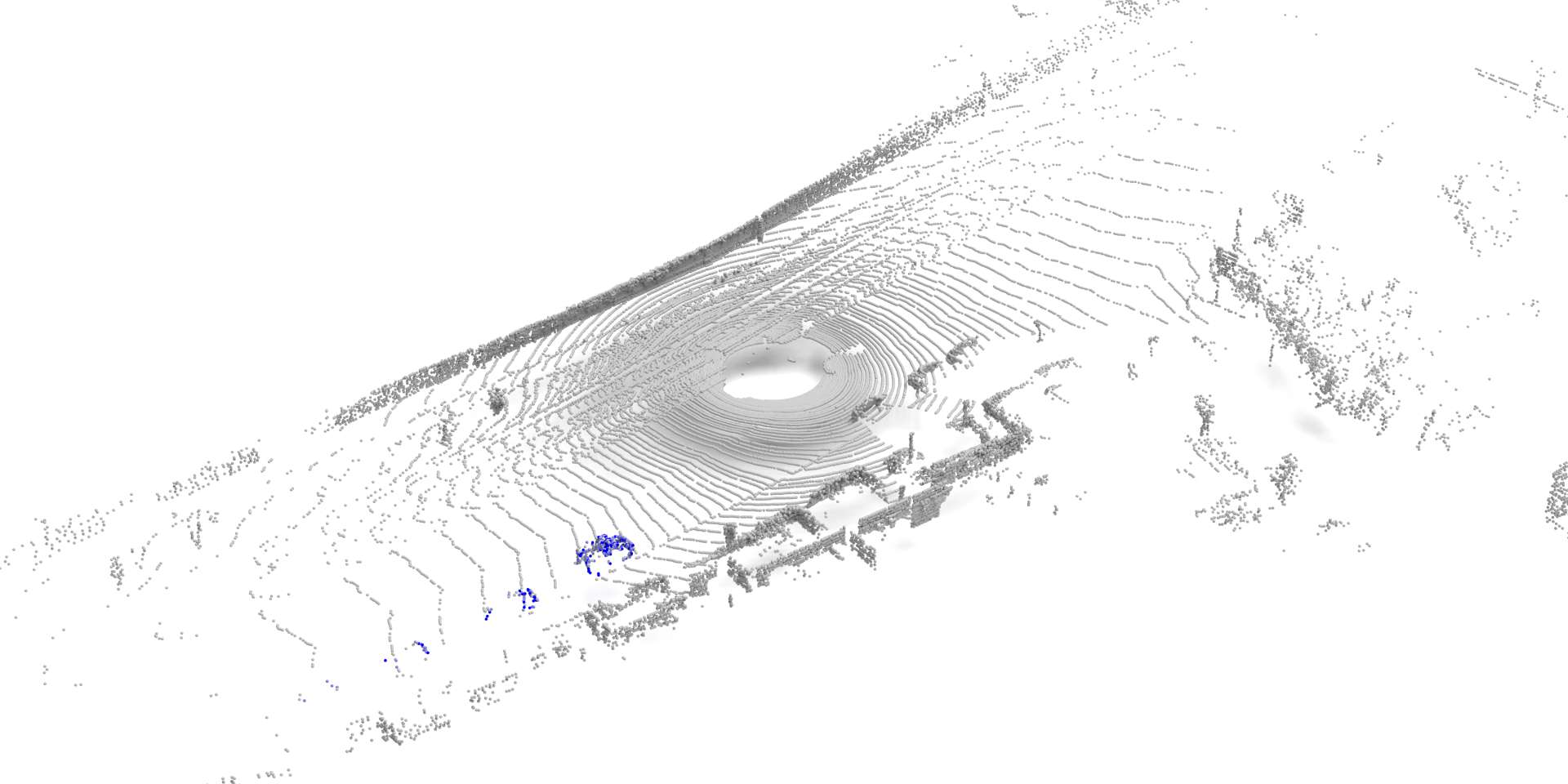}&
    \includegraphics[width=0.5\columnwidth,trim=40 40 40 40, clip]{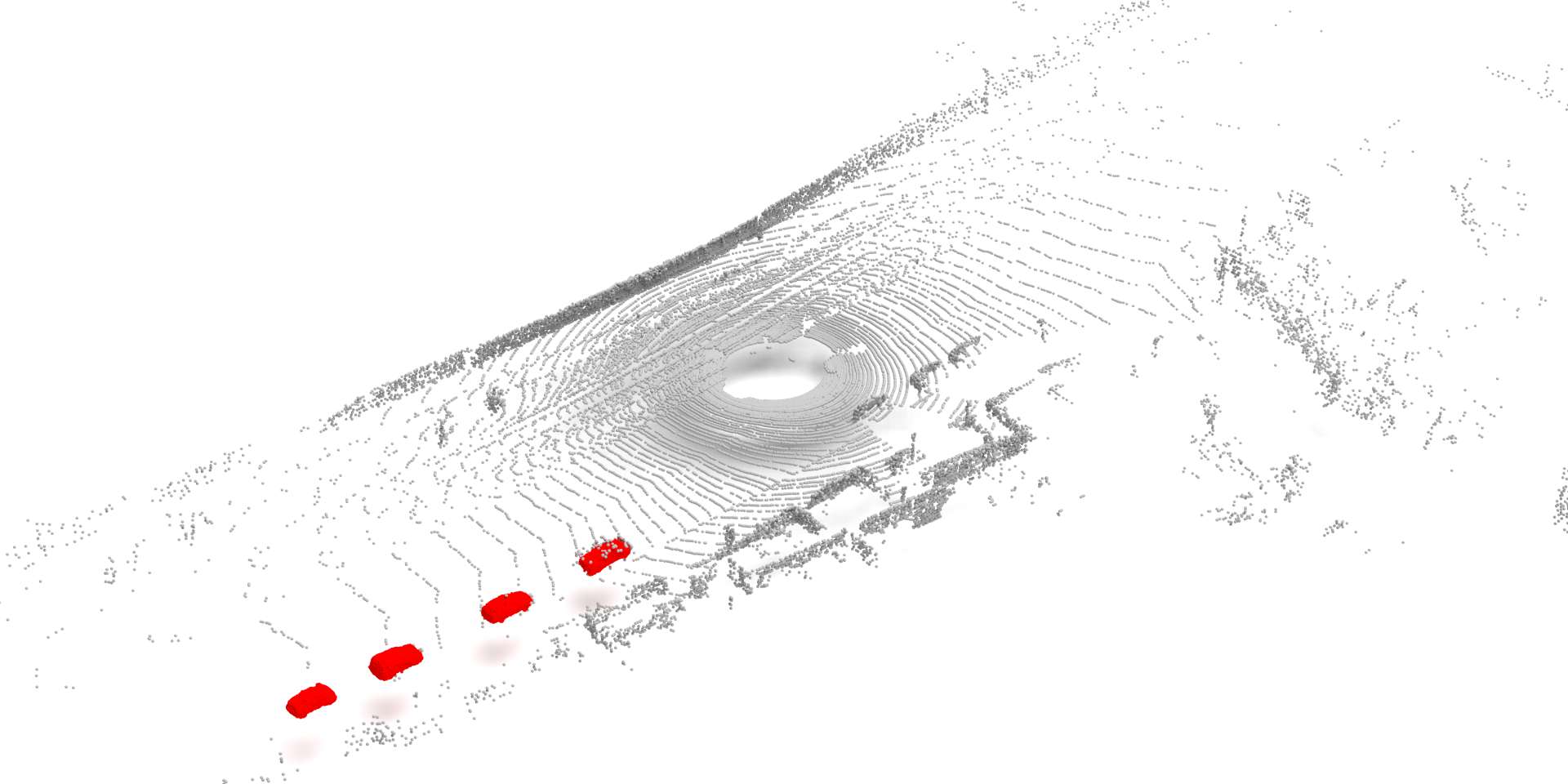}\\
    
    \includegraphics[width=0.5\columnwidth,trim=40 40 40 40, clip]{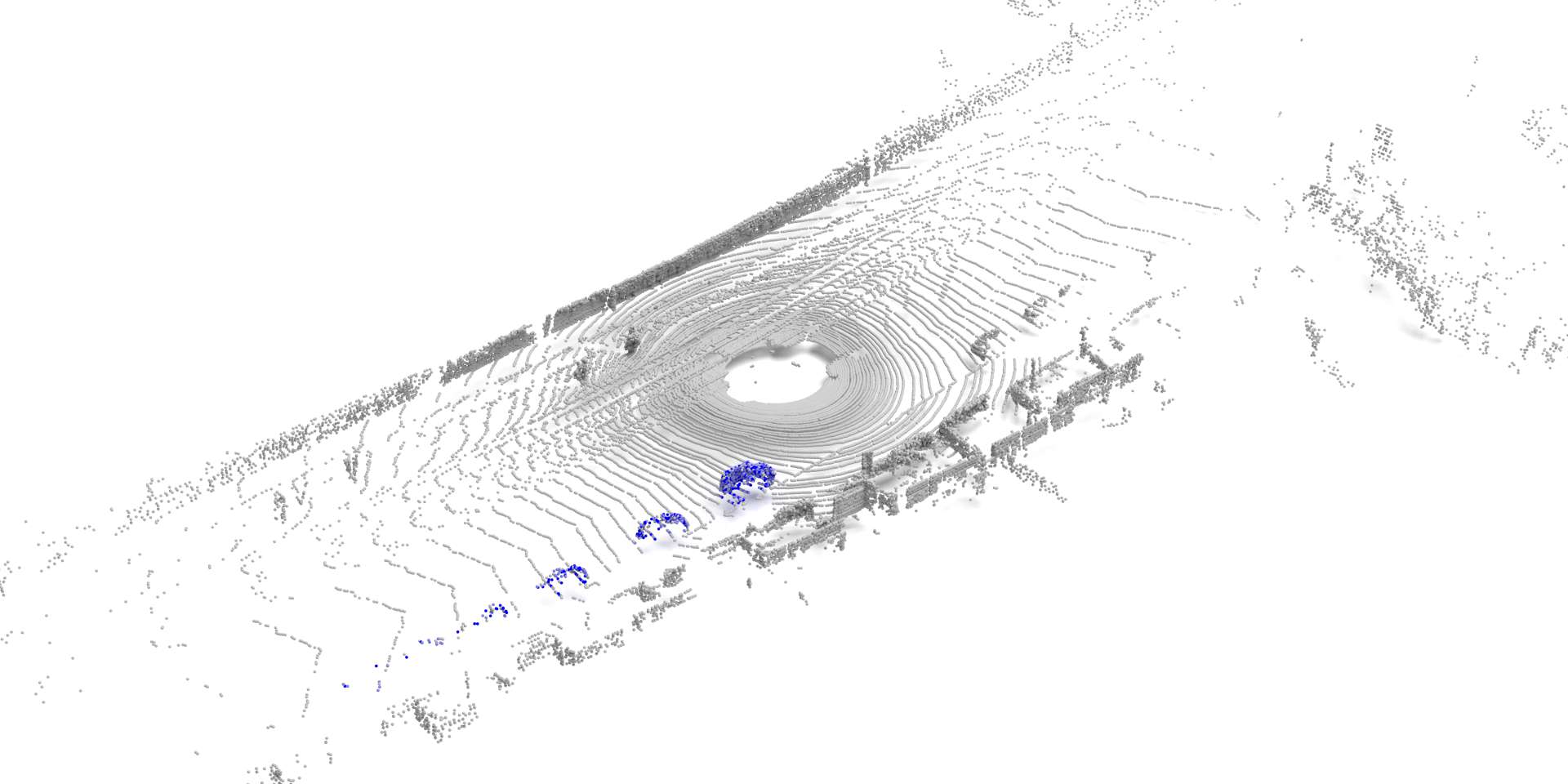}&
    \includegraphics[width=0.5\columnwidth,trim=40 40 40 40, clip]{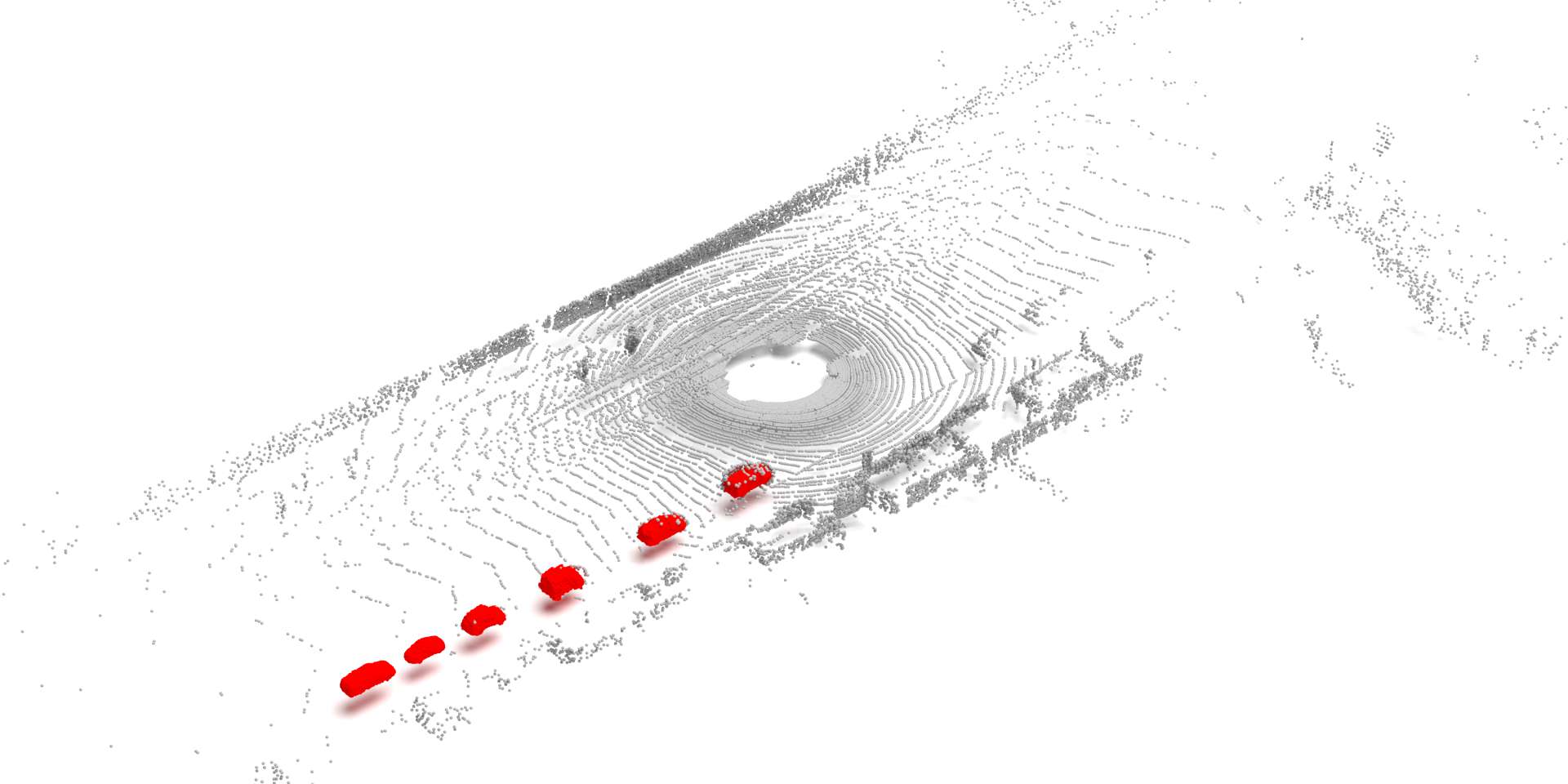}\\
    
    \includegraphics[width=0.5\columnwidth,trim=40 40 40 40, clip]{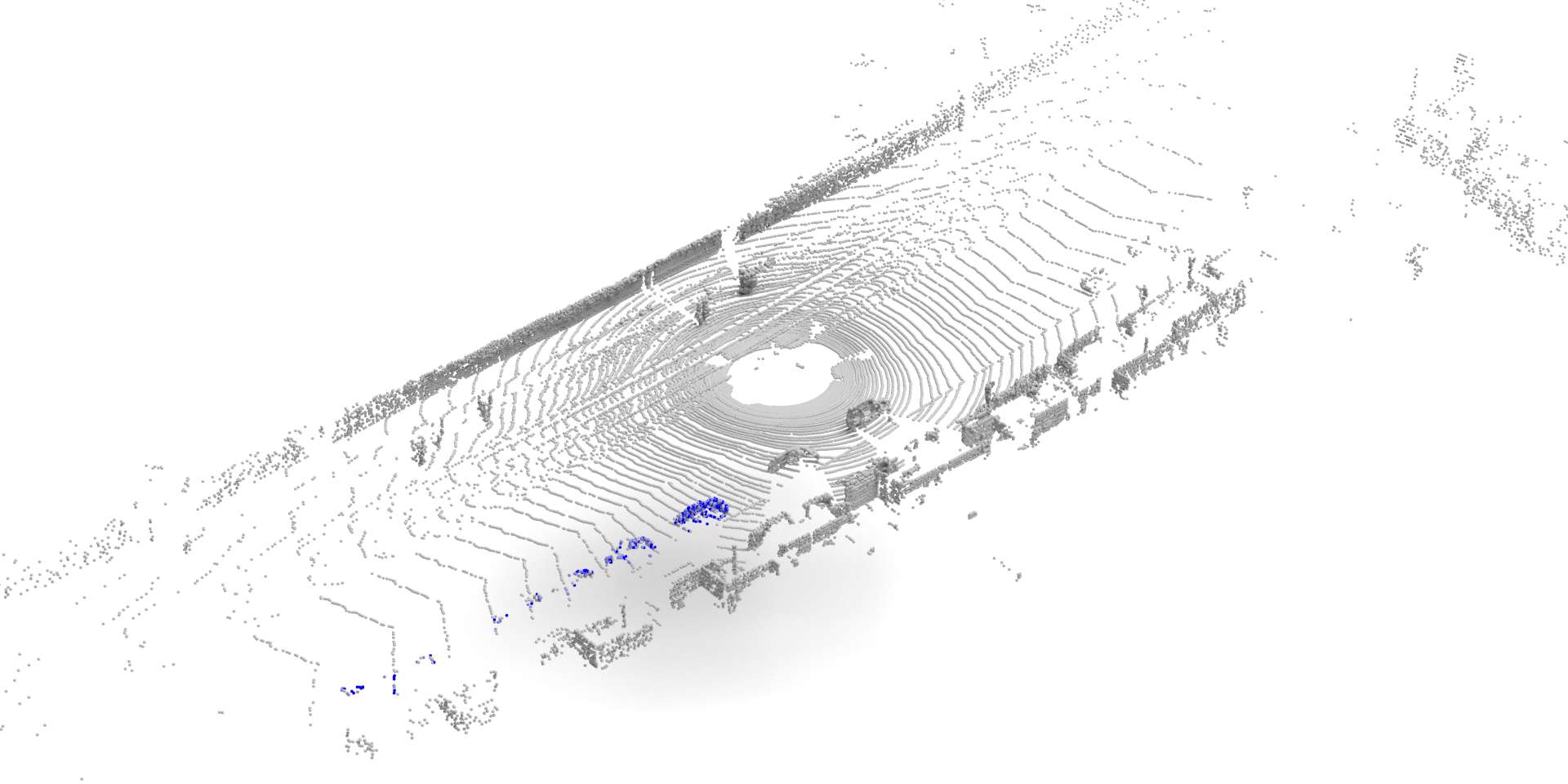}&
    \includegraphics[width=0.5\columnwidth,trim=40 40 40 40, clip]{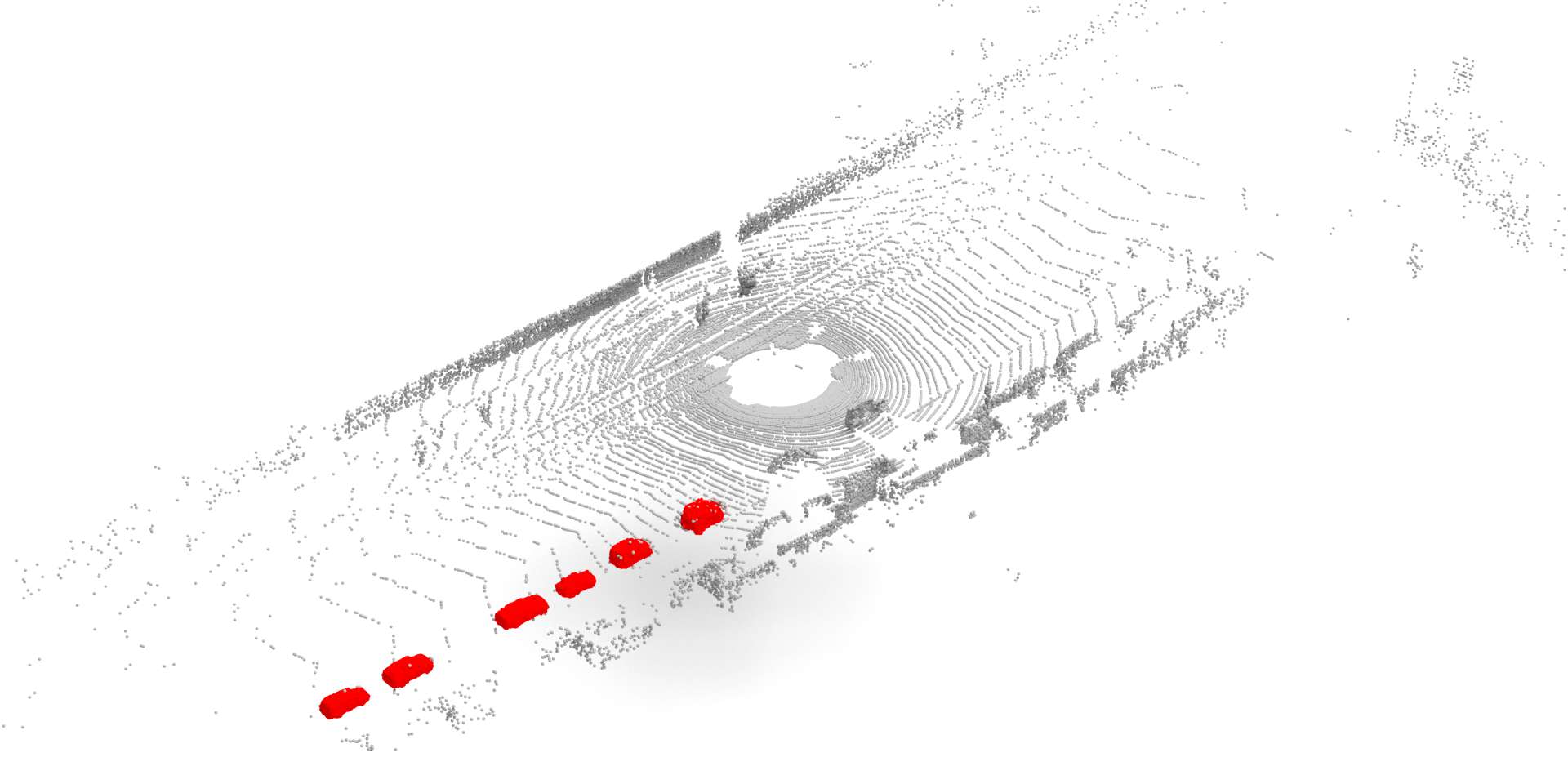}\\
    
    \includegraphics[width=0.5\columnwidth,trim=40 40 40 40, clip]{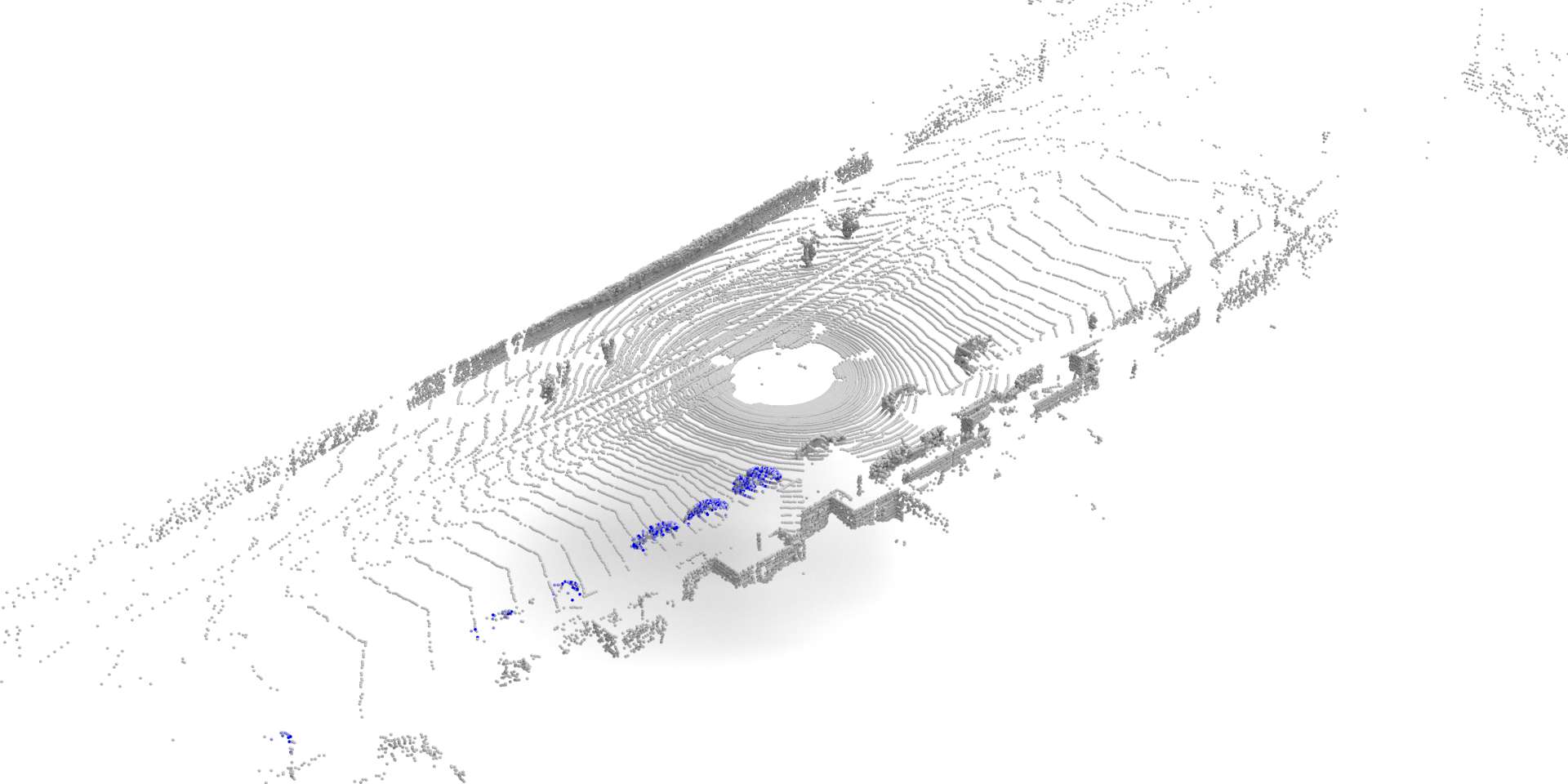}&
    \includegraphics[width=0.5\columnwidth,trim=40 40 40 40, clip]{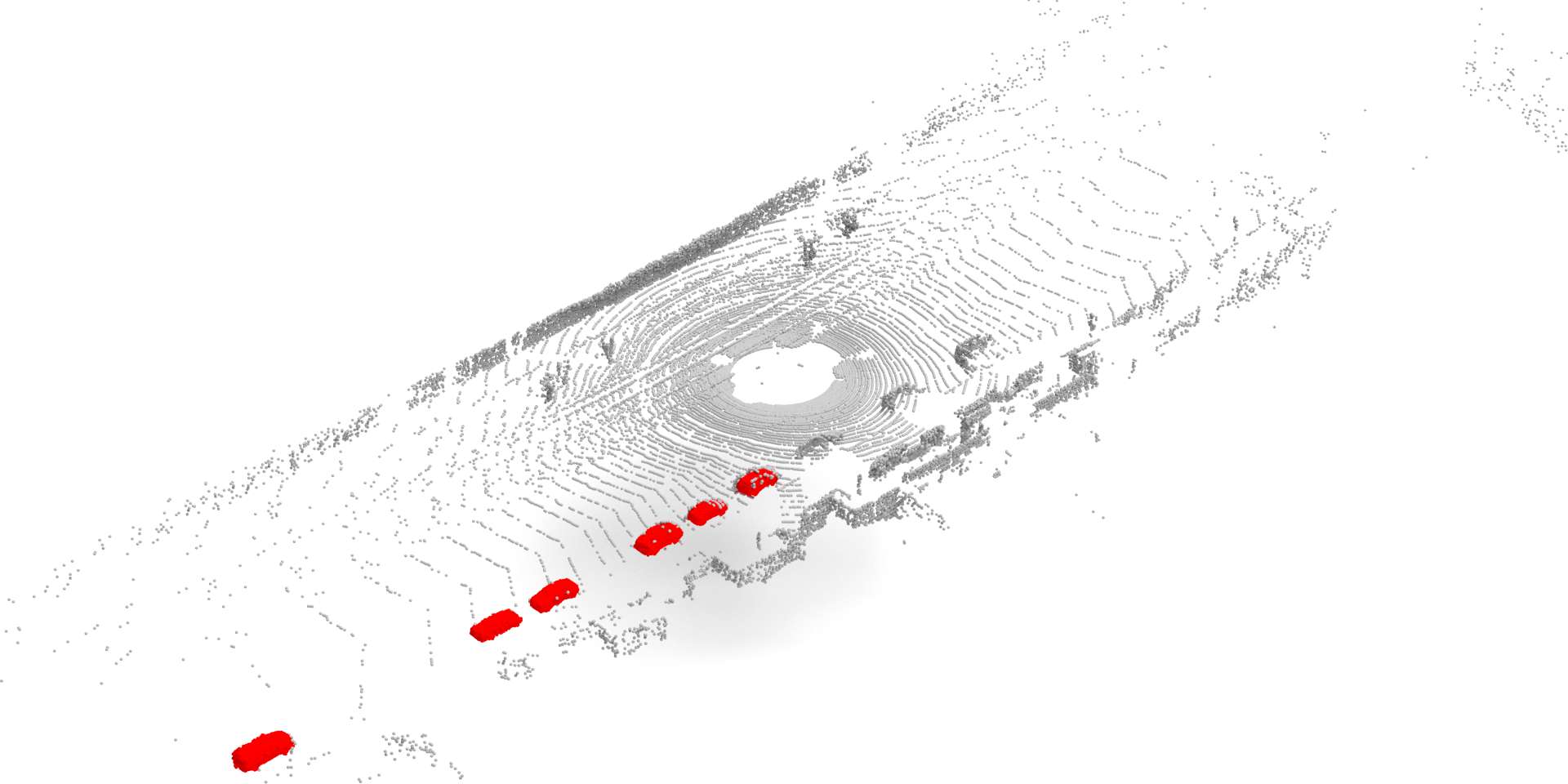}\\
       
\end{tabular}
}
\caption{Visualized car completion results based on real-world LiDAR scans from the KITTI dataset \cite{geiger2013vision}. The left frames show the input partial points in blue, the right frames show the completed point clouds of cars in red.}
\label{fig:kitti}
\end{figure*}

We show more qualitative completion results in Figures~\ref{fig:qualitative1},~\ref{fig:qualitative2},~\ref{fig:qualitative3}.
We also demonstrate more qualitative comparisons in Figure~\ref{fig:depthmaps} with respect to the rendered depth maps.
Moreover, in Figure~\ref{fig:kitti}, we illustrate the car completion results based on the real-world LiDAR scans from KITTI.

\begin{figure*}[t]
\center
\setlength\tabcolsep{0pt}
{
\renewcommand{\arraystretch}{0.0}
\small
\begin{tabular}{@{}rcccccc@{}}
    & Input & $w_{adv}=0.0$ & $w_{adv}=0.1$ & $w_{adv}=5$ & Point-based & Groundtruth\\
    
    \raisebox{1.0\height}{\rotatebox{90}{Car}}&
    \includegraphics[width=0.15\columnwidth,trim=30 30 30 30, clip]{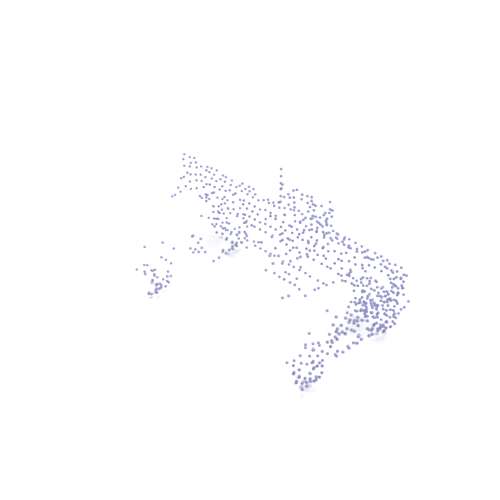}&
    \includegraphics[width=0.15\columnwidth,trim=30 30 30 30, clip]{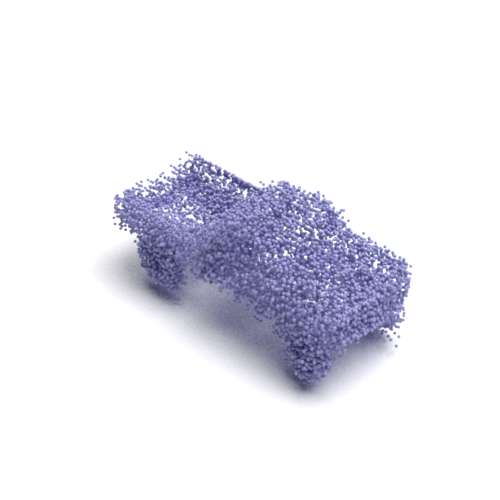}&
    \includegraphics[width=0.15\columnwidth,trim=30 30 30 30, clip]{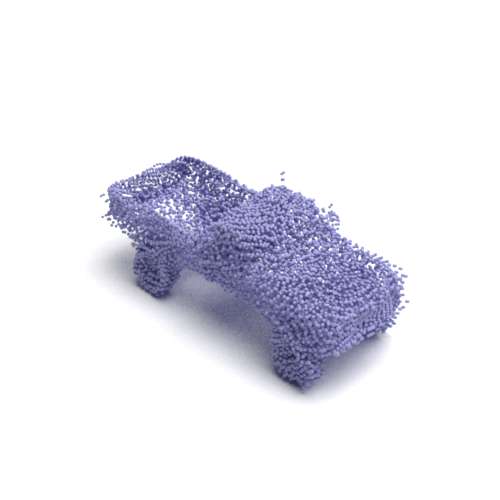}&
    \includegraphics[width=0.15\columnwidth,trim=30 30 30 30, clip]{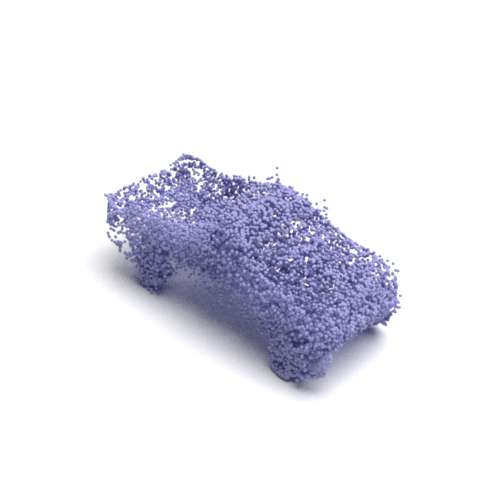}&
    \includegraphics[width=0.15\columnwidth,trim=30 30 30 30, clip]{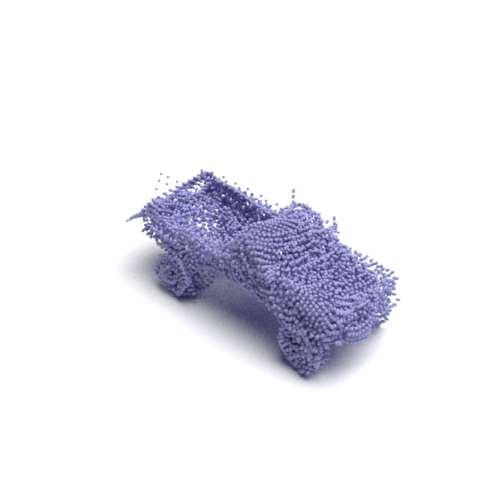}&
    \includegraphics[width=0.15\columnwidth,trim=30 30 30 30, clip]{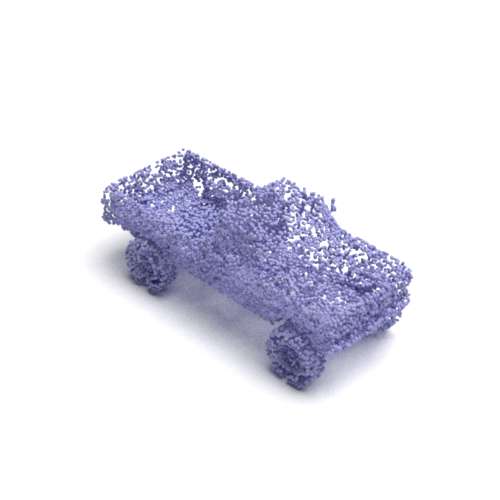}\\
    
    \raisebox{1\height}{\rotatebox{90}{Airplane}}&
    \includegraphics[width=0.15\columnwidth,trim=30 30 30 30, clip]{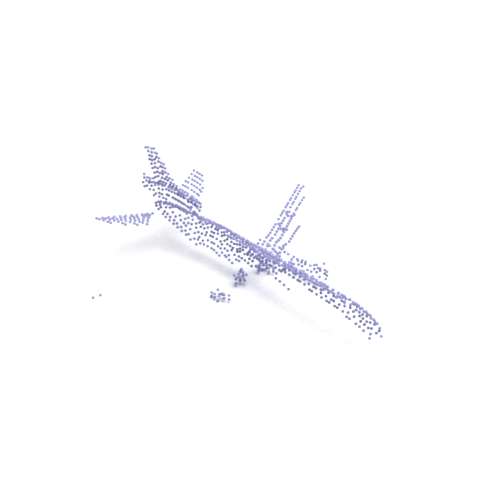}&
    \includegraphics[width=0.15\columnwidth,trim=30 30 30 30, clip]{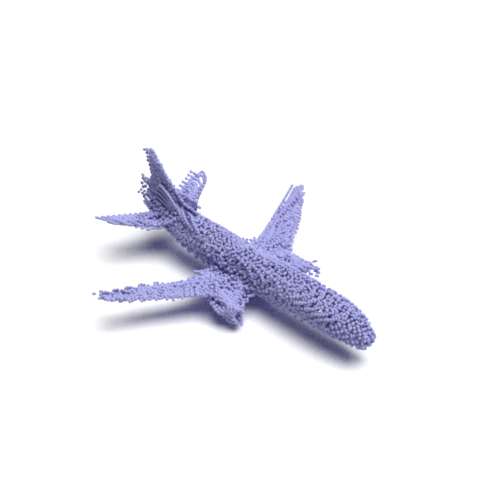}&
    \includegraphics[width=0.15\columnwidth,trim=30 30 30 30, clip]{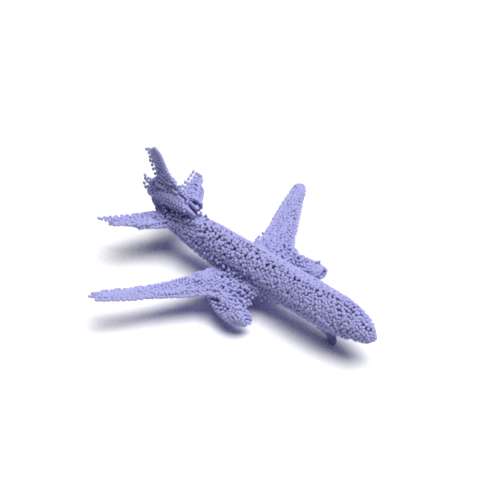}&
    \includegraphics[width=0.15\columnwidth,trim=30 30 30 30, clip]{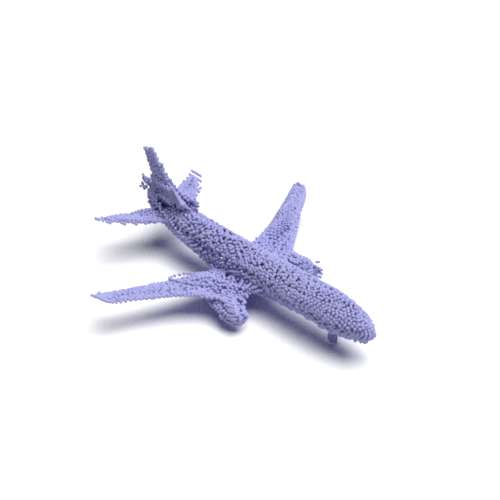}&
    \includegraphics[width=0.15\columnwidth,trim=30 30 30 30, clip]{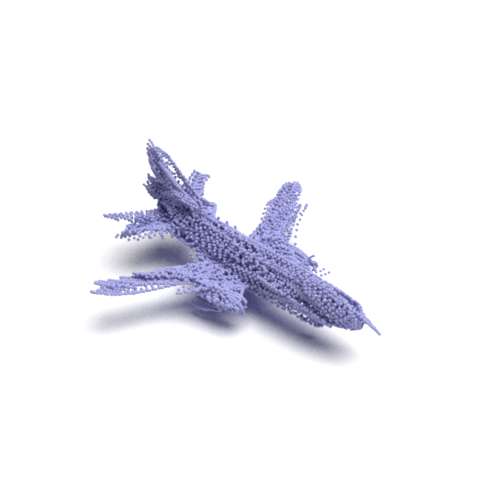}&
    \includegraphics[width=0.15\columnwidth,trim=30 30 30 30, clip]{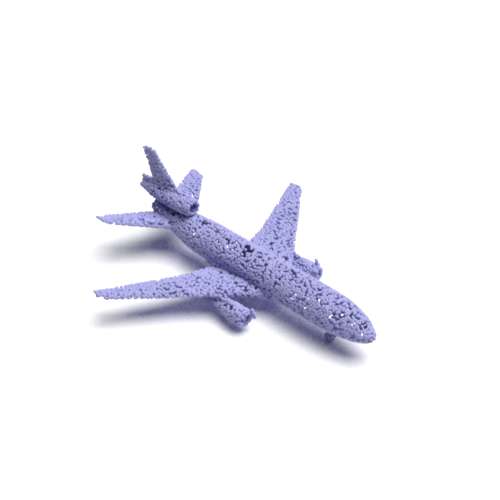}\\
    
    \raisebox{1.0\height}{\rotatebox{90}{Chair}}&
    \includegraphics[width=0.15\columnwidth,trim=30 30 30 30, clip]{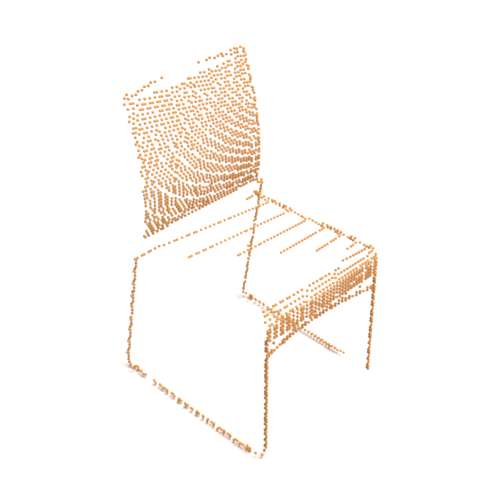}&
    \includegraphics[width=0.15\columnwidth,trim=30 30 30 30, clip]{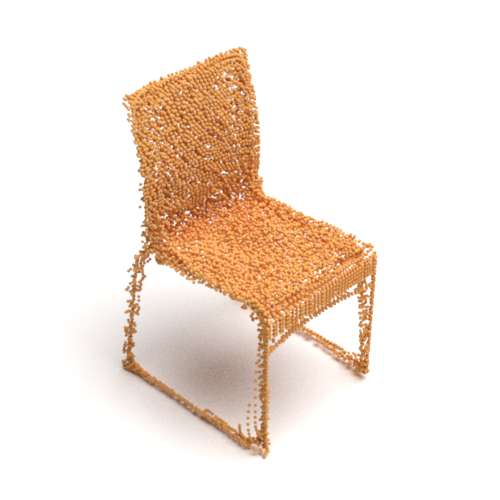}&
    \includegraphics[width=0.15\columnwidth,trim=30 30 30 30, clip]{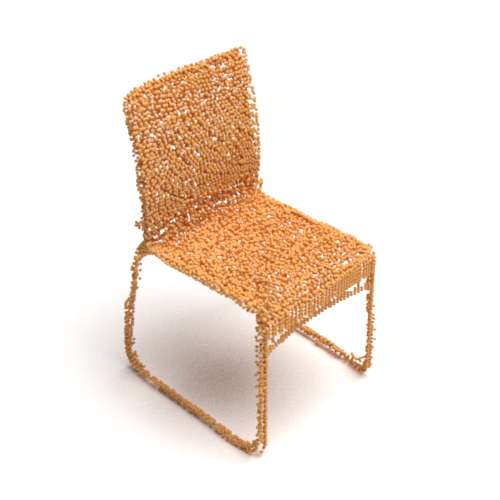}&
    \includegraphics[width=0.15\columnwidth,trim=30 30 30 30, clip]{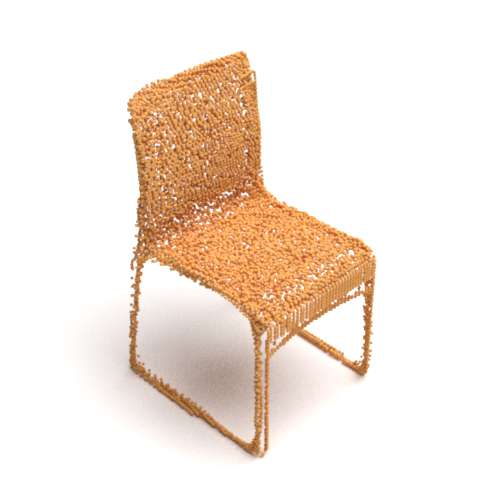}&
    \includegraphics[width=0.15\columnwidth,trim=30 30 30 30, clip]{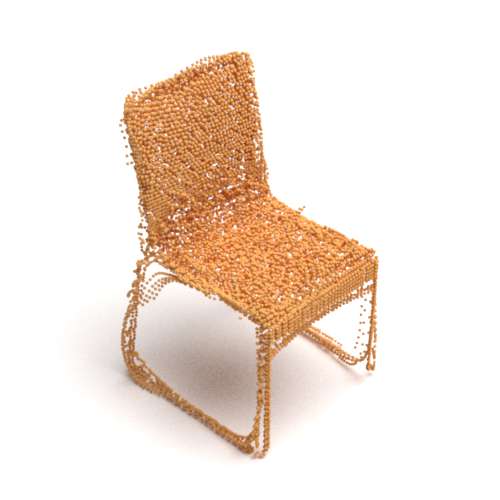}&
    \includegraphics[width=0.15\columnwidth,trim=30 30 30 30, clip]{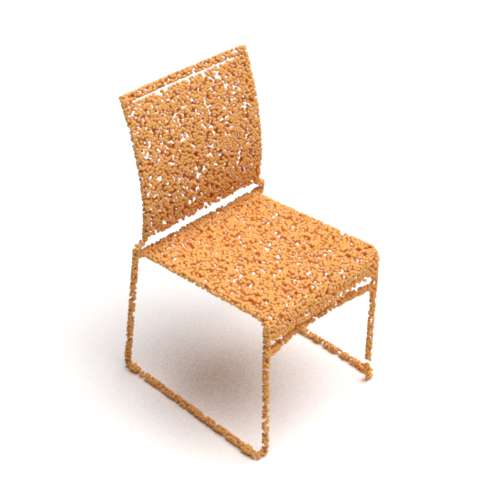}\\
    
    \raisebox{1\height}{\rotatebox{90}{Lamp}}&
    \includegraphics[width=0.15\columnwidth,trim=30 30 30 30, clip]{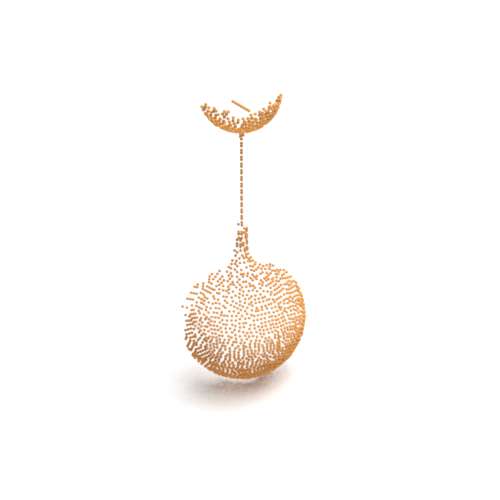}&
    \includegraphics[width=0.15\columnwidth,trim=30 30 30 30, clip]{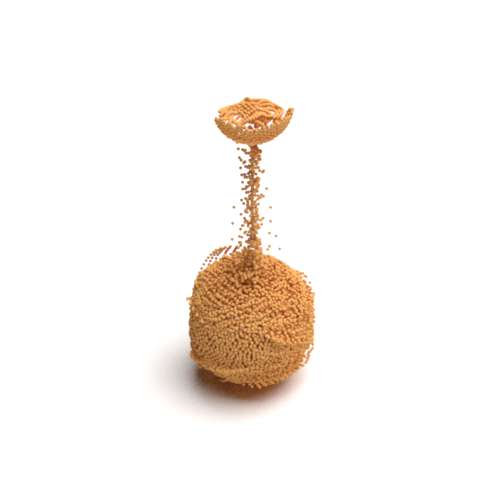}&
    \includegraphics[width=0.15\columnwidth,trim=30 30 30 30, clip]{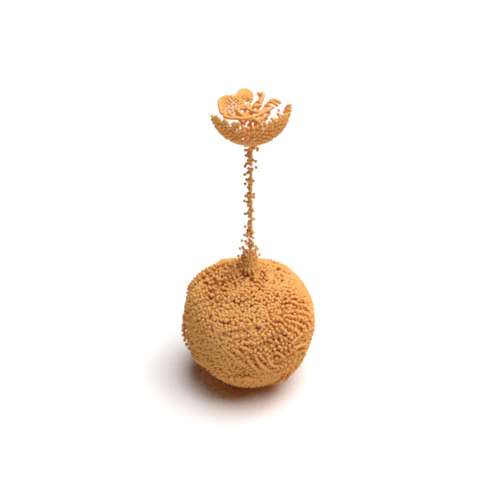}&
    \includegraphics[width=0.15\columnwidth,trim=30 30 30 30, clip]{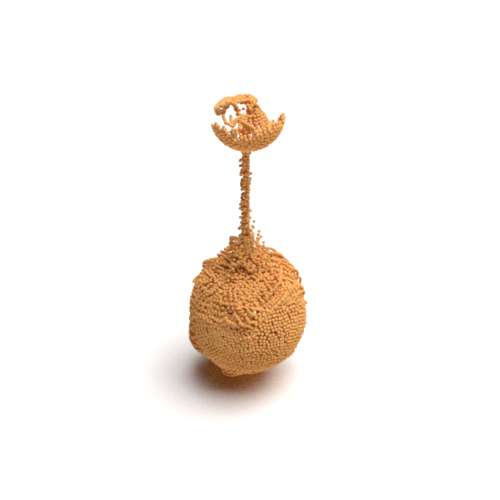}&
    \includegraphics[width=0.15\columnwidth,trim=30 30 30 30, clip]{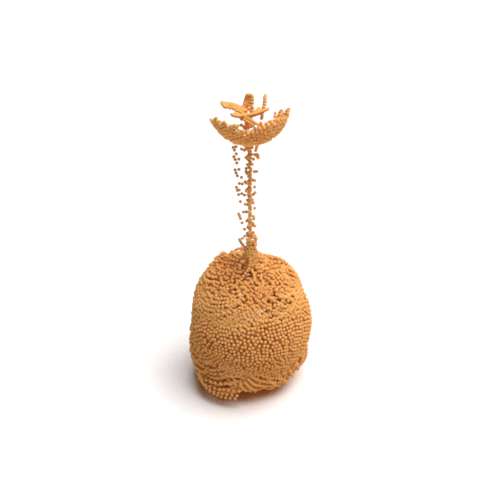}&
    \includegraphics[width=0.15\columnwidth,trim=30 30 30 30, clip]{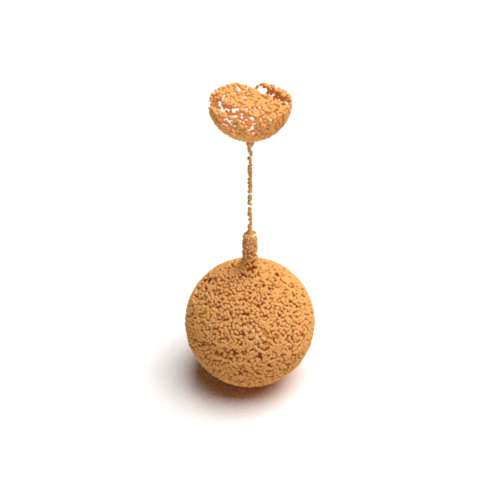}\\
    
    \raisebox{1\height}{\rotatebox{90}{Sofa}}&
    \includegraphics[width=0.15\columnwidth,trim=30 30 30 30, clip]{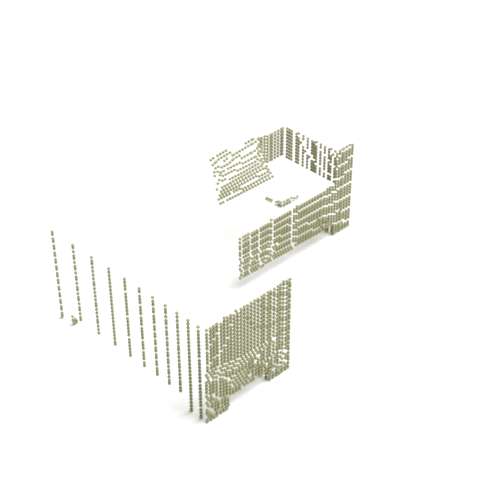}&
    \includegraphics[width=0.15\columnwidth,trim=30 30 30 30, clip]{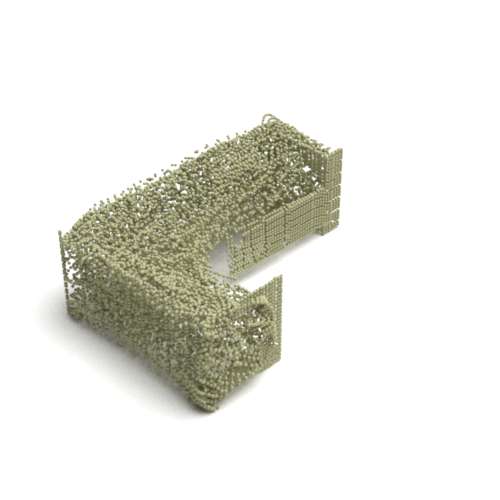}&
    \includegraphics[width=0.15\columnwidth,trim=30 30 30 30, clip]{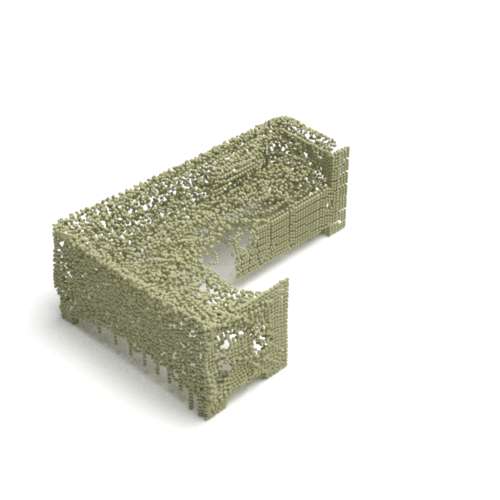}&
    \includegraphics[width=0.15\columnwidth,trim=30 30 30 30, clip]{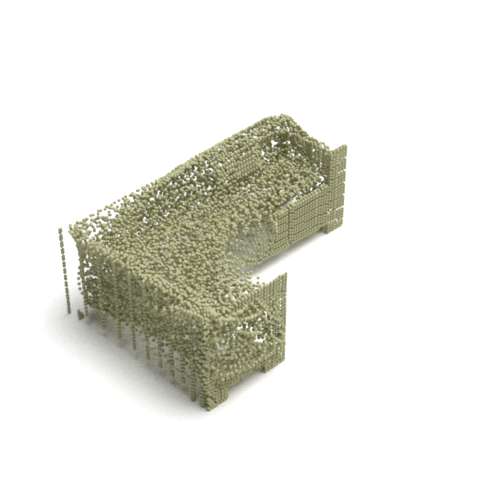}&
    \includegraphics[width=0.15\columnwidth,trim=30 30 30 30, clip]{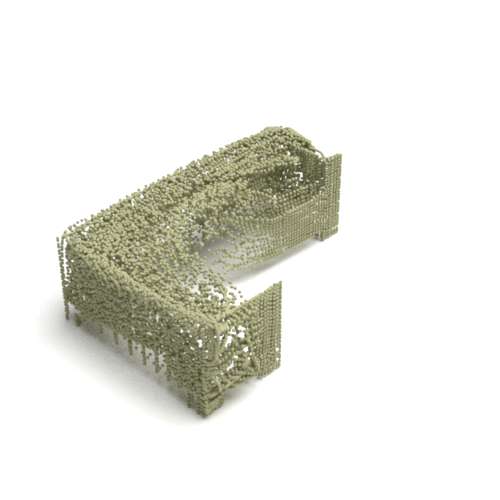}&
    \includegraphics[width=0.15\columnwidth,trim=30 30 30 30, clip]{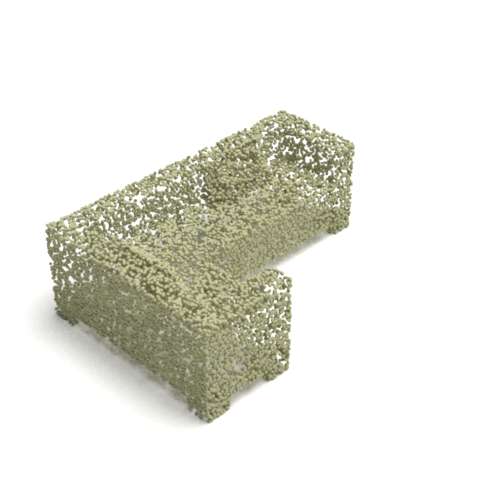}\\
    
    \raisebox{1\height}{\rotatebox{90}{Cabinet}}&
    \includegraphics[width=0.15\columnwidth,trim=30 30 30 30, clip]{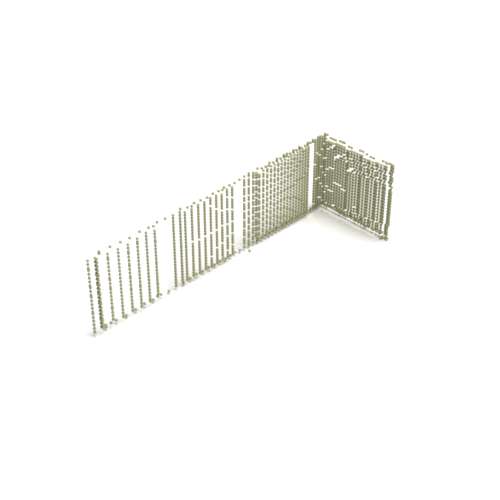}&
    \includegraphics[width=0.15\columnwidth,trim=30 30 30 30, clip]{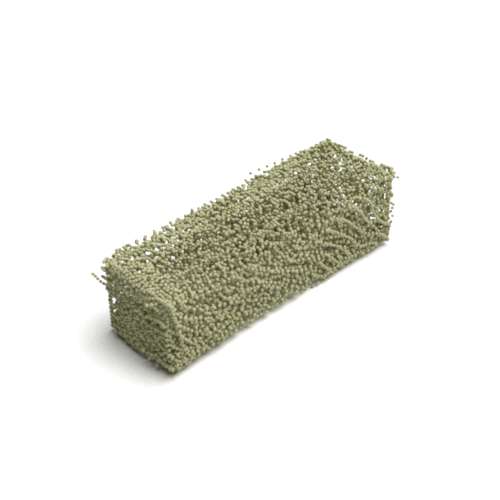}&
    \includegraphics[width=0.15\columnwidth,trim=30 30 30 30, clip]{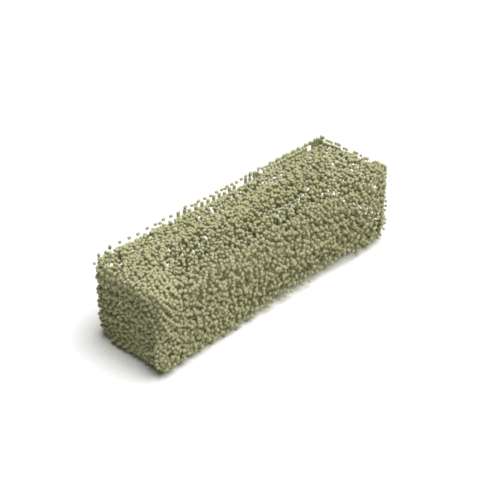}&
    \includegraphics[width=0.15\columnwidth,trim=30 30 30 30, clip]{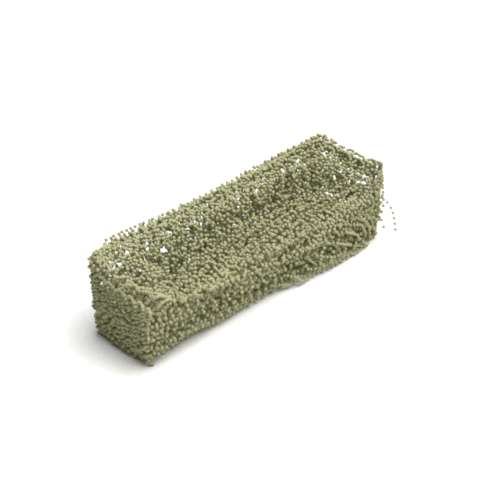}&
    \includegraphics[width=0.15\columnwidth,trim=30 30 30 30, clip]{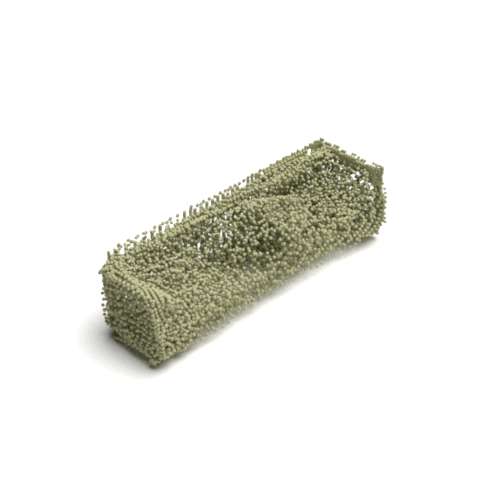}&
    \includegraphics[width=0.15\columnwidth,trim=30 30 30 30, clip]{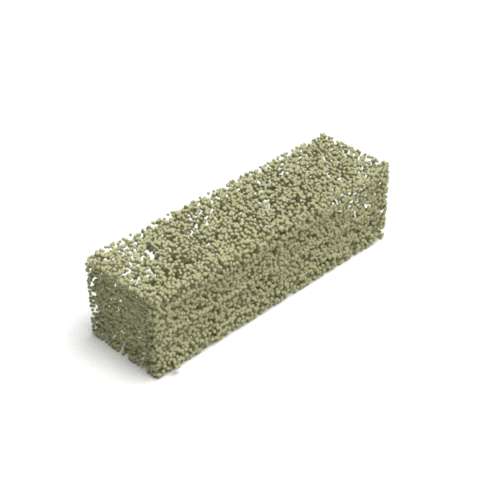}\\
       
\end{tabular}
}
\caption{Visualized comparison of models with or without rendering supervisions. In comparison, the proposed rendered discriminator is more capable to examine the local details than the point-based discriminator.}
\label{fig:image_domain}
\end{figure*}

\section{Rendering with Different Point Cloud Resolutions}
\begin{figure*}[t]
\center
\setlength\tabcolsep{2pt}
{
\renewcommand{\arraystretch}{1.0}
\small
\begin{tabular}{@{}rcccccccc@{}}
    \toprule
    & view1 & view2 & view3 & view4 & view5 & view6 & view7 & view8 \\
    \midrule
    \raisebox{\height}{\rotatebox{90}{2048}}&
    \includegraphics[width=0.11\columnwidth,trim=30 30 30 30, clip]{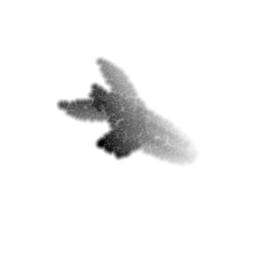}& \includegraphics[width=0.11\columnwidth,trim=30 30 30 30, clip]{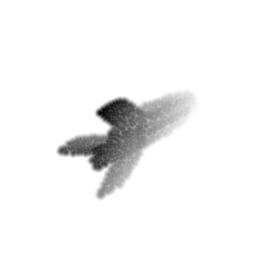}&
    \includegraphics[width=0.11\columnwidth,trim=30 30 30 30, clip]{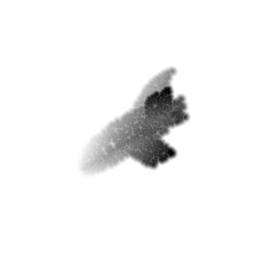}&
    \includegraphics[width=0.11\columnwidth,trim=30 30 30 30, clip]{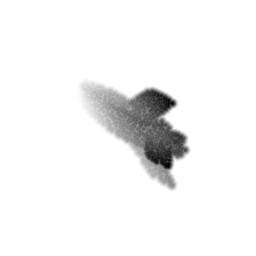}&
    \includegraphics[width=0.11\columnwidth,trim=30 30 30 30, clip]{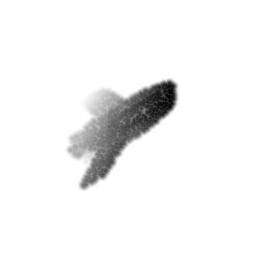}& \includegraphics[width=0.11\columnwidth,trim=30 30 30 30, clip]{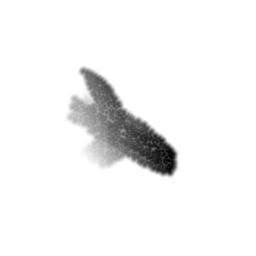}&
    \includegraphics[width=0.11\columnwidth,trim=30 30 30 30, clip]{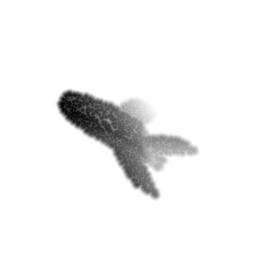}&
    \includegraphics[width=0.11\columnwidth,trim=30 30 30 30, clip]{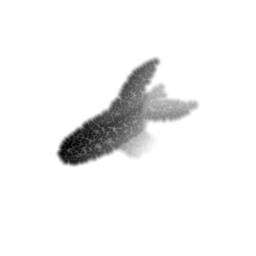}
    \\
    
    \raisebox{\height}{\rotatebox{90}{4096}}&
    \includegraphics[width=0.11\columnwidth,trim=30 30 30 30, clip]{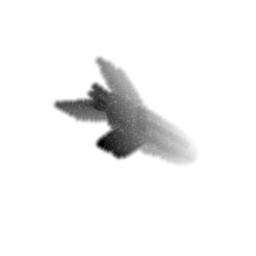}& \includegraphics[width=0.11\columnwidth,trim=30 30 30 30, clip]{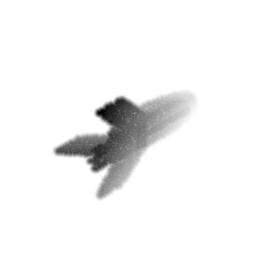}&
    \includegraphics[width=0.11\columnwidth,trim=30 30 30 30, clip]{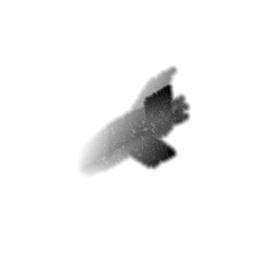}&
    \includegraphics[width=0.11\columnwidth,trim=30 30 30 30, clip]{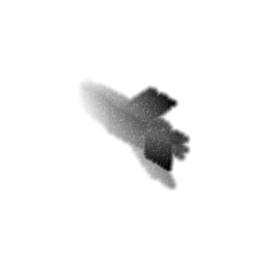}&
    \includegraphics[width=0.11\columnwidth,trim=30 30 30 30, clip]{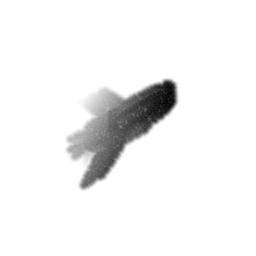}& \includegraphics[width=0.11\columnwidth,trim=30 30 30 30, clip]{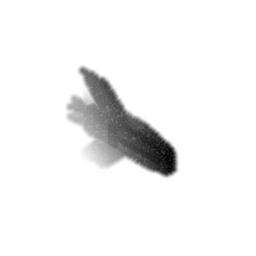}&
    \includegraphics[width=0.11\columnwidth,trim=30 30 30 30, clip]{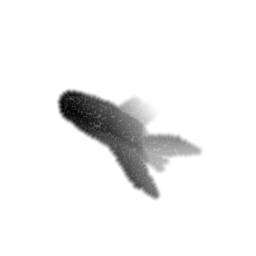}&
    \includegraphics[width=0.11\columnwidth,trim=30 30 30 30, clip]{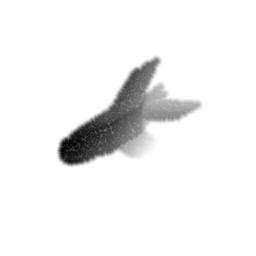}\\
       
    \raisebox{\height}{\rotatebox{90}{8192}}&
    \includegraphics[width=0.11\columnwidth,trim=30 30 30 30, clip]{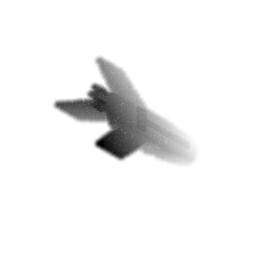}& \includegraphics[width=0.11\columnwidth,trim=30 30 30 30, clip]{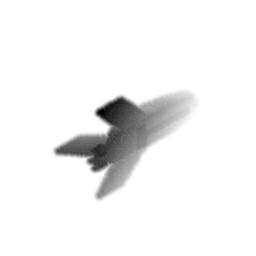}&
    \includegraphics[width=0.11\columnwidth,trim=30 30 30 30, clip]{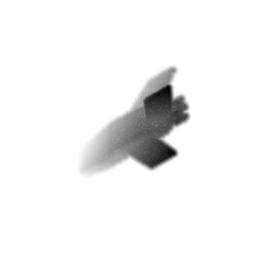}&
    \includegraphics[width=0.11\columnwidth,trim=30 30 30 30, clip]{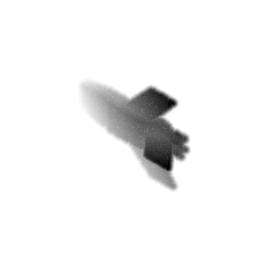}&
    \includegraphics[width=0.11\columnwidth,trim=30 30 30 30, clip]{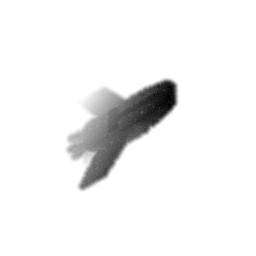}& \includegraphics[width=0.11\columnwidth,trim=30 30 30 30, clip]{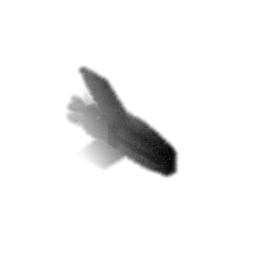}&
    \includegraphics[width=0.11\columnwidth,trim=30 30 30 30, clip]{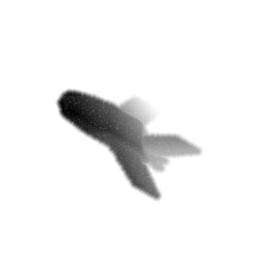}&
    \includegraphics[width=0.11\columnwidth,trim=30 30 30 30, clip]{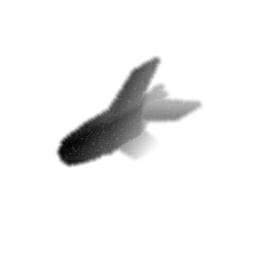}\\
    
    \raisebox{\height}{\rotatebox{90}{16384}}&
    \includegraphics[width=0.11\columnwidth,trim=30 30 30 30, clip]{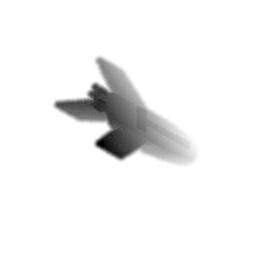}&
    \includegraphics[width=0.11\columnwidth,trim=30 30 30 30, clip]{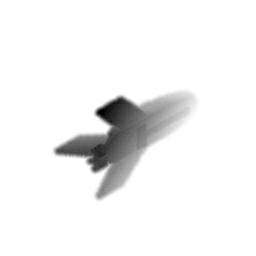}&
    \includegraphics[width=0.11\columnwidth,trim=30 30 30 30,  clip]{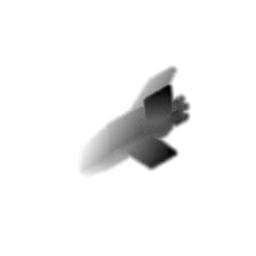}&
    \includegraphics[width=0.11\columnwidth,trim=30 30 30 30, clip]{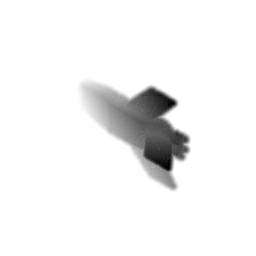}&
    \includegraphics[width=0.11\columnwidth,trim=30 30 30 30, clip]{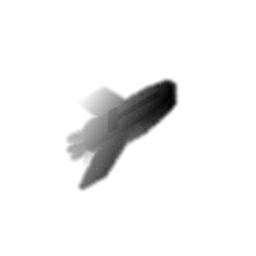}&
    \includegraphics[width=0.11\columnwidth,trim=30 30 30 30, clip]{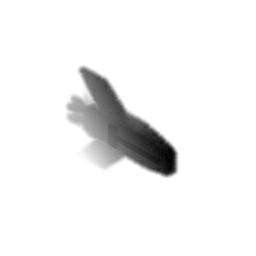}&
    \includegraphics[width=0.11\columnwidth,trim=30 30 30 30, clip]{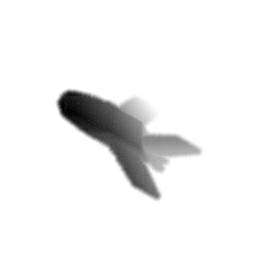}&
    \includegraphics[width=0.11\columnwidth,trim=30 30 30 30, clip]{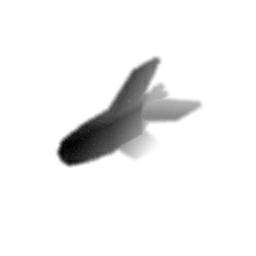}\\

    \midrule
    \raisebox{\height}{\rotatebox{90}{2048}}&    
    \includegraphics[width=0.11\columnwidth,trim=30 30 30 30, clip]{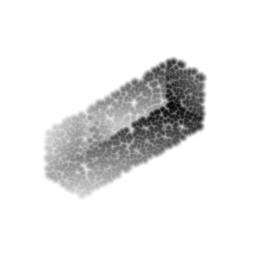}& 
    \includegraphics[width=0.11\columnwidth,trim=30 30 30 30, clip]{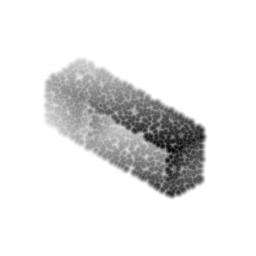}&
    \includegraphics[width=0.11\columnwidth,trim=30 30 30 30, clip]{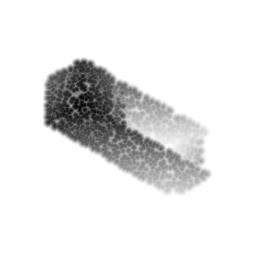}&
    \includegraphics[width=0.11\columnwidth,trim=30 30 30 30, clip]{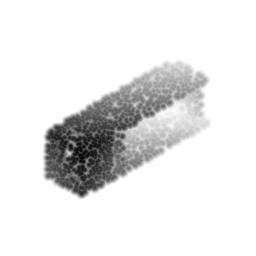}&
    \includegraphics[width=0.11\columnwidth,trim=30 30 30 30, clip]{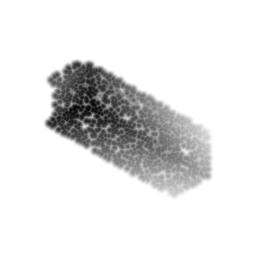}& 
    \includegraphics[width=0.11\columnwidth,trim=30 30 30 30, clip]{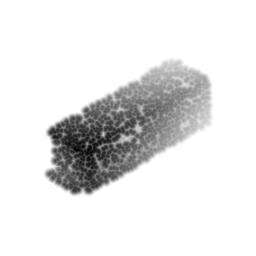}&
    \includegraphics[width=0.11\columnwidth,trim=30 30 30 30, clip]{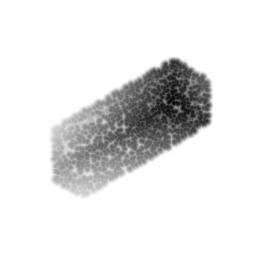}&
    \includegraphics[width=0.11\columnwidth,trim=30 30 30 30, clip]{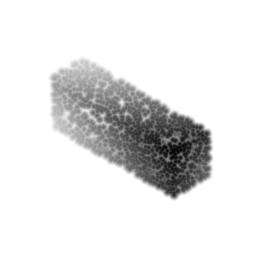}\\
    
    \raisebox{\height}{\rotatebox{90}{4096}}&
    \includegraphics[width=0.11\columnwidth,trim=30 30 30 30, clip]{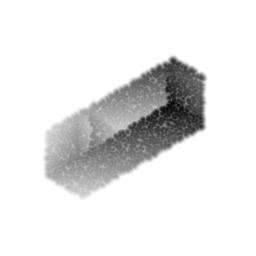}& 
    \includegraphics[width=0.11\columnwidth,trim=30 30 30 30, clip]{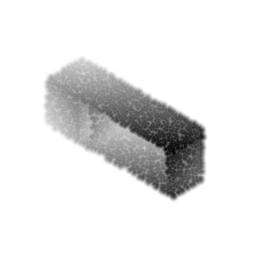}&
    \includegraphics[width=0.11\columnwidth,trim=30 30 30 30, clip]{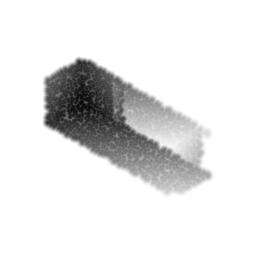}&
    \includegraphics[width=0.11\columnwidth,trim=30 30 30 30, clip]{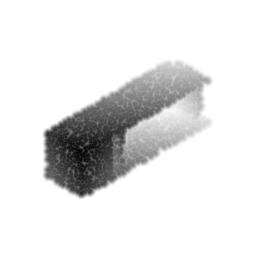}&
    \includegraphics[width=0.11\columnwidth,trim=30 30 30 30, clip]{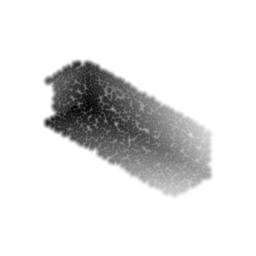}& 
    \includegraphics[width=0.11\columnwidth,trim=30 30 30 30, clip]{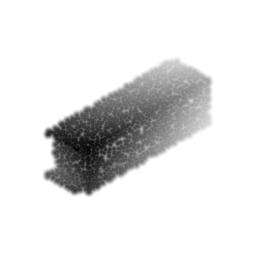}&
    \includegraphics[width=0.11\columnwidth,trim=30 30 30 30, clip]{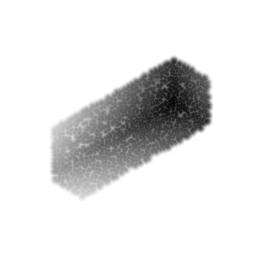}&
    \includegraphics[width=0.11\columnwidth,trim=30 30 30 30, clip]{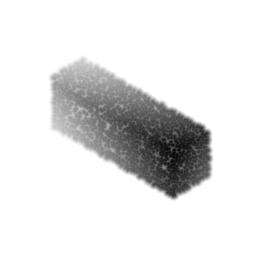}\\
    
    \raisebox{\height}{\rotatebox{90}{8192}}&
    \includegraphics[width=0.11\columnwidth,trim=30 30 30 30, clip]{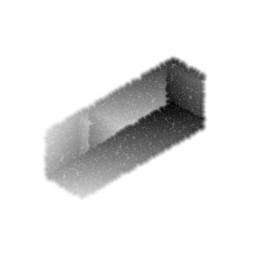}& 
    \includegraphics[width=0.11\columnwidth,trim=30 30 30 30, clip]{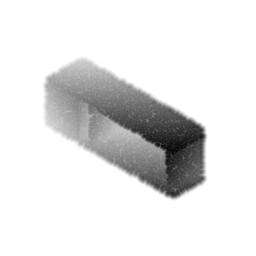}&
    \includegraphics[width=0.11\columnwidth,trim=30 30 30 30, clip]{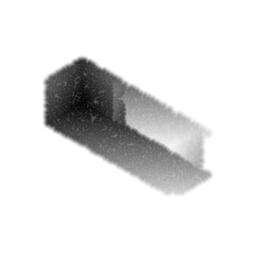}&
    \includegraphics[width=0.11\columnwidth,trim=30 30 30 30, clip]{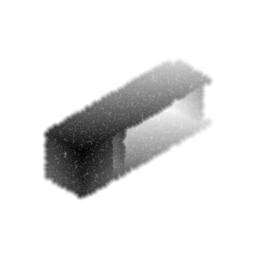}&
    \includegraphics[width=0.11\columnwidth,trim=30 30 30 30, clip]{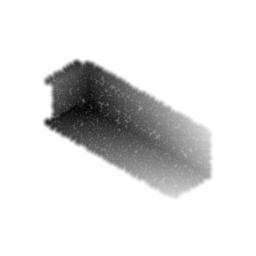}& 
    \includegraphics[width=0.11\columnwidth,trim=30 30 30 30, clip]{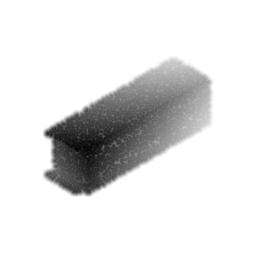}&
    \includegraphics[width=0.11\columnwidth,trim=30 30 30 30, clip]{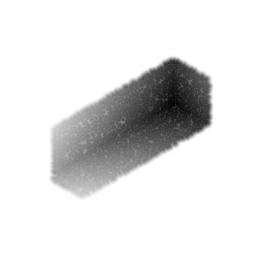}&
    \includegraphics[width=0.11\columnwidth,trim=30 30 30 30, clip]{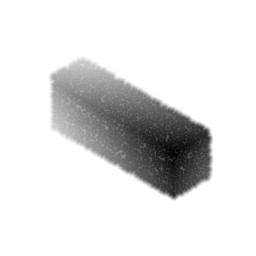}\\
    
    \raisebox{0.2\height}{\rotatebox{90}{16384}}&
    \includegraphics[width=0.11\columnwidth,trim=30 30 30 30, clip]{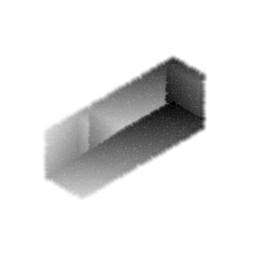}& 
    \includegraphics[width=0.11\columnwidth,trim=30 30 30 30, clip]{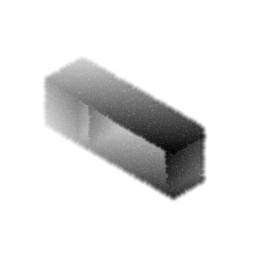}&
    \includegraphics[width=0.11\columnwidth,trim=30 30 30 30, clip]{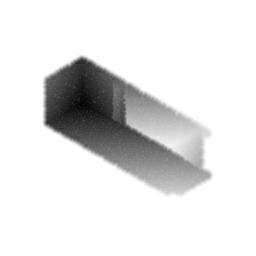}&
    \includegraphics[width=0.11\columnwidth,trim=30 30 30 30, clip]{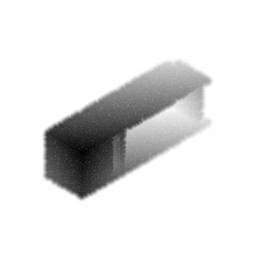}&
    \includegraphics[width=0.11\columnwidth,trim=30 30 30 30, clip]{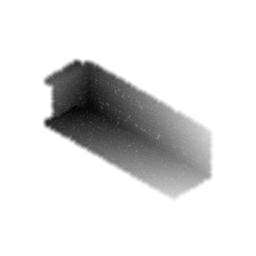}& 
    \includegraphics[width=0.11\columnwidth,trim=30 30 30 30, clip]{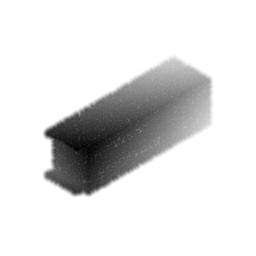}&
    \includegraphics[width=0.11\columnwidth,trim=30 30 30 30, clip]{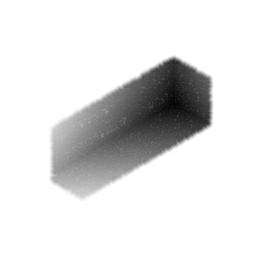}&
    \includegraphics[width=0.11\columnwidth,trim=30 30 30 30, clip]{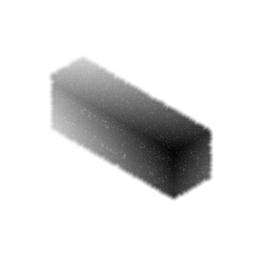}\\

    \bottomrule
\end{tabular}
}
\caption{Multi-view depth maps rendered with different point numbers. A denser point cloud can alleviate the point scattering artifacts in its rendered depth maps.}
\label{fig:compare_res}
\end{figure*}

We also illustrate the multi-view depth maps of the same shape that are rendered with different point numbers in Figure~\ref{fig:compare_res}. It demonstrates that a denser point cloud can alleviate the point scattering artifacts in rendered depth maps.

\section{Robustness Study}

\begin{figure*}
\center
\setlength\tabcolsep{0pt}
{
\renewcommand{\arraystretch}{0.0}
\small
\begin{tabular}{@{}rccc@{}}
    & Input & Ours & Groundtruth\\
    \raisebox{2.5\height}{\rotatebox{90}{Head}}~&
    \includegraphics[width=0.25\columnwidth,trim=30 30 30 30, clip]{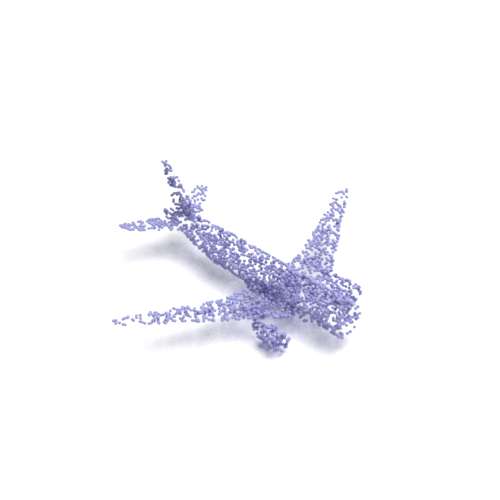}&
    \includegraphics[width=0.25\columnwidth,trim=30 30 30 30, clip]{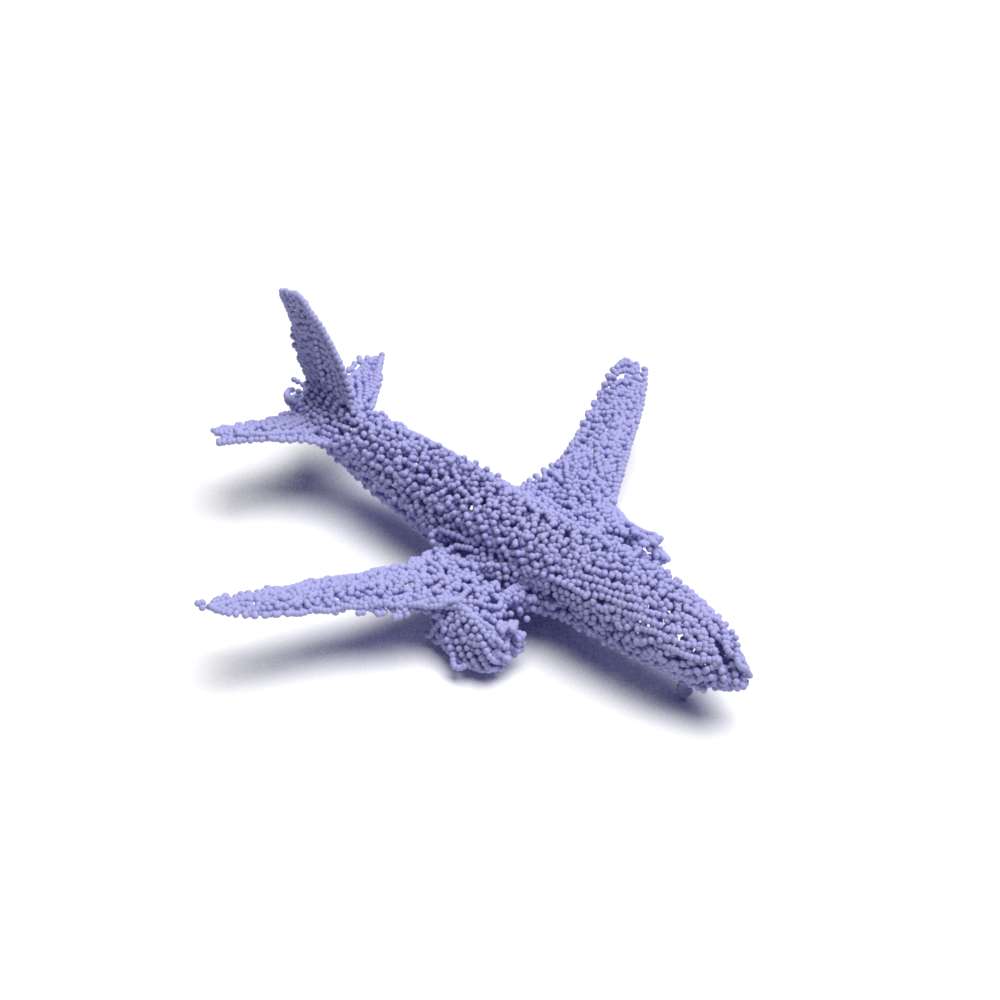}&
    \multirow{2}*{
    \includegraphics[width=0.4\columnwidth,trim=30 30 30 30, clip]{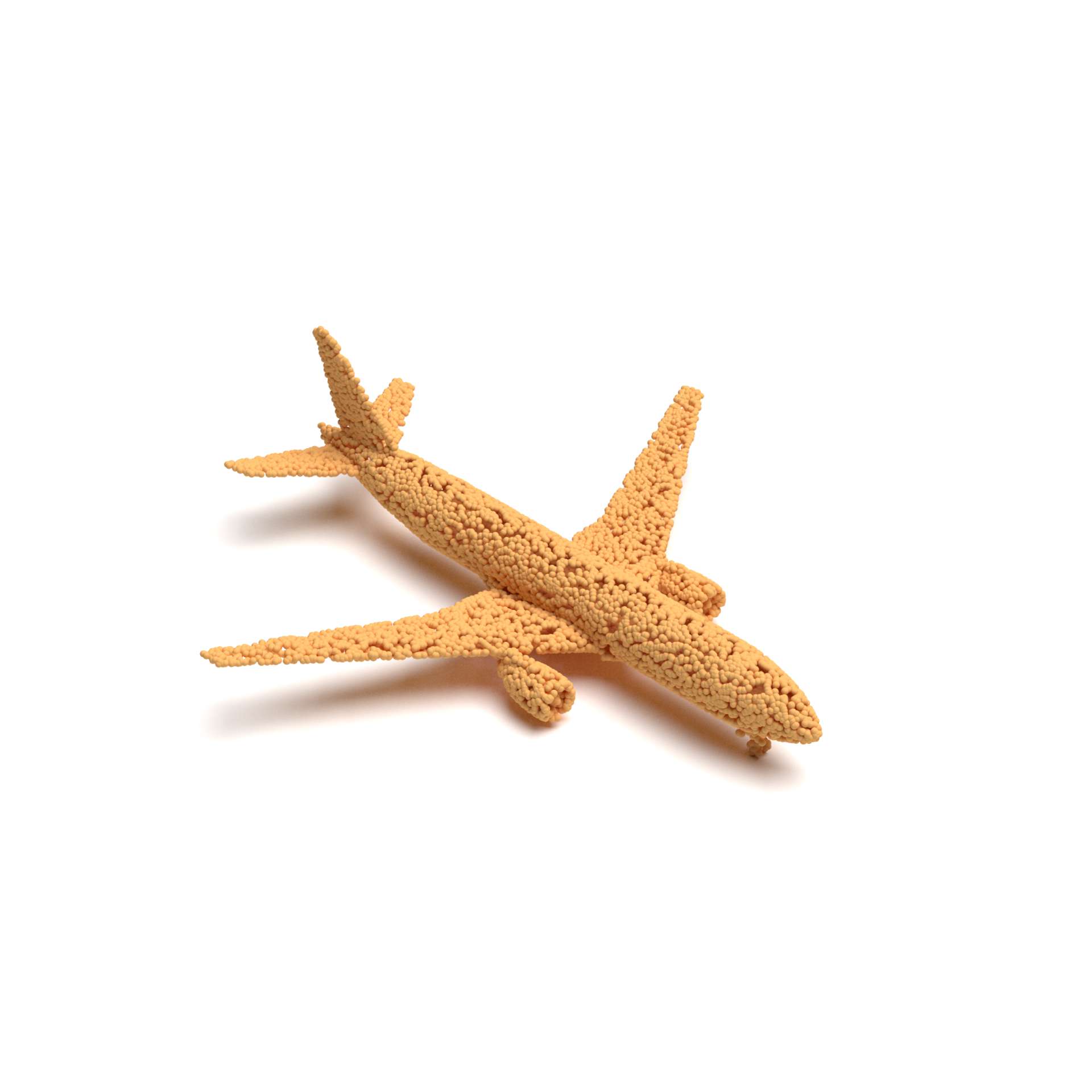}
    }\\
    
    \raisebox{2.5\height}{\rotatebox{90}{Tail}}~&
    \includegraphics[width=0.25\columnwidth,trim=30 30 30 30, clip]{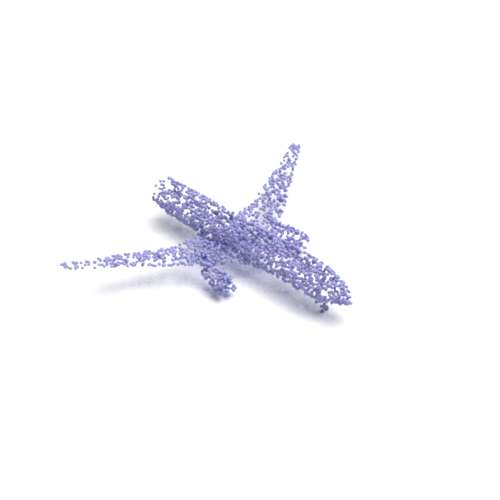}&
    \includegraphics[width=0.25\columnwidth,trim=30 30 30 30, clip]{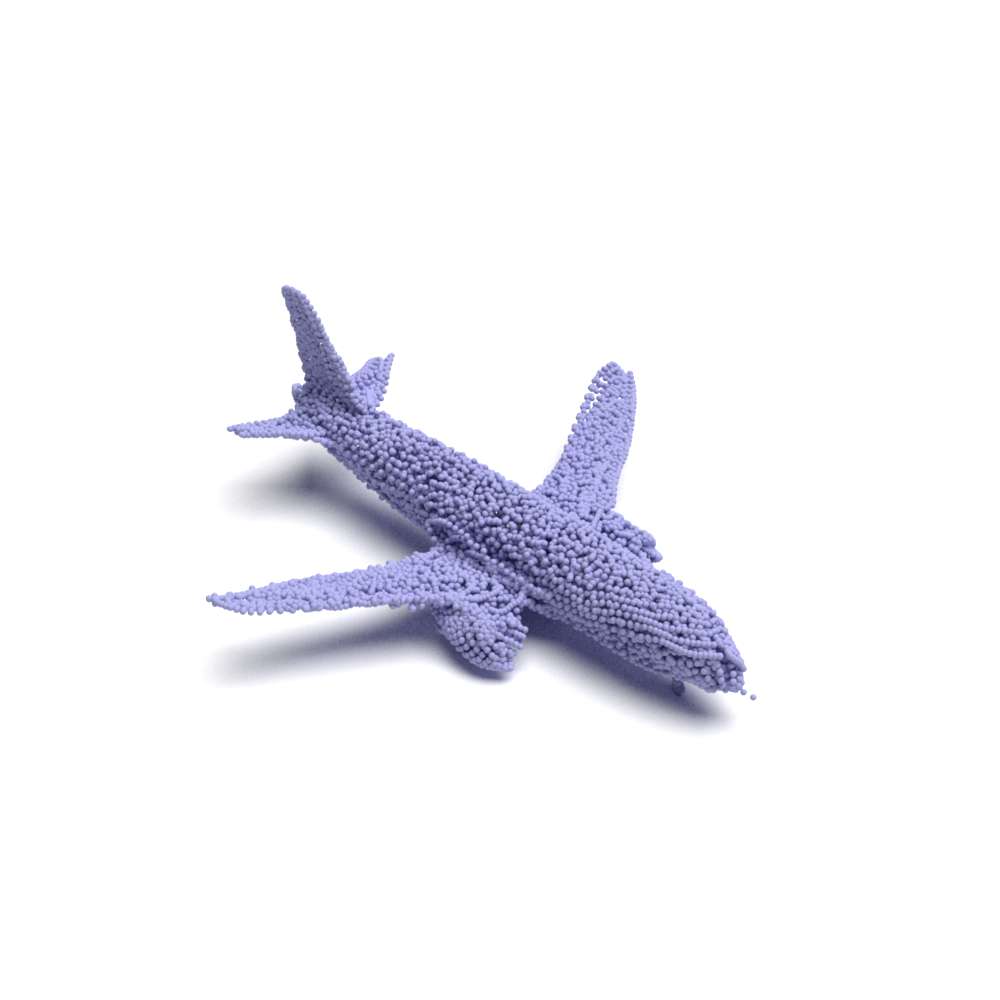}&\\
    
    \raisebox{2.5\height}{\rotatebox{90}{Right}}~&
    \includegraphics[width=0.25\columnwidth,trim=30 30 30 30, clip]{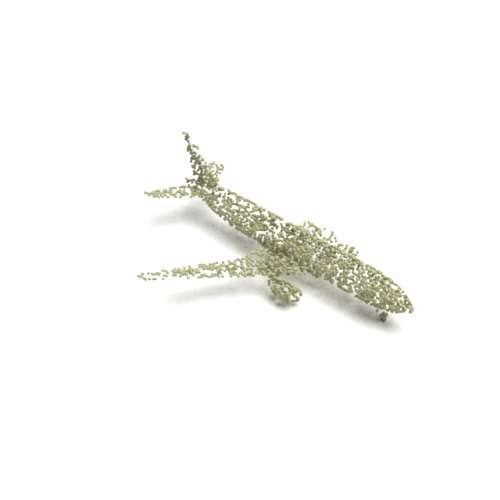}&
    \includegraphics[width=0.25\columnwidth,trim=30 30 30 30, clip]{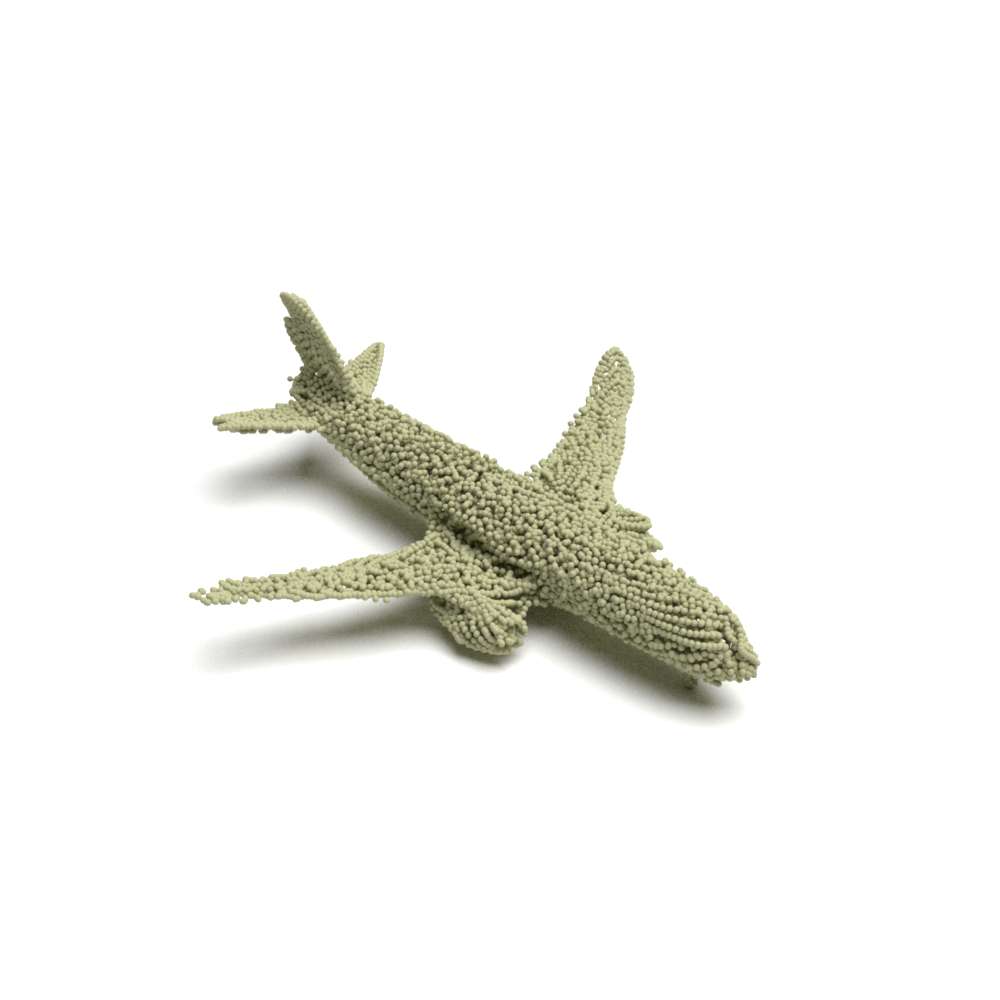}\\
    
    \raisebox{2.5\height}{\rotatebox{90}{Left}}~&
    \includegraphics[width=0.25\columnwidth,trim=30 30 30 30, clip]{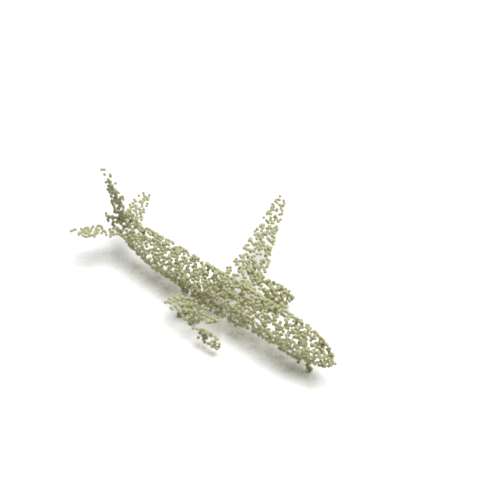}&
    \includegraphics[width=0.25\columnwidth,trim=30 30 30 30, clip]{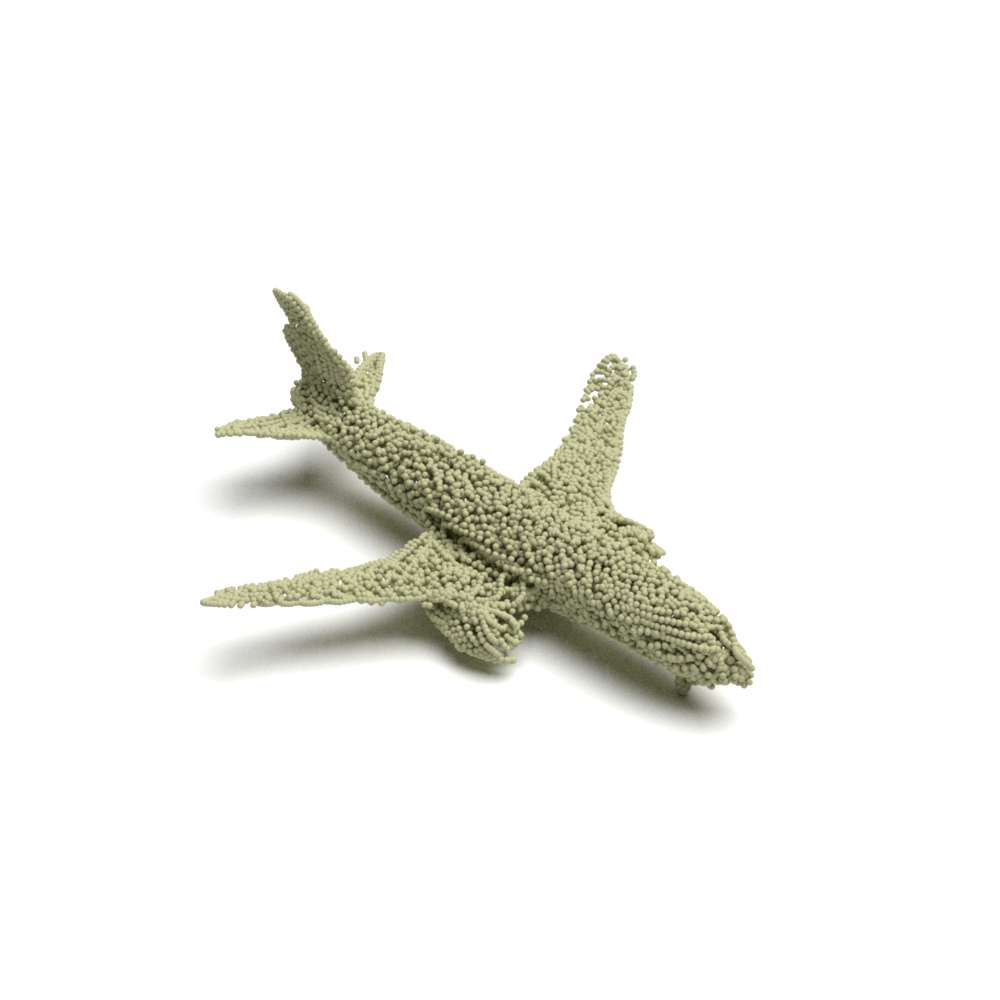}\\
       
\end{tabular}
}
\caption{Completing the same 3D shape from partial points that are sampled from four different view angles.}
\label{fig:robustness1}
\end{figure*}

\begin{figure*}
\center
\setlength\tabcolsep{0pt}
{
\renewcommand{\arraystretch}{0.0}
\small
\begin{tabular}{@{}rccc@{}}
    & Input & Ours & Groundtruth\\
    \raisebox{2.5\height}{\rotatebox{90}{1500}}~&
    \includegraphics[width=0.25\columnwidth,trim=30 30 30 30, clip]{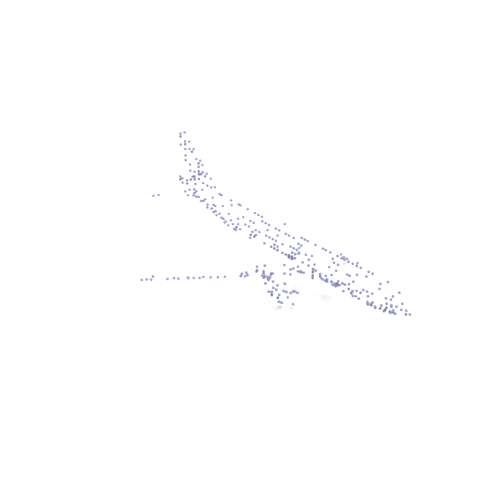}&
    \includegraphics[width=0.25\columnwidth,trim=30 30 30 30, clip]{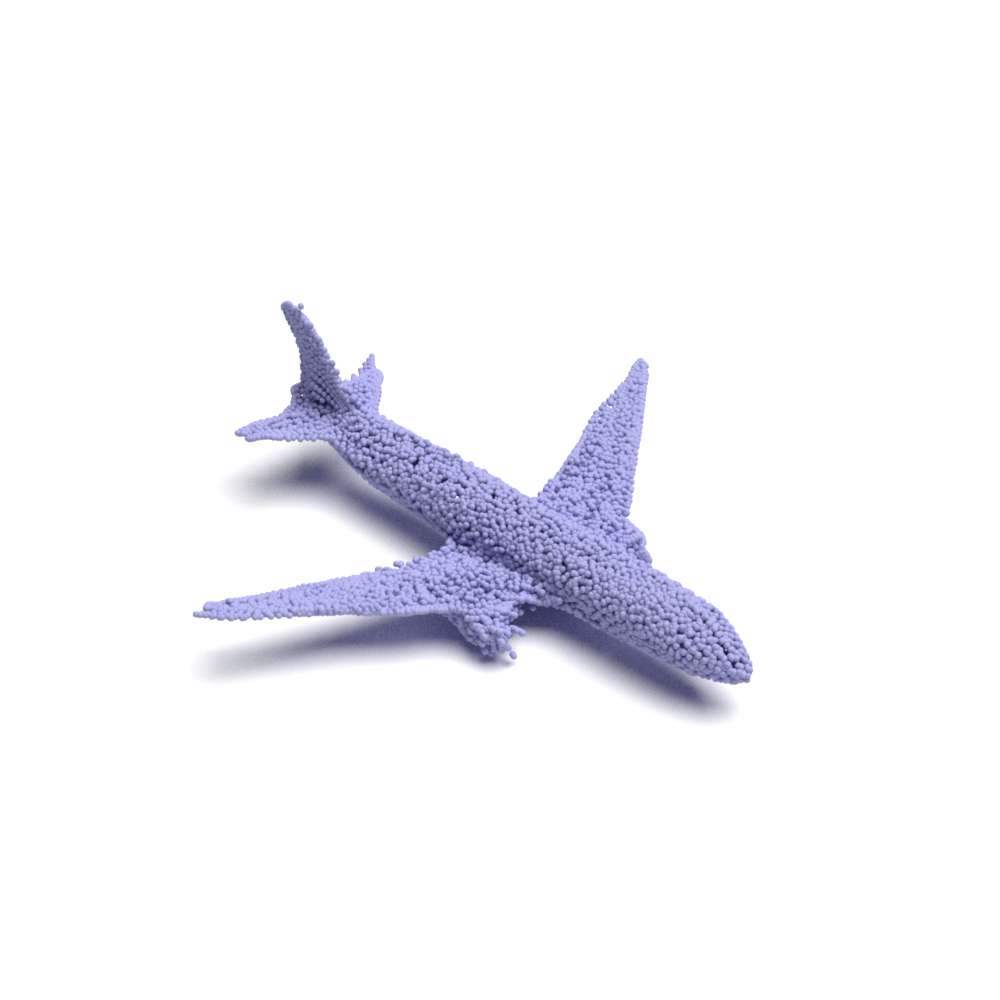}&
    \multirow{2}*{
    \includegraphics[width=0.4\columnwidth,trim=30 30 30 30, clip]{imgs/supplementary/robust1/gt.jpg}
    }\\

    \raisebox{2.5\height}{\rotatebox{90}{2000}}~&
    \includegraphics[width=0.25\columnwidth,trim=30 30 30 30, clip]{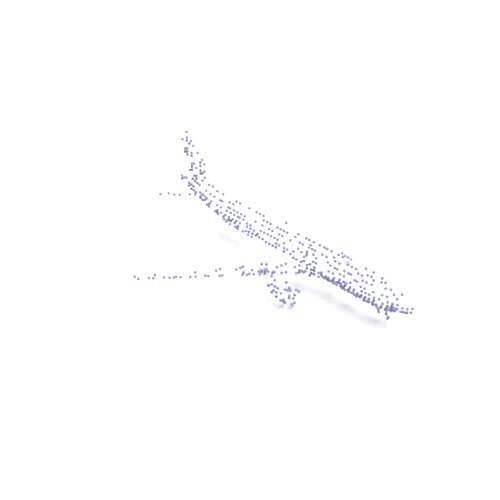}&
    \includegraphics[width=0.25\columnwidth,trim=30 30 30 30, clip]{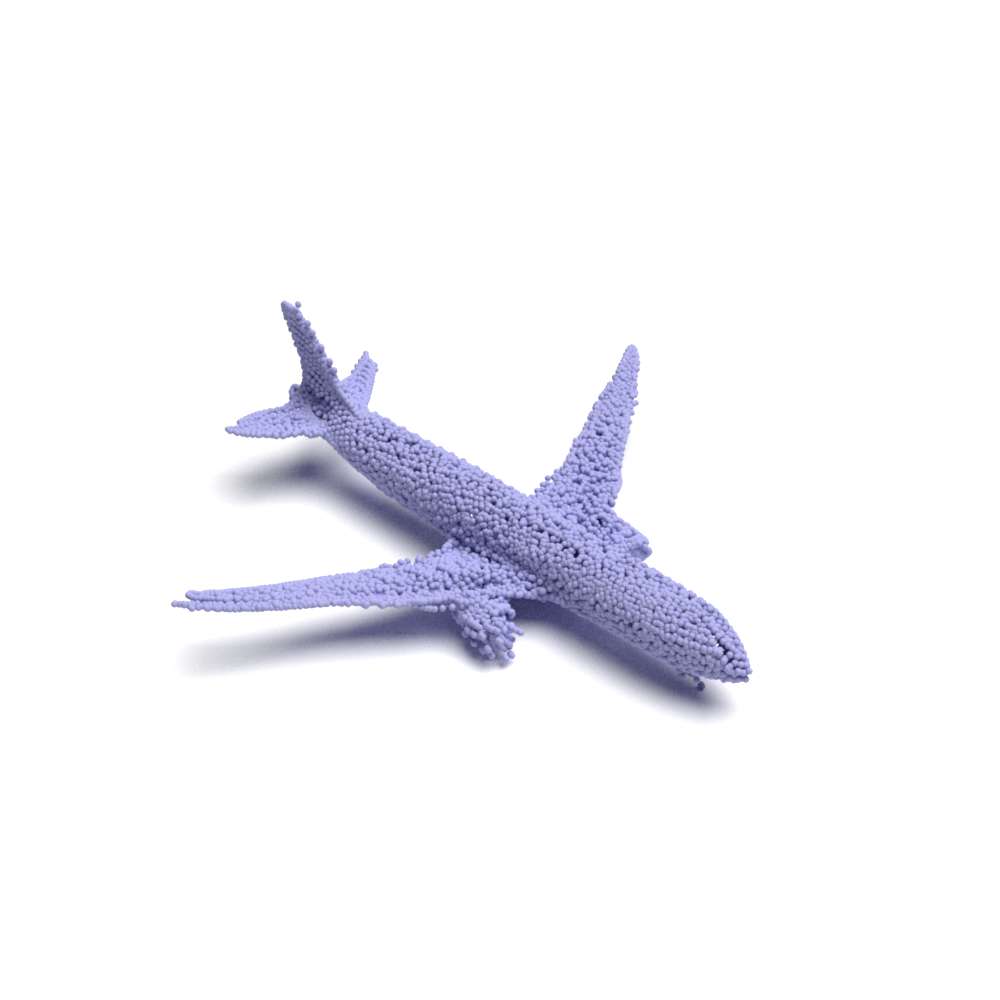}&\\
    
    \raisebox{2.5\height}{\rotatebox{90}{2500}}~&
    \includegraphics[width=0.25\columnwidth,trim=30 30 30 30, clip]{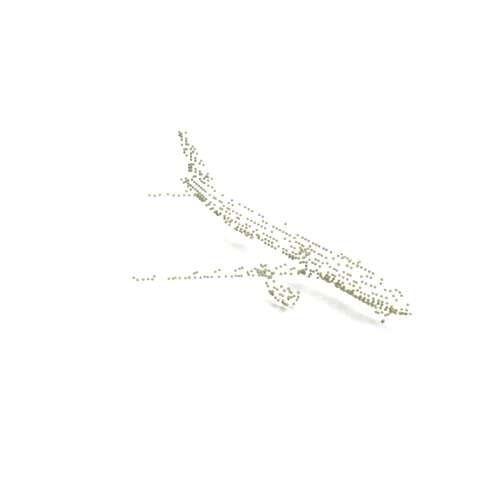}&
    \includegraphics[width=0.25\columnwidth,trim=30 30 30 30, clip]{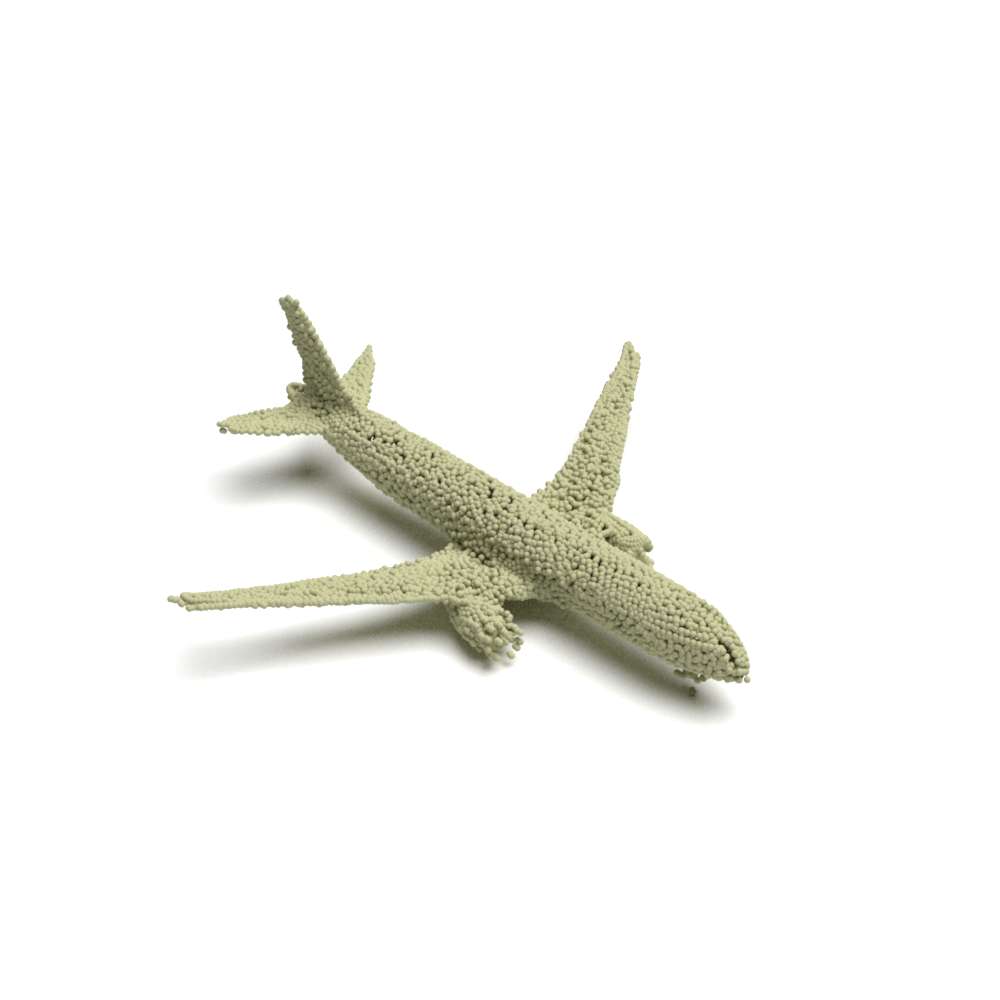}\\
    
    \raisebox{2.5\height}{\rotatebox{90}{3000}}~&
    \includegraphics[width=0.25\columnwidth,trim=30 30 30 30, clip]{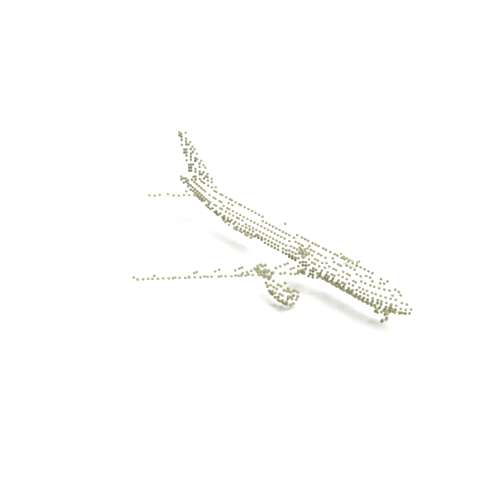}&
    \includegraphics[width=0.25\columnwidth,trim=30 30 30 30, clip]{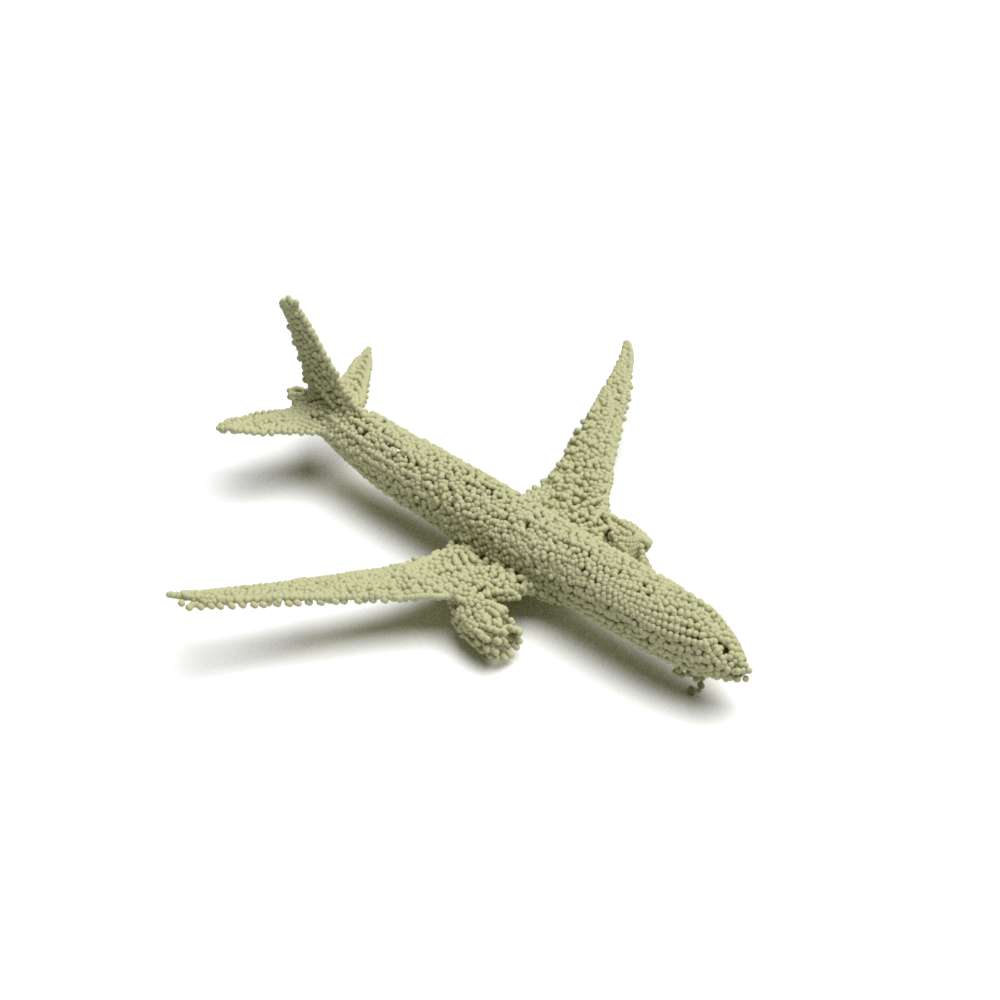}\\
       
\end{tabular}
}
\caption{Completing the same 3D shape from partial points of various densities (from 1500 points to 3000 points).}
\label{fig:robustness2}
\end{figure*}

We finally visualize point completion results of the same 3D shape, but from partial points that are sampled from different view angles (Figure~\ref{fig:robustness1}) or with different sampling densities (Figure~\ref{fig:robustness2}). These results verify that our method is robust to various acquisition conditions, such as different viewpoint and point density.

\end{document}